\documentclass[10pt,logo,copyright]{nvidiatechreport}
\usepackage[authoryear,round]{natbib}

\usepackage[utf8]{inputenc} 
\usepackage[T1]{fontenc}    

\usepackage{parskip}        
\usepackage{url}            
\usepackage{booktabs}       
\usepackage{amsfonts}       
\usepackage{nicefrac}       
\usepackage{microtype}      
\usepackage{xcolor}         
\usepackage[dvipsnames]{xcolor} 
\usepackage{graphicx}
\usepackage{animate}        
\usepackage{subcaption}
\usepackage{tabularx}
\usepackage{makecell}
\usepackage{adjustbox}
\usepackage{setspace}
\newcolumntype{M}[1]{>{\centering\arraybackslash}m{#1}}
\usepackage{float}
\usepackage{tikz}
\usetikzlibrary{positioning,shapes,arrows}
\usepackage{amsmath,amsfonts,bm, bbm,leftindex}
\usepackage{multirow}
\usepackage{comment}
\usepackage{gensymb}
\usepackage{lipsum}
\usetikzlibrary{arrows.meta, positioning, fit}
\usepackage[para]{threeparttable}
\usepackage{tikz}
\usetikzlibrary{tikzmark}

\usepackage[most]{tcolorbox}
\usepackage{fancyvrb}
\usepackage{fvextra}
\usepackage{dashrule}
\usepackage{nicematrix}

\usepackage{multicol}

\def\eqref#1{equation~\ref{#1}}

\def\1{\bm{1}}

\DeclareMathAlphabet{\mathsfit}{\encodingdefault}{\sfdefault}{m}{sl}
\SetMathAlphabet{\mathsfit}{bold}{\encodingdefault}{\sfdefault}{bx}{n}

\makeatletter
\let\save@mathaccent\mathaccent
\newcommand*\if@single[3]{%
  \setbox0\hbox{${\mathaccent"0362{#1}}^H$}%
  \setbox2\hbox{${\mathaccent"0362{\kern0pt#1}}^H$}%
  \ifdim\ht0=\ht2 #3\else #2\fi
  }
\newcommand*\rel@kern[1]{\kern#1\dimexpr\macc@kerna}
\newcommand*\widebar[1]{\@ifnextchar^{{\wide@bar{#1}{0}}}{\wide@bar{#1}{1}}}
\newcommand*\wide@bar[2]{\if@single{#1}{\wide@bar@{#1}{#2}{1}}{\wide@bar@{#1}{#2}{2}}}
\newcommand*\wide@bar@[3]{%
  \begingroup
  \def\mathaccent##1##2{%
    \let\mathaccent\save@mathaccent
    \if#32 \let\macc@nucleus\first@char \fi
    \setbox\z@\hbox{$\macc@style{\macc@nucleus}_{}$}%
    \setbox\tw@\hbox{$\macc@style{\macc@nucleus}{}_{}$}%
    \dimen@\wd\tw@
    \advance\dimen@-\wd\z@
    \divide\dimen@ 3
    \@tempdima\wd\tw@
    \advance\@tempdima-\scriptspace
    \divide\@tempdima 10
    \advance\dimen@-\@tempdima
    \ifdim\dimen@>\z@ \dimen@0pt\fi
    \rel@kern{0.6}\kern-\dimen@
    \if#31
      \overline{\rel@kern{-0.6}\kern\dimen@\macc@nucleus\rel@kern{0.4}\kern\dimen@}%
      \advance\dimen@0.4\dimexpr\macc@kerna
      \let\final@kern#2%
      \ifdim\dimen@<\z@ \let\final@kern1\fi
      \if\final@kern1 \kern-\dimen@\fi
    \else
      \overline{\rel@kern{-0.6}\kern\dimen@#1}%
    \fi
  }%
  \macc@depth\@ne
  \let\math@bgroup\@empty \let\math@egroup\macc@set@skewchar
  \mathsurround\z@ \frozen@everymath{\mathgroup\macc@group\relax}%
  \macc@set@skewchar\relax
  \let\mathaccentV\macc@nested@a
  \if#31
    \macc@nested@a\relax111{#1}%
  \else
    \def\gobble@till@marker##1\endmarker{}%
    \futurelet\first@char\gobble@till@marker#1\endmarker
    \ifcat\noexpand\first@char A\else
      \def\first@char{}%
    \fi
    \macc@nested@a\relax111{\first@char}%
  \fi
  \endgroup
}
\makeatother

\usepackage{pifont}
\usepackage{diagbox}

\usepackage[nameinlink]{cleveref}
\crefname{equation}{Eq.}{Eqs.}
\crefname{figure}{Fig.}{Figs.}
\crefname{section}{Sec.}{Sec.}
\crefname{appendix}{App.}{App.}
\crefname{table}{Tab.}{Tabs.}
\crefname{algorithm}{Algo}{Algo}
\crefname{thm}{Thm}{Thm}
\Crefname{thm}{Thm}{Thm}
\crefname{prop}{Prop}{Prop}

\definecolor{darkred}{rgb}{0.7, 0.0, 0.0}

\newcommand{\crefnames}[3]{%
  \@for\next:=#1\do{%
    \expandafter\crefname\expandafter{\next}{#2}{#3}%
  }%
}

\title{World Simulation with Video Foundation Models for Physical AI}

\author{NVIDIA\footnote{A detailed list of contributors and acknowledgments can be found in~\cref{sec::contributors} of this paper.}}

\begin{abstract}
We introduce [Cosmos-Predict2.5], the latest generation of the Cosmos World Foundation Models for Physical AI. Built on a flow-based architecture, [Cosmos-Predict2.5] unifies Text2World, Image2World, and Video2World generation in a single model and leverages [Cosmos-Reason1], a Physical AI vision-language model, to provide richer text grounding and finer control of world simulation. Trained on 200M curated video clips and refined with reinforcement learning-based post-training, [Cosmos-Predict2.5] achieves substantial improvements over [Cosmos-Predict1] in video quality and instruction alignment, with models released at 2B and 14B scales. These capabilities enable more reliable synthetic data generation, policy evaluation, and closed-loop simulation for robotics and autonomous systems. We further extend the family with [Cosmos-Transfer2.5], a control-net style framework for Sim2Real and Real2Real world translation. Despite being 3.5× smaller than [Cosmos-Transfer1], it delivers higher fidelity and robust long-horizon video generation. Together, these advances establish [Cosmos-Predict2.5] and [Cosmos-Transfer2.5] as versatile tools for scaling embodied intelligence. To accelerate research and deployment in Physical AI, we release source code, pretrained checkpoints, and curated benchmarks under the NVIDIA Open Model License at \url{https://github.com/nvidia-cosmos/cosmos-predict2.5} and \url{https://github.com/nvidia-cosmos/cosmos-transfer2.5}. We hope these open resources lower the barrier to adoption and foster innovation in building the next generation of embodied intelligence.
\end{abstract}

\begin{document}
\maketitle
\abscontent

\newpage
\tableofcontents
\newpage

\section{Introduction}
\label{sec::intro}

Physical AI systems---embodied agents equipped with sensors and actuators---assist humans in carrying out physical tasks by interacting with the environment: their actuators act on the world in response to sensory inputs. Training such systems directly in the real world, however, is often slow, costly, and risky. This is particularly true in the early stages, when system imperfections may lead to unsafe actions that damage either the agent, the environment, or both. A world simulator that can generate high-quality, diverse visual environments based on the actions of a Physical AI agent can serve as a safe proxy for the physical world. Such simulators enable agents to acquire perception and control skills entirely in silicon before deployment in the real world~\citep {cosmos_v1}. In this paper, we introduce [Cosmos-Predict2.5], our latest world foundation model based on flow matching that significantly enhances simulation fidelity across diverse Physical AI domains.

[Cosmos-Predict2.5] leapfrogs the diffusion video world model in [Cosmos-Predict1]~\citep{cosmos_v1} via three key improvements. First, we strengthen the data filtering pipeline to produce higher-quality pre-training datasets and manually curate specialized post-training data tailored for Physical AI. Second, we simplify the model architecture and combine Text2World, Image2World, and Video2World capabilities into a single model. Third, we improve the training recipe, leveraging model merging and a novel reinforcement learning algorithm for post-training, and replace the T5 text encoder used in [Cosmos-Predict1] with [Cosmos-Reason1]~\citep{azzolini2025cosmos}, a modern decoder-only VLM architecture specialized for Physical AI, providing richer text representations and enabling finer-grained control over world generation. Through extensive experiments, we demonstrate that [Cosmos-Predict2.5] delivers substantial gains over [Cosmos-Predict1] in both output quality and prompt alignment.

We further demonstrate that these advancements yield broad and practical benefits across diverse downstream Physical AI applications. In particular, they enable more efficient synthetic data generation for Vision-Language-Action (VLA) model training~\citep{jang2025dreamgen}, a key ingredient for scaling embodied intelligence. Beyond this, [Cosmos-Predict2.5] improves action-conditioned video world generation for policy validation, enhances coherent multi-view video world generation for autonomous driving simulation, and supports camera-controllable multi-view video world generation for robotic manipulation.

Beyond these use cases, we expand [Cosmos-Predict2.5] into a broader family of control-net models, termed [Cosmos-Transfer2.5], designed for diverse visual world-translation tasks. These include enhancing the photorealism of physical simulator outputs~\citep{cosmos_transfer1}, augmenting real-world videos~\citep{ren2025cosmos}, and converting semantic world scenarios into realistic, multi-view sensory inputs for Physical AI agents~\citep{cosmos_v1}. Compared to its predecessor [Cosmos-Transfer1], the new framework delivers substantially higher quality while being 3.5× smaller in size. Moreover, [Cosmos-Transfer2.5] demonstrates the ability to generate robust long-horizon video translations and enable closed-loop simulation---two essential capabilities for advancing the next stage of Physical AI research and deployment.

To further accelerate progress in this domain, we are releasing our source code, pre-trained checkpoints, and curated post-training examples to the community. By providing these open resources, we aim to lower the barrier for practitioners to adapt and specialize the pre-trained models for their own targeted Physical AI setups---whether in robotics, autonomous systems, or embodied reasoning. \cref{tab:model_inventory} provides a clear mapping of the released models and their corresponding capabilities, offering a practical guide for researchers and developers. We hope that by sharing these assets, we can foster broader adoption, reproducibility, and innovation in Physical AI.

\begin{table}[t]
    \centering
    \setlength{\tabcolsep}{12pt} 
    \footnotesize
    \captionsetup{justification=centering}
    \begin{threeparttable}
        \caption{List of released models with their corresponding capabilities and inputs.}
        \label{tab:model_inventory}
        \begin{tabular}{lll}
            \toprule
            \textbf{Model Name} & \textbf{Capability} & \textbf{Input} \\
            \midrule
            \multicolumn{3}{c}{\textbf{Cosmos-Predict2.5 base}} \\
            \midrule
            Cosmos-Predict2.5-2B/pre-trained & pre-trained base & text + image or video \\
            Cosmos-Predict2.5-14B/pre-trained & pre-trained base & text + image or video \\
            Cosmos-Predict2.5-2B/post-trained & post-trained base & text + image or video \\
            Cosmos-Predict2.5-14B/post-trained & post-trained base & text + image or video \\
            \midrule
            \multicolumn{3}{c}{\textbf{Cosmos-Predict2.5 domain specialized}} \\
            \midrule
            Cosmos-Predict2.5-2B/auto/multiview & driving, 7-camera view & text + image or video \\
            Cosmos-Predict2.5-2B/robot/action-cond & robotic, action-conditioned & action \\
            Cosmos-Predict2.5-2B/robot/multiview-agibot & robotic, AgiBot data, 3-camera view & text + image \\
            Cosmos-Predict2.5-14B/robot/gr00tdream-gr1 & robotic, GR00T GR1 data & text + image or video \\
            \midrule
            \multicolumn{3}{c}{\textbf{Cosmos-Transfer2.5}} \\
            \midrule
            Cosmos-Transfer2.5-2B/general & controlnet & edge, blur, segmentation, depth \\
            Cosmos-Transfer2.5-2B/auto/multiview & driving, multiview controlnet & world scenario map \\  
            Cosmos-Transfer2.5-2B/robot/multiview & robotic, 3-camera view & text + third-person video \\
            Cosmos-Transfer2.5-2B/robot/multiview-agibot & robotic, AgiBot data, 3-camera view & text + head-view video \\
            \bottomrule
        \end{tabular}
    \end{threeparttable}
\end{table}

\section{Data}
\label{sec::data}

We improve upon the data pipeline in~\cite{cosmos_v1} in two aspects. First, we upgrade the components in the filtering pipeline for general data processing. Second, we curate a set of high-quality Physical AI data to strengthen the capability of our models in this domain.
\subsection{Video Curation Pipeline}
\label{subsec:video_curation_pipeline}

\begin{figure}[htbp]
    \centering
    \includegraphics[width=0.95\textwidth]{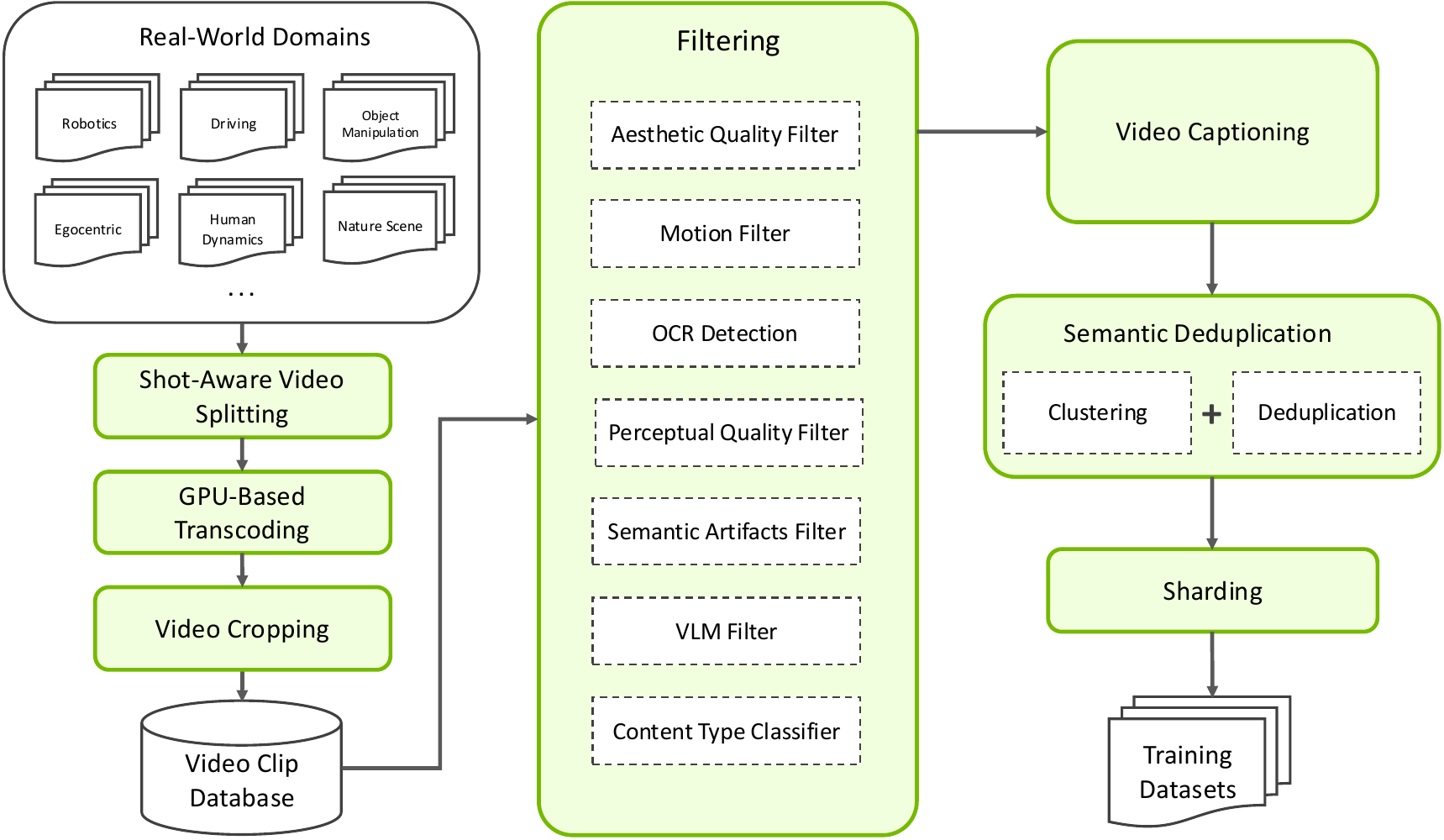}
    \caption{
    Our video curation pipeline transforms raw, unstructured video data from diverse real-world sources into a high-quality, annotated, deduplicated, and sharded dataset optimized for large-scale training of video world foundation models.
    }
    \label{fig:3_curation_pipeline}
\end{figure}

Our video curation pipeline is shown in~\cref{fig:3_curation_pipeline}. It comprises seven stages: 1) shot-aware video splitting, 2) GPU-based transcoding, 3) video cropping, 4) filtering, 5) captioning, 6) semantic deduplication, and 7) sharding. Each stage is designed for high-throughput processing of enormous amounts of video data, with an emphasis on obtaining large-scale, high-quality, and semantically diverse data.

Building on the foundation established in~[Cosmos-Predict1]~\citep{cosmos_v1}, we refine and substantially scale up our video data curation pipeline in [Cosmos-Predict2.5]. We processed over 200 million raw videos sourced from both proprietary datasets and open internet platforms. These videos cover domains such as driving, object manipulation, spatial navigation, human interaction, and nature scenes, among others, ensuring broad generalization across Physical AI use cases.

The pipeline begins by segmenting long-form videos into shots using high-accuracy boundary detection models, ensuring that raw shot transitions are excluded. Each segment is then GPU-accelerated, transcoded, and cropped to eliminate black borders and spatial padding. Very short clips under 5 seconds are discarded, while the remaining segments yield over 6 billion curated clips ranging from 5 to 60 seconds in length.

A multi-stage filtering process then removes low-quality or unsuitable data. Filters target motion artifacts, distortion, visual noise, overlay text, content that is unsuitable for training, and mismatched video types. A deduplication step is further applied to remove videos that are semantically similar. Only about 4\% of the initial clips pass all filters, producing a curated dataset of approximately 200 million trainable clips. These 200 million clips form our pre-training dataset.

This multi-stage filtering pipeline comprises several key components, each serving a unique purpose. To begin with, we apply an aesthetic scoring filter, which grades the inputs by their aesthetic quality. Following this, we apply a motion filter, which quantifies and removes clips based on their degree of motion. The third stage is an OCR filter that attempts to remove clips with excessive text overlay. In the fourth stage, we apply a perceptual quality filter (akin to DOVER~\citep{dover}) to weed out clips with technical distortions and perceptual artifacts. Next, we use a ``semantic artifacts'' filter (akin to VTSS~\citep{koala}) that aims to filter out clips with semantic artifacts (video-in-video, poor transitions, \etc). Finally, we use a vision language model (VLM)~\citep{qwen2p5vl} to further remove clips with a set of undesirable issues with higher precision. We apply the VLM at the very end of filtering because it is computationally more expensive. Surviving clips are subsequently categorized via a video content-type classifier, which enables structured downstream use of the dataset. At this stage, we further exclude content depicting physically unrealistic phenomena---such as video games, synthetic visual patterns, animations, or cartoons---to maintain alignment with the physical world distributions.

In the video captioning stage, we segment each clip into 5-second windows and caption each window using a Qwen2.5-VL-7B vision-language model~\citep{qwen2p5vl}, prompting it to generate factual, context-aware captions. We apply targeted prompt engineering to guide the model toward descriptions that emphasize the primary object, its motion, and key semantic details in the scene. Captions are produced at multiple lengths (short, medium, and long) to support diverse use cases, serving as both supervision signals and conditioning prompts for the model training.

In the semantic deduplication stage, we first assign video clips to clusters using embedding-based similarity. Within each cluster, clips are compared pairwise to detect semantically similar content, and the highest-resolution version is retained, as higher resolution preserves finer visual details and provides a richer signal for training. To support incremental and large-scale data curation, we adopt an online deduplication strategy: each new clip is compared against previously retained clips, with preference given to older and higher-resolution clips during tie-breaking. This enables scalable deduplication across growing datasets while maintaining semantic consistency across the full corpus.

To support scalable and flexible training, we implement a top-down sharding strategy. Using internally trained content type classifiers, each clip is assigned a semantic label from a custom-built taxonomy of 26 video types. The dataset is then sharded along multiple axes: content type, resolution, aspect ratio, and length. This structured sharding enables efficient sampling, curriculum-based training, and fine-grained domain balancing.

At the same time, the underlying infrastructure has been upgraded to handle petabyte-scale data processing, providing the capacity required for massive video corpora. The pipeline is built on highly parallelized workflows with dynamic auto-scaling of CPU and GPU worker allocation, ensuring workloads are efficiently balanced across heterogeneous resources. To further improve throughput, we employ video chunking and frame-rate control during inference, which reduces redundant computation while preserving semantic fidelity. Beyond ingestion and processing, the infrastructure integrates with a Delta Lake–based lakehouse~\citep{delta_lake_databricks_2019} for large-scale SQL analytics and Milvus~\citep{milvus_zilliz_2019}, an open-source vector database for embedding-based search, enabling advanced semantic video-content similarity search and caption-level text embedding retrieval. Together, these analytical capabilities not only improve current training efficiency but also lay the foundation for large-scale dataset exploration, retrieval-augmented training, and fine-grained knowledge mining.

In contrast to the pipeline in~[Cosmos-Predict1], the [Cosmos-Predict2.5] pipeline scales to a much larger volume, processing 35 million hours of raw video compared to 20 million hours, and producing over 6 billion clips from which 200 million high-quality clips are retained. At the same time, it achieves improved data quality control through a far stricter multi-stage filtering pipeline, which reduces survival from 30\% of clips to only 4\%. This pipeline systematically removes motion artifacts, distortions, overlay text, semantic artifacts, and other undesirable issues, with a final high-precision pass by a vision-language model. The pipeline further introduces finer content granularity by segmenting clips into shorter temporal windows, generating captions at multiple levels of detail, and structuring the dataset through semantic deduplication and sharding, resulting in richer and more precise supervision signals. These advances are supported by a more robust and scalable infrastructure, designed for petabyte-scale processing, flexible resource allocation, and advanced analytics. Together, these advances yield a dataset that is larger, cleaner, and semantically richer, underpinned by scalable infrastructure that facilitates enhanced pre-training efficiency and improved downstream generalization across diverse Physical AI domains.

\subsection{Domain Specific Data}
\label{subsec:domain_specific}

To curate high-quality data across diverse Physical AI domains, we design domain-specific pipelines that collect and annotate visual data tailored to each domain. We focus on five target domains: Robotics, Autonomous Driving, Smart Spaces, Human Dynamics, and Physics. The combined output is added to the general pre-training data.

Each domain follows a curation process similar to that used in pre-training (\cref{fig:3_curation_pipeline}), but with two key differences in filtering and captioning. For filtering, we omit the VLM filter and instead apply a domain-specific subset of filters with adjusted hyperparameter values. For captioning, we employ a larger VLM model \citep{qwen2p5vl}, incorporating customized prompts tailored to each domain. The following sub-sections provide detailed descriptions of the curation process for each domain.

\subsubsection{Robotics}

We sourced robotics datasets spanning diverse settings. For each dataset, we filtered out low-resolution and near-static videos. To ensure a consistent pace of action across the datasets, we increased the playback speed for videos featuring overly slow robotic movements. The resulting statistics are summarized in~\cref{tab:robot_datastats}.

\begin{table}[ht!]
    \centering
    \setlength{\tabcolsep}{6pt} 
    \small
    \captionsetup{justification=centering}
    \begin{threeparttable}
        \caption{Overview of high-quality robotics datasets with video counts by camera perspective.}
        \label{tab:robot_datastats}
        \begin{tabular}{l|cccc}
            \toprule
            \textbf{Dataset} & \textbf{Embodiment} & \textbf{Central-view} & \textbf{Left-view} & \textbf{Right-view} \\
            \midrule
            AgiBot-Beta~\citep{bu2025agibot} & Bimanual & 194k & 30k & 30k \\
            Bridge~\citep{walke2023bridgedata} & Single-arm & 36k & - & - \\
            DROID~\citep{khazatsky2024droid} & Single-arm & 39k (wrist) & 51k & 51k \\
            GR00T~\citep{bjorck2025gr00t} & Bimanual & 3k & - & - \\
            1X~\citep{1XTechno4} & Bimanual & 17k & - & - \\
            OpenX~\citep{vuong2023open} & Single-arm & 500 & - & - \\
            RoboMIND~\citep{wu2024robomind} & Dual-arm/Humanoid & 16k & 6k & 7k \\
            \bottomrule
        \end{tabular}
    \end{threeparttable}
\end{table}

We design dataset‑aware caption prompts that enforce task‑centric, grounded descriptions while normalizing viewpoint and embodiment. Prompts require enumerating the initial scene, then describing the robot’s actions chronologically with explicit motion types (\eg, linear, rotational), involved parts (arm, wrist, gripper), camera motion, and fine‑grained object attributes. 
To improve caption accuracy and reduce hallucination, we inject available dataset-specific metadata into the prompt. For example, we include task description with human-labeled success ratings for GR00T, step‑level instructions for Bridge, initial scene description for AgiBot, and unified camera perspectives across multiple dataset sources. 

\subsubsection{Autonomous Driving}

We built a proprietary dataset consisting of approximately 3.1M 20-second surround-view video clips collected using NVIDIA’s internal driving platform. Each clip includes recordings from seven synchronized cameras: front-wide, front-tele, left, right, rear, rear-left, and rear-right.

The dataset is sampled from a large-scale corpus to align with a target distribution of diverse driving attributes. The selected attributes encompass a wide range of conditions, including geographic regions (\eg, USA and Europe), traffic density (\eg, light and heavy), ego-vehicle speed (\eg, local roads and highways), ego-vehicle acceleration (\eg, constant and fast acceleration), ego-vehicle maneuvers (\eg, slow curves and sharp turns), road types (\eg, urban and rural), uncommon road structures (\eg, tunnels and tollbooths), visibility conditions (\eg, clear and foggy), weather (\eg, dry and snowy), and illumination (\eg, daytime and nighttime).

We design prompts for captioning autonomous driving scenarios by explicitly defining task requirements and emphasizing driving-relevant information. Specifically, the captions focus on the following aspects:
\begin{enumerate}
    \item Various agents (\eg, vehicles, pedestrians, cyclists) and traffic elements (\eg, traffic lights, traffic signs) that the ego vehicle should be aware of for safe navigation.
    \item Global environmental factors (\eg, weather, time of day, road conditions) that could influence driving behavior.
    \item Meta actions in both longitudinal and lateral of ego vehicle and surrounding vehicles.
    \item Speed of ego vehicle and surrounding vehicles.
    \item Dynamic actions or state transitions of other objects. 
    \item Interactions between key objects.
\end{enumerate}
To capture varying levels of detail, captions of each video are produced in three lengths: short, medium, and long.

\subsubsection{Smart Spaces}

We curate videos featuring scenarios that take place in warehouses, factories, construction sites, and other similar settings. We use the same pipeline for splitting these videos into individual shots as that used for the pretraining dataset. We use search keywords to find an initial set of videos that may be relevant to a smart space. For each video, we used a VLM~\citep{qwen2p5vl} to verify its relevance. After clipping and filtering, approximately 40K video clips survive. These clips are then captioned by a VLM~\citep{qwen2p5vl}. We prompt the VLM by specifying that the videos focus on smart spaces (factories, warehouses, industrial facilities, automobiles, and other manufacturing environments) and also tailor the language and style of the generated captions accordingly.

\subsubsection{Human Dynamics}

We curated a human-dynamics video dataset by retaining clips of at least 5 seconds and processing each video with YOLOX \citep{yolox2021} for human detection and RTMPose \citep{jiang2023rtmpose} for full-body keypoints and facial landmark estimation. A clip is included only when people appear in more than 40\% of its frames, no more than eight individuals are visible in any frame, and at least one person occupies 3\% percent or more of the image area. We generated captions with the VLM using prompts that emphasize detailed descriptions of human motion and dynamics. We include this dataset to enhance the simulation capabilities of Physical AI agents, enabling them to simulate human behavior for improved action planning.

\subsubsection{Physics}

We curate a dataset that aims at improving the physical plausibility of generated videos by systematically emphasizing real-world dynamics. To achieve this, we first define a taxonomy of visually observable physical phenomena spanning core domains such as classical mechanics and fluid mechanics. This taxonomy provides a principled framework for identifying and categorizing key behaviors and interactions—such as shattering glass, colliding rolling balls, or flowing water. Using this structure, we curate a diverse set of videos that foreground the dynamic properties of these phenomena. In addition, we design tailored captioning prompts that guide the VLM to generate accurate, detailed descriptions of both the underlying physical processes and the associated object interactions. Together, these elements produce a dataset that is systematically organized and tightly aligned with the goal of advancing physically grounded video generation.

\section{Method}
\label{sec::method}

In this section, we first discuss our flow-matching formulation and then present the network architecture. 

\subsection{Flow Matching}
\label{subsec::flowmatching}
We adopt flow matching (FM)~\citep{lipman2022flow} for training diffusion models because of its conceptual simplicity and practical effectiveness. While FM and the Elucidated Diffusion Model (EDM)~\citep{karras2022elucidating}, which was used in [Cosmos-Predict1]~\citep{cosmos_v1}, are mathematically equivalent in terms of their forward and backward diffusion processes, they differ in how the denoising network is parameterized~\citep{gao2025diffusion}. In EDM, the preconditioning coefficients are chosen so that both the inputs and outputs of the denoising network are approximately standardized Gaussians, which simplifies training and improves stability. In contrast, FM selects coefficients that make the denoising network predict the velocity of the diffusion trajectory. This velocity-based formulation not only provides a more direct training target but also tends to yield smoother optimization and improved sample quality in practice.

Formally, given a data sample $\mathbf{x}$ (image or video), a noise vector $\epsilon \sim \mathcal{N}(0, I)$, and a timestep $t \in [0,1]$ drawn from a logit-normal distribution, the interpolated latent $\mathbf{x}_t$ is defined as
\begin{equation}
    \mathbf{x}_t = (1-t)\mathbf{x} + t \epsilon.
\end{equation}
The corresponding ground-truth velocity is
\begin{equation}
    \mathbf{v}_t = \epsilon - \mathbf{x}.
\end{equation}
The model is trained to predict $\mathbf{v}_t$ by minimizing the mean squared error (MSE) between the prediction and ground truth:
\begin{equation}
    \mathcal{L}(\theta) = \mathbb{E}_{\mathbf{x}, \epsilon, \mathbf{c}, t} \left\| \mathbf{u}(\mathbf{x}_t, t, \mathbf{c}; \theta) - \mathbf{v}_t \right\|^2,
\end{equation}
where $\mathbf{c}$ denotes conditioning information associated with $\mathbf{x}$ (\eg, text embeddings, reference frames, and other conditional inputs), $\theta$ represents the model parameters, and $\mathbf{u}(\cdot; \theta)$ is the predicted velocity function.

High-resolution content often contains significant redundancy, since nearby pixels are highly correlated. As a result, if the level of injected noise is too small, the model may fail to “break apart” this correlation, making it harder for the FM model to learn meaningful structure~\citep{esser2024scaling,hoogeboom2023simple,chen2023importance,atzmon2024edify}. To address this, we deliberately bias the training process toward higher noise levels. Specifically, we adopt the shifted logit-normal distribution~\citep{esser2024scaling}. In practice, we first sample $t$ from a logit-normal distribution, and then apply the monotone transformation
\begin{equation}
t_{s} = \frac{\beta t}{1 + (\beta - 1) t}
\end{equation}
where $\beta$ is a shift hyper-parameter. This transformation reweights the distribution so that $t_{s}$ values are skewed toward higher noise. Intuitively, increasing $\beta$ pushes the model to encounter noisier inputs more frequently, which helps it learn to reconstruct signals even when correlations are heavily disrupted. When $\beta = 1$, no shift is applied and $t_{s} = t$.

\subsection{Network Architecture}
\label{subsec::network}

\begin{figure}[ht!]
    \centering
    \includegraphics[width=0.95\textwidth]{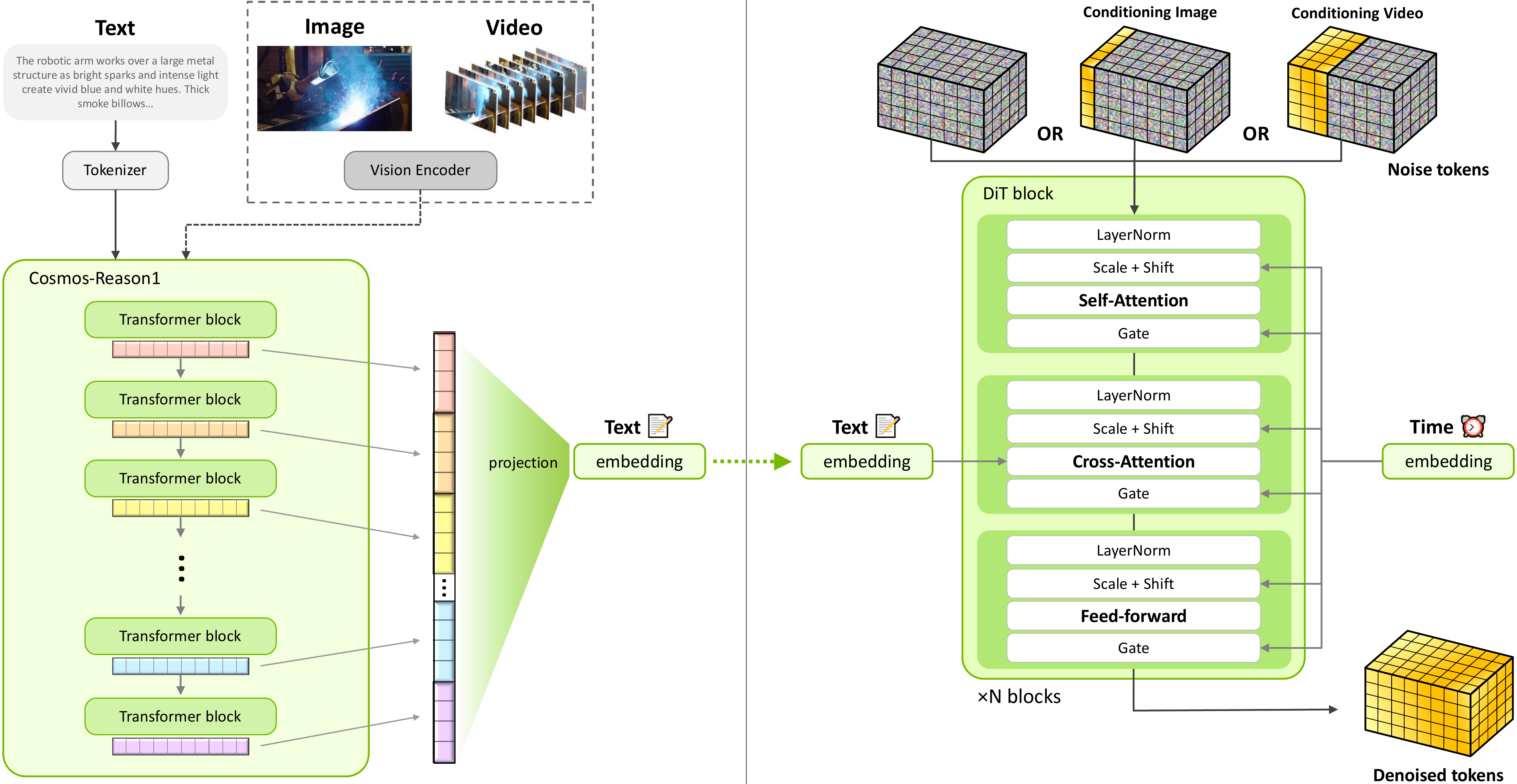}
    \caption{\textbf{Overall architecture of [Cosmos-Predict2.5]}. As shown on the right, in the latent space, the model applies repeated blocks of self-attention, cross-attention, and feed-forward MLP layers, modulated by adaptive layer normalization (scale, shift, gate) for a given time step $t$. We leverage [Cosmos-Reason1] as the text encoder (shown on the left). [Cosmos-Reason1] can also accommodate visual inputs (image and video) beyond text, which we leave for future work.}
    \label{fig:diffusion_architecture}
\end{figure}

In [Cosmos-Predict2.5], we largely reuse the denoising network $\mathbf{u}(\cdot, \theta)$ introduced in [Cosmos-Predict1]'s DiT~\citep{cosmos_v1}, which is based on a latent diffusion model. The main architectural change is the removal of the absolute positional embeddings and only keeping the relative positional embeddings. While absolute embeddings provide a fixed spatial or temporal reference, they limit the model’s ability to generalize to resolutions or sequence lengths not seen during training. By removing them, [Cosmos-Predict2.5] gains greater flexibility for handling higher-resolution content and longer video sequences during post-training. This design choice is motivated by recent progress in long-context large language models, where alternative positional encoding strategies~\citep{peng2023yarn,bloc97_ntkaware_scaled_rope_2023} have proven effective at extending context length without sacrificing performance. The overall velocity prediction network design is illustrated in~\cref{fig:diffusion_architecture}.

We adopt a different set of auxiliary models in [Cosmos-Predict2.5] compared to [Cosmos-Predict1], with improvements in both visual and textual representations. For the visual tokenizer, we use WAN2.1 VAE~\citep{wan2025}, a causal variational autoencoder that compresses video sequences with a compression rate of $4\times8\times8$ across the time, height, and width dimensions, respectively. This compression greatly reduces the computational cost while preserving essential spatiotemporal structure. On top of this representation, we apply the same $1\times2\times2$ patchification strategy to compress latent features further. We train our model to generate 93 frames, which corresponds to 24 latent frames, at a time using 16 fps videos. Each of the generated videos is about 5.8 seconds long.

For the text encoder, we leverage [Cosmos-Reason1]~\citep{azzolini2025cosmos} instead of the T5 encoder used in [Cosmos-Predict1]. Unlike standard approaches that rely on the output of a single transformer layer, we concatenate activations across multiple blocks for each token and project them into a 1024-dimensional space inspired by~\citet{wang2025comprehensivestudydecoderonlyllms}. This yields a sequence of embedding vectors that more faithfully captures both local and global linguistic context. During training, these embeddings are integrated into the denoising process via cross-attention layers, enabling textual prompts to directly guide video generation. Moreover, the vision encoder in [Cosmos-Reason1] supports additional visual conditional inputs for style control, which we leave as an exciting direction for future exploration.

\begin{table}[t]
    \setlength{\tabcolsep}{8pt} 
    \small
    \captionsetup{justification=centering}
    \caption{Configuration details of [Cosmos-Predict2.5] models.}
    \centering
    \begin{tabular}{l|cc}
        \toprule
         \textbf{Configuration} & \textbf{Cosmos-Predict2.5-2B} & \textbf{Cosmos-Predict2.5-14B}\\
        \midrule
        Number of Layers & $32$ & $36$ \\
        Model Dimension & $2{,}048$ & $5{,}120$\\
        FFN Hidden Dimension & $8{,}192$ & $20{,}480$\\
        AdaLN-LoRA Dimension & $256$ & $256$\\
        Number of Attention Heads & $16$ & $40$\\
        Head Dimension & $128$ & $128$\\ 
        MLP Activation & \multicolumn{2}{c}{GELU} \\
        Positional Embedding & \multicolumn{2}{c}{3D RoPE} \\
        \bottomrule
    \end{tabular}
    \label{tab:diffusion_model_specs}
\end{table}

Each [Cosmos-Predict2.5] model is designed to operate in three modes: Text2World, Image2World, and Video2World. In the Text2World setting, generation is guided solely by a text prompt. In Image2World, the model receives both a text prompt and a reference image, allowing it to ground the generated video in specific visual content. In Video2World, the model further extends this conditioning to video sequences, enabling temporally coherent continuation or transformation of input clips.

For both Image2World and Video2World, we employ a frame-replacement strategy, where the initial frames of the generated sequence are consistently substituted with the conditioned frames. This approach serves two purposes: (1) it provides flexibility, since the number of conditioned frames can be adjusted depending on the task, and (2) it strengthens temporal consistency by ensuring that early frames remain faithful to the conditioning input. As a result, visual cues from the input image or video propagate more smoothly across subsequent frames, leading to more coherent world generation.

\section{Training}
\label{sec::training}

We employ a progressive training strategy that balances efficiency with model quality. The process begins with multi-stage pretraining, where training difficulty is gradually increased along two axes: pixel resolution and task diversity. After pretraining, we perform supervised fine-tuning (SFT) on carefully curated, high-quality Physical AI datasets to strengthen the model’s capabilities in specialized domains, before merging them into a unified model. Finally, we further enhance generation quality by applying a reinforcement learning (RL) algorithm on top of the merged model.

\subsection{Pre-training}
\label{subsec::pretraining}

We describe the multi-stage pretraining procedure in~\cref{tab:diffusion_training_strategy}. Training begins with the Text2Image task at a resolution of 256p. This stage allows the model to learn to generate high-quality individual frames before addressing motion and temporal consistency. We then introduce the Image2World and Video2World tasks to support joint image–video training. In this setting, we randomly sample either 1 or 5 conditioning frames and require the model to generate the remaining 92 or 88 frames, respectively (for a total of 93 pixel frames, corresponding to 24 latent video frames). The DiT is conditioned by concatenating ground-truth frames with noisy frames. To specify which inputs are conditional, we apply a masking scheme: each input token is formed by concatenating the original token with a mask token, where the mask serves as a binary flag indicating whether the inputs are conditional inputs. The denoising loss is applied only to the designated frames, ensuring gradients propagate correctly. After this, we progressively increase the resolution from 256p to 480p and then to 720p, advancing to the next stage once the model converges and visual quality plateaus. Finally, we add the Text2World task, where zero conditioning frames are provided. At this stage, we sample 0, 1, or 2 condition frames with probabilities of 0.5, 0.25, and 0.25, respectively.

\begin{table}[t]
    \centering
    \setlength{\tabcolsep}{6pt} 
    \small
    \captionsetup{justification=centering}
    \begin{threeparttable}
        \caption{Stages of progressive pretraining and their specifications.}
        \label{tab:diffusion_training_strategy}
        \begin{tabular}{l|cl}
            \toprule
            \textbf{Task} & \textbf{Resolution} & \textbf{Number of Frames} \\
            \midrule
            Text2Image & 256p (320$\times$192) & 1 \\
            Text2Image | Video2World & 256p (320$\times$192) & 1 | 93  \\
            Text2Image | Video2World & 480p (832$\times$480) & 1 | 93  \\
            Text2Image | Video2World & 720p (1280$\times$704) & 1 | 93  \\
            Text2Image | Video2World | Text2World & 720p (1280$\times$704) & 1 | 93 | 93 \\
            \bottomrule
        \end{tabular}
    \end{threeparttable}
\end{table}

We draw training timesteps from a logit-normal distribution, which, as shown in \citep{esser2024scaling}, places higher probability mass in the middle range of $[0,1]$. Consistent with their approach, we apply a progressive timestep shift that grows with training resolution—starting with a shift of $\beta=1$ at 256p and gradually increasing to $\beta=5$ at 720p. Despite these adjustments, we observed artifacts in the generated videos, specifically abrupt and unnatural transitions between frames. We hypothesized that this instability arose because the model received too few training examples in the high-noise region, leaving it underexposed to such conditions. To address this imbalance, we modified the scheduler so that 5\% of training samples are drawn explicitly from the highest 2\% of the noise distribution. This targeted sampling strategy significantly reduced the transition artifacts and improved temporal consistency across generated sequences.

We train using the AdamW optimizer with $\beta_1=0.9$ and $\beta_2=0.999$. We set the learning rate to $3\times10^{-5}$ for [Cosmos-Predict2.5-2B] and $1.3\times10^{-5}$ for [Cosmos-Predict2.5-14B]. The weight decay is set as 0.001. To stabilize optimization, we apply a linearly decaying learning rate scheduler that includes an initial warmup phase with 2000 iterations.

\subsection{Post-training}
\label{subsec::posttraining}

\subsubsection{Supervised Fine-tuning}
We further conduct supervised fine-tuning (SFT) on a collection of curated, high-quality Physical AI datasets. To construct these datasets, we train a multi-head classifier on InternVideo2 embeddings~\citep{wang2024internvideo2}, which enables us to categorize samples into five domains: object permanence, high motion, complex scenes, driving, and robotic manipulation. The distribution of samples across these domains is summarized in \cref{tab:post-train-stat}.

\begin{table}[h]
\centering
\captionsetup{justification=centering}
\caption{Video statistics across different post-train domains.}
\small
\label{tab:post-train-stat}
\begin{NiceTabular}{l|cccccc}[baseline=2,cell-space-limits=1pt]
\toprule
\RowStyle{\bfseries}
 Domain & Object Permanence & High Motion & Complex Scenes & Driving & Robotic Manipulation & 4K \\
\midrule
\#Videos & 10.4M & 1.0M & 1.6M & 3.1M & 730K & 388K \\
\bottomrule
\end{NiceTabular}
\end{table}

\begin{figure}[t]
    \centering
    \captionsetup{justification=centering}
    \caption{Domain-specific SFT training improves the performance of the pretrained model on each domain.}
    \includegraphics[width=0.8\textwidth]{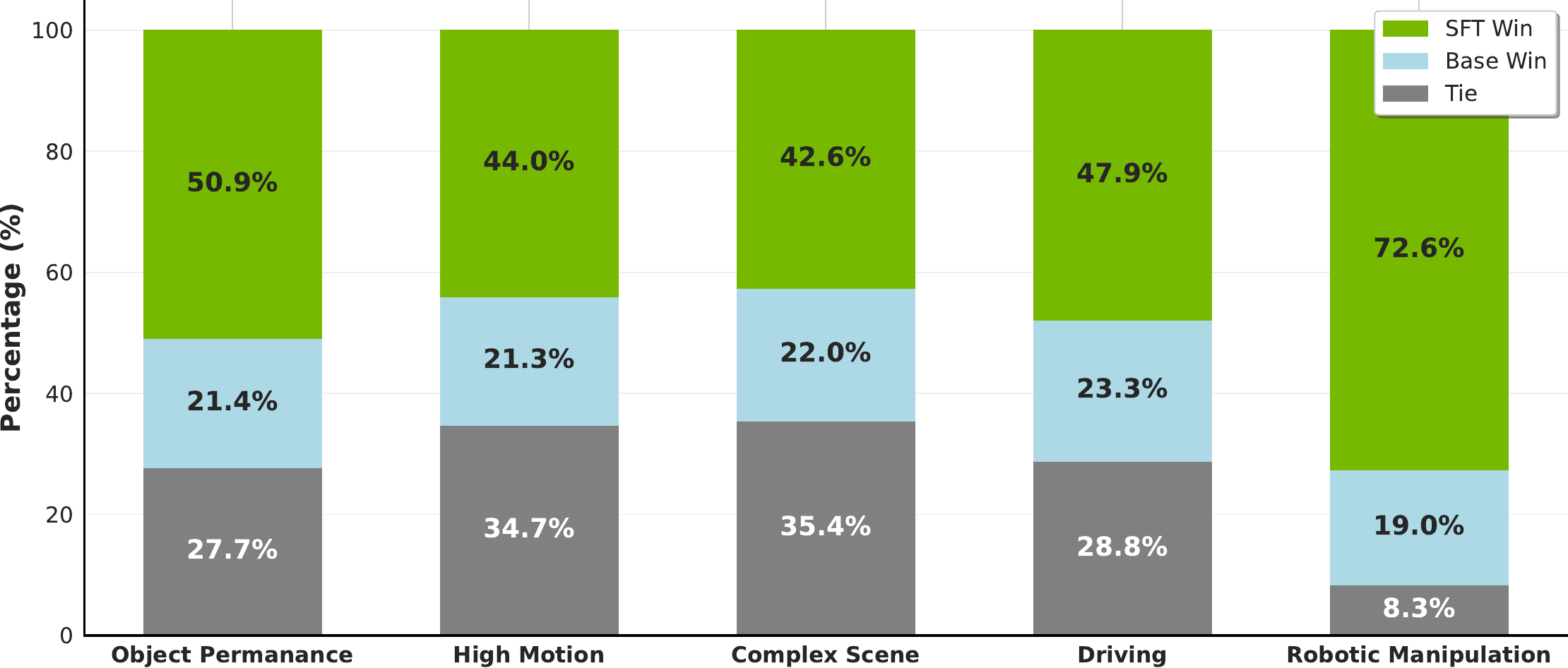}
    \label{fig:domain_win_rates}
\end{figure}

\begin{figure}[tbh!]
    \centering
        \captionsetup{justification=centering}
        \caption{\textbf{Merged model gets the best of all the worlds while maintaining performance on the general domain.} Win rate for pretrained is average across three comparisons.}
        \includegraphics[width=.5\textwidth]{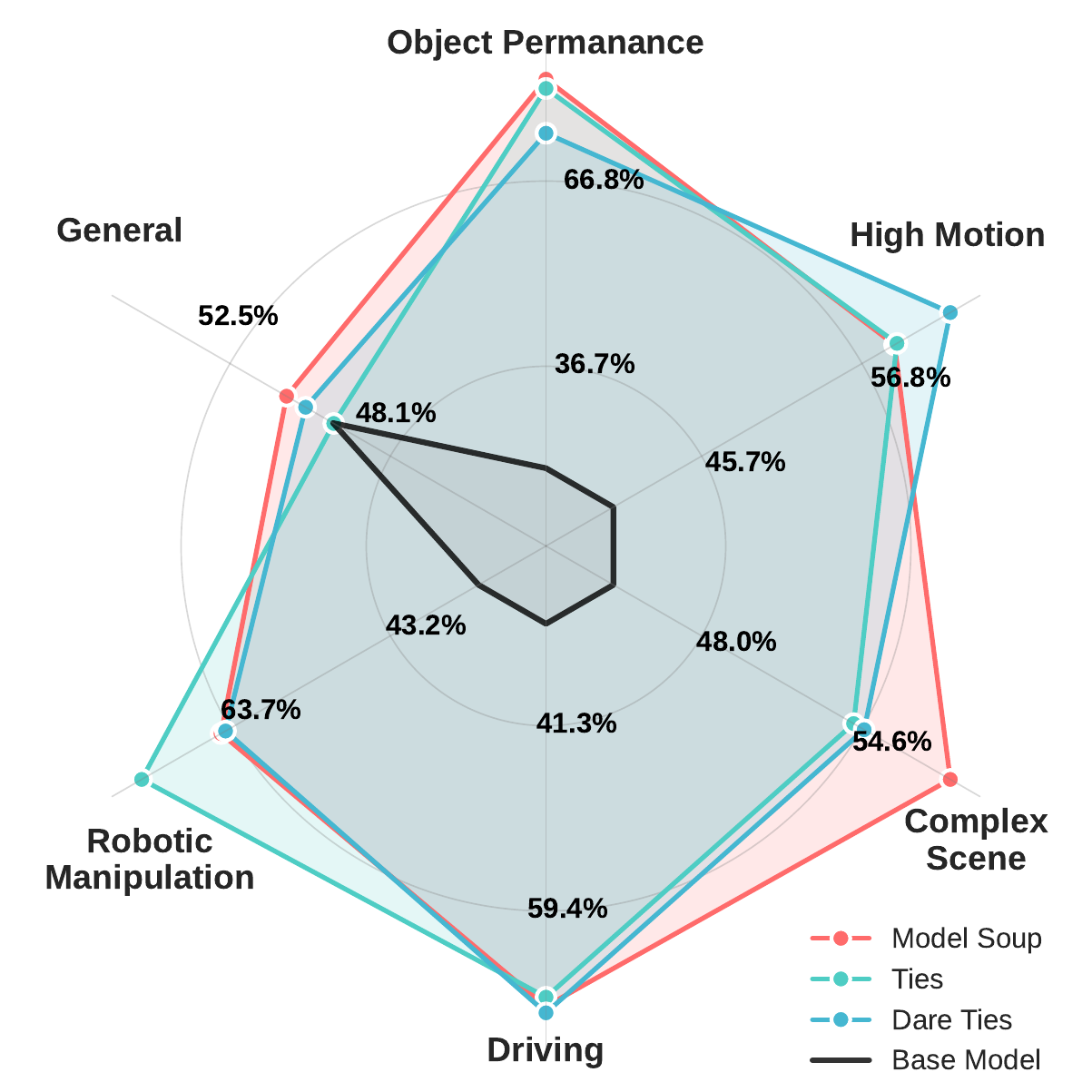}
        \label{fig:posttrain_vs_pretrain}
\end{figure}

\begin{figure}
        \centering
        \caption{Human voting shows that RL can effectively improve the quality of the generated videos.}
        \includegraphics[width=.5\textwidth]{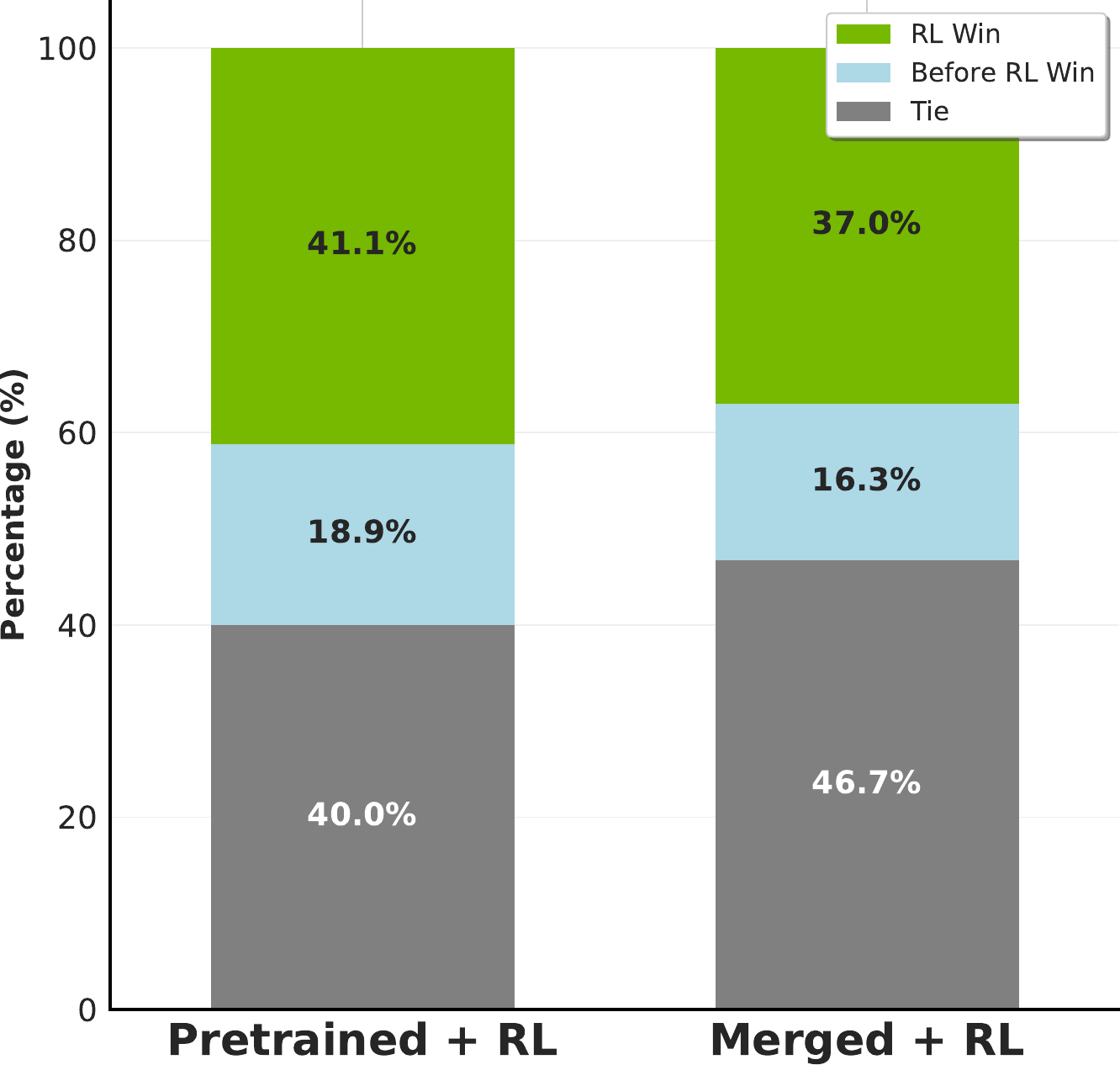}
        \label{fig:win_rates_rl_is_effective}
\end{figure}

We fine-tune a separate model for each domain rather than training a single model jointly across all domains. This domain-specific strategy enables us to fully leverage the available data without the need to balance mixture ratios across a combined dataset. In practice, we find that domain-specific SFT substantially improves performance on specialized domains, while causing only minimal degradation on general-domain tasks. Moreover, these slight degradations can be further mitigated through the model-merging approach described later. Each specialized model is trained for 30k iterations with a batch size of 256, using the same hyperparameter settings as the final stage of pretraining.

To evaluate these models, we construct a domain-specific test set for each category and conduct human preference studies. As shown in \Cref{fig:domain_win_rates}, every SFT model achieves a significantly higher win rate than the pretrained baseline on its target domain.

In addition, we apply a cooldown stage to the pretrained model using a curated set of high-quality 4K videos, where the learning rate is linearly decayed to zero. This step enhances fine-grained visual detail and produces smoother motion.

To unify the strengths of both the domain-specific SFT models and the cooldown model, we adopt model merging \citep{yang2024model}. We experiment with four approaches: model soup \citep{wortsman2022model}, TIES \citep{yadav2023ties}, DARE-Linear \citep{yu2024language}, and DARE-TIES \citep{yu2024language}. For each method, we run hyperparameter sweeps and generate more than 20 merged models. From these candidates, we select the best-performing model based on quality assessments over a small, hand-picked set of challenging examples. We then validate the selected models on a larger evaluation set using human preference voting to ensure robust performance across both domain-specific and general tasks.

Interestingly, we find that simple grid search over hyperparameters consistently outperforms heuristic selection based on individual fine-tuned models’ win rates. As illustrated in \Cref{fig:posttrain_vs_pretrain}, all methods achieve comparable performance with the exception of DARE-Linear. Given its effectiveness and simplicity, we select the model soup variant as our final post-trained model.

\subsubsection{Reinforcement Learning}

Reinforcement learning has been widely adopted in post-training to align model outputs with human preferences, either represented by human feedback~\citep{ouyang2022training} or by reward models~\citep{schulman2017proximal,guo2025deepseek}. For flow-based world generation, we can similarly view conditions as states and the entire denoising trajectories as actions, and leverage the reinforcement learning framework to post-train the model. 
We briefly introduce our RL mechanism below, with comprehensive techniques presented in \cite{ye2025data}.

We adopt VideoAlign~\citep{liu2025improving}, a VLM-based reward model that evaluates text alignment, motion quality, and visual quality to post-train [Cosmos-Predict2.5-2B] (both the pre-trained and merged model). We generate eight outputs with 20 diffusion steps for each input condition and then compute the advantage of each output by normalizing the reward within its rollout group, following GRPO~\citep{guo2025deepseek}. Due to the GPU memory constraint, the probability of each trajectory is computed by decomposing it into the sum of conditional probabilities at each step, and in practice, we compute the gradient of every two conditional probabilities based on the advantages and accumulate the gradient of the probability over the entire trajectory (ten steps in total) for one parameter update. The model is trained for 256 steps with a batch size of 32. As discussed in \cite{ye2025data}, we use diffusion loss on the fine-tuning dataset for regularization, which effectively alleviates the reward hacking phenomenon. We release the EMA weight after post-training on the merged model as our final [Cosmos-Predict2.5] post-train checkpoint.

We present the reward scores on PAI-Bench before and after RL post-training in~\Cref{tab:rl_2b}. Both in Text2World and Image2World scenarios, and both for the pre-trained model and the merged model of the various SFT models, the reward increases by a large margin. We additionally conduct human voting between the videos generated by models before and after reinforcement learning, and the results are presented in \Cref{fig:win_rates_rl_is_effective}. In all cases, videos generated by the RL models are preferred on average. In summary, reinforcement learning is proven effective in improving model quality, both in terms of the reward scores and of the human voting results.

\begin{table}[ht]
\centering
\small
\caption{Rewards of [Cosmos-Predict2.5-2B], before and after reinforcement learning on VideoAlign, for Text2World and Image2World settings.}
\label{tab:rl_2b}
\begin{tabular}{lcccccccc}
\toprule

& \multicolumn{4}{c}{\textbf{Text2World}} & \multicolumn{4}{c}{\textbf{Image2World}} \\
\cmidrule(lr){2-5} \cmidrule(lr){6-9}


\multirow{-3}{*}{\diagbox{Model}{Rewards}}  & {\shortstack{Text \\ Alignment}} & {\shortstack{Motion \\ Quality}} & {\shortstack{Visual \\ Quality}} & \shortstack{ \\ Sum} & {\shortstack{Text \\ Alignment}} & {\shortstack{Motion \\ Quality}} & {\shortstack{Visual \\ Quality}} & \shortstack{ \\ Sum} \\
\midrule

Predict2.5-2B [pre-train] & {1.55} & {-0.43} & {-0.05} & 1.08 & {1.48} & {-0.76} & {-0.49} & 0.23 \\
+ RL & {1.69} & {-0.19} & {0.19} & 1.69 & {1.57} & {-0.70} & {-0.45} & 0.42 \\
\midrule
Predict2.5-2B [merged] & {1.69} & {-0.46} & {-0.01} & 1.23 & {1.57} & {-0.82} & {-0.52} & 0.24 \\
+ RL & {1.75} & {-0.18} & {0.18} & 1.74 & {1.57} & {-0.68} & {-0.44} & 0.45 \\
\bottomrule
\end{tabular}
\end{table}

\subsubsection{Timestep Distillation}
The inference of diffusion-based world generation can be substantially accelerated through timestep distillation. We adopt a hybrid forward-reverse joint distillation framework, rCM~\citep{zheng2025large}, which integrates continuous-time consistency distillation with distribution matching distillation. To support this framework, we build dedicated infrastructure, including fused flash attention with Jacobian–vector product (JVP) support, as well as adaptations for FSDP2 and context parallelism. The distilled models are capable of producing high-fidelity samples in only 4 steps with quantitative results similar to the teacher model (\Cref{tab:PaiBench_T2W_distill}, \Cref{tab:PaiBench_I2W_distill}).

\begin{table*}[t]
\centering
\small
\caption{Distillation results on PAI-Bench-Predict-Text2World Benchmark.}
\label{tab:PaiBench_T2W_distill}
\begin{tabular}{ccccccc}
\toprule
\textbf{Model} &
\textbf{Domain Score} &
\textbf{Quality Score} &
\textbf{Overall Score} \\
\midrule
Cosmos-Predict2.5-2B [teacher]  & 0.804 & 0.732 & 0.768 \\
Cosmos-Predict2.5-2B [distilled]  & 0.797 & 0.731 & 0.764 \\
\bottomrule
\end{tabular}
\end{table*}

\begin{table*}[t]
\centering
\small
\caption{Distillation results on PAI-Bench-Predict-Image2World benchmark.}
\label{tab:PaiBench_I2W_distill}
\begin{tabular}{ccccccc}
\toprule
\textbf{Model} &
\textbf{Domain Score} &
\textbf{Quality Score} &
\textbf{Overall Score} \\
\midrule
Cosmos-Predict2.5-2B [teacher]  & 0.840 & 0.779 & 0.810 \\
Cosmos-Predict2.5-2B [distilled]  & 0.842 & 0.790 & 0.816 \\
\bottomrule
\end{tabular}
\end{table*}

\subsection{Infrastructure}
\label{subsec::training_infra}

{\bf Hybrid Sharded Mode of FSDP2.} We use FSDP2 as our primary distributed training framework because of its ability to shard model weights, gradients, and optimizer states while efficiently overlapping communication with computation. Unlike FSDP1, which relies on a bucket-based sharding strategy, FSDP2 performs per-parameter sharding. This finer-grained design enables more efficient memory management by releasing memory promptly, thereby reducing overhead and improving utilization—an especially critical factor in video model training, where a single sequence can produce hundreds of thousands of tokens. These capabilities make FSDP2 a more scalable and flexible solution for large-scale distributed training. In addition, we incorporate several FSDP2-related optimizations from TorchTitan~\citep{liang2025torchtitan}, including asynchronous distributed checkpointing and meta-device initialization, to further enhance training efficiency.

{\bf Flexible Context Parallelism.} When training on high-resolution or long-duration videos, the input sequence length can easily grow to hundreds of thousands of tokens. To control per-GPU memory usage and distribute the computation of a single sample across multiple devices, we employ context parallelism. For added flexibility, we adopt the Ulysses-style parallelism approach~\citep{rasley2020deepspeed}. Compared with the ring-attention strategy used in the diffusion world model of [Cosmos-Predict1], this method is both simpler and more communication-efficient, leveraging intra-node all-to-all collectives on NVIDIA GPUs. It also offers greater adaptability: for example, it better supports video post-training and diffusion distillation workloads that require advanced mechanisms such as NATTEN sparse attention~\citep{hassani2025generalized} and fused flash attention with Jacobian–vector product (JVP) support~\citep{lu2024simplifying}. Achieving these capabilities with ring attention would be far more difficult while keeping computation balanced. To enable joint training across images and videos, we dynamically disable context parallelism during image iterations and re-enable it for video batches.

{\bf Selective Activation Checkpointing.} To balance memory usage with computational efficiency, we apply torch Selective Activation Checkpointing (SAC) using a fine-grained policy. Lightweight operators—such as element-wise functions and normalization layers—are prioritized for recomputation, since they introduce minimal overhead while yielding significant memory savings. For large-scale video training workloads, we further extend checkpointing to portions of linear layers once all memory-intensive but computation-light operators have been covered, enabling additional reductions in memory consumption.

{\bf Elastic Reward Service.} To handle a large amount of input and different reward models in the RL post-training, we rely on an efficient and flexible external service. The service supports VideoAlign and other reward functions, and can be dynamically scaled up or down according to the input traffic. Decoded latent is used to send the video for evaluation, enabling data compression during transfer. The service is pipelined in a producer-consumer fashion: a decode stage decodes the video from the received latent, while several reward models compute different rewards simultaneously in the inference stage. The decode and inference stages process different videos in a pipeline to fully utilize compute capacity. Each stage runs in a separate process to satisfy different environment requirements and support scalability. Data sharing between stages is achieved via CUDA inter-process communication (IPC) in a zero-copy manner, further enhancing efficiency. The reward calculation is handled in an asynchronous way. A task UUID is returned immediately after the video with its desired reward types is enqueued. A Redis server stores the computed rewards, which can be retrieved later using the UUID. Each task also supports batch processing of multiple videos. Between the interval of enqueue and result fetching, other actions can proceed asynchronously to maximize the system utilization.

\begin{table}[htb!]
\centering
\caption{Training efficiency with 4096 NVIDIA H100 GPUs where the video resolution is 720p and number of frames is 93.}
\label{tbl:mfu}
\begin{tabular}{lcccccc}
\toprule

Model & \multicolumn{2}{c}{Context Parallelism Size} & \multicolumn{2}{c}{MFU} \\
\midrule
Cosmos-Predict2.5-2B & \multicolumn{2}{c}{2} & \multicolumn{2}{c}{36.49\%} \\
Cosmos-Predict2.5-14B & \multicolumn{2}{c}{8} & \multicolumn{2}{c}{33.08\%} \\
\bottomrule
\end{tabular}
\end{table}

\cref{tbl:mfu} shows the Model Flops Utilization (MFU) of our video model training infrastructure. For [Cosmos-Predict2.5-2B], the MFU is 36.49\%. For [Cosmos-Predict2.5-14B], the MFU drops to 33.08\%. The drop is due to large context parallelism, which introduces more communication cost.

\section{Results}
\label{sec::results}

{\bf Benchmarking.} 
We report the performance of [Cosmos-Predict2.5-2B] models on PAI-Bench~\citep{PAI-Bench}, a recently proposed benchmark designed to assess physical AI generation and understanding capabilities.

We evaluate on PAI-Bench's predict task and report two main scores: the \textit{Domain Score}, which measures performance on domain-specific physical AI tasks, and the \textit{Quality Score}, which reflects the quality of generated videos. The \textit{Quality Score} is derived from eight text-to-video and image-to-video metrics adapted from VBench. In contrast, the \textit{Domain Score} is obtained through VQA-based evaluation across seven domains: av, common, human, industry, misc, physics, and robotics. The final \textit{PAI-Bench Overall Score} is computed as the average of the Quality and Domain scores.

The PAI-Bench T2W and I2W quantitative results are shown in \Cref{tab:PaiBench_T2W} and \Cref{tab:PaiBench_I2W}, respectively. Both the [Cosmos-Predict2.5-2B] and Cosmos-Predict2.5-14B] post-trained models perform similarly to the larger Wan2.2 27B-A14B model in T2W, and are the best-performing models in I2W.

\begin{table*}[t]
\centering
\small
\caption{Results on PAI-Bench-Predict-Text2World Benchmark.}
\label{tab:PaiBench_T2W}
\begin{tabular}{ccccccc}
\toprule
\textbf{Model} &
\textbf{Domain Score} &
\textbf{Quality Score} &
\textbf{Overall Score} \\
\midrule
Cosmos-Predict2.5-2B [pre-train] & 0.782 & 0.720 & 0.751 \\
Cosmos-Predict2.5-2B [post-train]  & 0.804 & \textbf{0.732} & 0.768 \\
\midrule
Cosmos-Predict2.5-14B [pre-train] & 0.791 & 0.722 & 0.757 \\
Cosmos-Predict2.5-14B [post-train] & 0.803 & \textbf{0.732} & 0.768 \\
\midrule
Wan2.1-1.3B & 0.786 & 0.726 & 0.756 \\
Wan2.1-14B & 0.794 & 0.727 & 0.761 \\
\midrule
Wan2.2-5B & 0.797 & 0.730 & 0.764 \\
Wan2.2-27B-A14B & \textbf{0.810} & 0.728 & \textbf{0.769} \\
\bottomrule
\end{tabular}
\end{table*}

\begin{table*}[t]
\centering
\small
\caption{Results on PAI-Bench-Predict-Image2World benchmark.}
\label{tab:PaiBench_I2W}
\begin{tabular}{ccccccc}
\toprule
\textbf{Model} &
\textbf{Domain Score} &
\textbf{Quality Score} &
\textbf{Overall Score} \\
\midrule
Cosmos-Predict2.5-2B [pre-train] & 0.824 & 0.775 & 0.799 \\
Cosmos-Predict2.5-2B [post-train]  & 0.840 & 0.779 & \textbf{0.810} \\
\midrule
Cosmos-Predict2.5-14B [pre-train] & 0.835 & 0.777 & 0.806 \\
Cosmos-Predict2.5-14B [post-train]  & 0.838 & \textbf{0.781} & \textbf{0.810} \\
\midrule
Wan2.1-14B & 0.827 & 0.768 & 0.797 \\
\midrule
Wan2.2-5B & 0.834 & 0.774 & 0.804 \\
Wan2.2-27B-A14B & \textbf{0.841} & 0.772 & 0.806 \\
\bottomrule
\end{tabular}
\end{table*}

\begin{figure}[tbh!]
    \centering
    \caption{Despite being of smaller size, post-trained [Cosmos-Predict2.5-2B] is more preferred by humans over Wan2.2 5B and is on par with Wan2.1 14B across a diverse set of prompts.}
    \includegraphics[width=0.5\textwidth]{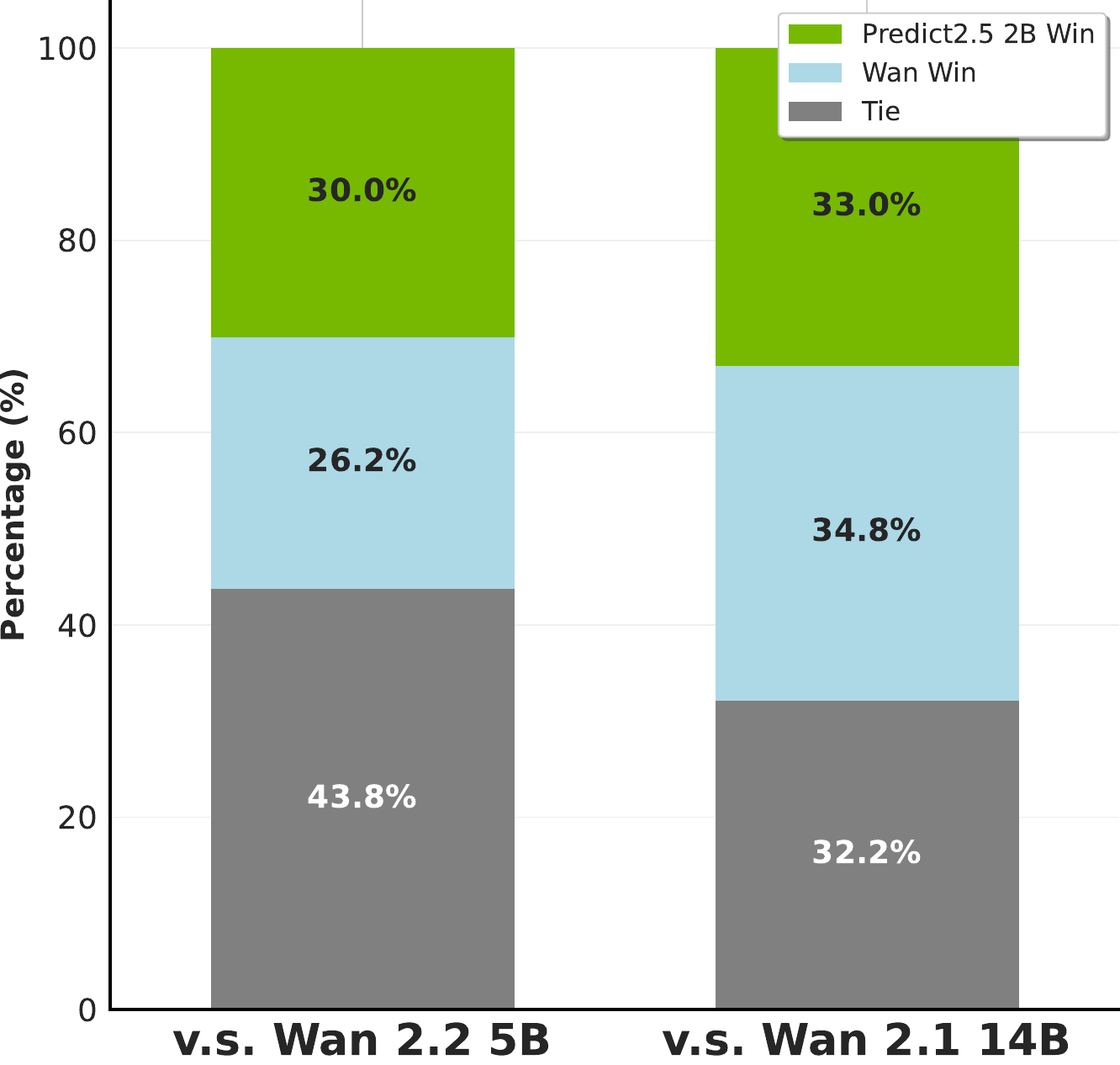}
    \label{fig:win_rates_vs_wan2.2}
\end{figure}

{\bf Human Evaluation.} Alongside automated metrics, we include human evaluation to capture aspects of video quality that are difficult to quantify and that better reflect human preference. Annotators compare pairs of generated videos, assessing criteria such as realism, visual quality, temporal consistency, and alignment with conditioning inputs. In \Cref{fig:win_rates_vs_wan2.2}, results are summarized using win ratios, defined as the proportion of comparisons in which a model’s output is preferred over a baseline. Despite being 60.0\% and 85.7\% smaller compared with Wan 2.2 5B and Wan 2.1 14B, human votings on PAI-Bench I2W and T2W settings show that [Cosmos-Predict2.5-2B] is more preferred over Wan2.2 5B (30.0\% \vs 26.2\%) and comparable with Wan 2.1 14B (33.0\% \vs 34.8\%).

\begin{figure}[tbh!]
    \centering
    \caption{Across a diverse set of prompts, post-trained [Cosmos-Predict2.5-14B] is preferred more often than Wan 2.1 14B, and achieves on par performance to Wan 2.2 27B-A14B, despite having only half the parameter count.}
    \includegraphics[width=0.5\textwidth]{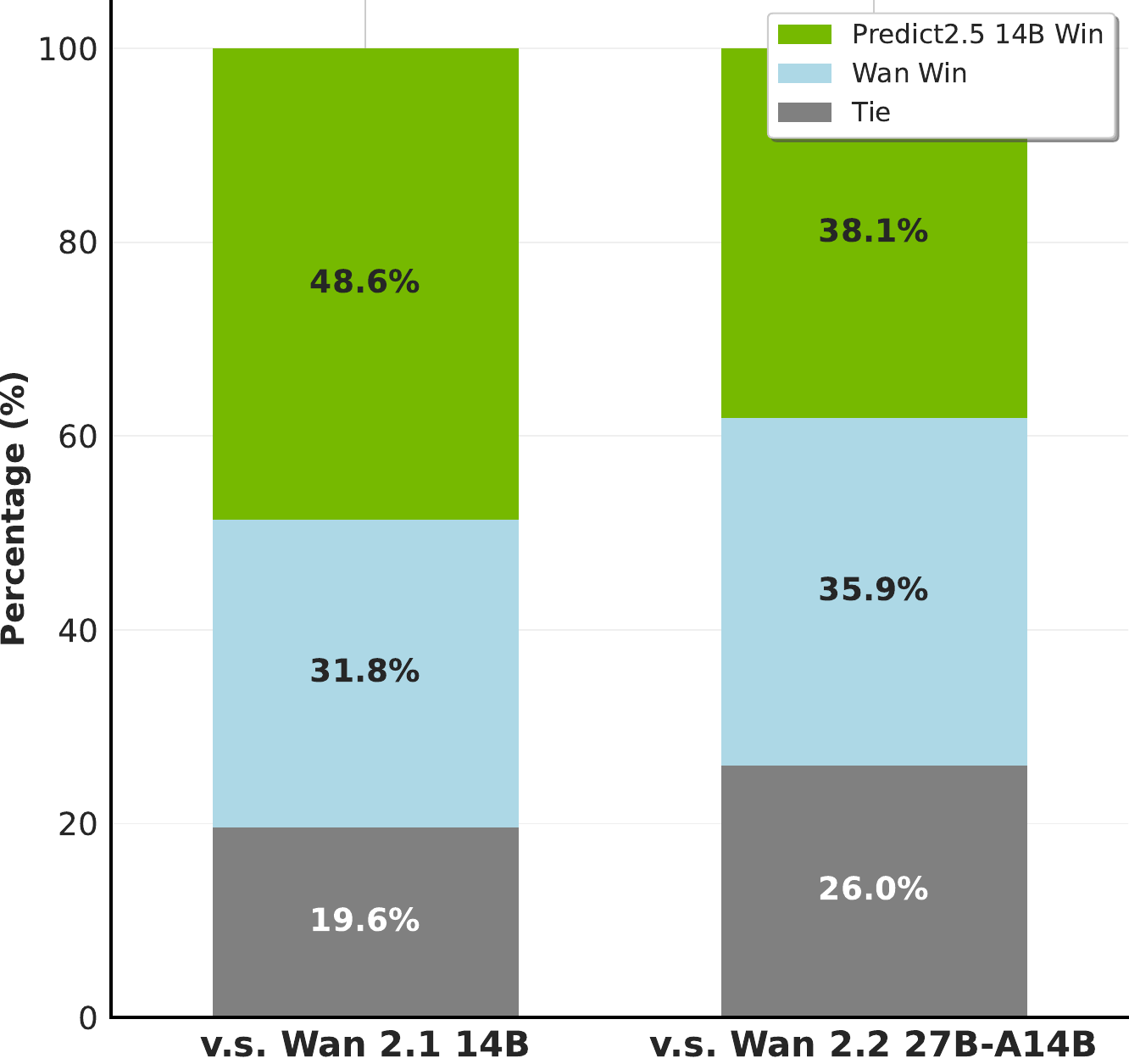}
    \label{fig:win_rates_vs_wan2.2_14B}
\end{figure}

Human evaluation results in \Cref{fig:win_rates_vs_wan2.2_14B} show that our post-trained 14B is preferred more often than Wan 2.1 14B (48.6\% \vs 31.8\%), and achieves on par performance against Wan 2.2 27B-A14B (38.1\% \vs 35.9\%), despite having only half of the parameter counts. In addition, we find that the benefit of our 14B model over our 2B model is more evident in the human evaluation results. When comparing against the same Wan 2.1 14B model, human preference increases significantly from 33.0\% to 48.6\% when the model size increases from 2B to 14B.

{\bf Qualitative Examples.} Evaluation of generative video models requires both quantitative and qualitative perspectives. Automated benchmarks and human evaluation yield measurable results, but qualitative inspection reveals model behaviors that are difficult to capture numerically. We present high-quality sample videos generated by [Cosmos-Predict2.5-2B], focusing on physical AI. These examples complement benchmark results by illustrating the model’s ability to generate realistic, high-quality, and physically coherent world simulations.

\begin{figure*}[th!]
    \centering
    \setlength{\tabcolsep}{2pt}
    \renewcommand{\arraystretch}{0.9} 
    \begin{tabular}{cccccc} 
         \footnotesize{Input frame} & & \multicolumn{4}{c}{\footnotesize Predicted frames} \\
         \includegraphics[width=0.19\textwidth]{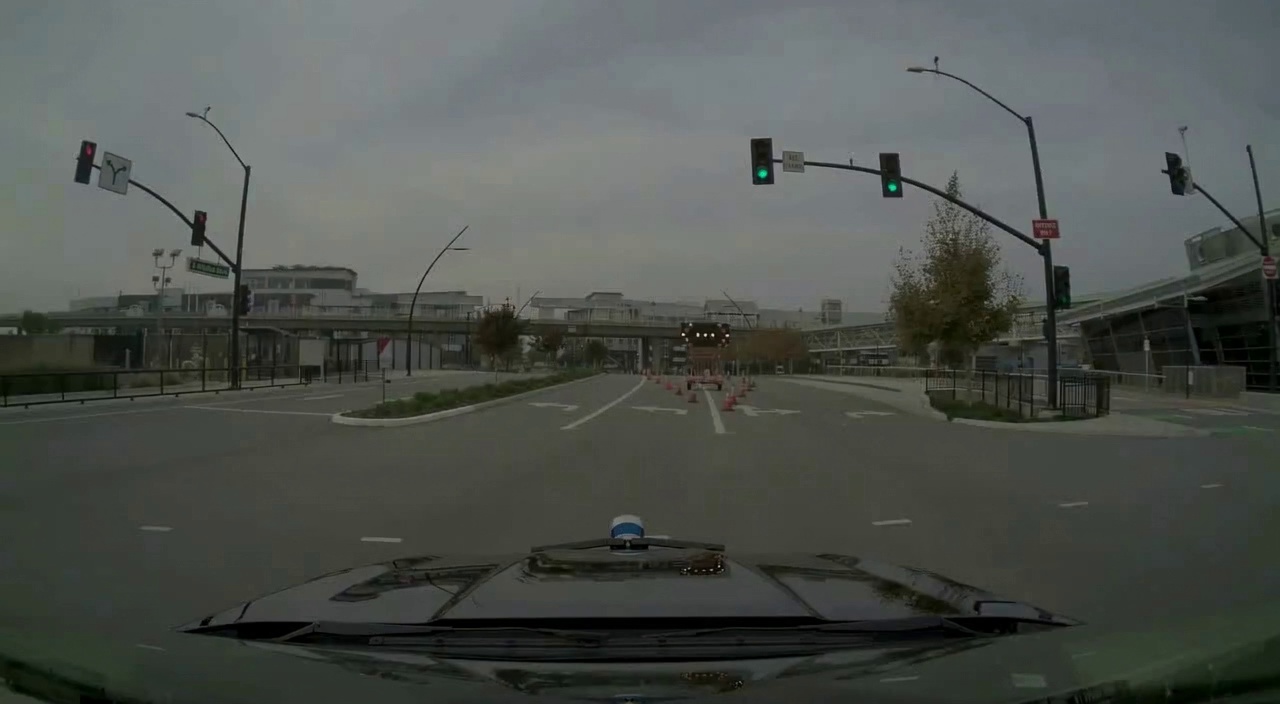} & &
         \includegraphics[width=0.19\textwidth]{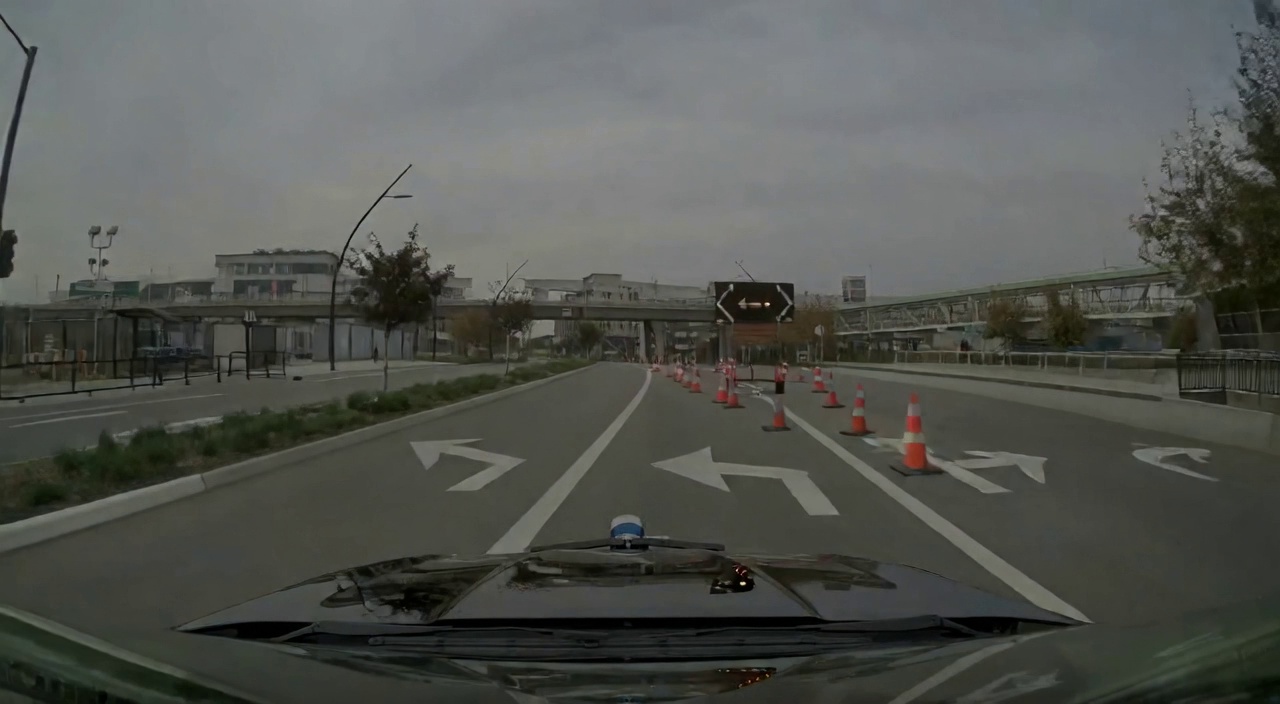} &
         \includegraphics[width=0.19\textwidth]{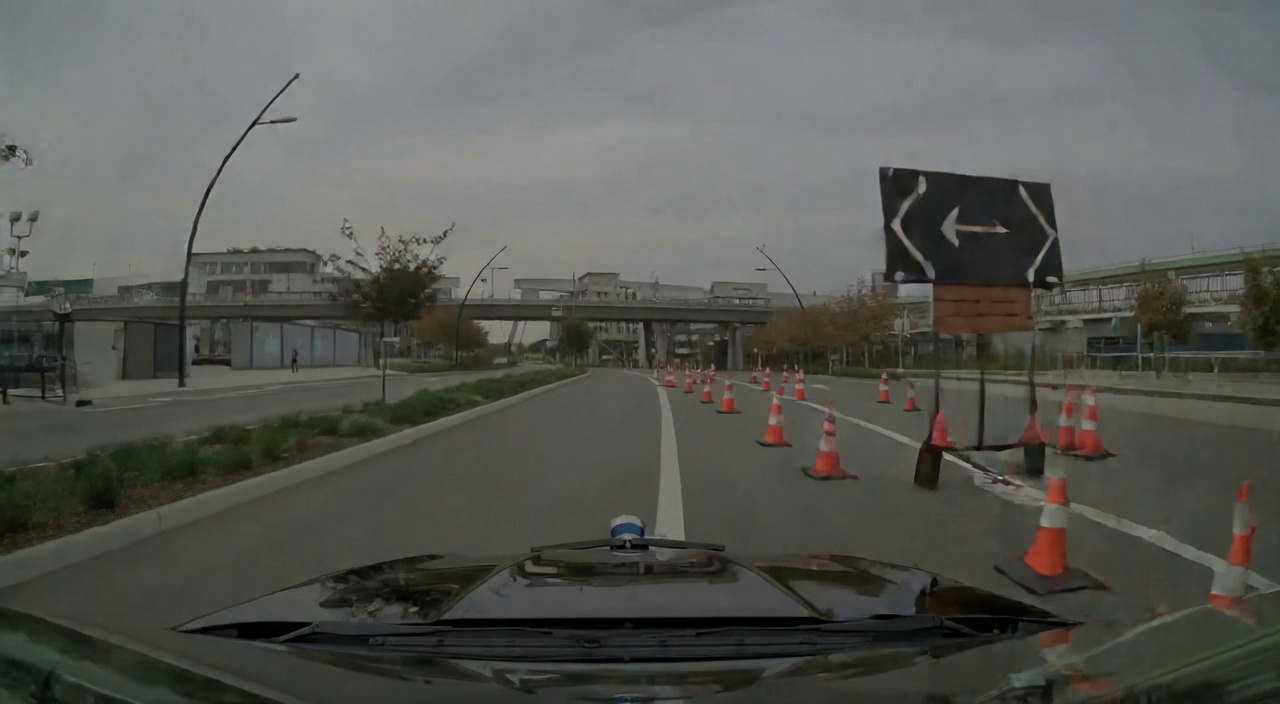} &
         \includegraphics[width=0.19\textwidth]{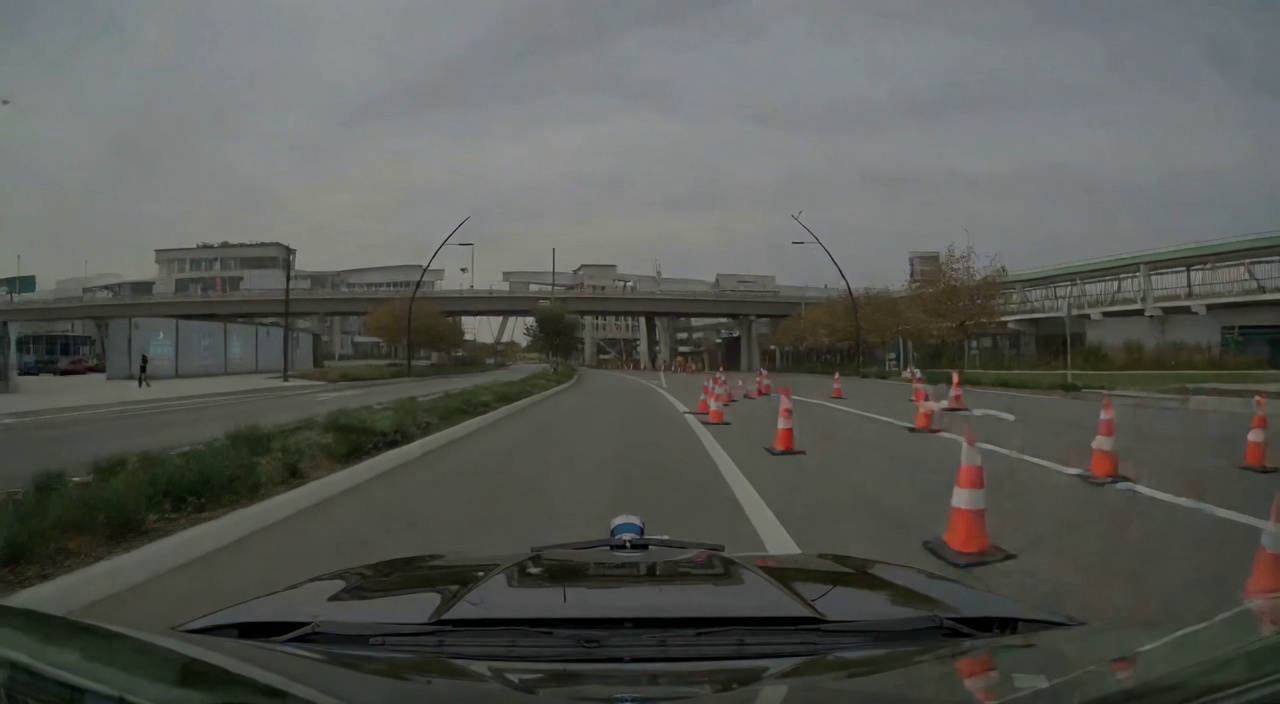} &
         \includegraphics[width=0.19\textwidth]{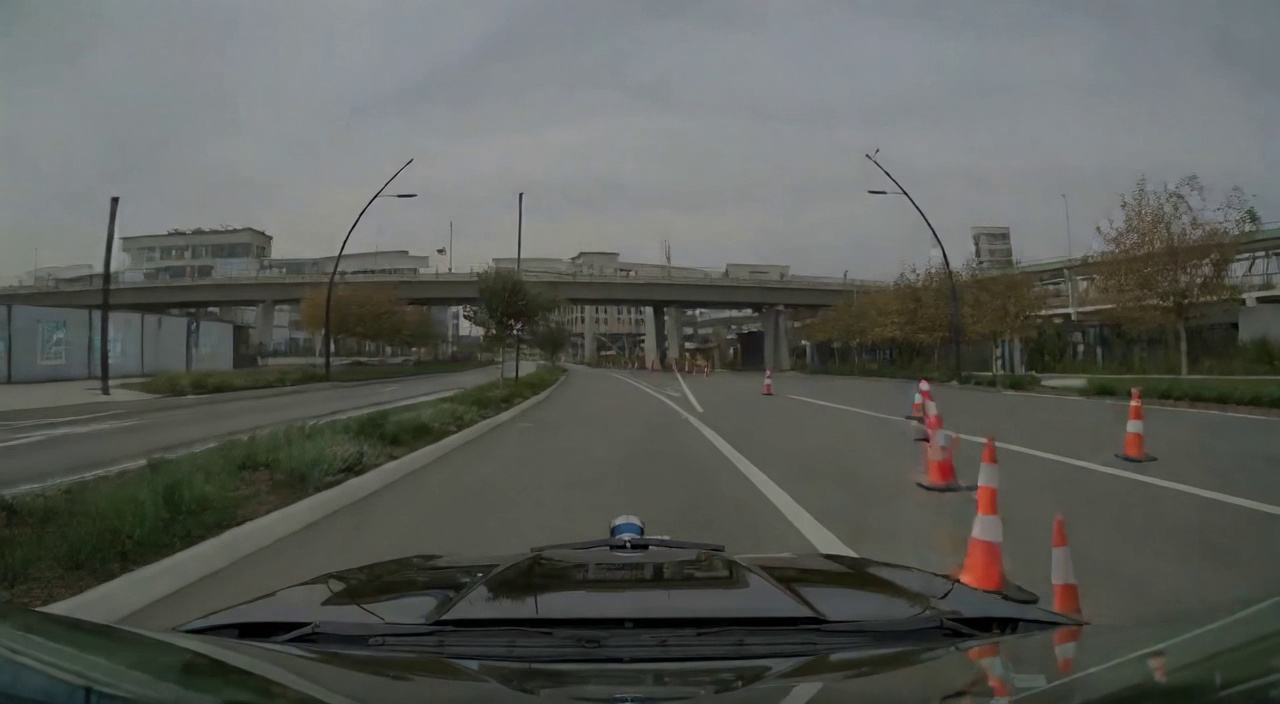} \\
         \includegraphics[width=0.19\textwidth]{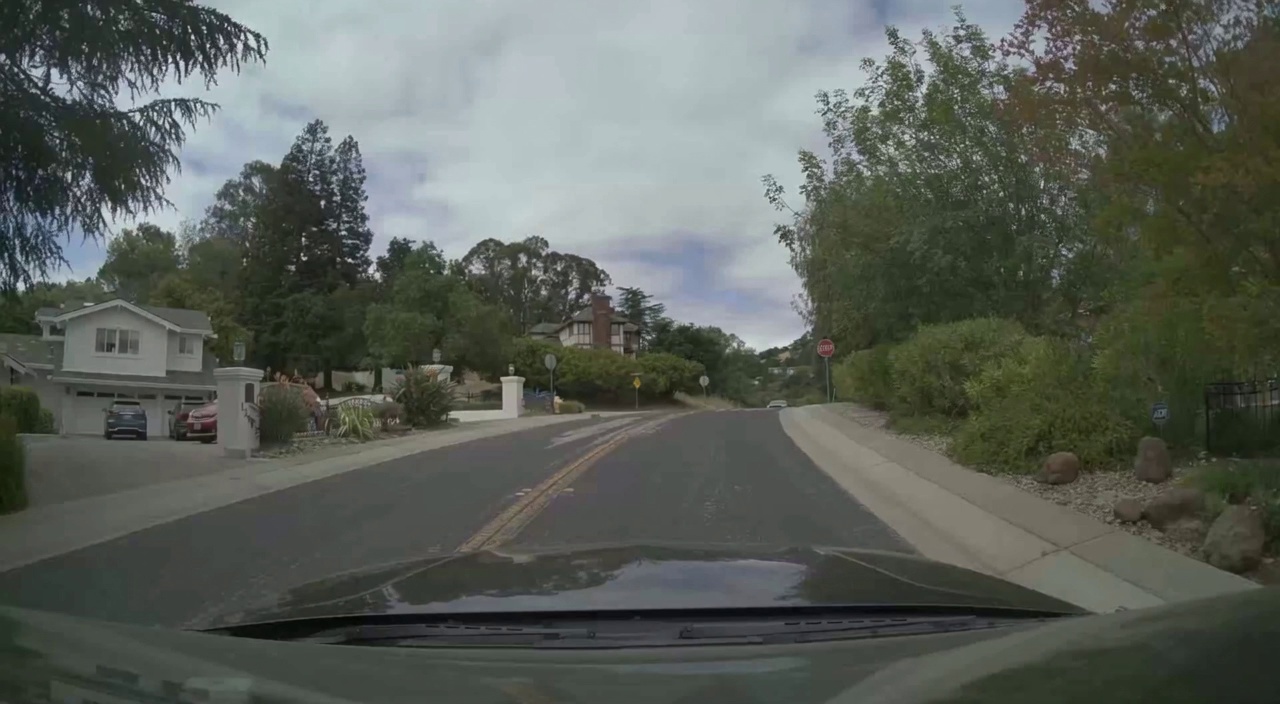} & &
         \includegraphics[width=0.19\textwidth]{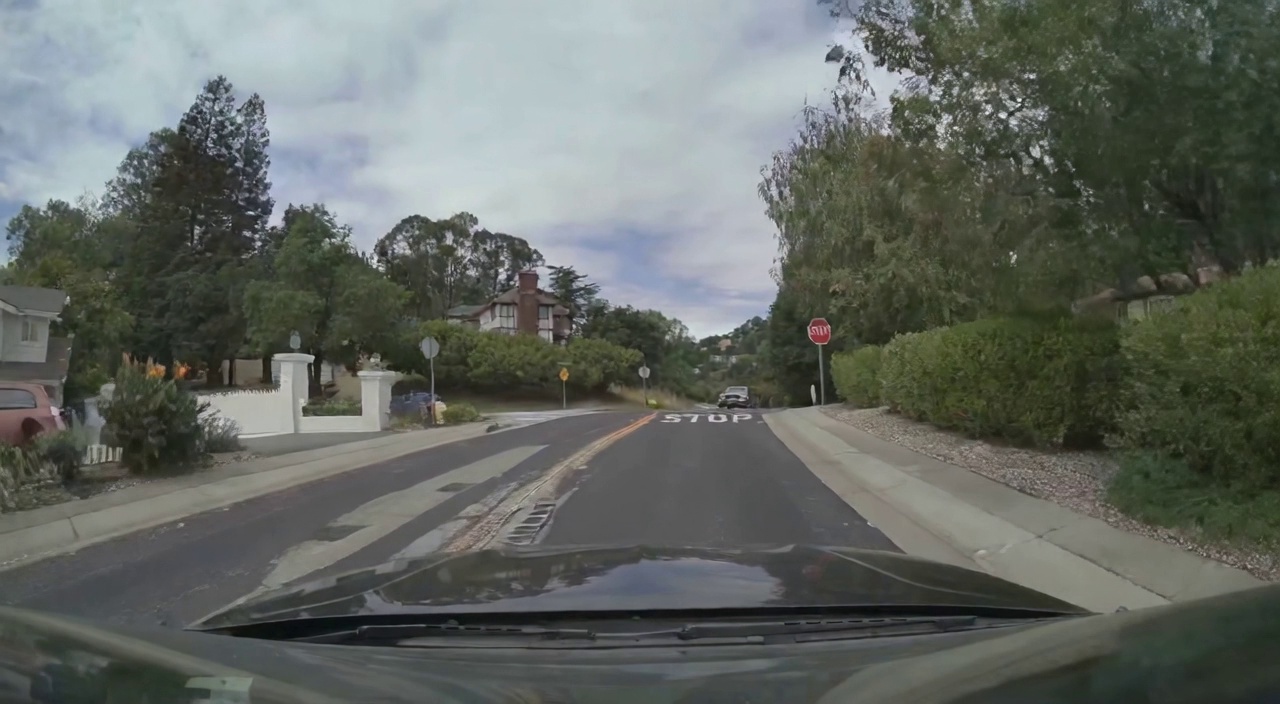} &
         \includegraphics[width=0.19\textwidth]{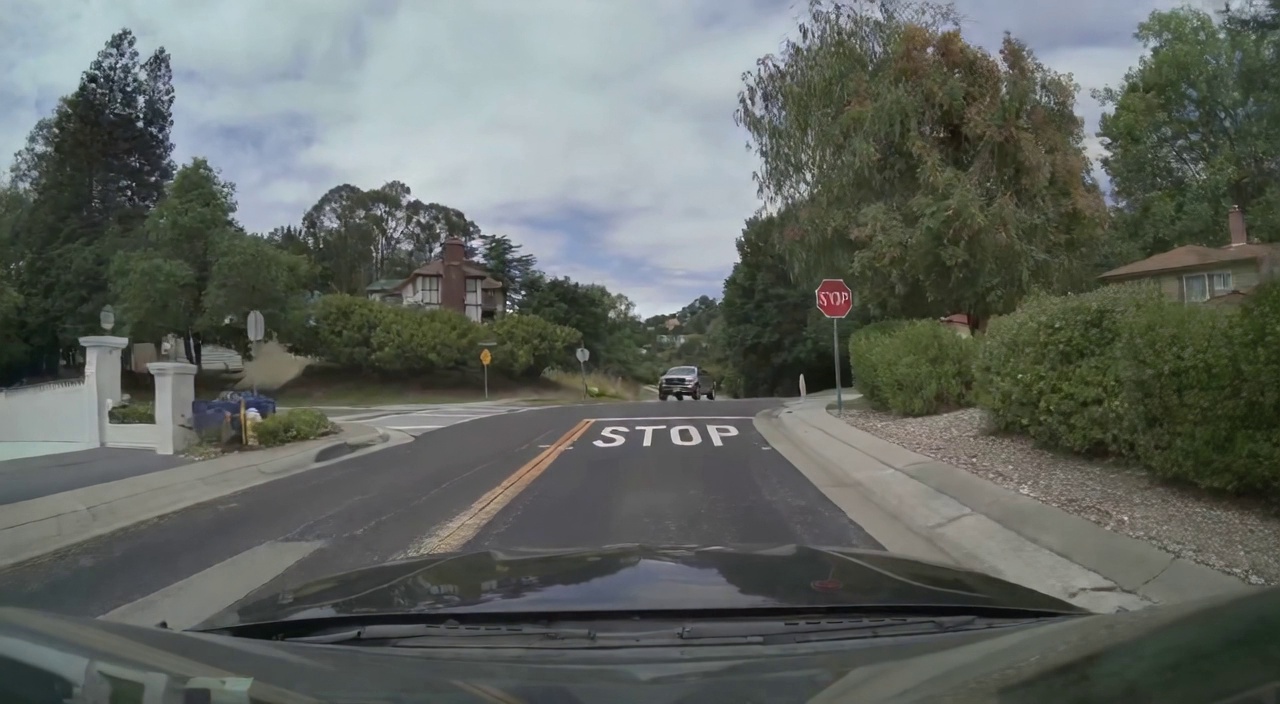} &
         \includegraphics[width=0.19\textwidth]{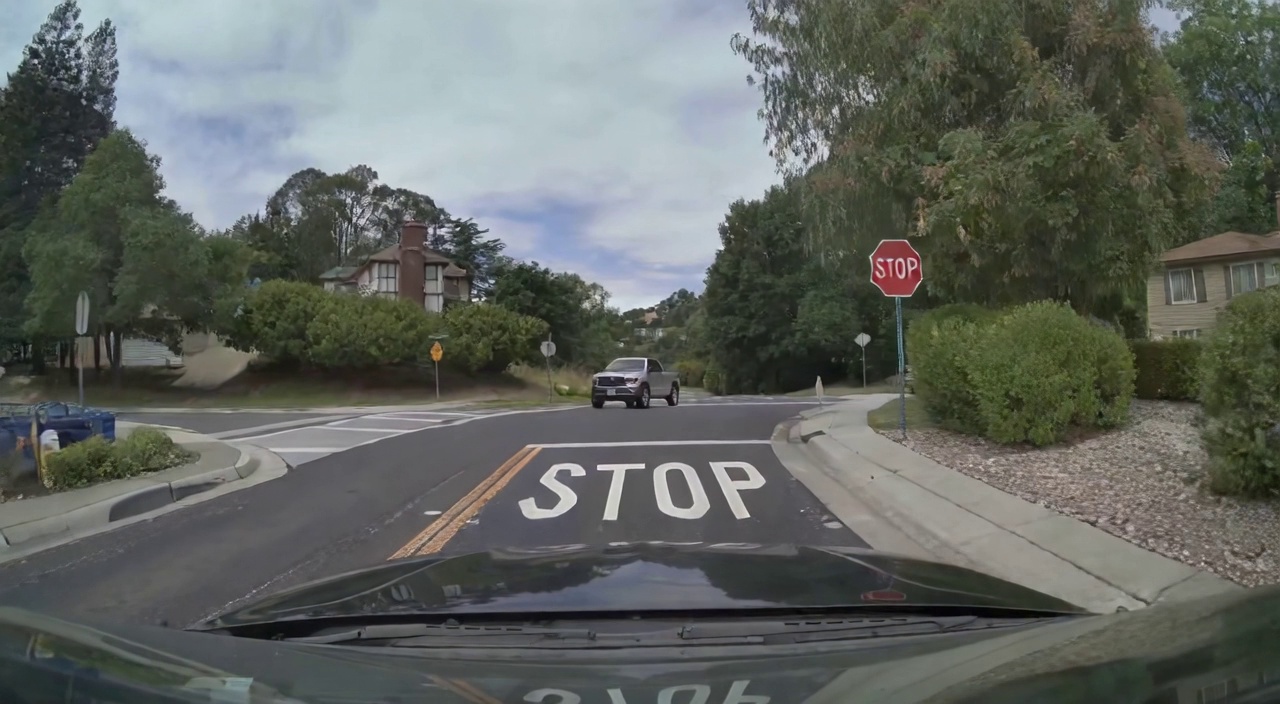} &
         \includegraphics[width=0.19\textwidth]{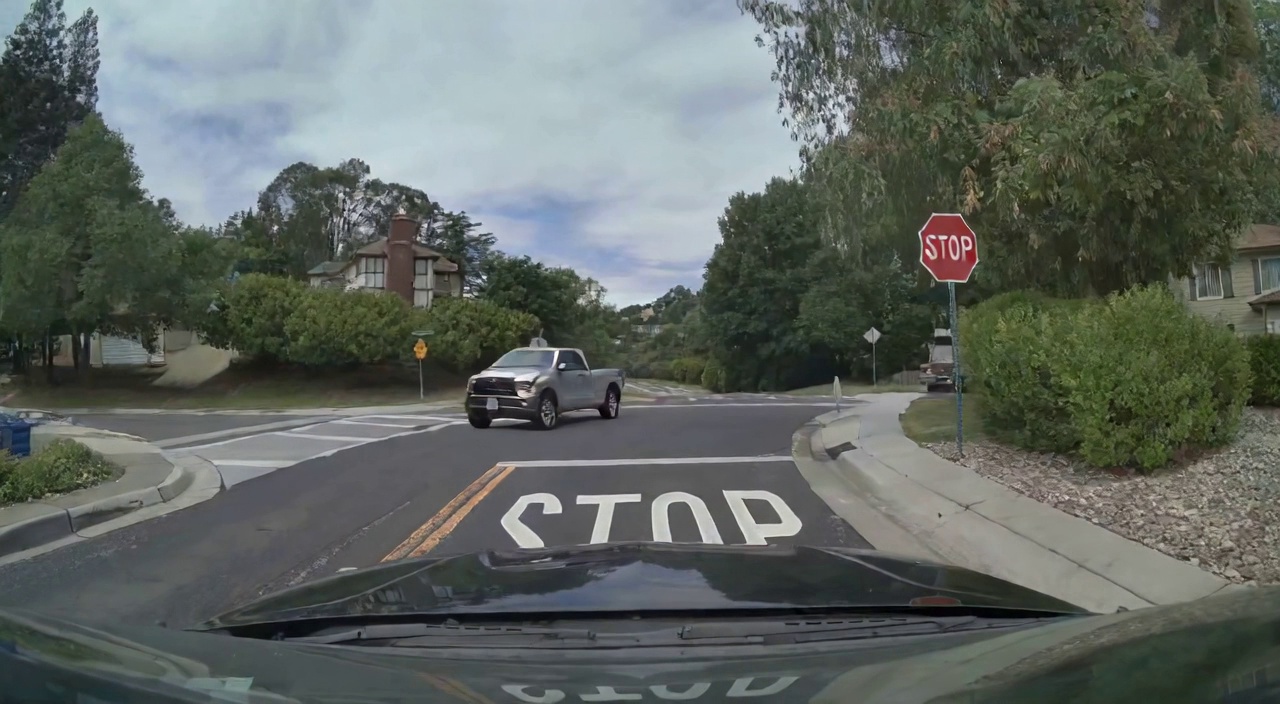} \\
         \includegraphics[width=0.19\textwidth]{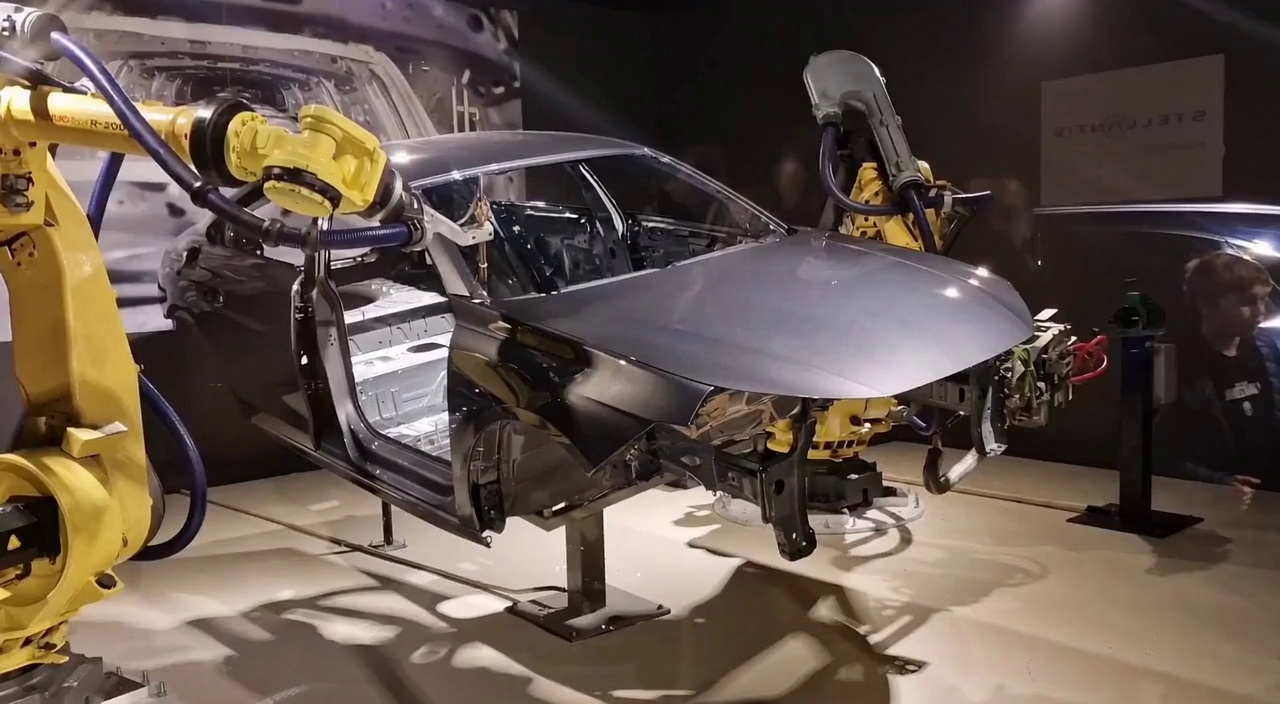} & &
         \includegraphics[width=0.19\textwidth]{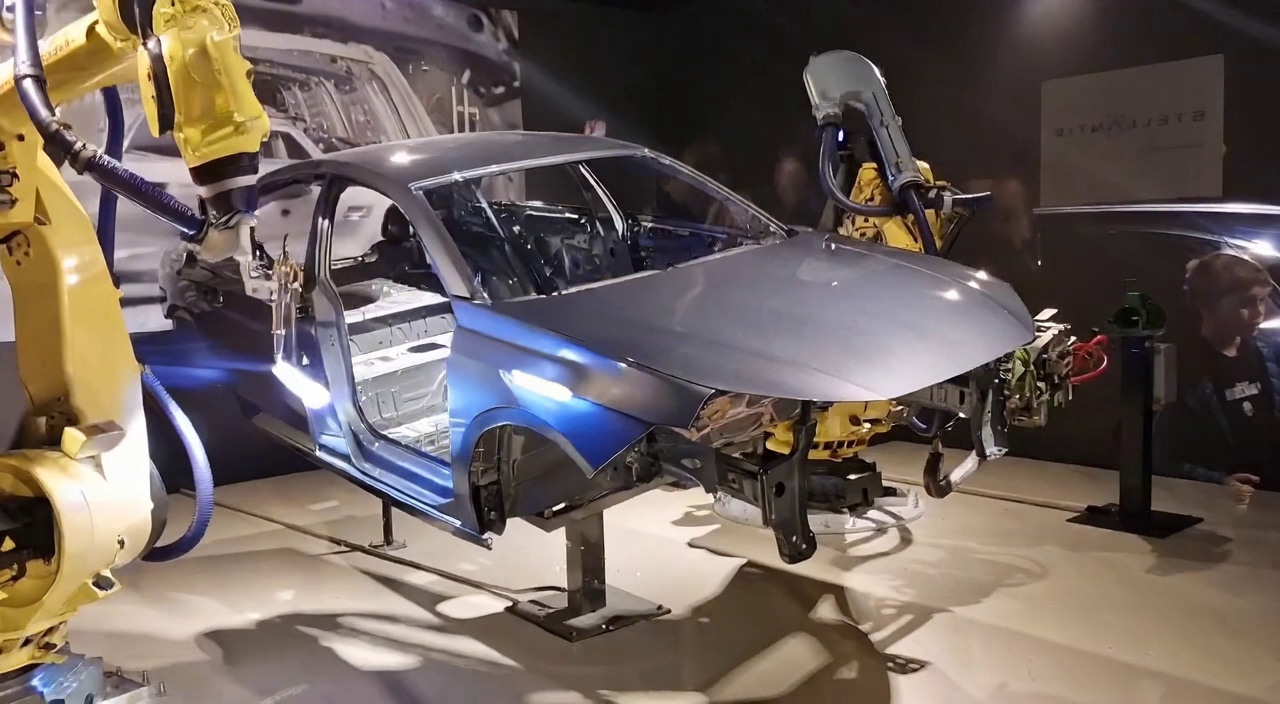} &
         \includegraphics[width=0.19\textwidth]{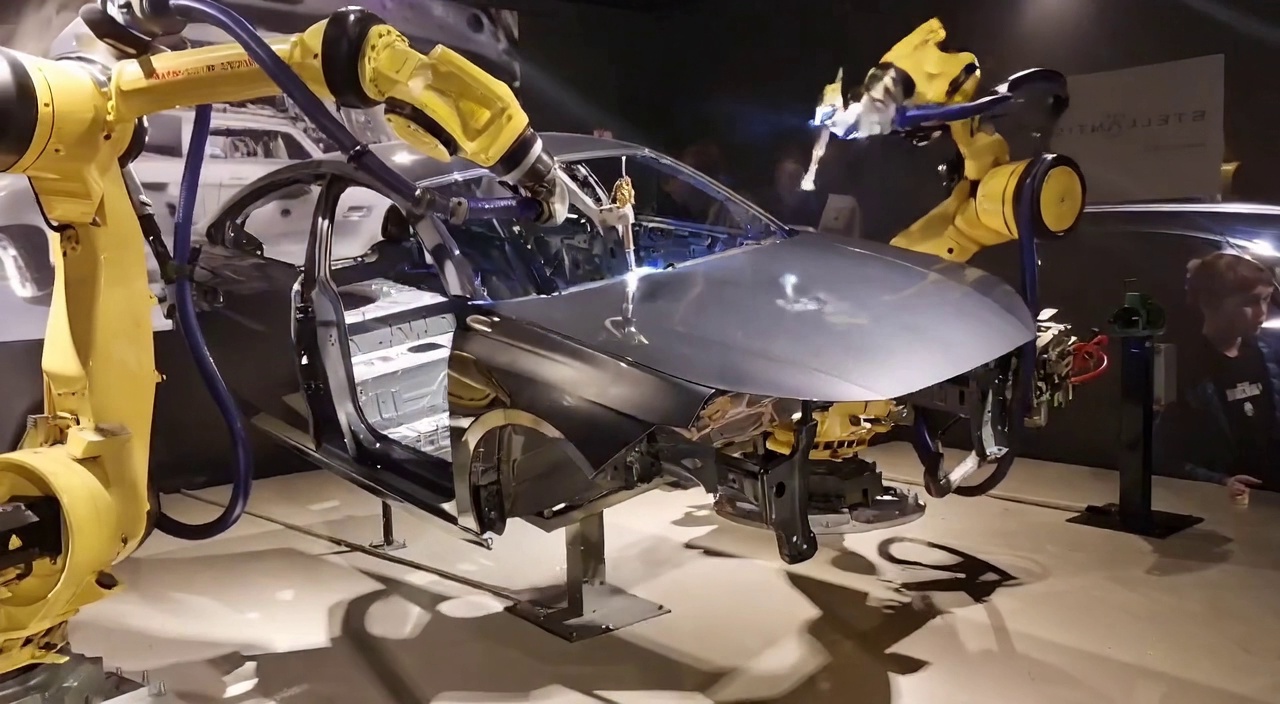} &
         \includegraphics[width=0.19\textwidth]{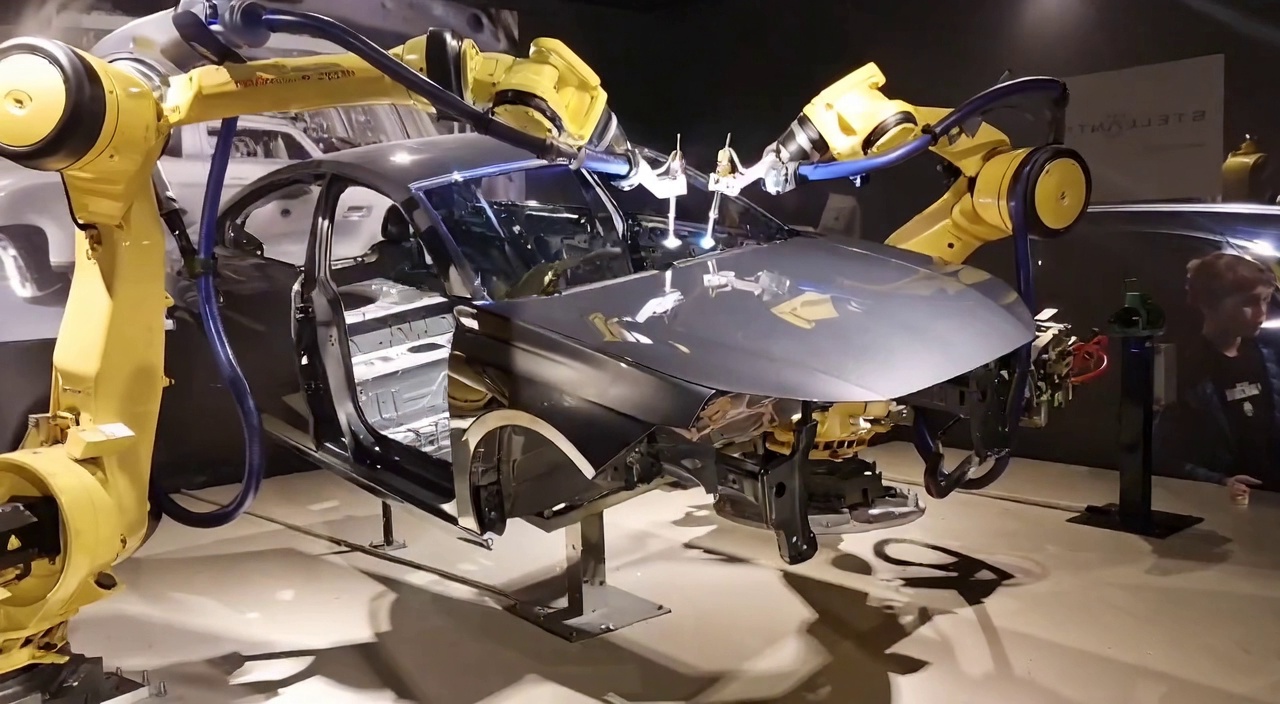} &
         \includegraphics[width=0.19\textwidth]{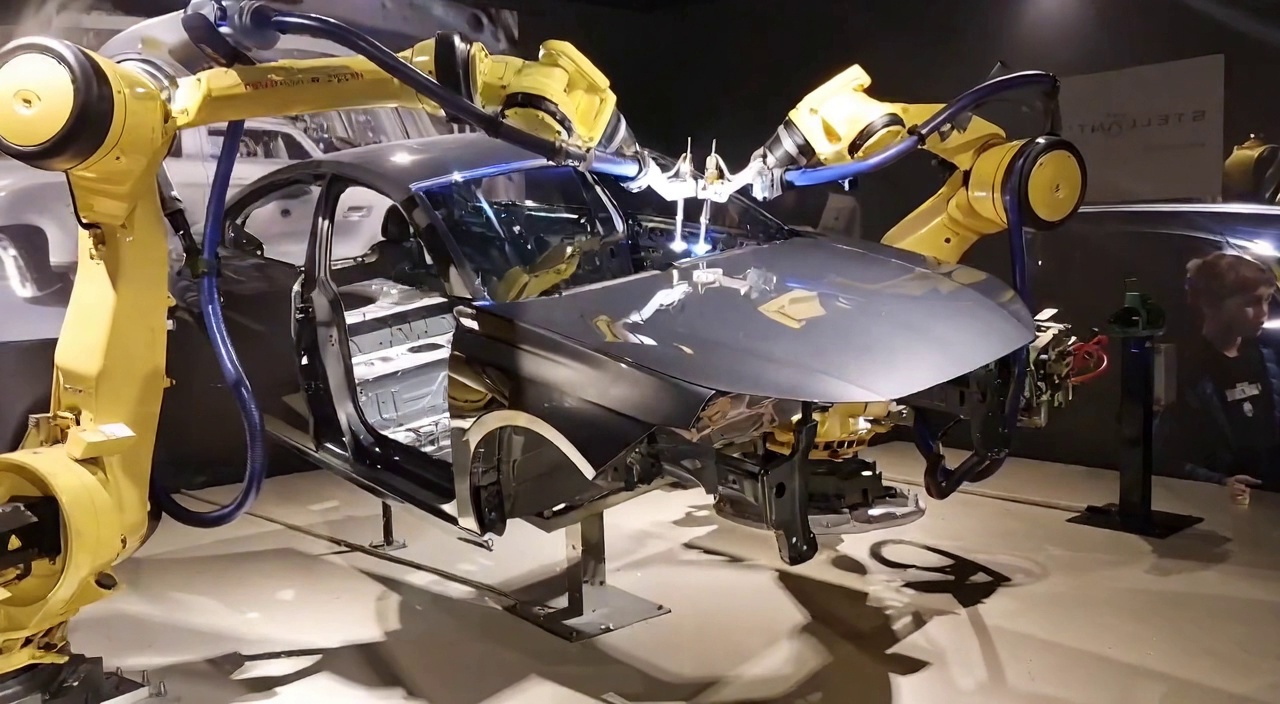} \\
         \includegraphics[width=0.19\textwidth]{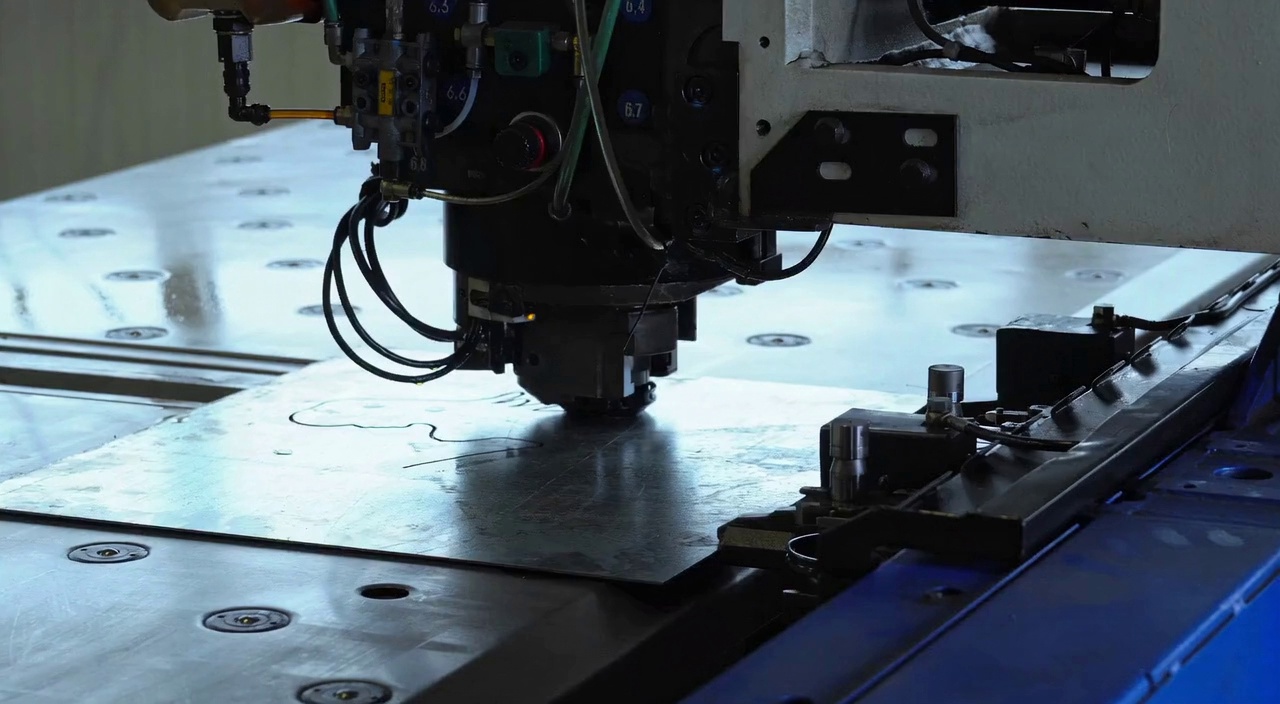} & &
         \includegraphics[width=0.19\textwidth]{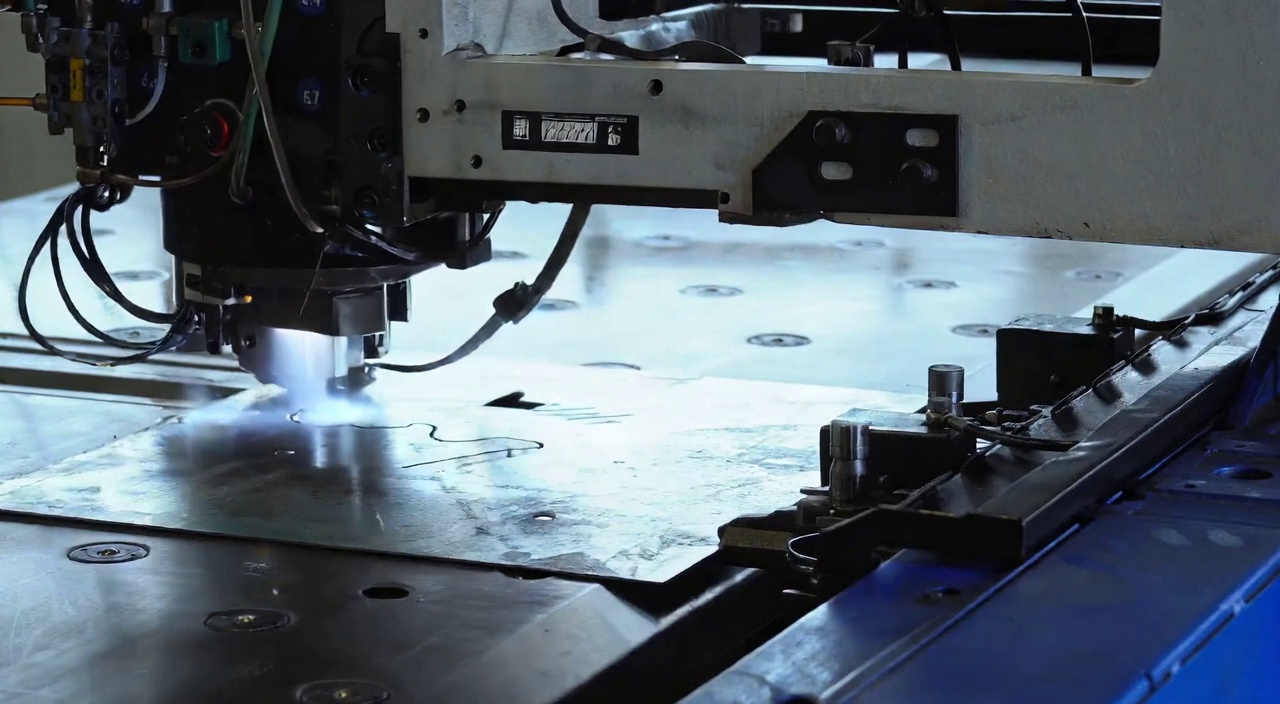} &
         \includegraphics[width=0.19\textwidth]{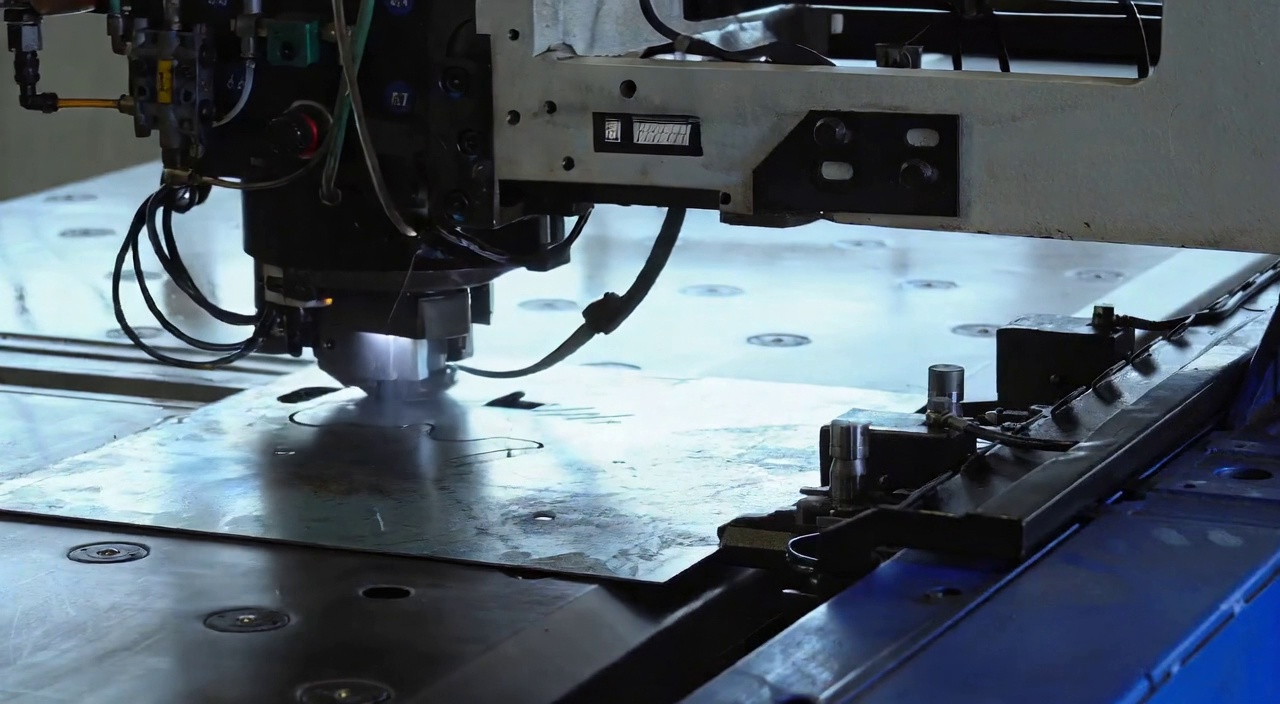} &
         \includegraphics[width=0.19\textwidth]{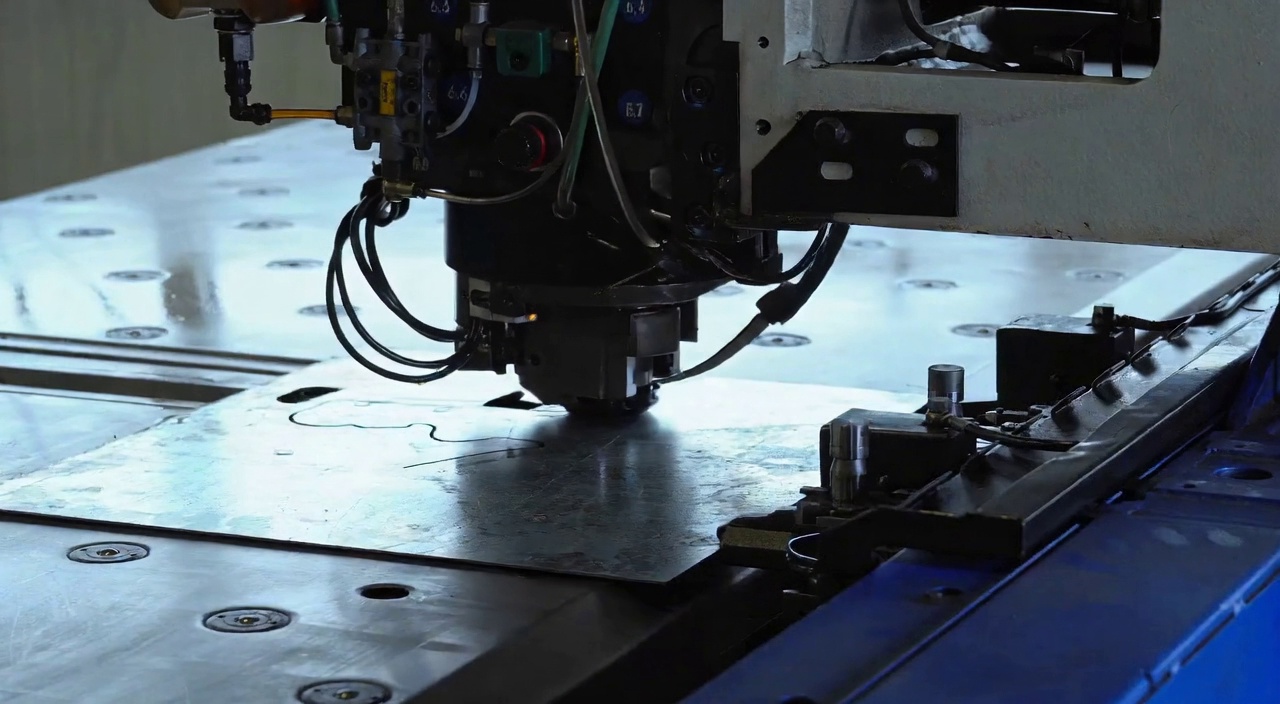} &
         \includegraphics[width=0.19\textwidth]{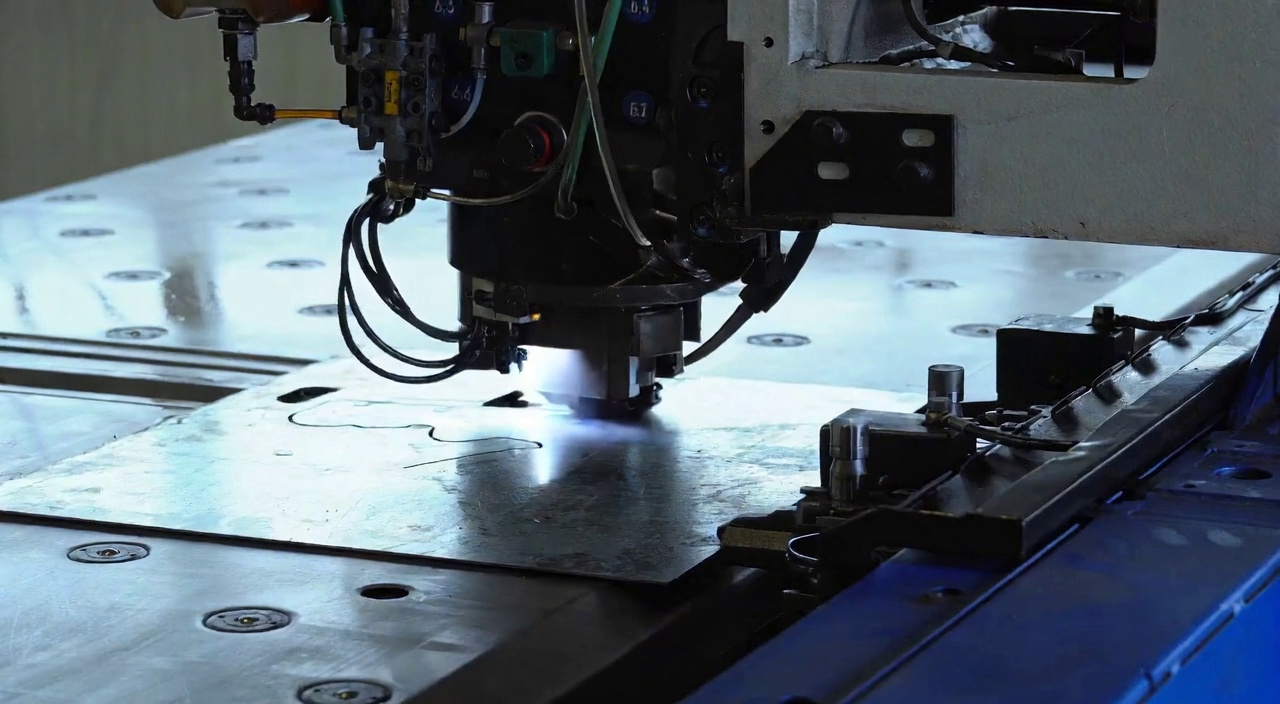} \\
         \includegraphics[width=0.19\textwidth]{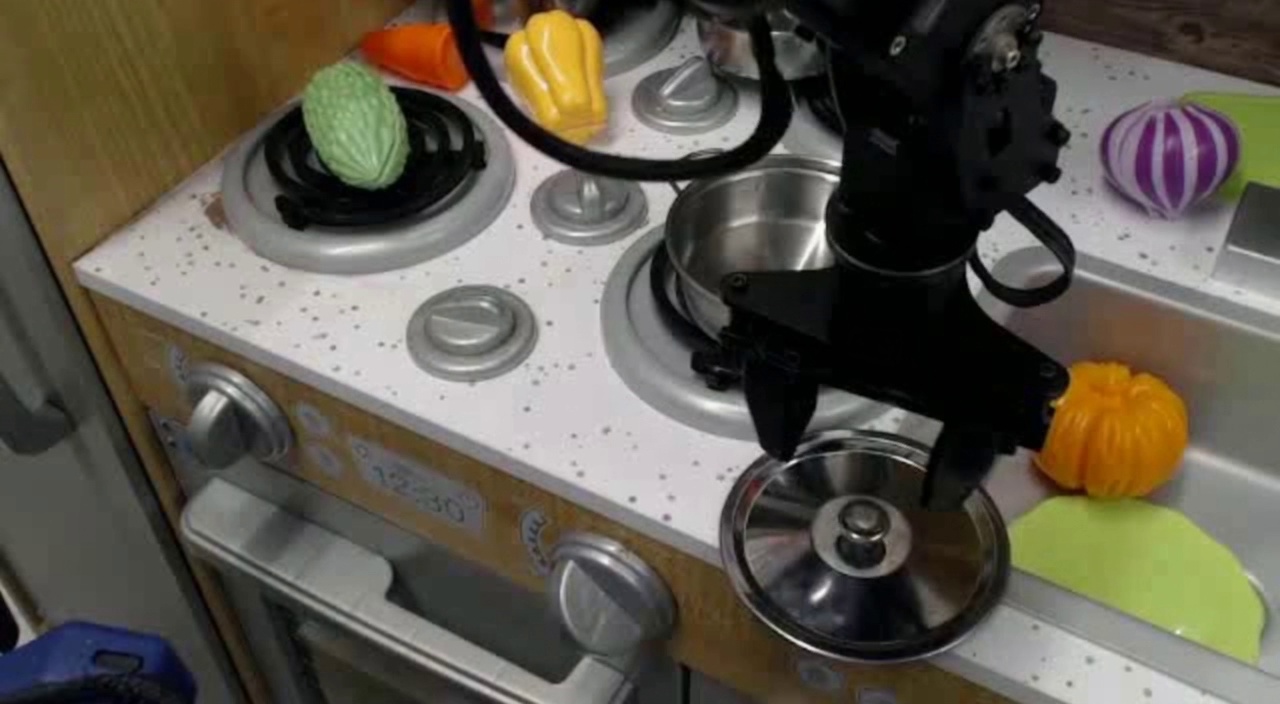} & &
         \includegraphics[width=0.19\textwidth]{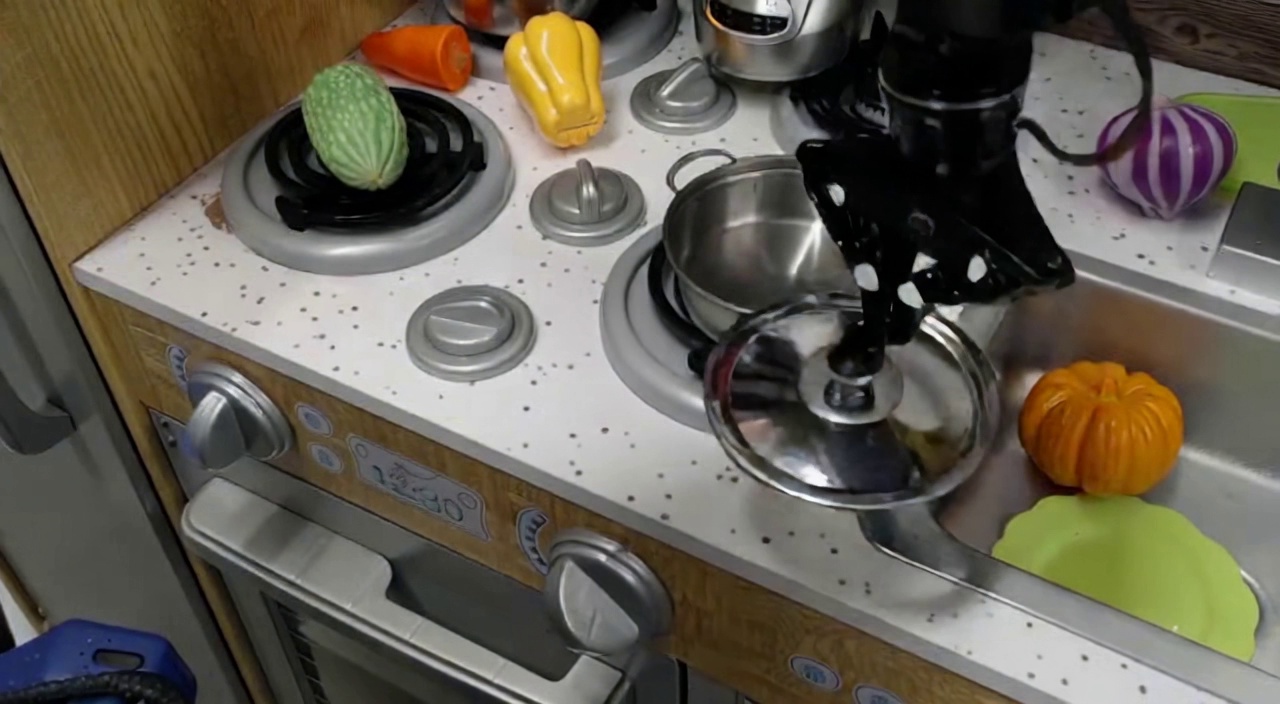} &
         \includegraphics[width=0.19\textwidth]{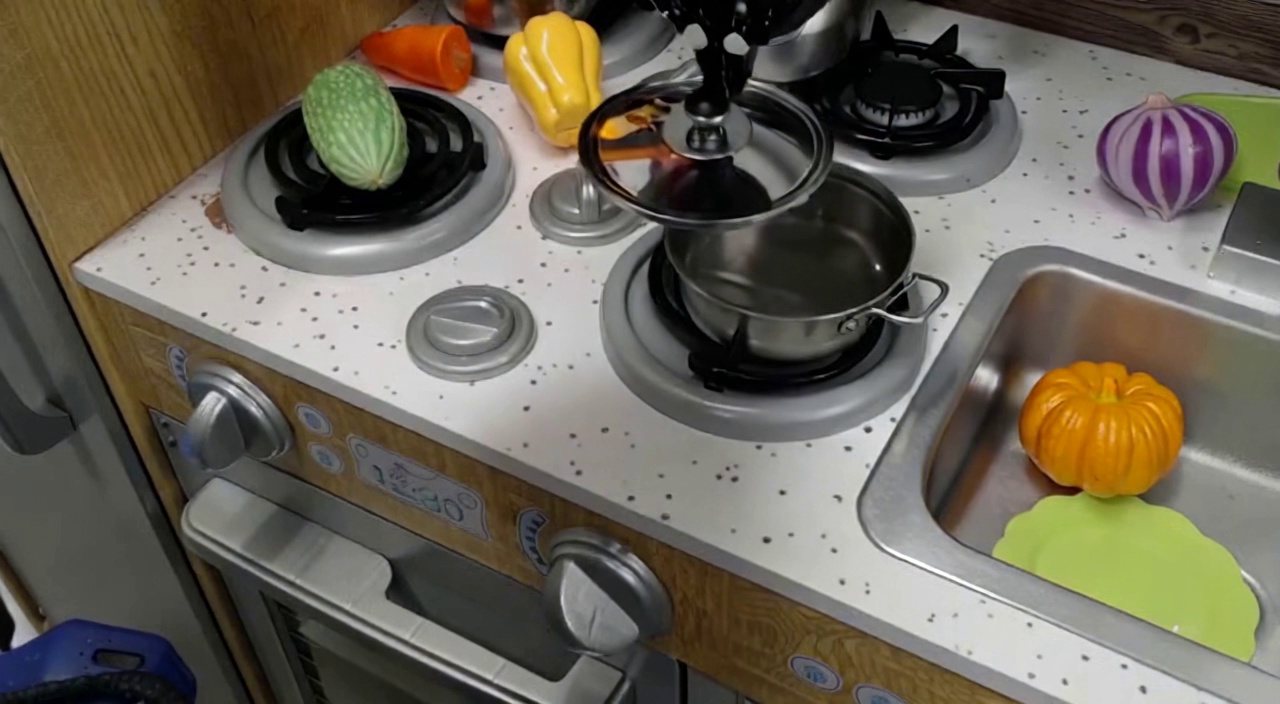} &
         \includegraphics[width=0.19\textwidth]{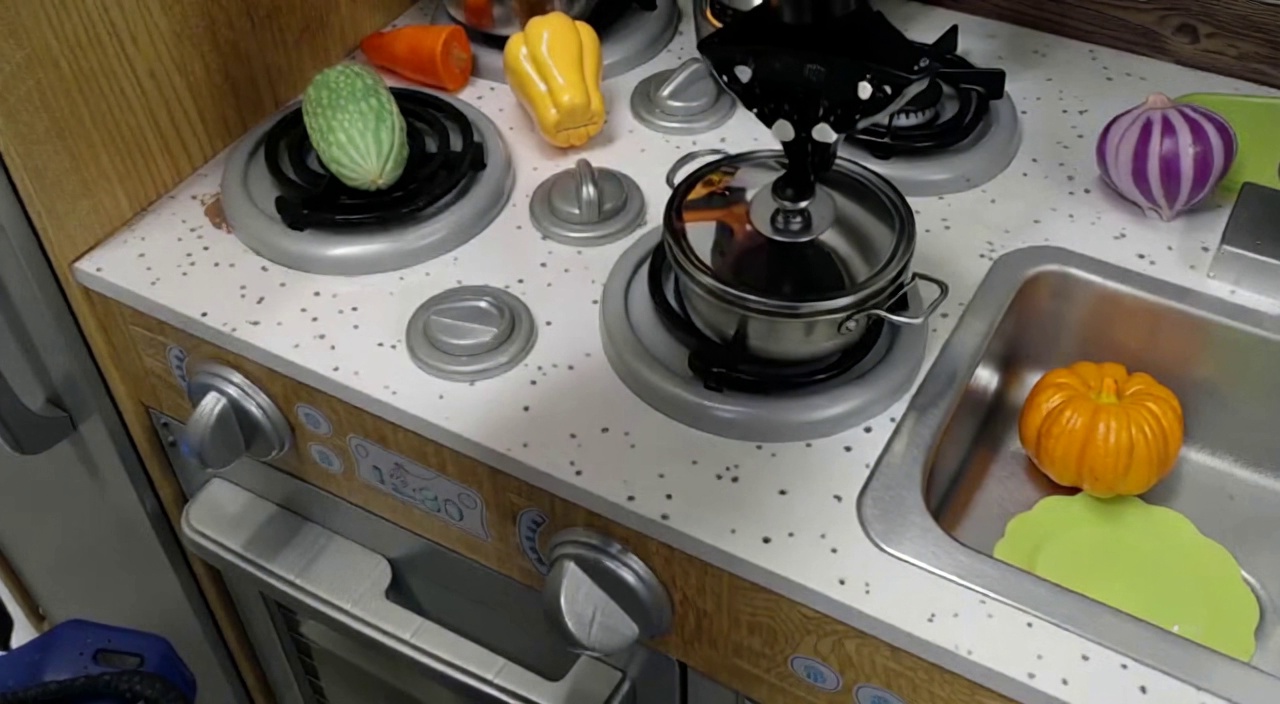} &
         \includegraphics[width=0.19\textwidth]{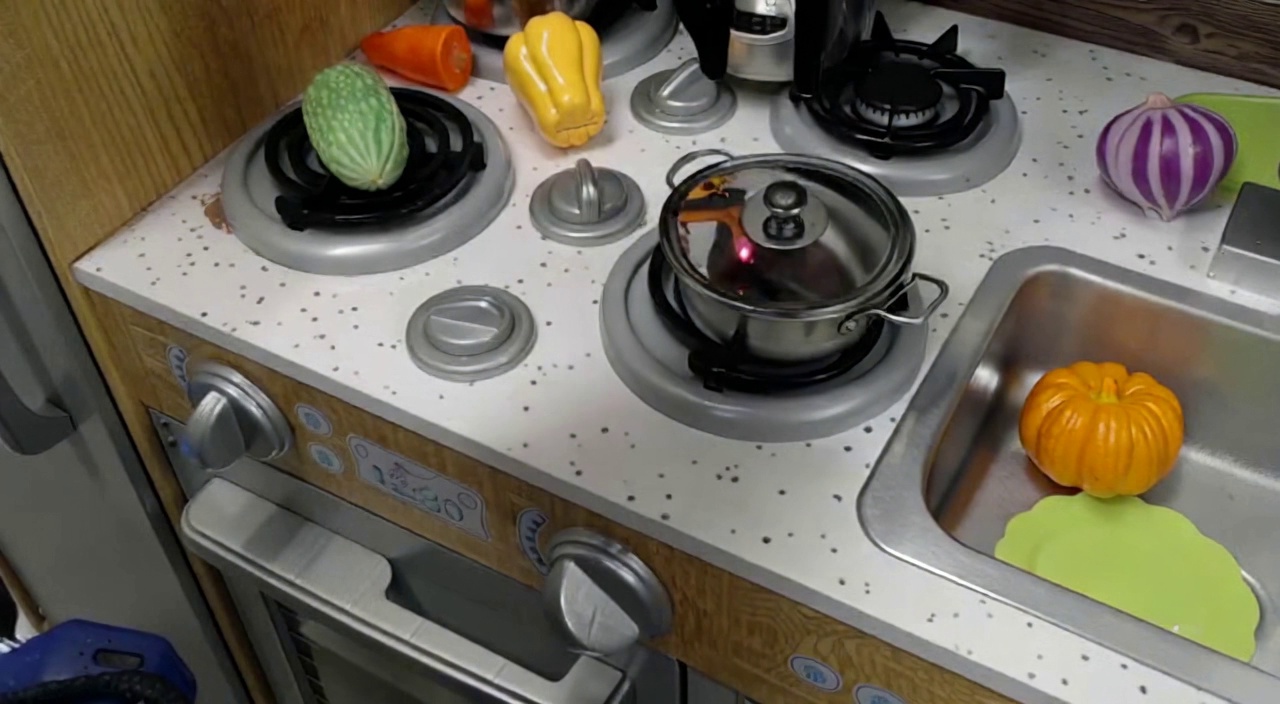} \\
         \includegraphics[width=0.19\textwidth]{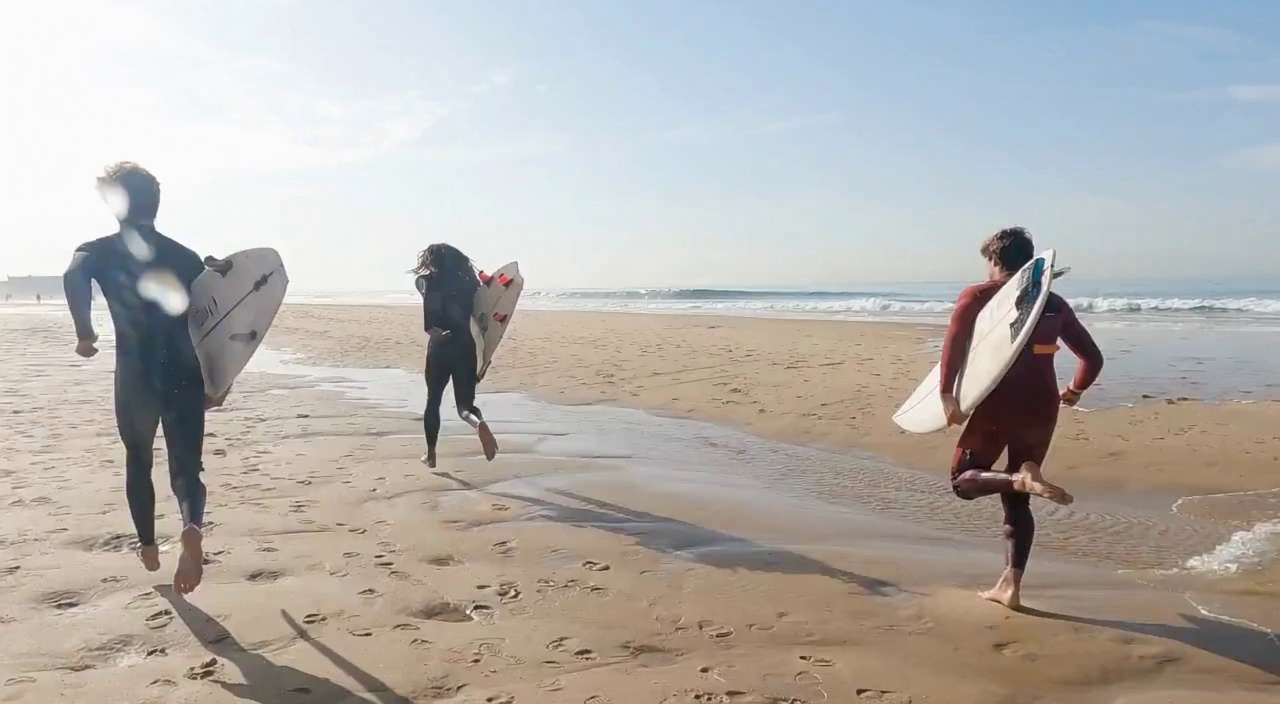} & &
         \includegraphics[width=0.19\textwidth]{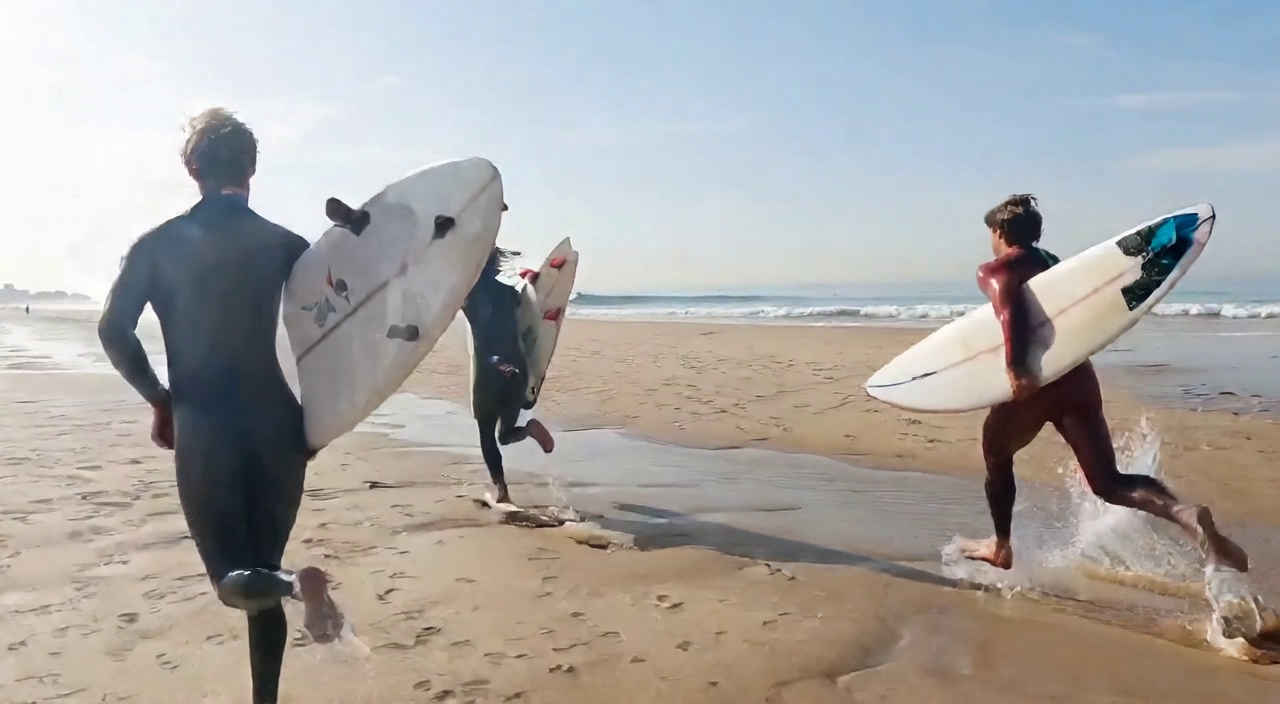} &
         \includegraphics[width=0.19\textwidth]{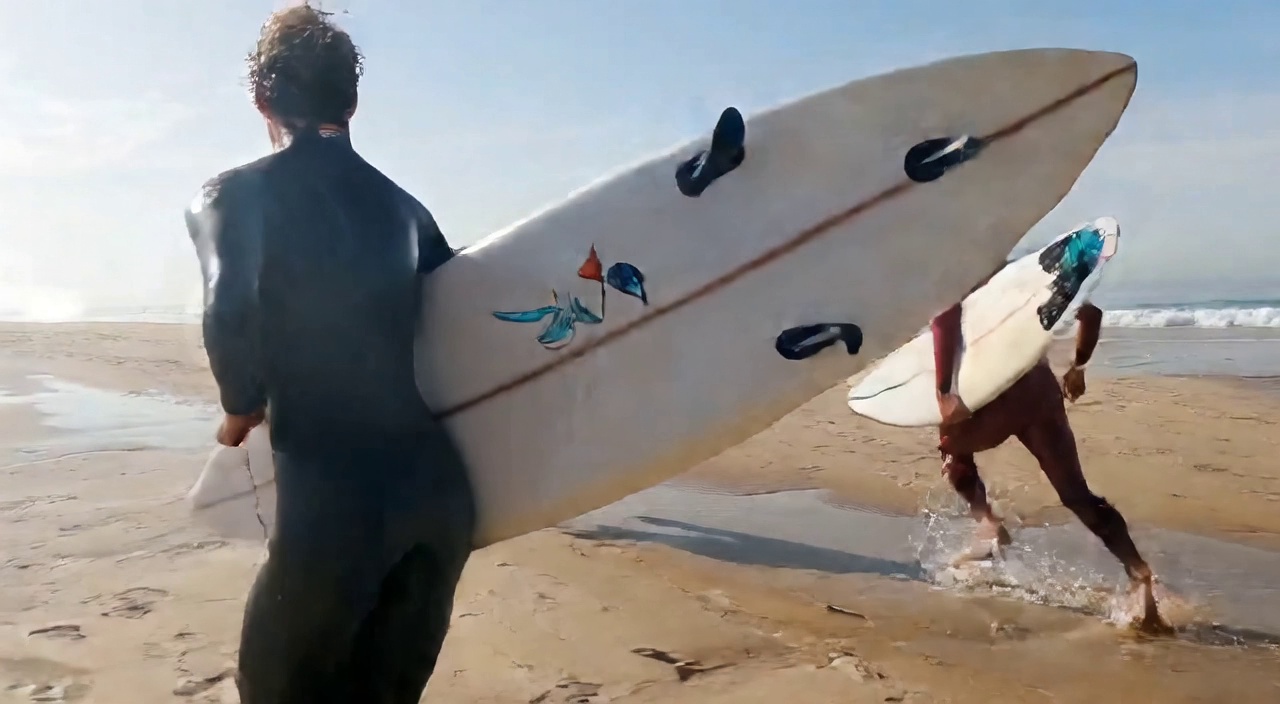} &
         \includegraphics[width=0.19\textwidth]{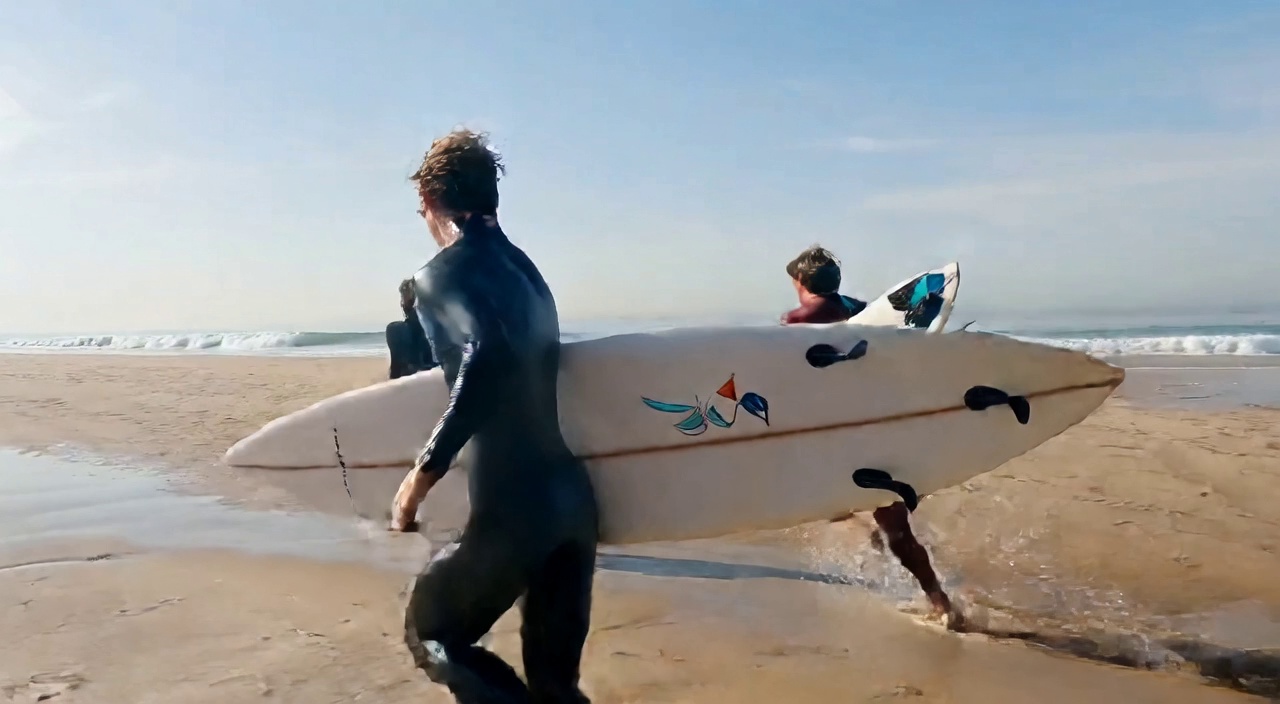} &
         \includegraphics[width=0.19\textwidth]{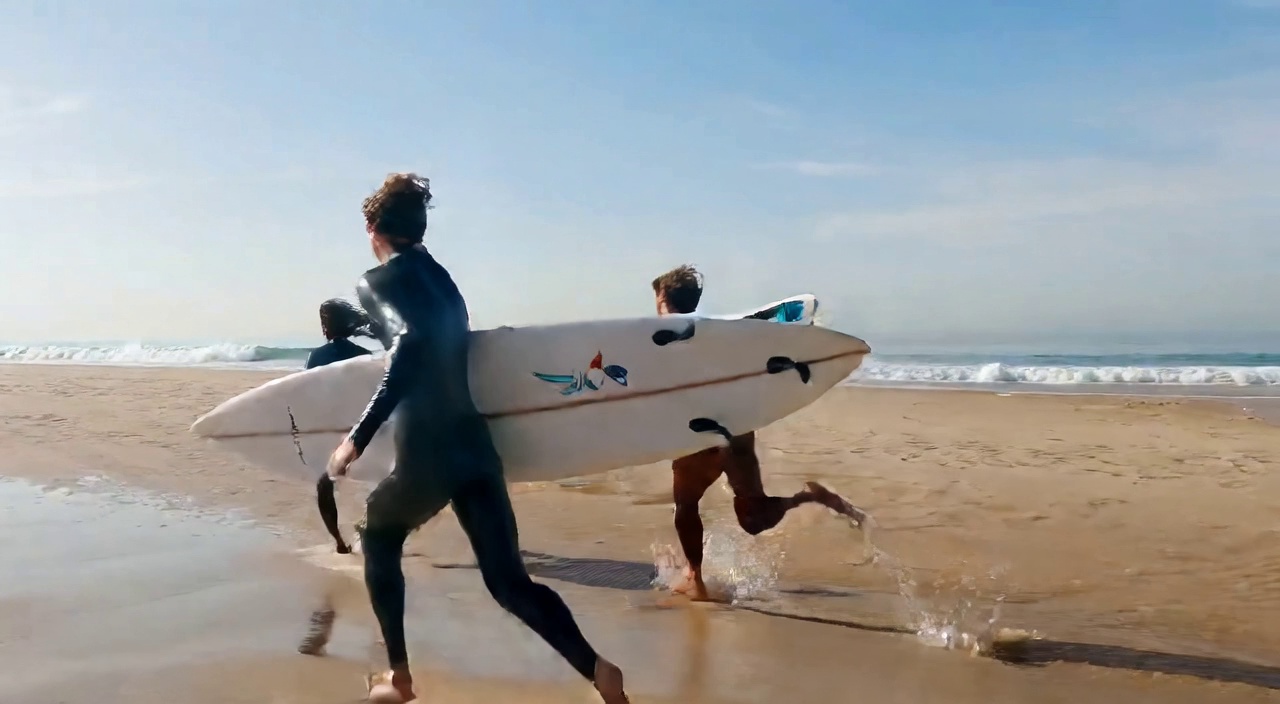} \\
    \end{tabular}
    \caption{[Cosmos-Predict2.5-2B] post-trained prediction samples on the PAI-Bench dataset.}
    \label{fig:2B_visual_results}
\end{figure*}

We show visual samples in \Cref{fig:2B_visual_results} as representative examples of physical AI scenarios. The [Cosmos-Predict2.5-2B] post-trained model is able to simulate accurate behaviors in driving, generate realistic industrial and robotics scenes, and produce physically coherent motion.

\section{Applications}
\label{sec::applications}

We demonstrate the versatility of [Cosmos-Predict2.5] across multiple Physical AI applications. First, we introduce [Cosmos-Transfer2.5], which provides control-net style generation capability to Physical AI applications~(\cref{subsec::cosmos-transfer2p5}). Compared to [Cosmos-Transfer1], the new model is substantially more effective while being 3.5× smaller. We further show that [Cosmos-Transfer2.5] enables Real2Real augmentation for policy learning~(\cref{subsec::robotpolicy}), and that the same paradigm applies to autonomous driving, where we construct multiview world models conditioned on world scenario maps for realistic driving simulation~(\cref{subsec::cosmos-auto}).

We also extend [Cosmos-Predict2.5] to support camera-pose–controllable multiview generation~(\cref{subsec::camera-control}) and apply it to synthetic data generation for VLA training~(\cref{subsec::gr00tdream}). Finally, in~\cref{subsec::action-cond}, we post-train [Cosmos-Predict2.5] into an action-conditioned world model that is particularly well-suited for policy evaluation.

\subsection{Cosmos-Transfer2.5}
\label{subsec::cosmos-transfer2p5}

We develop a conditional world generation model, [Cosmos-Transfer2.5-2B], built on top of [Cosmos-Predict2.5-2B], that produces high-quality world simulations conditioned on \textit{multiple spatial control inputs}. These inputs can take different modalities—including edges, blurred video, segmentation maps, and depth maps—and may originate from either a physics simulation engine, such as NVIDIA IsaacSim, or from real-world video data.

In terms of architecture, [Cosmos-Transfer2.5-2B] follows the general design of [Cosmos-Transfer1-7B]~\citep{cosmos_transfer1}, but with a key modification. Whereas [Cosmos-Transfer1-7B] inserts four control blocks sequentially at the start of the main branch, [Cosmos-Transfer2.5-2B] distributes its four control blocks more evenly by inserting one after every seven blocks in the main branch. This design preserves the total number of control blocks while integrating conditioning information more gradually throughout the network. For additional architectural details, please refer to [Cosmos-Transfer1]\citep{cosmos_transfer1}.

To train [Cosmos-Transfer2.5-2B], we curate high-quality, control-condition data from our pre-training video dataset, with a particular emphasis on Physics AI domains such as autonomous driving, robotics, smart spaces, and physics. World generations in these domains require precise spatial and temporal understanding, making them ideal for testing the effectiveness of different control modalities.

Depth information is crucial for capturing geometric structure and 3D reasoning. We use Video Depth Anything~\citep{chen2025videodepth} to generate depth maps for 10 million videos for depth conditioning. Semantic segmentation provides fine-grained object-level and region-level cues that are essential for tasks like robotics and scene interaction. We apply SAMv2~\citep{ravi2024sam} on 3 million videos for segmentation conditioning. In addition, following the pipeline of [Cosmos-Transfer1-7B]~\citep{cosmos_transfer1}, we curate 14 million videos with edge and blur conditions. Edge maps highlight object boundaries that aid perception, while blurred videos serve as a robust training signal, forcing the model to recover sharp details.

Each control branch corresponding to a modality is trained independently for 100,000 iterations with an effective batch size of 64, allowing the model to specialize in extracting useful representations from each type of input before integration. For all other hyperparameters, we adopt the same settings as those used in [Cosmos-Predict2.5-2B], ensuring consistency across models.

\begin{table}[htb!]
\centering
\caption{\textbf{Quantitative evaluation on transfer models for various configurations.} We compare single control models (each conditioned on a single modality) and multi-modal variants that use spatially uniform weights. For the multi-modal cases, ``Uniform Weights'' denotes the full model that integrates all four control modalities (each weighted at 0.25). Best results are in bold; second-best are underlined.}
\label{tab:eval_transfer2}
\resizebox{\textwidth}{!}{
\begin{tabular}{l|cccc|c}
\toprule
\multicolumn{1}{c|}{\multirow{4}{*}{Model}} & \makecell{Blur \\Alignment} & \makecell{Edge\\Alignment} & \makecell{Depth\\Alignment} & \makecell{Segmentation\\Alignment} & \makecell{Overall\\Quality} \\
\cmidrule(lr){2-6}
          & \makecell{Blur\\SSIM $\uparrow$} & \makecell{Edge\\F1 $\uparrow$} & \makecell{Depth\\si-RMSE $\downarrow$} & \makecell{Mask\\mIoU $\uparrow$} & \makecell{Quality\\Score $\uparrow$}\\
\midrule
Cosmos-Transfer1-7B [Blur] & \underline{0.89} & 0.20 & \underline{0.66} & 0.73 &  6.56 \\
Cosmos-Transfer1-7B [Edge] & 0.77 & 0.38 & 0.85 & 0.73 & 6.76  \\
Cosmos-Transfer1-7B [Depth] & 0.67 & 0.15 & 0.76 & 0.71  & 6.89 \\
Cosmos-Transfer1-7B [Seg] & 0.62 & 0.11 & 1.13 & 0.70  & 6.02 \\
Cosmos-Transfer1-7B Uniform Weights & 0.82 & 0.26 & 0.70 & 0.74 & 9.24 \\
\midrule
Cosmos-Transfer2.5-2B [Blur] & \textbf{0.90} & 0.26 & \textbf{0.59} & \underline{0.75} &  \textbf{9.75} \\
Cosmos-Transfer2.5-2B [Edge] & 0.79 & \textbf{0.49} & 0.76 & 0.75 & 8.73  \\
Cosmos-Transfer2.5-2B [Depth] & 0.71 & 0.19 & 0.70 & 0.73  & 8.85 \\
Cosmos-Transfer2.5-2B [Seg] & 0.68 & 0.14 & 1.02 & 0.71  & 8.81 \\
Cosmos-Transfer2.5-2B Uniform Weights & 0.87 & \underline{0.41} & 0.67 & \textbf{0.76} & \underline{9.31} \\
\bottomrule
\end{tabular}
}
\end{table}
\begin{figure}[htb!]
  \centering
  \includegraphics[width=\textwidth]{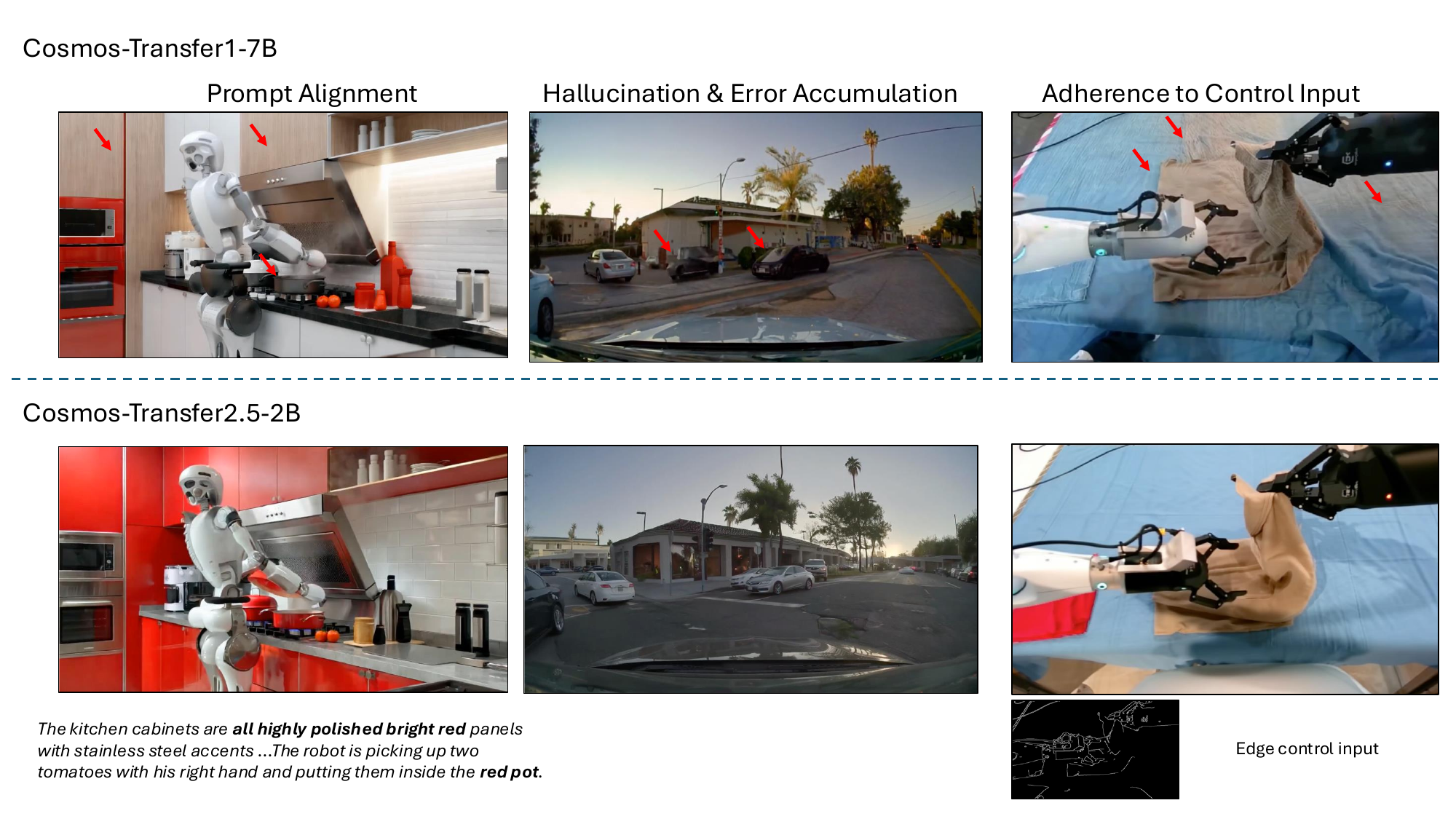}          
  \caption{\textbf{Sample comparison results of [Cosmos-Transfer2.5-2B]}. Compared to [Cosmos-Transfer1-7B], [Cosmos-Transfer2.5-2B] has better prompt alignment, better adherence to control input, and less hallucination and error accumulation (especially for long videos).}
  \label{fig:transfer2_results}
\end{figure}
\begin{figure}[htb!]
  \centering
  \begin{subfigure}{0.45\textwidth}
    \captionsetup{justification=centering, labelformat=empty}
    \includegraphics[width=\linewidth]{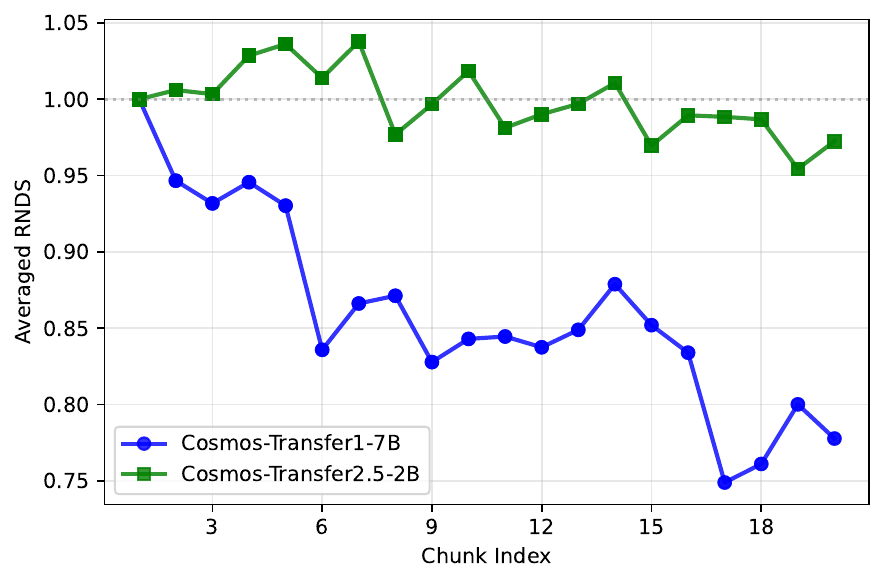}
    \caption{Edge Control}
  \end{subfigure}
  \hfill
  \begin{subfigure}{0.45\textwidth}
    \captionsetup{justification=centering, labelformat=empty}
    \includegraphics[width=\linewidth]{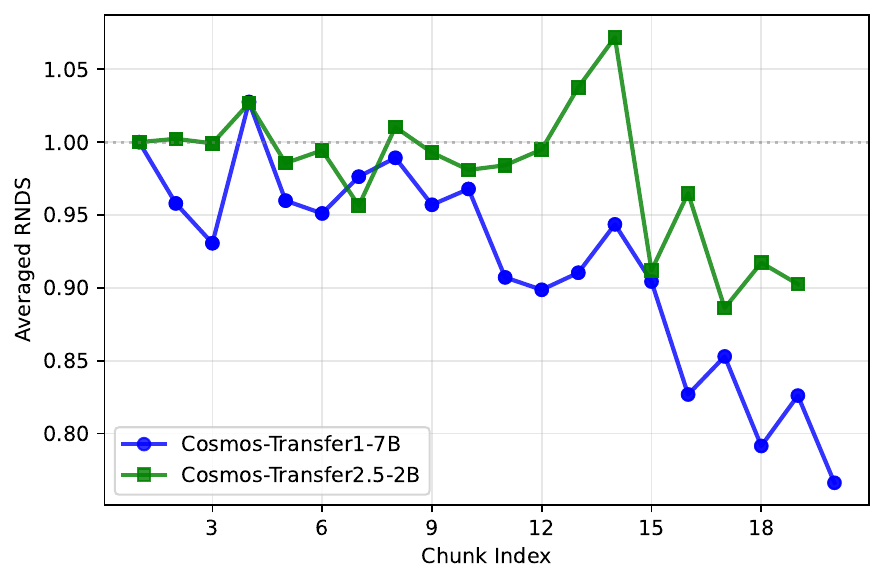}
    \caption{Blur Control}    
  \end{subfigure}

  \vskip\baselineskip   
 
  \begin{subfigure}{0.45\textwidth}
    \captionsetup{justification=centering, labelformat=empty}
    \includegraphics[width=\linewidth]{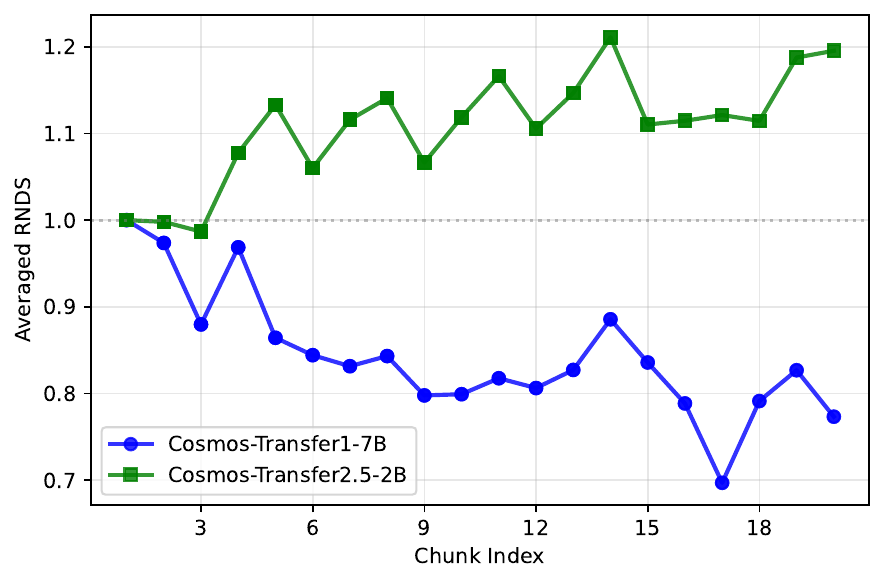}
    \caption{Depth Control}    
  \end{subfigure}
  \hfill
  \begin{subfigure}{0.45\textwidth}
    \captionsetup{justification=centering, labelformat=empty}
    \includegraphics[width=\linewidth]{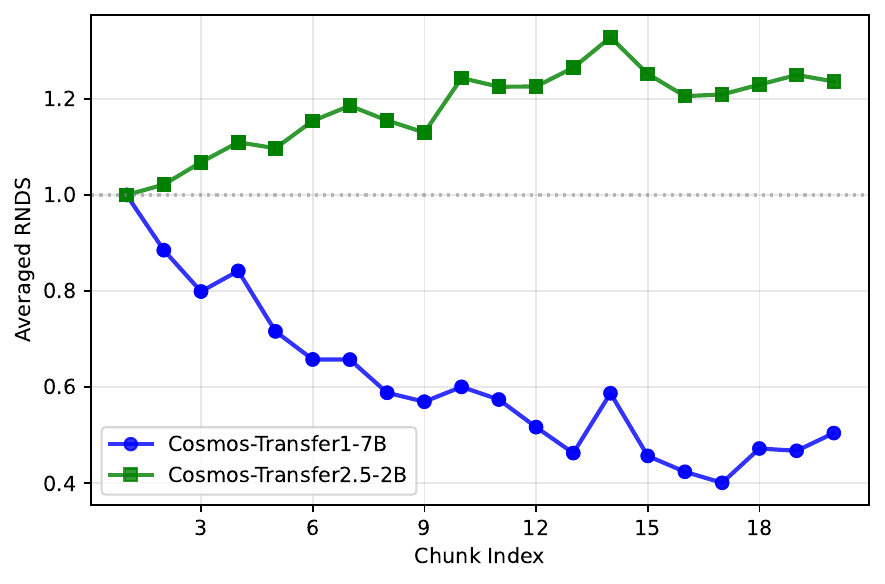}
    \caption{Segmentation Control}     
  \end{subfigure}
  \caption{\textbf{Error accumulation for long video generations.} These plots show the Normalized Relative Dover Score vs Chunk Index for auto-regressive multi-trunk long video generation where each trunk is 93 frames. As shown, for all four control modalities (edge/blur/depth/seg), compared to [Cosmos-Transfer1-7B] (blue curves), [Cosmos-Transfer2.5-2B] (green curves) has much less reduction in RNDS along the chunk index dimension, which shows less hallucination and error accumulation for long videos.}
  \label{fig:5_1_transfer2_error_accumulation}
\end{figure}

\subsubsection{Results} 

For evaluation, we use PAIBench-Transfer~\citep{PAI-Bench}, a benchmark dataset containing 600 videos spanning diverse domains such as driving and robotics. The evaluation is structured around two key dimensions: adherence to control inputs (how well the generated video follows the provided conditions) and overall video quality (measuring realism and consistency). The quantitative results are summarized in \cref{tab:eval_transfer2}.

As shown in the table, [Cosmos-Transfer2.5-2B] outperforms [Cosmos-Transfer1-7B] on both metrics, despite being 3.5 times smaller in size. This improvement can be attributed to two factors: (1) stronger [Cosmos-Predict2.5-2B] as the base model, and (2) the use of more carefully curated, Physics-AI-focused training data, which better aligns with the benchmark domains. Visual comparisons highlighting these gains are provided in \cref{fig:transfer2_results}.

\subsubsection{Long Video Generation} 

In addition, we introduce a new metric designed to evaluate error accumulation in long-video generation. Since DiT-based video generation models are constrained by limited context length, they typically generate long videos autoregressively, producing one chunk at a time. This chunked generation process inevitably leads to error accumulation, where artifacts and inconsistencies increase as the video length grows

To study this effect, we curate a set of 17 evaluation videos ranging from 30 to 120 seconds in length. We then propose the averaged Relative Normalized Dover Score (RNDS) as a quantitative measure of how video quality degrades across chunks. RNDS is defined as a curve over chunk indices: 
\begin{equation}
    \mbox{RNDS}[i]=\left(\frac{\mbox{DOVER}[i]}{\mbox{DOVER}_{\mbox{GT}}[i]}\right)/\left(\frac{\mbox{DOVER}[1]}{\mbox{DOVER}_{\mbox{GT}}[1]}\right),
\end{equation}
where $i = 1, \ldots, T$ denotes the chunk index, $\text{DOVER}[i]$ is the Dover score~\citep{dover} of the $i$-th generated chunk, and $\text{DOVER}_{\text{GT}}[i]$ is the corresponding Dover score for the ground-truth video. This normalization ensures that the RNDS curve always starts at $(1,1)$, making it easy to compare degradation trends across models. The averaged RNDS is then obtained by averaging curves over all evaluation videos.

As shown in~\cref{fig:5_1_transfer2_error_accumulation}, the RNDS curves reveal that [Cosmos-Transfer2.5-2B] exhibits far less reduction in RNDS over time compared to [Cosmos-Transfer1-7B]. This indicates that our smaller model accumulates fewer errors, demonstrates less hallucination, and maintains higher fidelity over long video sequences.

\subsection{Cosmos-Transfer2.5 for Robot Policy Learning}
\label{subsec::robotpolicy}

We aim to investigate the following question: Can [Cosmos-Transfer2.5-2B] be used as a visual synthetic data generator to augment robot policy training and enable generalization to unseen visual scenarios?

Our setup follows a standard real-world imitation learning pipeline. Using a bimanual robot equipped with an egocentric camera, we first collect human teleoperation demonstrations for table-top manipulation tasks. From these demonstrations, we train a vision-based policy that maps image observations and proprioception to action chunks using state-of-the-art behavioral cloning techniques. The trained policy is then deployed back on the same platform for evaluation.

Unlike conventional imitation learning benchmarks, however, we introduce adversarial visual perturbations during evaluation---for example, modifying object appearances, changing scene backgrounds, or placing distractor objects on the table. This setting reflects a realistic deployment scenario, where a policy must operate in environments that differ drastically from the conditions in which the demonstrations were collected. Such domain shifts often involve structured visual changes that cannot be easily synthesized using standard image augmentation methods.

Here, [Cosmos-Transfer2.5-2B] offers a unique advantage: it not only enables the generation of these structured variations for visual data augmentation, but also provides controllability through text prompts that specify the desired visual conditions. This enables the systematic simulation of challenging out-of-domain scenarios and the testing of policy robustness in a controlled yet flexible manner.

\subsubsection{System and Task Settings} 
We conduct our experiments on a semi-humanoid robotic platform equipped with two 7-DoF Kinova Gen3 arms, each fitted with a Robotiq 2F-140 gripper. An Intel RealSense D455 camera is mounted on the robot’s head to capture egocentric image observations. The robot’s base is fixed in front of a table and remains stationary throughout all experiments to ensure consistency.

For teleoperation, we use Meta Quest 2 controllers to track the 6D target poses of the left and right end effectors. These 6D poses are converted into target joint positions and velocities via a GPU-accelerated model predictive control (MPC) framework from cuRobo~\citep{sundaralingam:arxiv2023}. The resulting commands are then executed by the robot’s low-level joint impedance controller, enabling smooth and responsive teleoperation.

The demonstration task is a bimanual pick-and-place scenario involving two objects: an apple and a bowl, placed randomly on the table at the start of each trial. The task requires the robot to grasp the apple and the bowl with separate arms, place the apple into the bowl while holding it, and finally set the bowl back on the table as if serving. Across trials, only the positions of the apple and bowl are varied, while the objects themselves (a gray apple and bowl), the table surface, and the background remain fixed.

\begin{figure*}[htb!]
 \centering
 \small
 \includegraphics[width=0.162\linewidth]{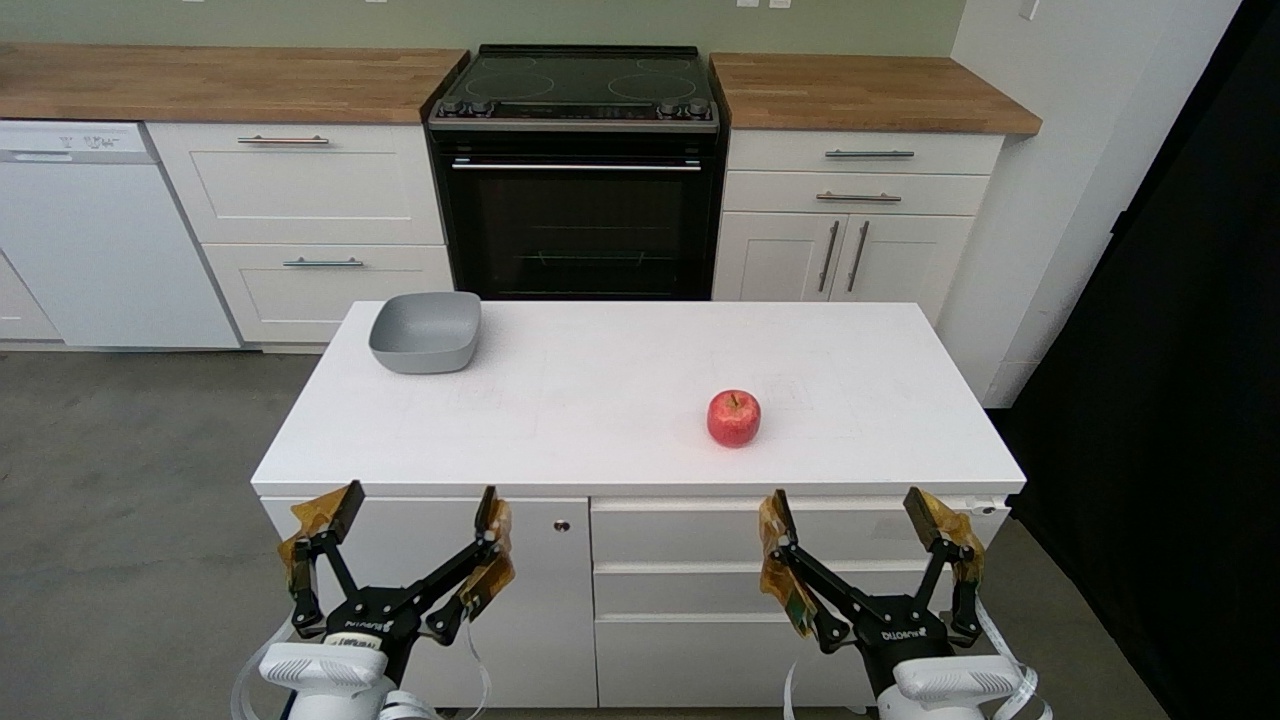}
 \includegraphics[width=0.162\linewidth]{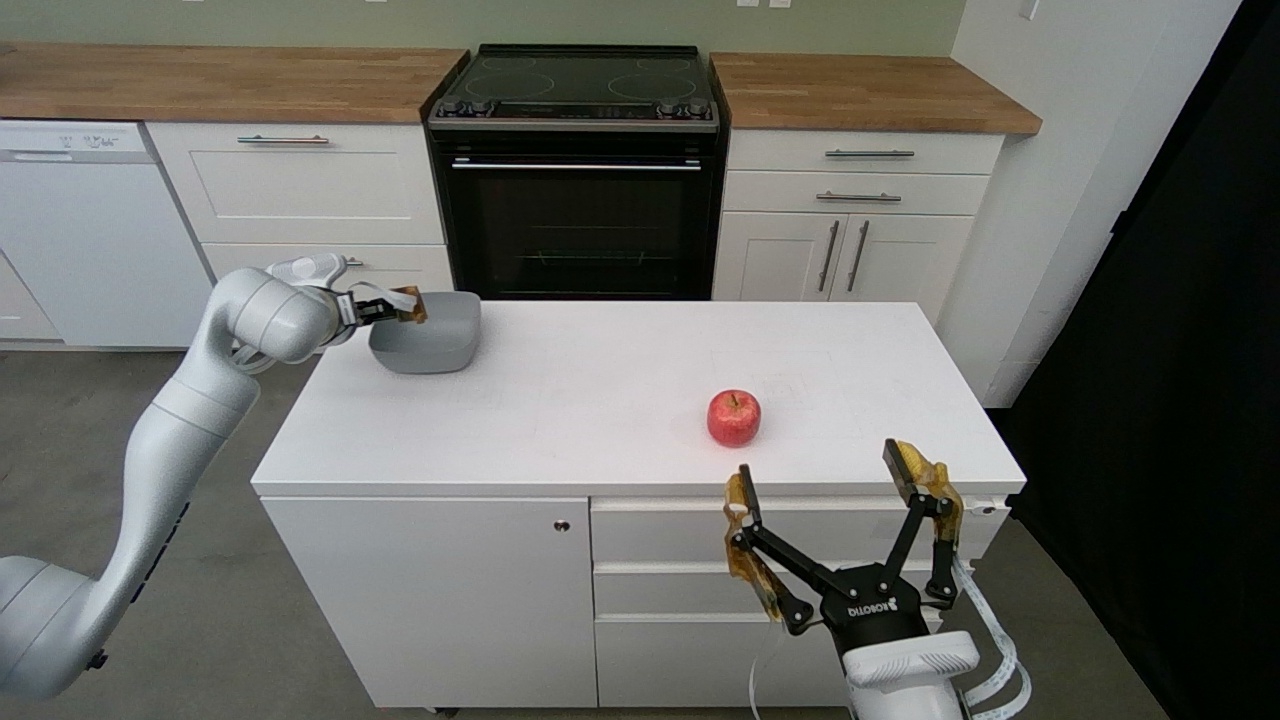}
 \includegraphics[width=0.162\linewidth]{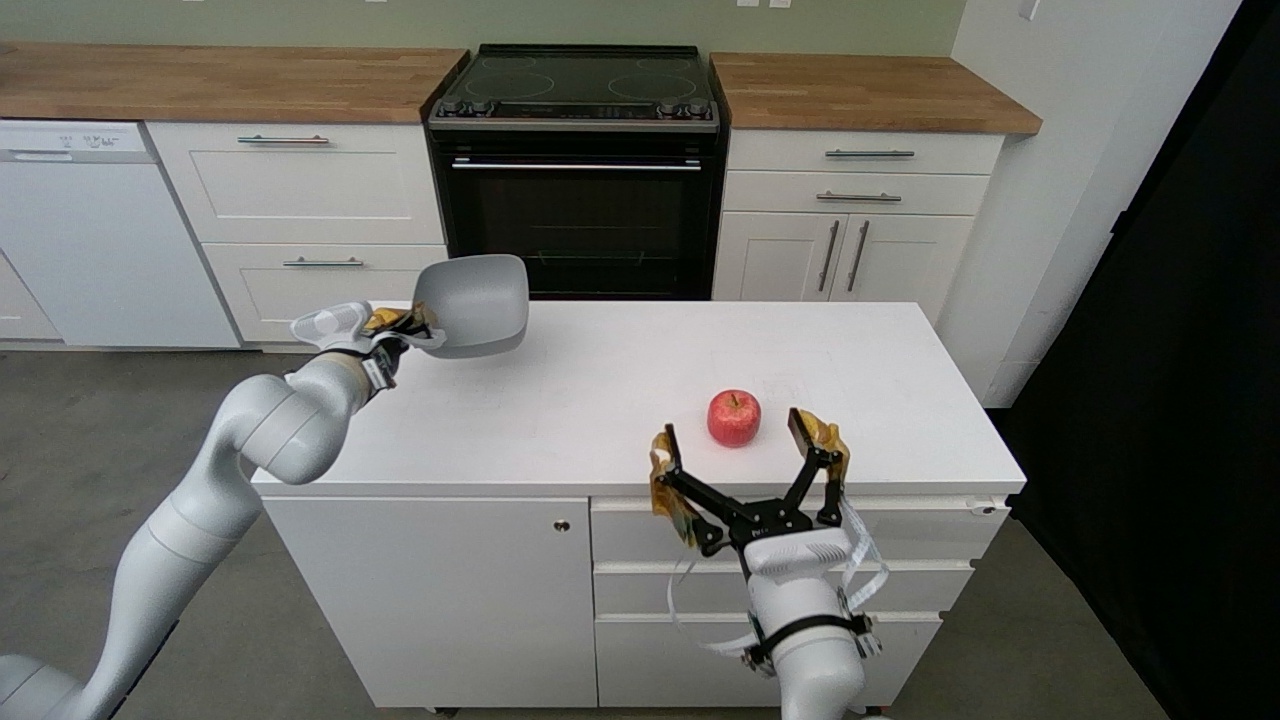}
 \includegraphics[width=0.162\linewidth]{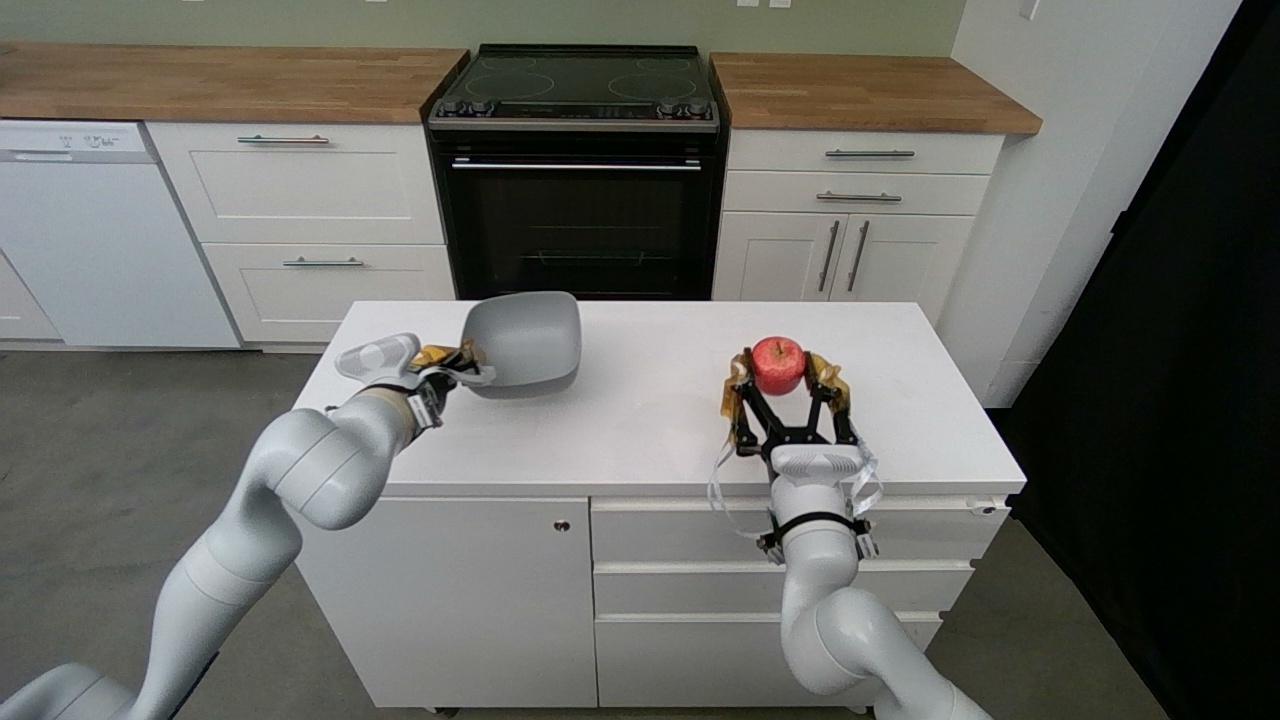}
 \includegraphics[width=0.162\linewidth]{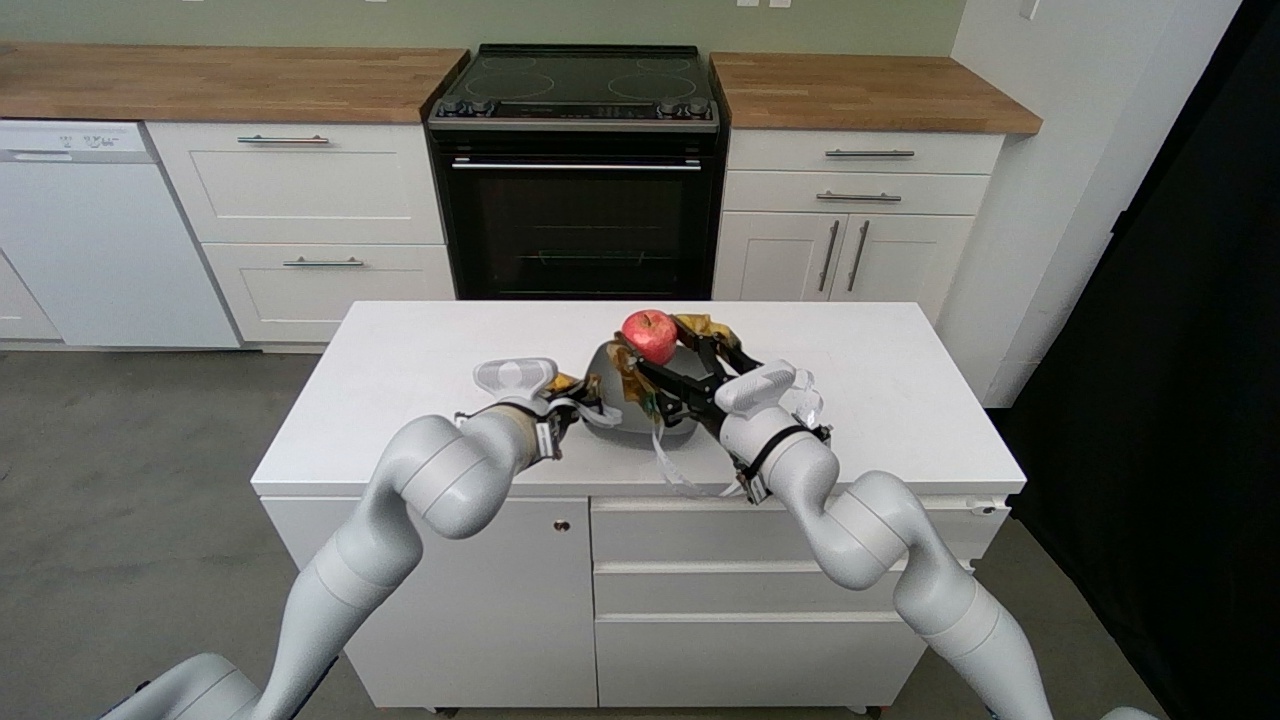}
 \includegraphics[width=0.162\linewidth]{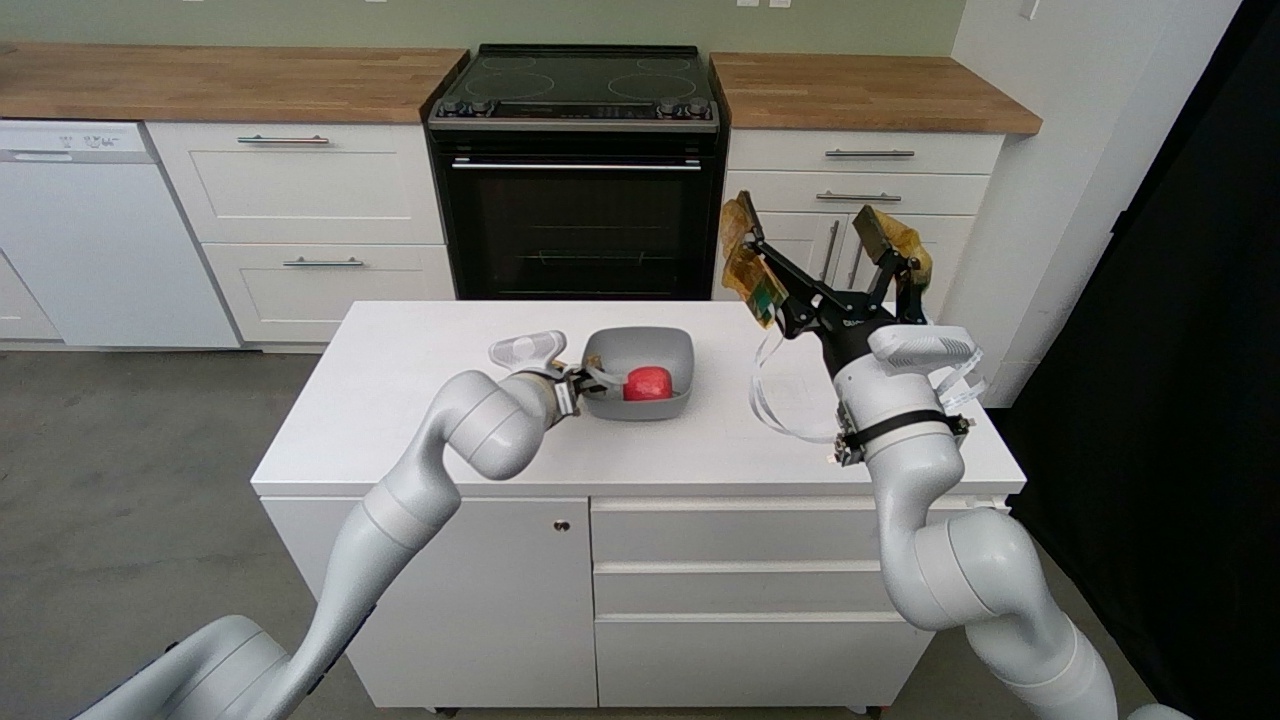}
 \\ \vspace{1mm}
 \includegraphics[width=0.162\linewidth]{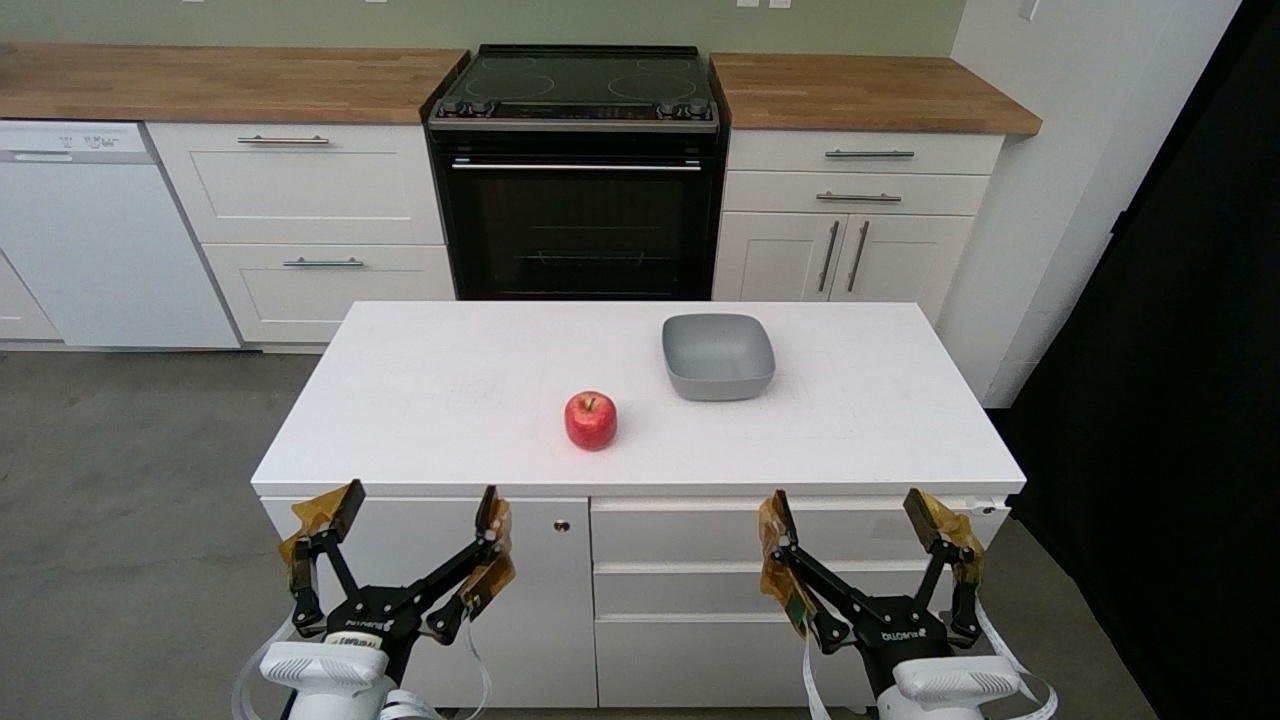}
 \includegraphics[width=0.162\linewidth]{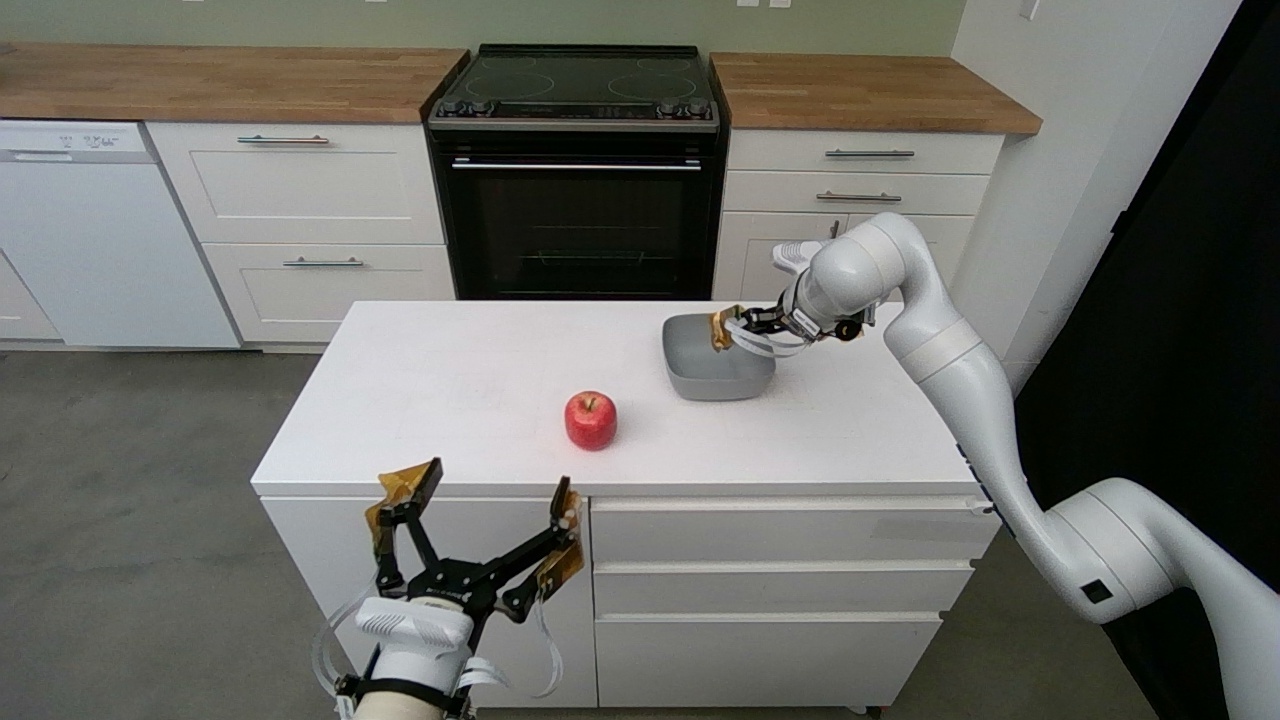}
 \includegraphics[width=0.162\linewidth]{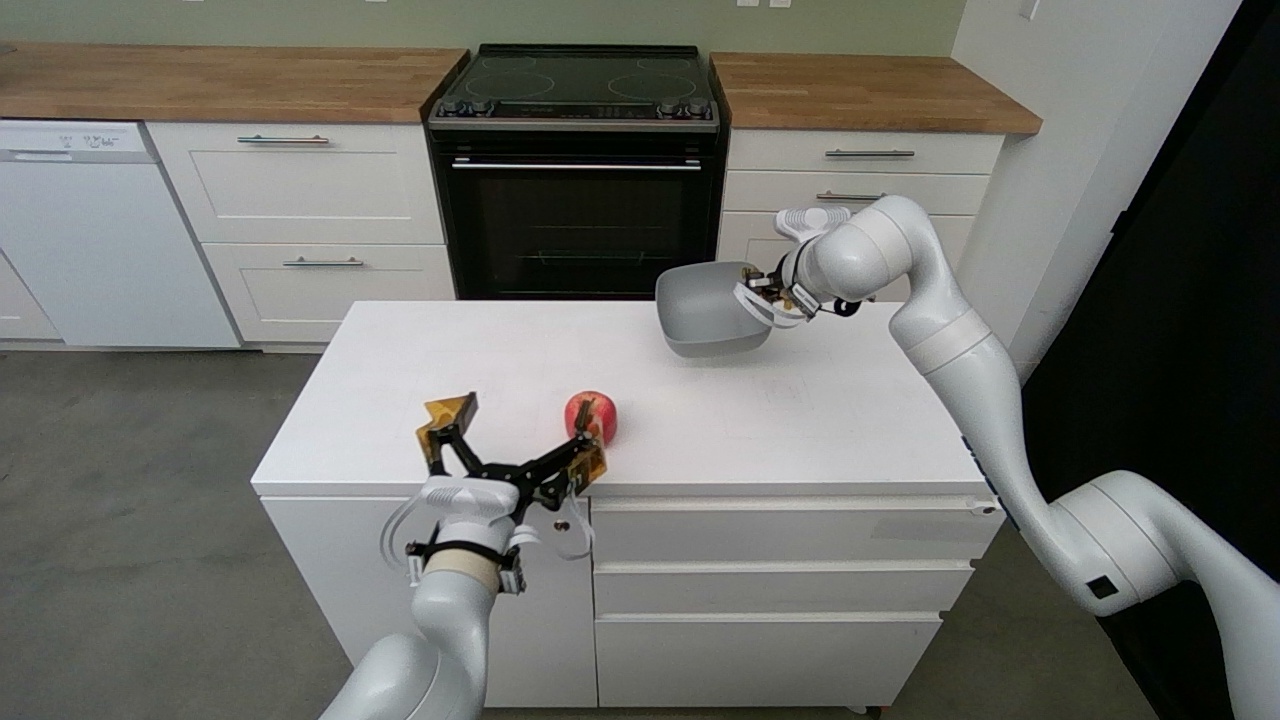}
 \includegraphics[width=0.162\linewidth]{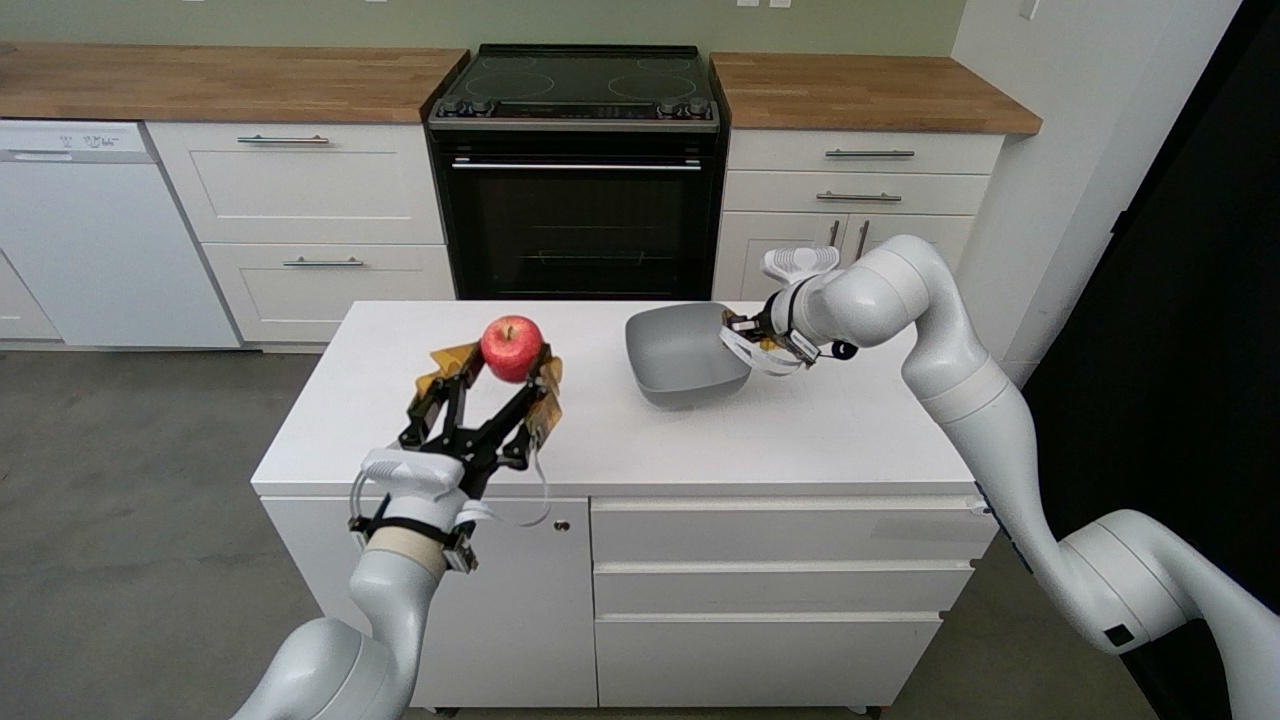}
 \includegraphics[width=0.162\linewidth]{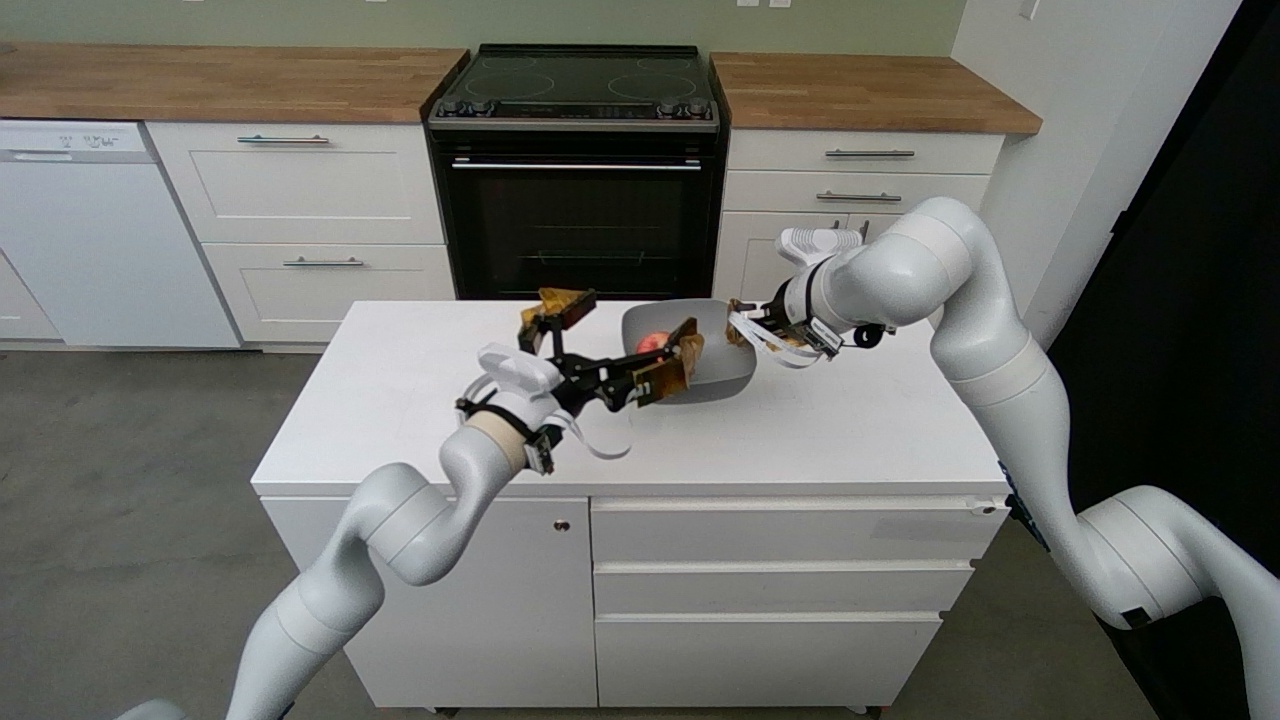}
 \includegraphics[width=0.162\linewidth]{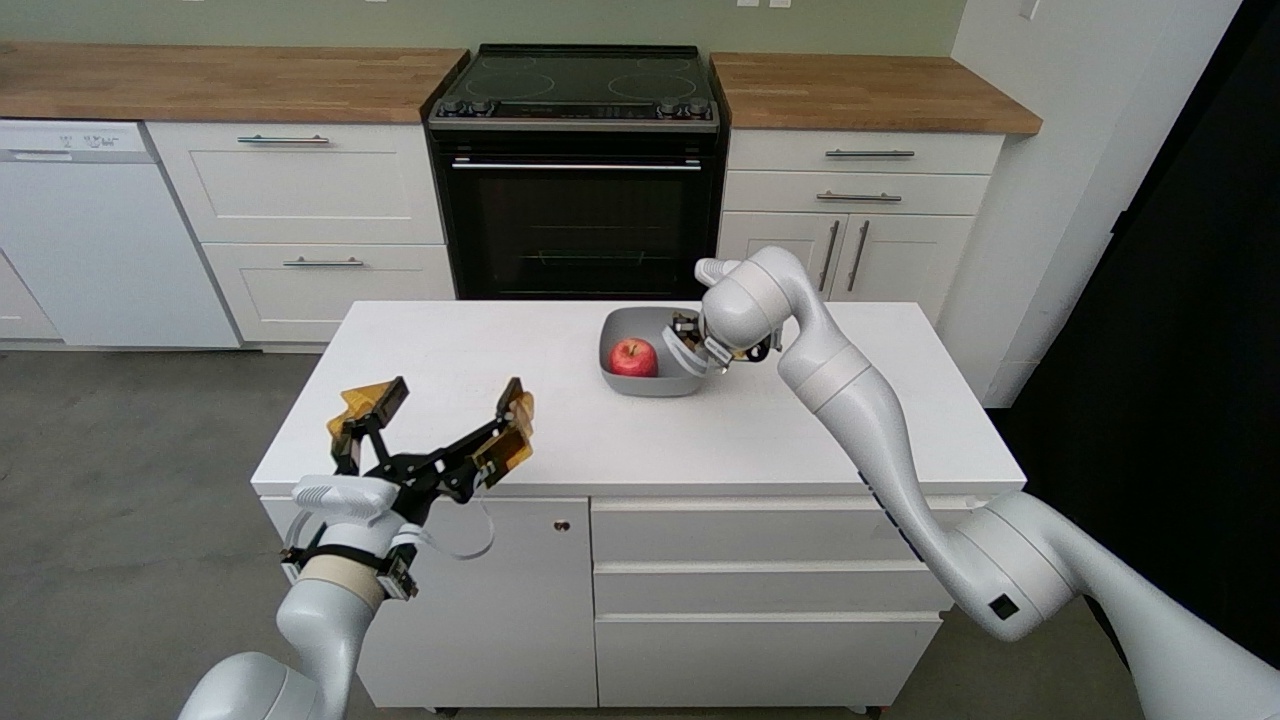}
 \caption{\textbf{Real-Robot Teleoperation Samples.} Two episodes of image observations captured from the egocentric camera during demonstration collection. We keep the object instances and scene fixed and only change the objects' (apple and bowl) poses.}
 \label{fig:real_robot_demo}
\end{figure*}

In total, we collect 100 human teleoperation demonstrations of this task. Using these demonstrations, we train a UNet-based Diffusion Policy~\citep{ren:iclr2025,chi:rss2023}, which takes the single-image observation (processed via a small ViT) with gripper joint state and predicts chunks of actions consisting of the target end-effector poses and gripper commands for both arms. Each chunk spans a horizon of 8 timesteps sampled at 10 FPS. Examples of egocentric image observations recorded during the demonstrations are shown in~\cref{fig:real_robot_demo}, illustrating the consistency of the setup and the controlled variability introduced by object placement.

\subsubsection{Data Augmentation Strategy} 

We use [Cosmos-Transfer2.5-2B] to generate diverse and realistic visually augmented videos that expand the training set, improving the diffusion policy’s ability to generalize to test-time variations. Our augmentation strategy applies global edge control across the entire image, while restricting blur control to robot pixels. To isolate the robot in each frame, we combine Grounding DINO and SAMv2~\citep{liu2023grounding,ravi2024sam,ren2024grounded} for detection and pixel-level segmentation. We set the edge threshold to medium, the blur threshold to very low, and the classifier-free guidance scale to 3, while keeping all other parameters at their default values.

We design a prompt template that diversifies the appearance of synthetic videos while preserving the underlying scene and task structure. The process begins by providing an example video to a VLM, which generates a detailed caption of the scene. We then iteratively refine this caption by prompting [Cosmos-Transfer2.5-2B] and checking whether the generated video faithfully resembles the original.

From the refined caption, we construct a formatted prompt that marks which components can vary. An LLM is then used to generate candidate variations for these components. Below is the full formatted prompt:

\textit{The scene depicts a bright, modern kitchen with plenty of ambient light. From a first-person perspective, a robot faces [TABLE]. On the table rest [COLOR\_APPLE] apple and [COLOR\_BOWL] bowl. [SENTENCE\_LIGHT] In the background are a black cooking range featuring a black stovetop, wooden countertops, and cabinetry with white doors and drawers, including a built-in white dishwasher on the left. [SENTENCE\_BACKGROUND] A wide black curtain hangs vertically on the right side, covering a large portion of the space. As the video progresses, the robot picks up the apple, then the bowl, places the apple into the bowl, and sets the bowl down on the table.}

\begin{figure*}[htb!]
  \centering
  \small
  \renewcommand{\arraystretch}{1.2} 
  \setlength{\tabcolsep}{1pt}       

  \begin{tabular}{c ccccccc}
    \multirow{1}{*}[3.2em]{\rotatebox[origin=c]{90}{Baseline}} &
    \includegraphics[width=0.16\linewidth]{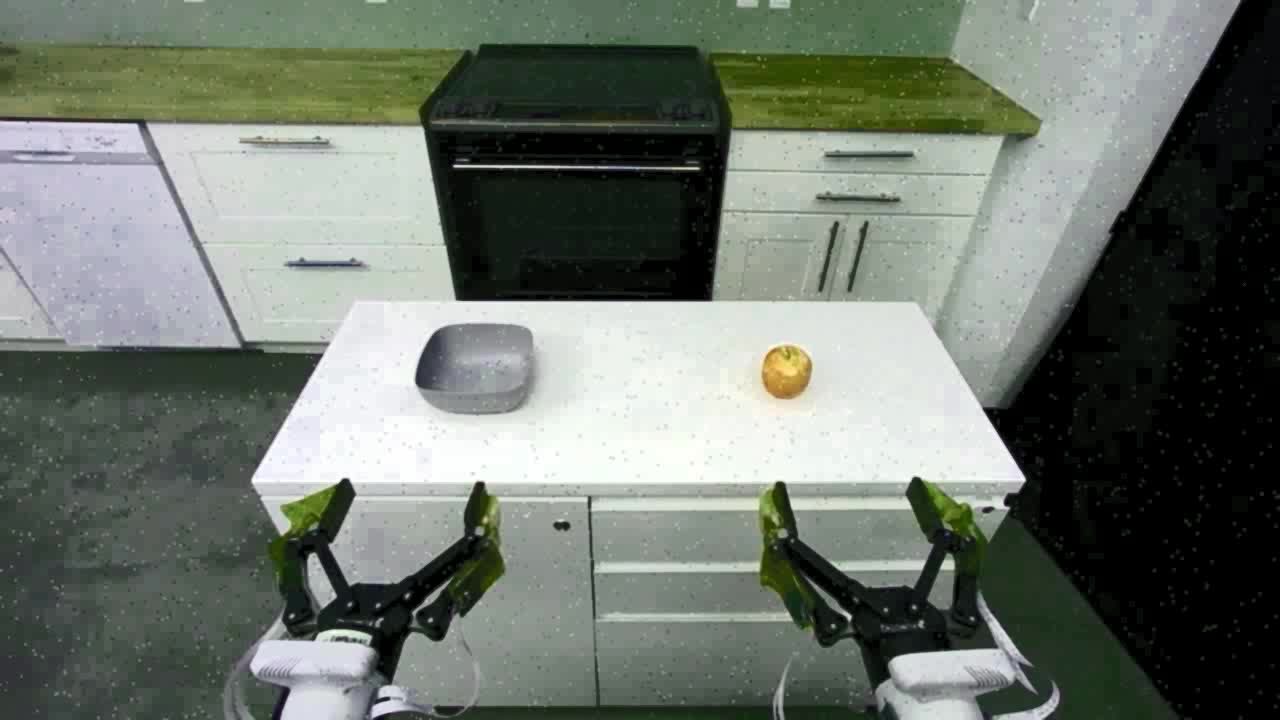} &
    \includegraphics[width=0.16\linewidth]{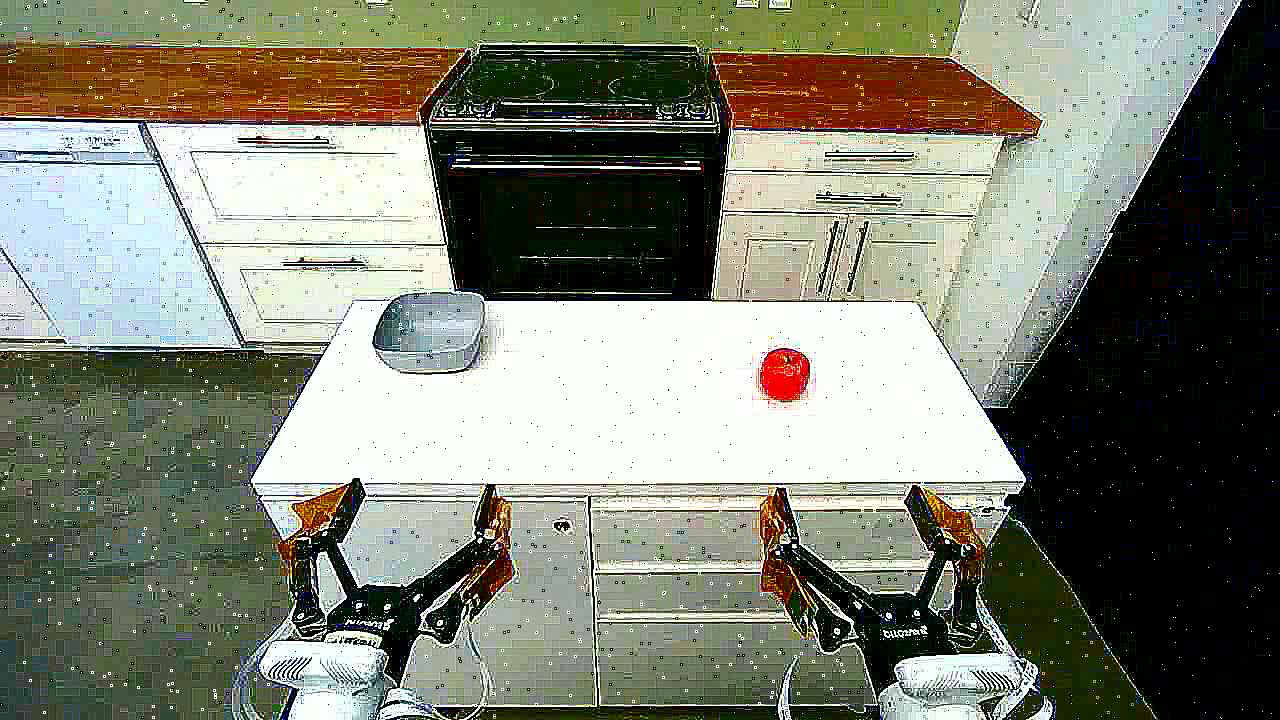} &
    \includegraphics[width=0.16\linewidth]{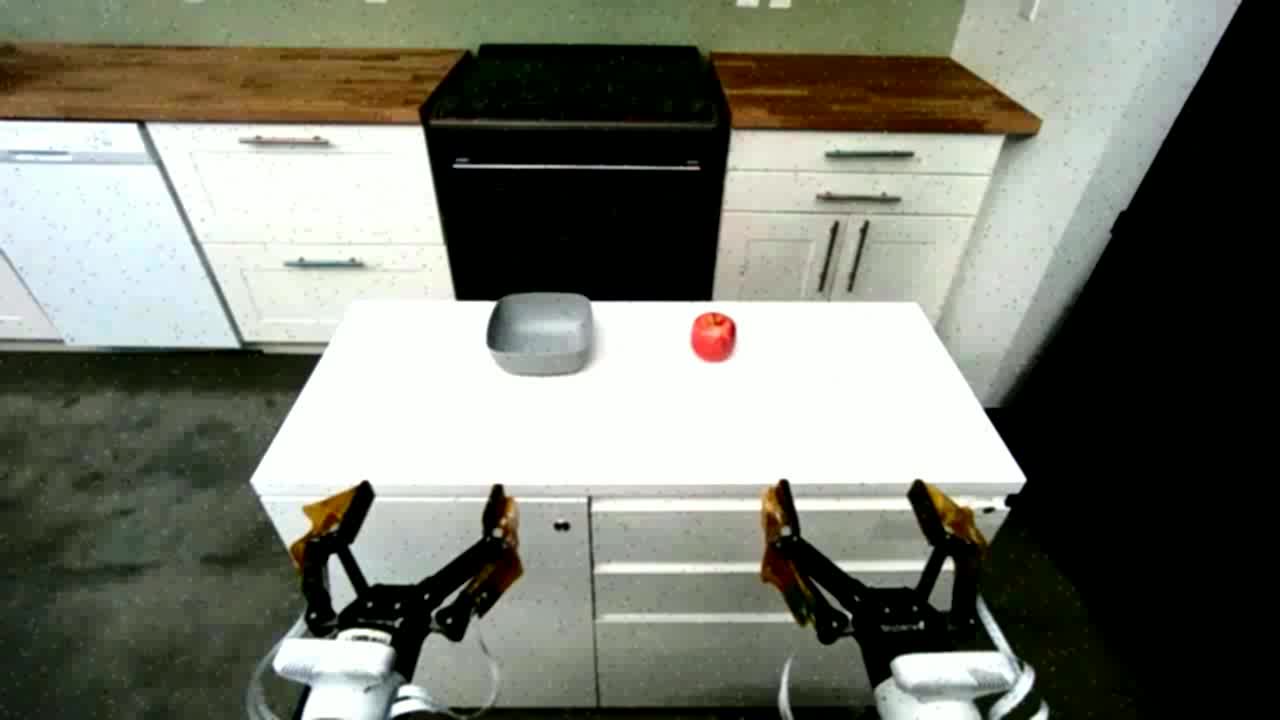} &
    \includegraphics[width=0.16\linewidth]{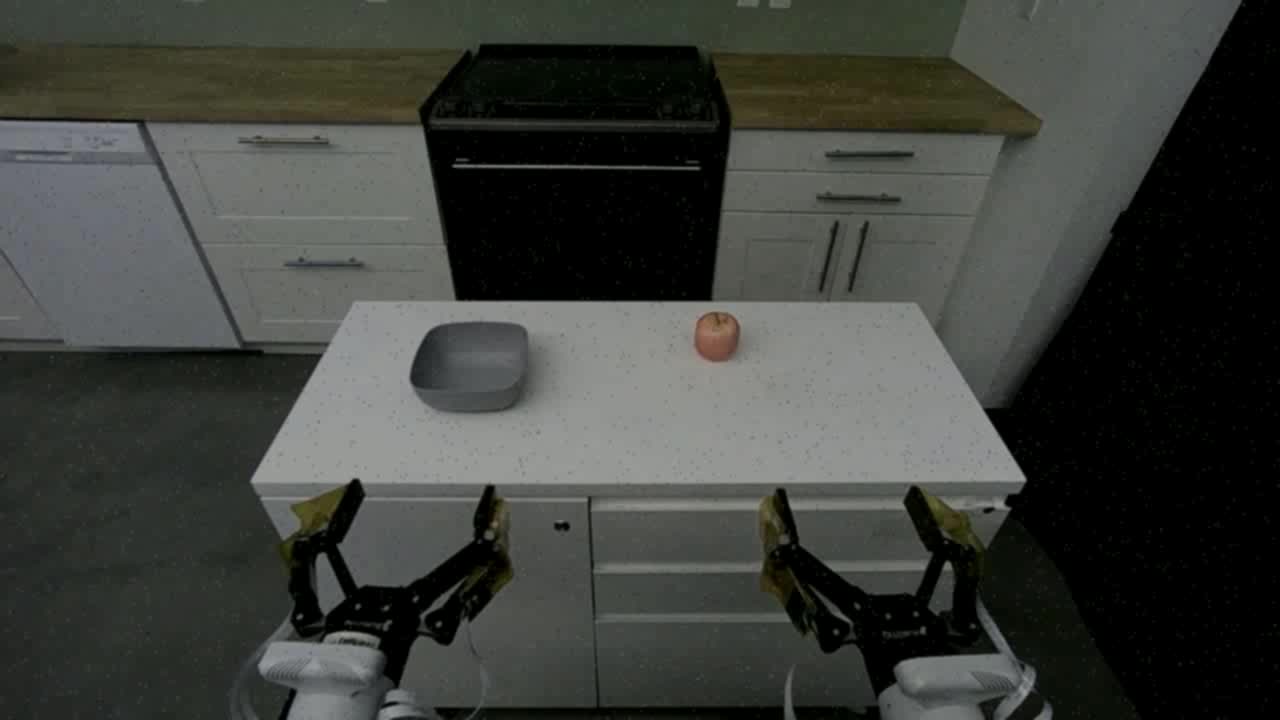} &
    \includegraphics[width=0.16\linewidth]{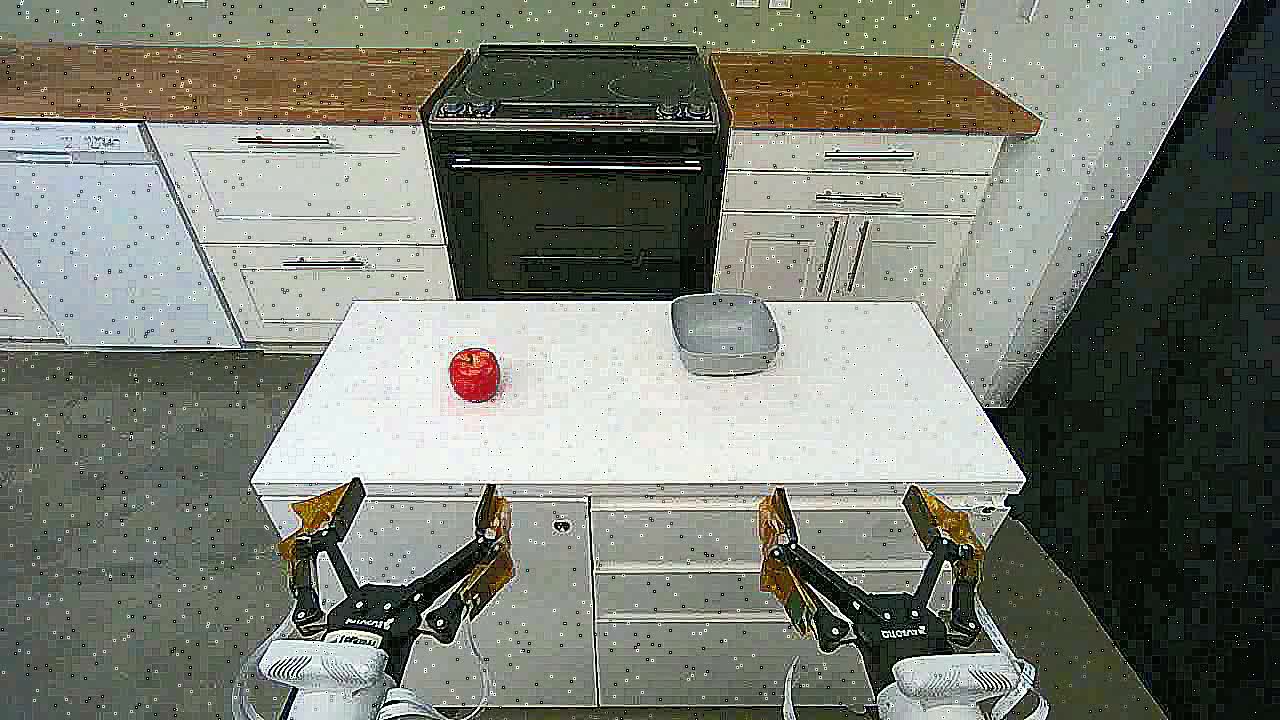} &
    \includegraphics[width=0.16\linewidth]{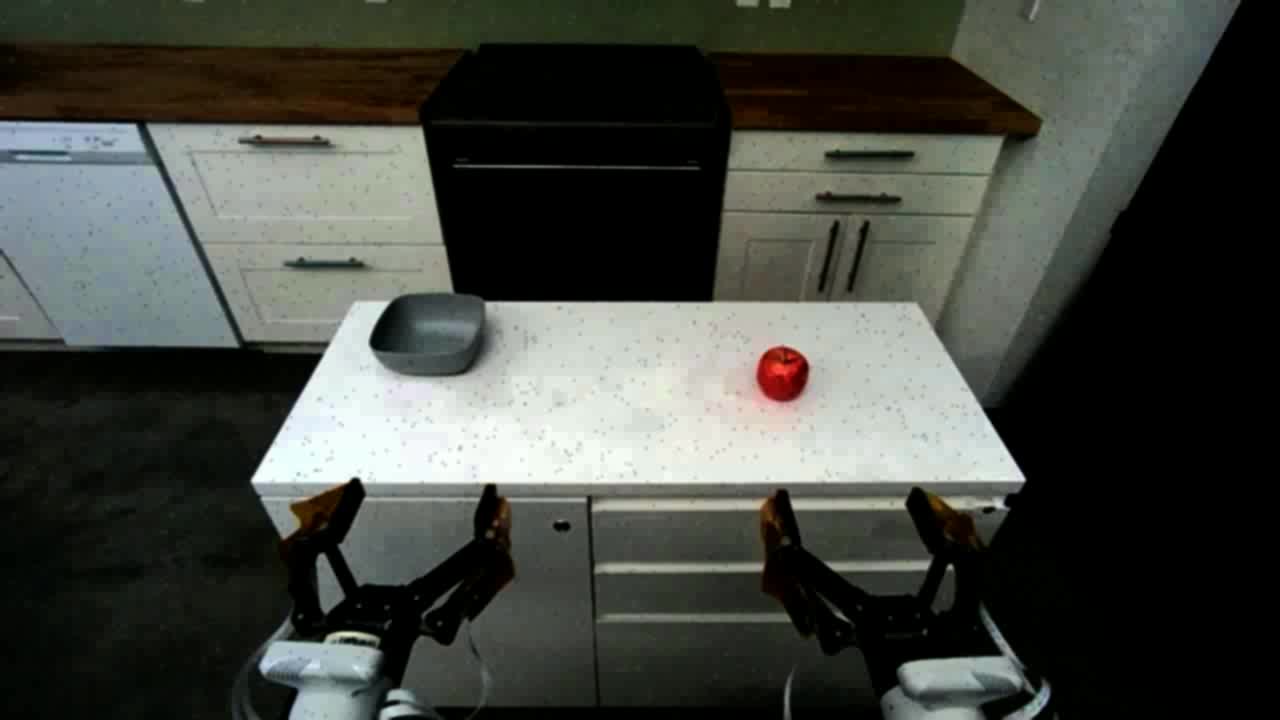} \\

    \cmidrule(lr){2-7}
    
    \multirow{2}{*}[3.2em]{\rotatebox[origin=c]{90}{Cosmos-Transfer2.5}} &
    \includegraphics[width=0.16\linewidth]{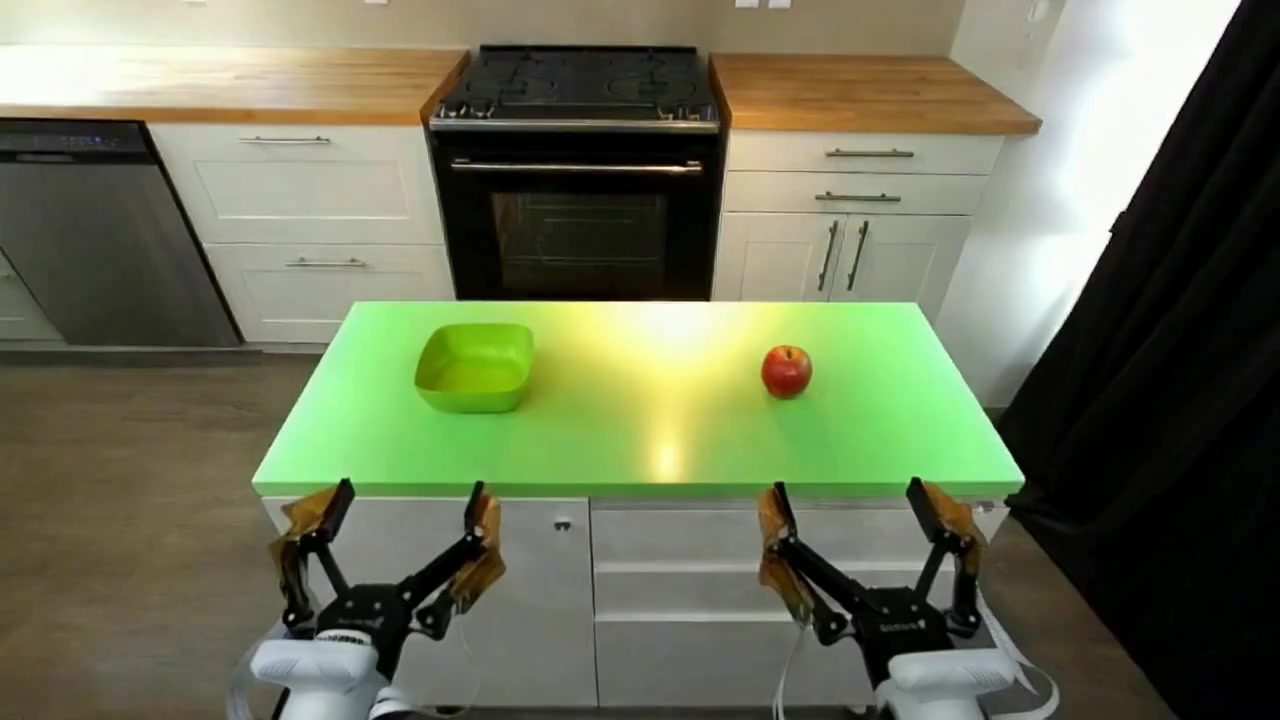} &
    \includegraphics[width=0.16\linewidth]{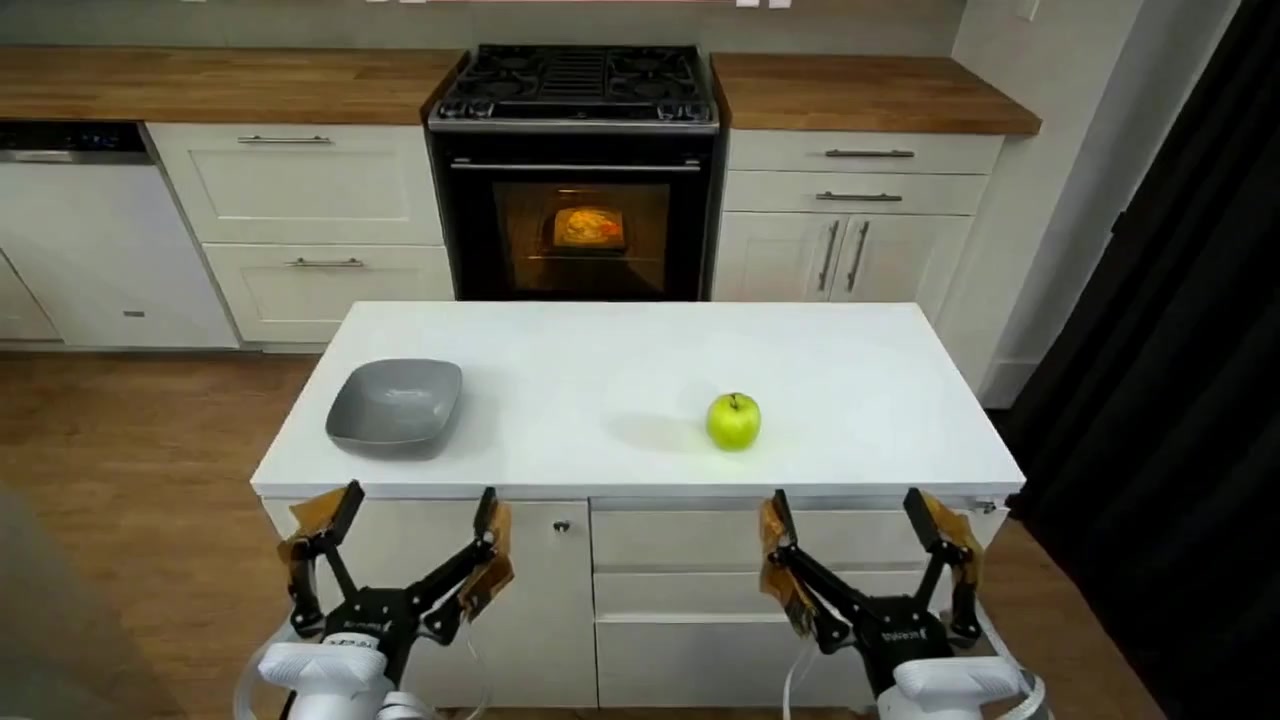} &
    \includegraphics[width=0.16\linewidth]{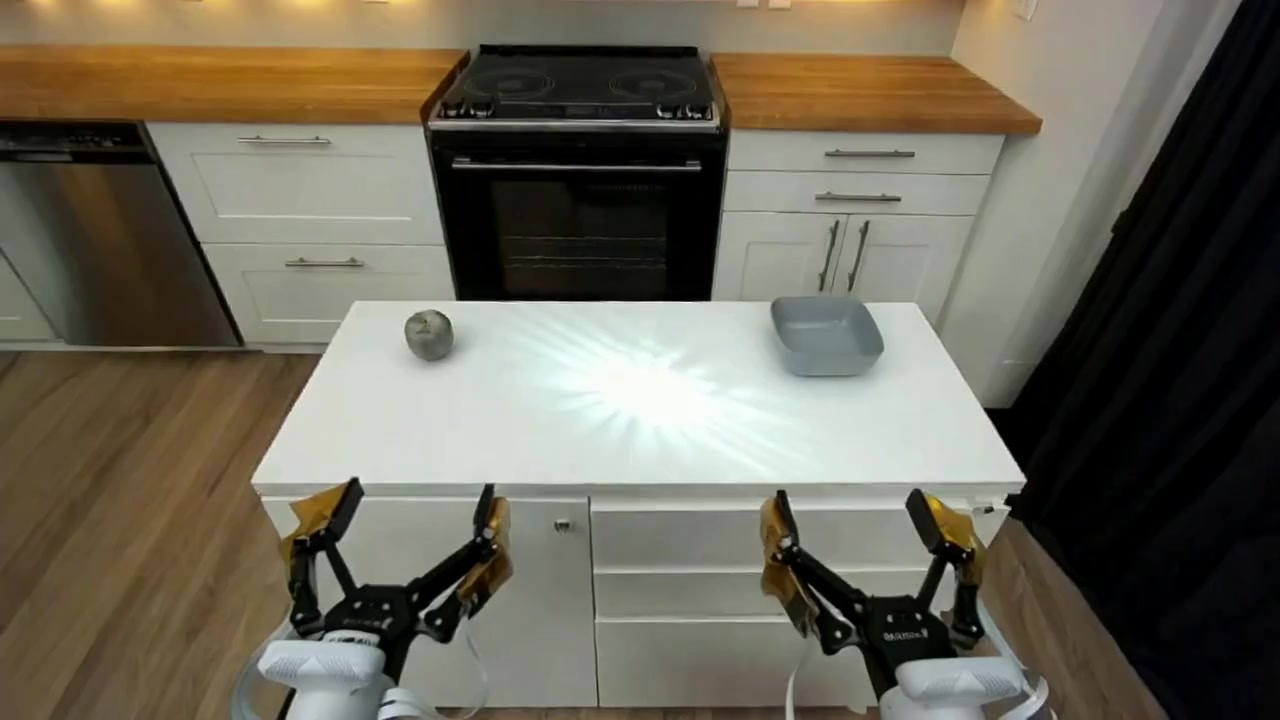} &
    \includegraphics[width=0.16\linewidth]{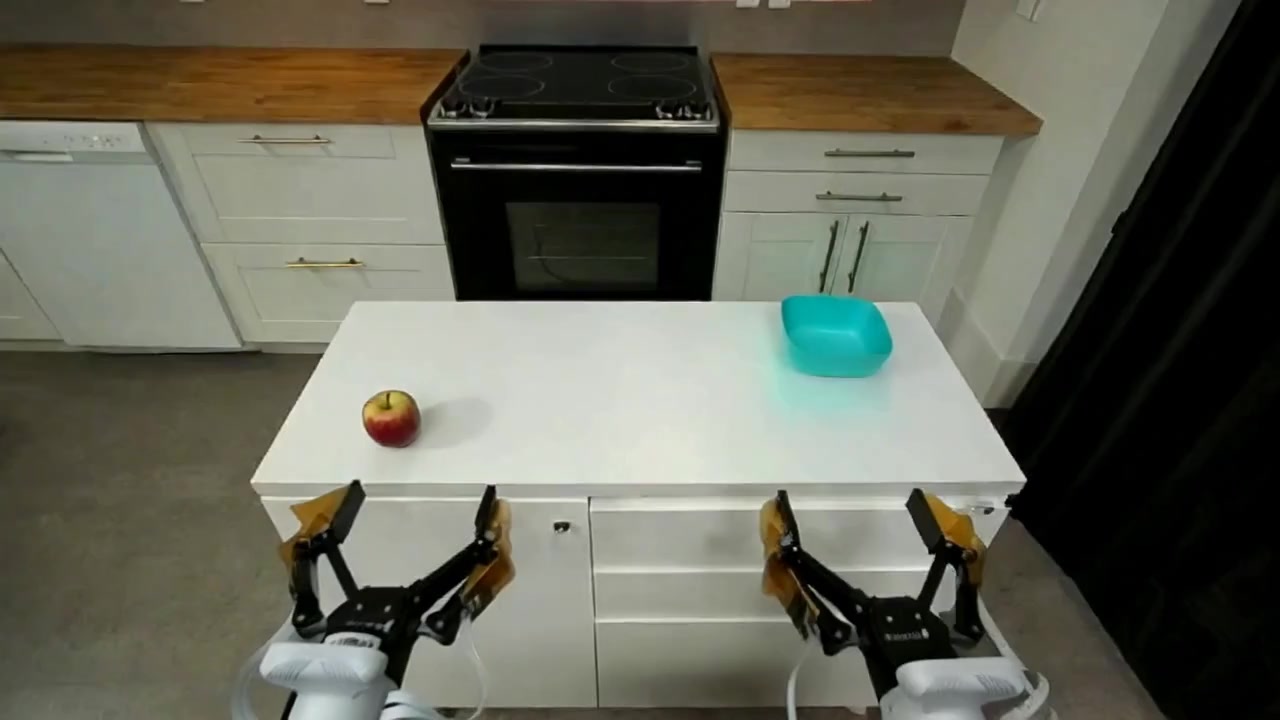} &
    \includegraphics[width=0.16\linewidth]{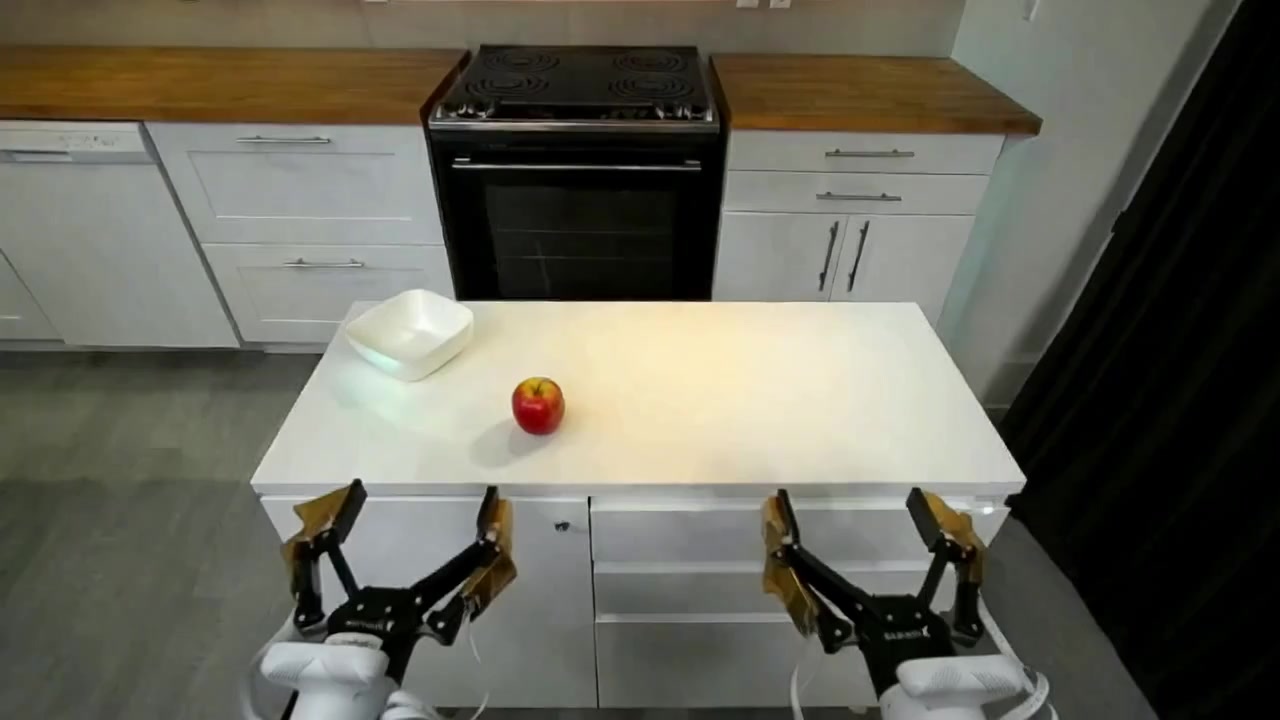} &
    \includegraphics[width=0.16\linewidth]{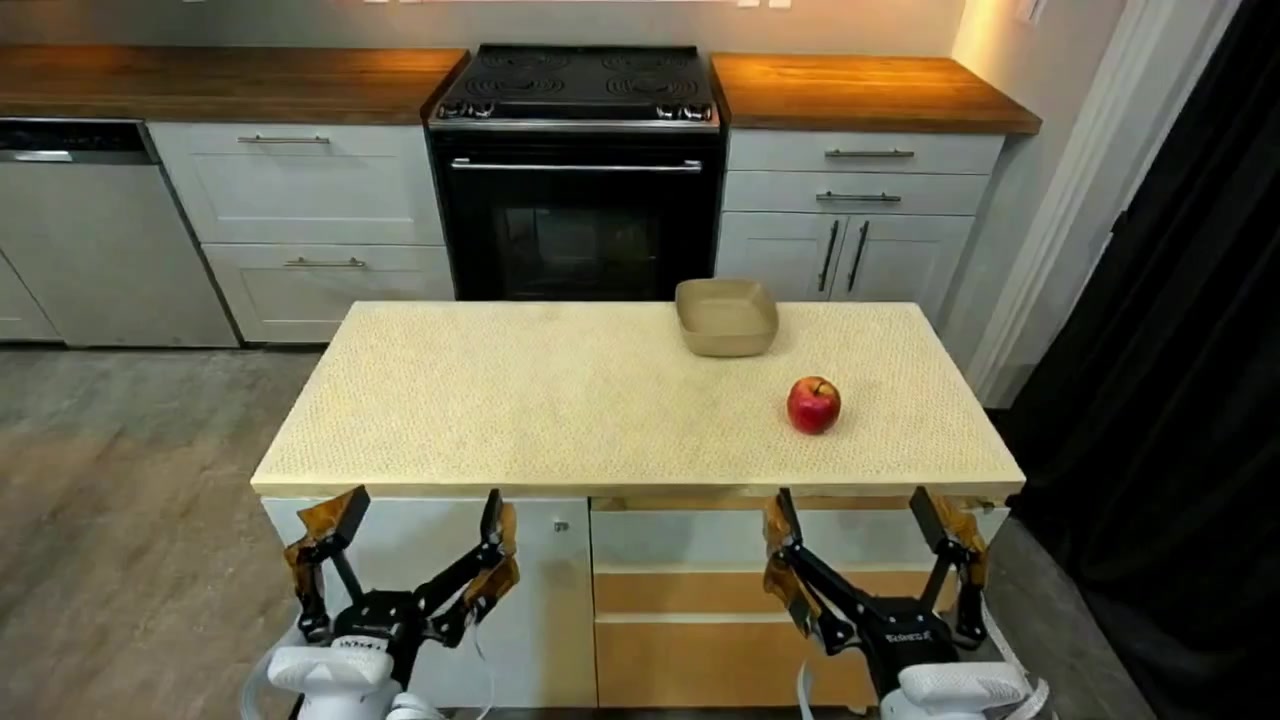} \\

    & \includegraphics[width=0.16\linewidth]{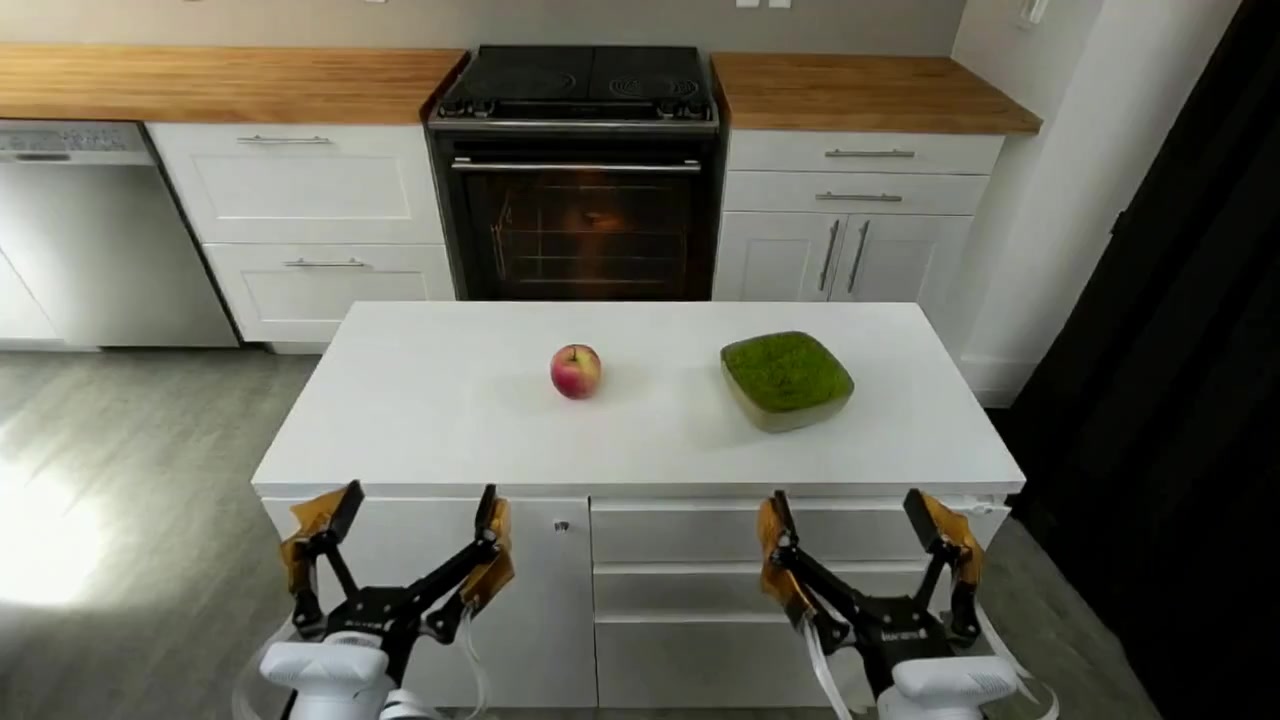} &
    \includegraphics[width=0.16\linewidth]{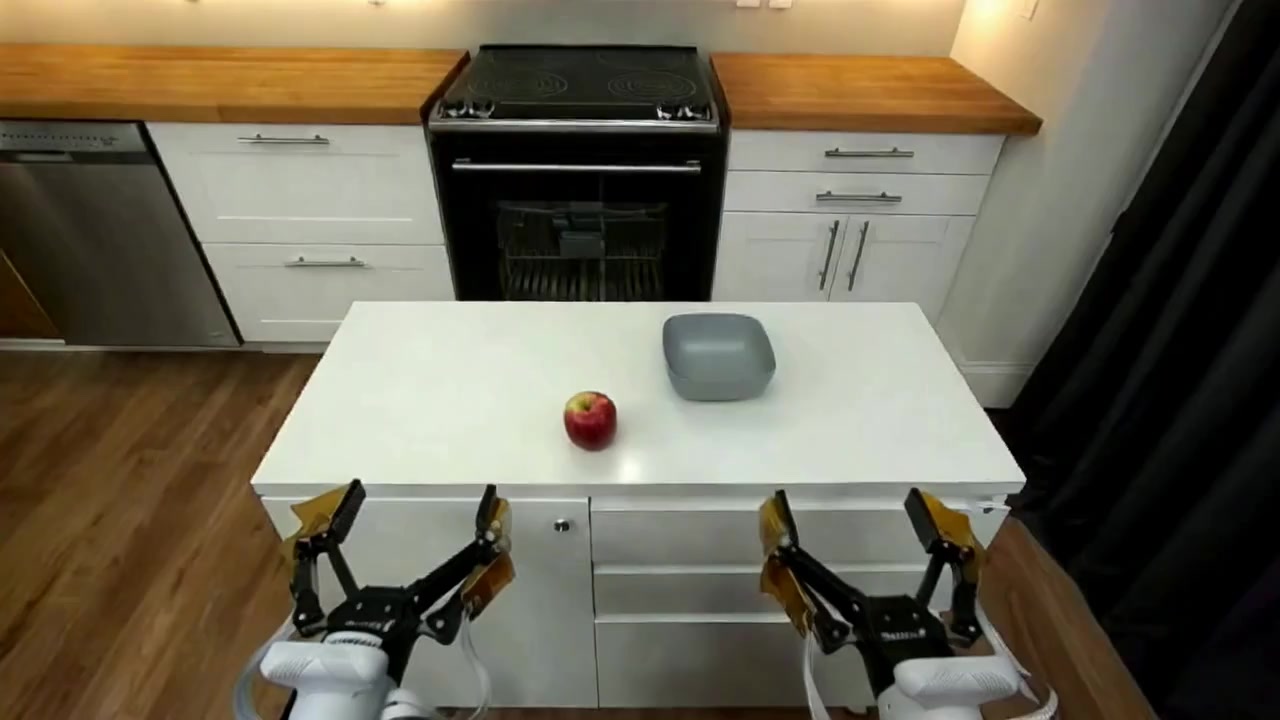} &
    \includegraphics[width=0.16\linewidth]{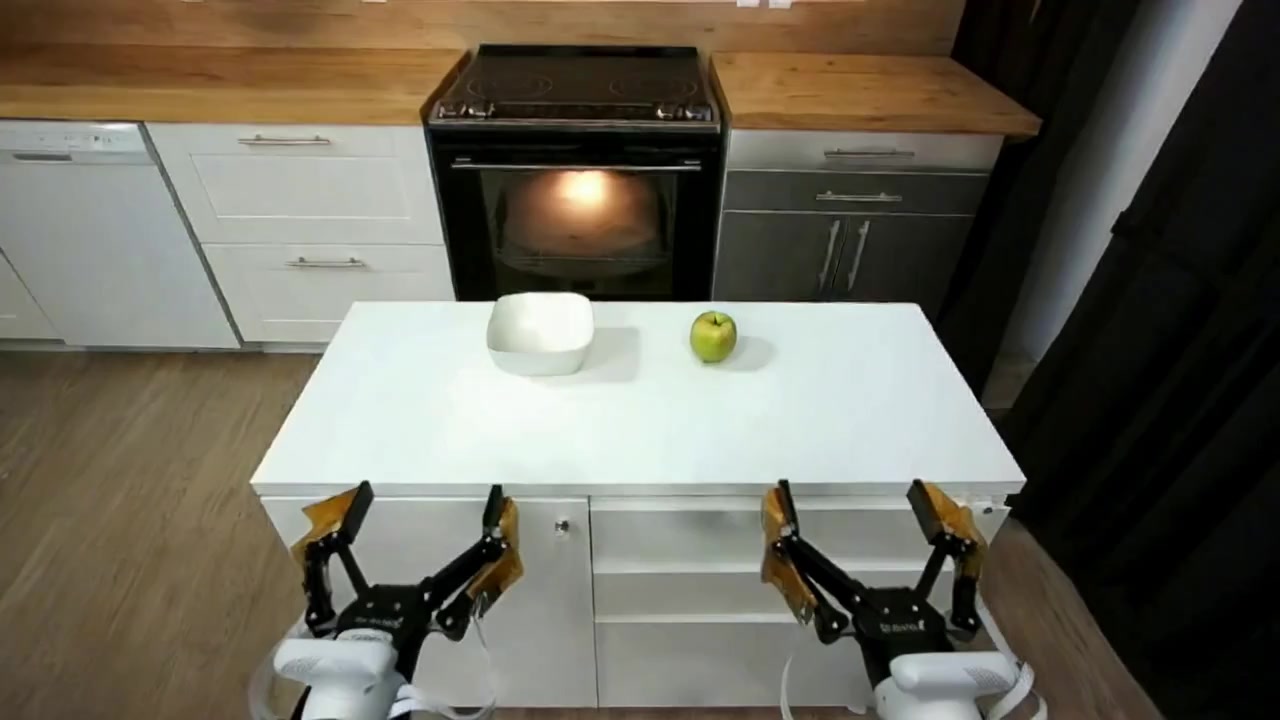} &
    \includegraphics[width=0.16\linewidth]{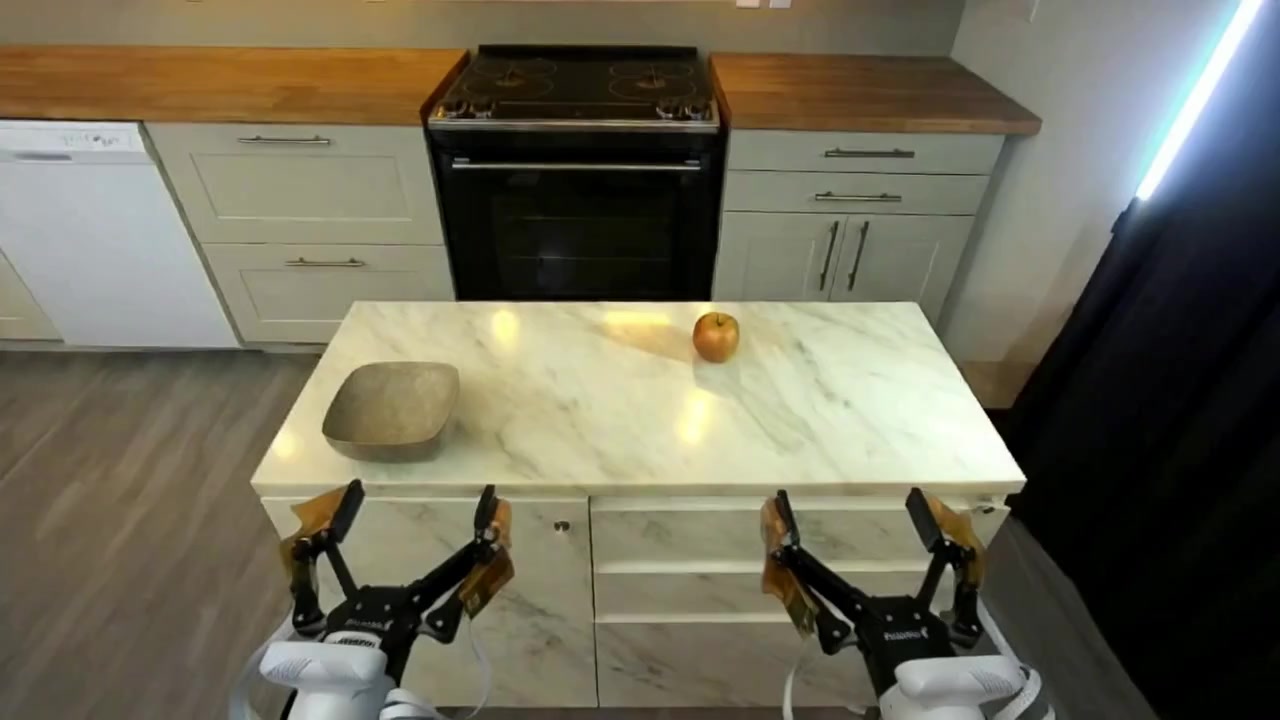} &
    \includegraphics[width=0.16\linewidth]{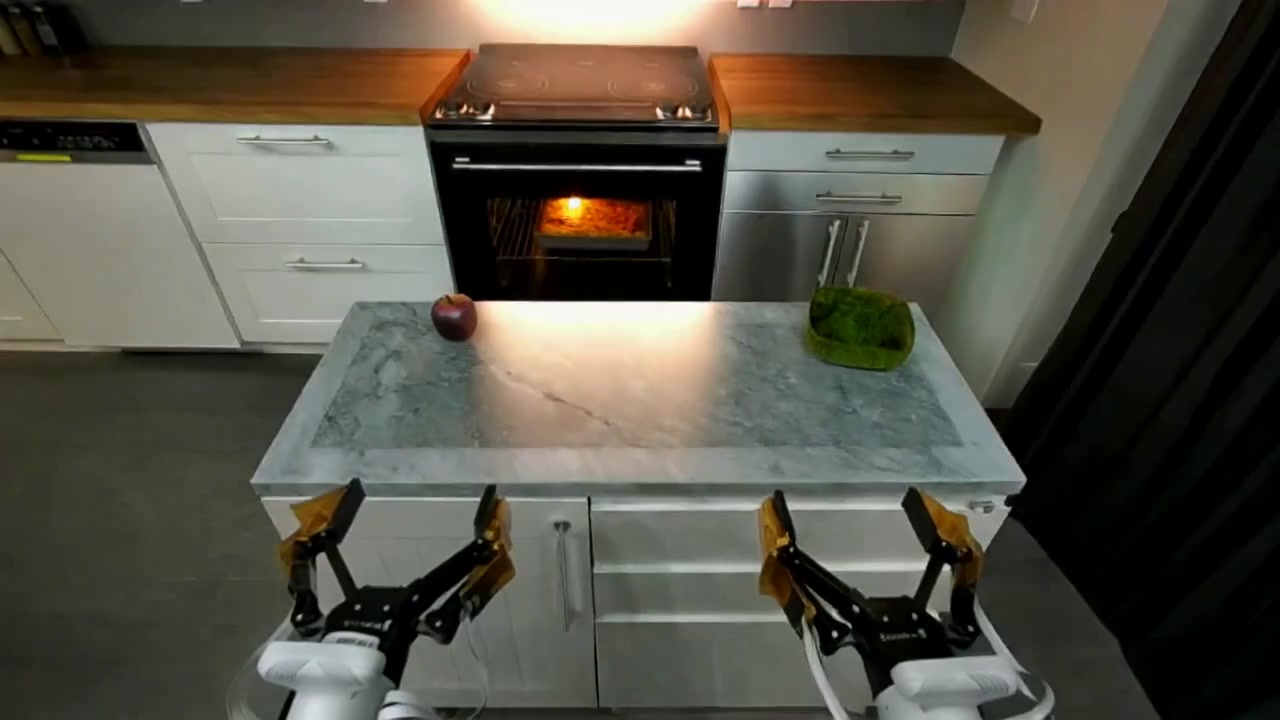} &
    \includegraphics[width=0.16\linewidth]{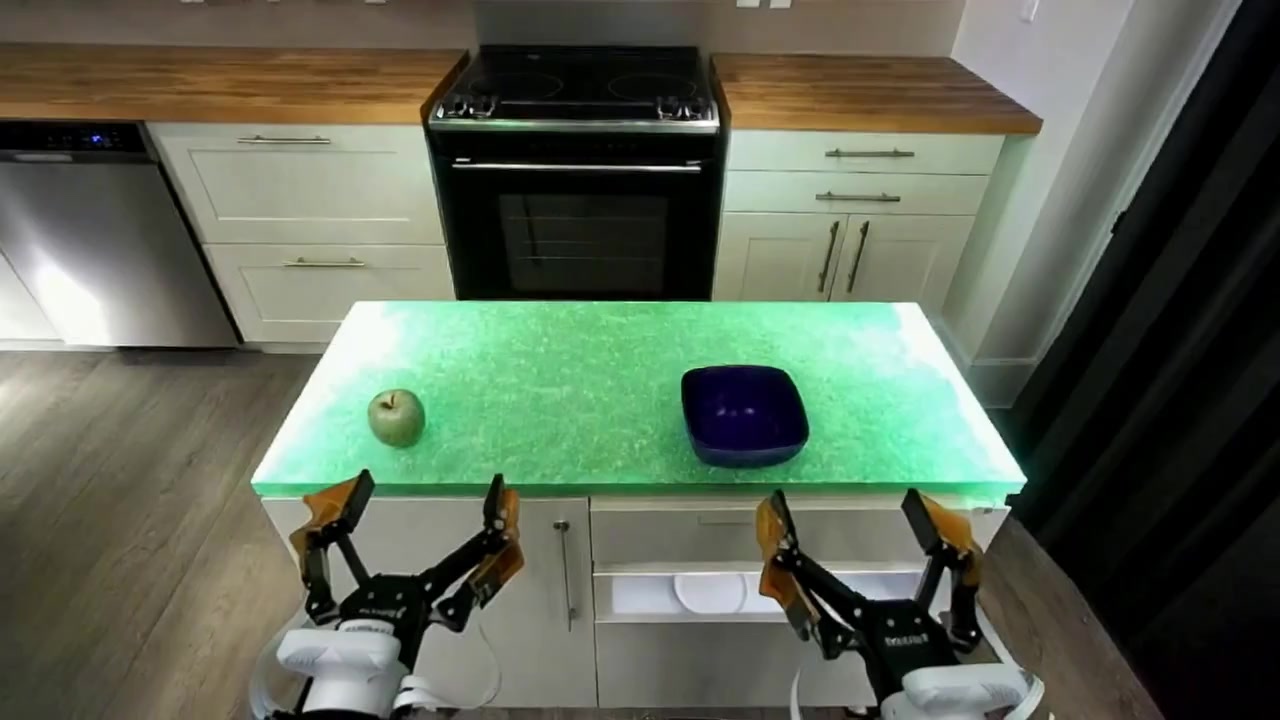} \\
  \end{tabular}

  \caption{\textbf{Real-Robot Data Augmentation Gallery.} We show the baseline (top row) and [Cosmos-Transfer2.5-2B] (bottom two rows) data augmentation samples.}
  \label{fig:real_robot_gallery}
\end{figure*}

\begin{figure*}[t!]
  \centering
  \small
  \renewcommand{\arraystretch}{1.2} 
  \setlength{\tabcolsep}{1pt}       

  \begin{tabular}{c cccccc}
    \multirow{1}{*}[2.1em]{\rotatebox[origin=c]{90}{\scriptsize Base}} &
    \includegraphics[width=0.16\linewidth]{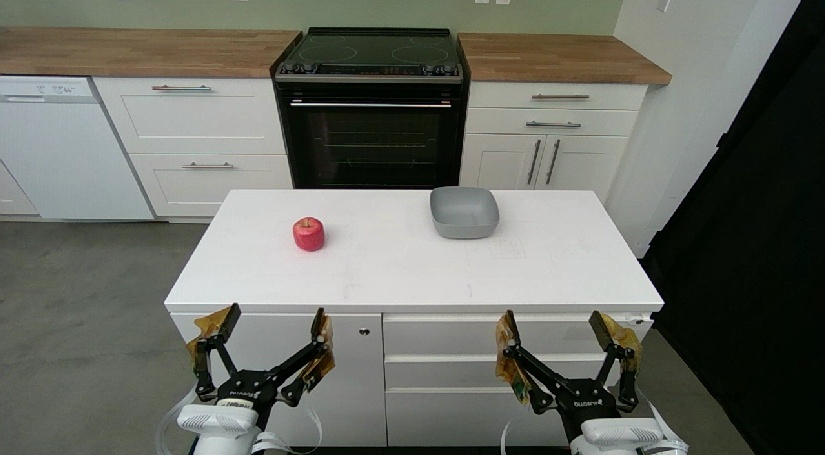} &
    \includegraphics[width=0.16\linewidth]{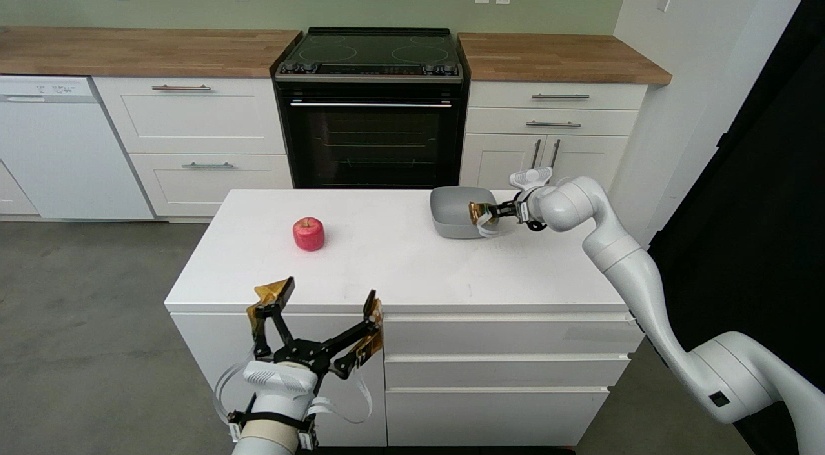} &
    \includegraphics[width=0.16\linewidth]{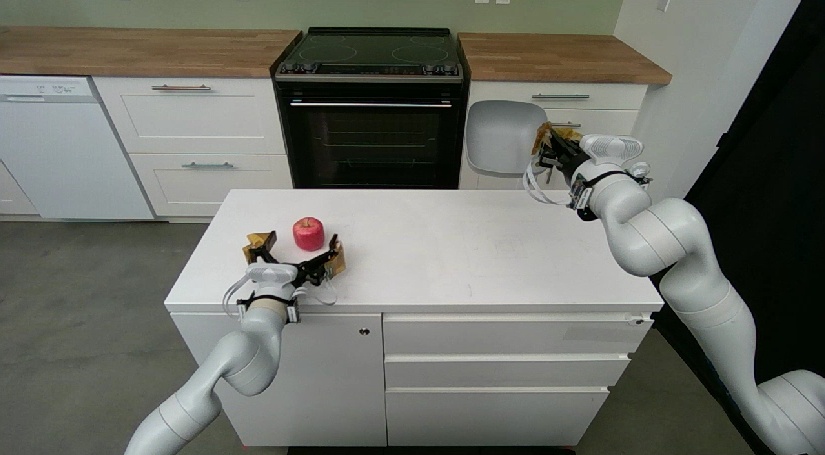} &
    \includegraphics[width=0.16\linewidth]{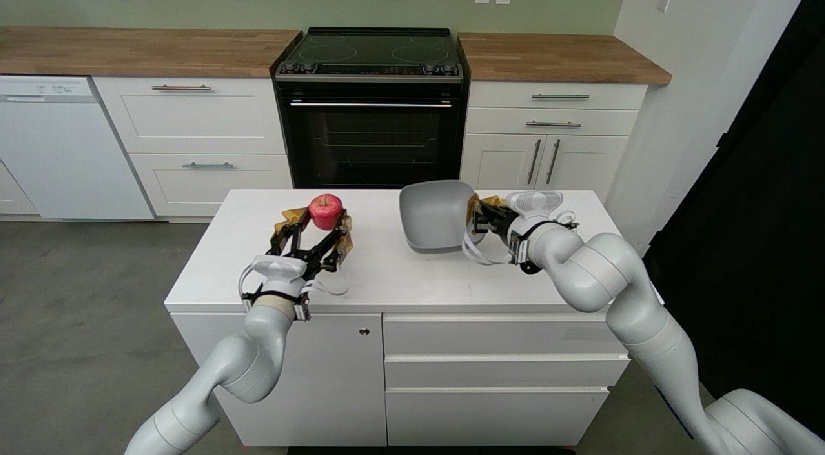} &
    \includegraphics[width=0.16\linewidth]{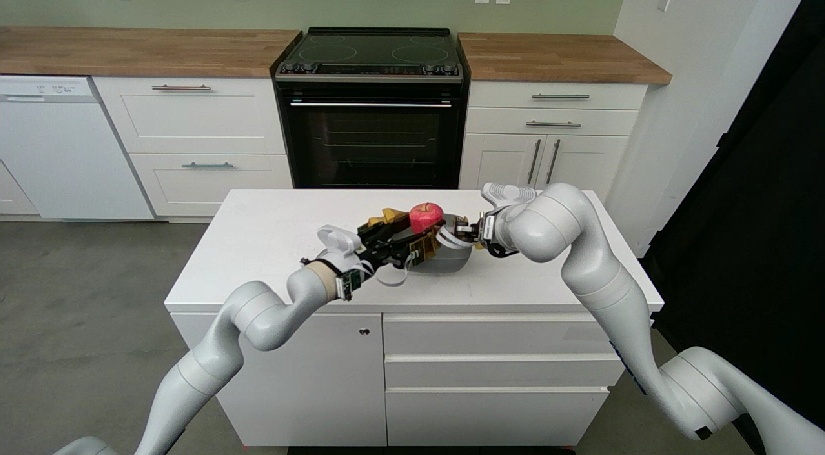} &
    \includegraphics[width=0.16\linewidth]{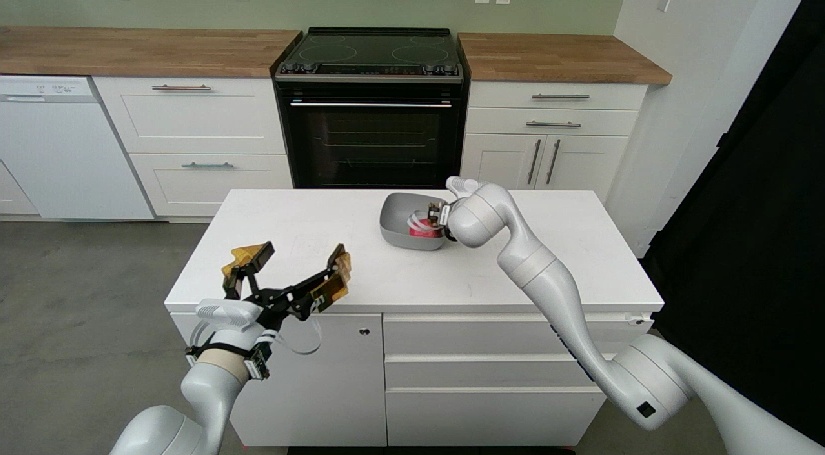} \\

    \multirow{1}{*}[3.3em]{\rotatebox[origin=c]{90}{\scriptsize Mangosteen}} &
    \includegraphics[width=0.16\linewidth]{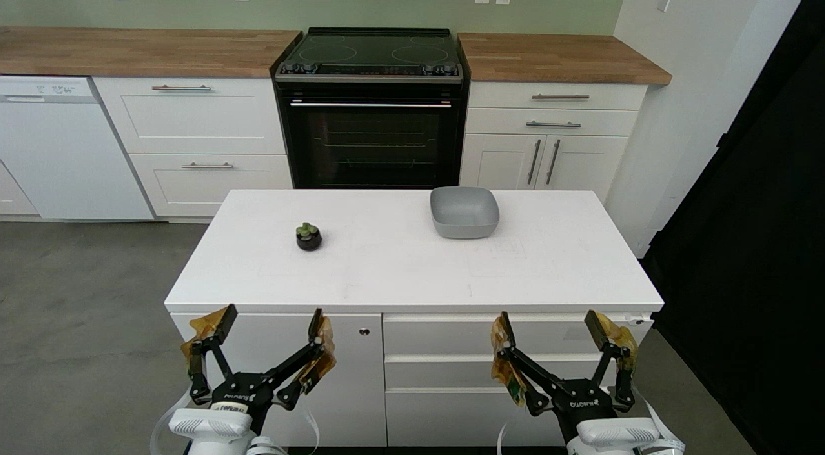} &
    \includegraphics[width=0.16\linewidth]{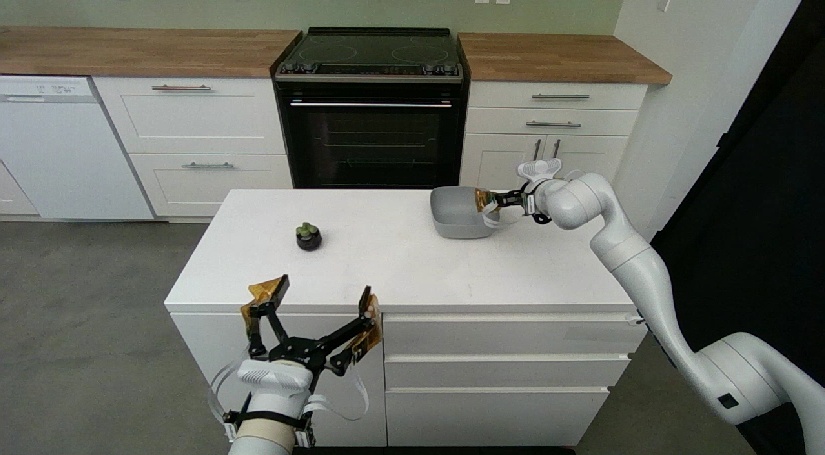} &
    \includegraphics[width=0.16\linewidth]{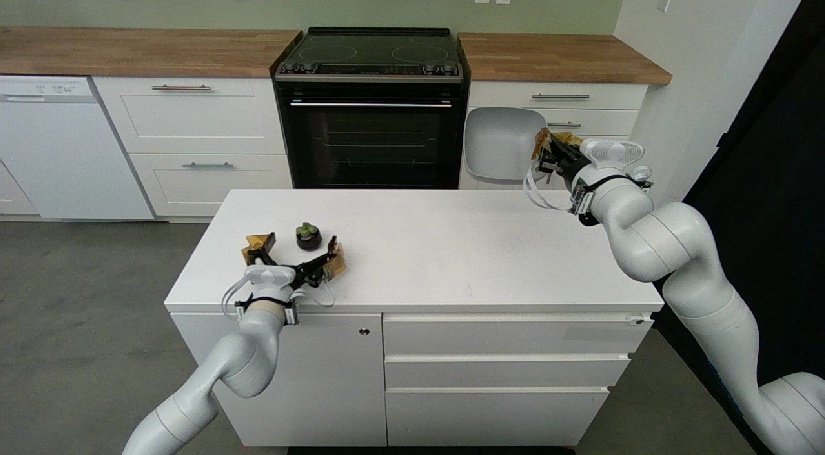} &
    \includegraphics[width=0.16\linewidth]{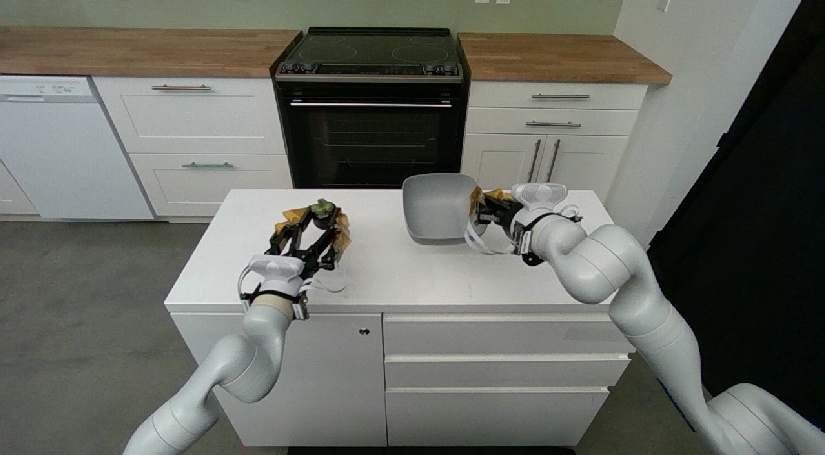} &
    \includegraphics[width=0.16\linewidth]{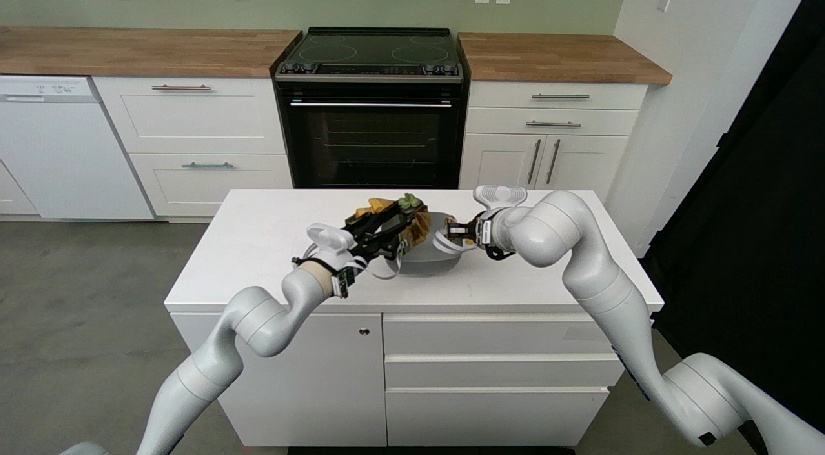} &
    \includegraphics[width=0.16\linewidth]{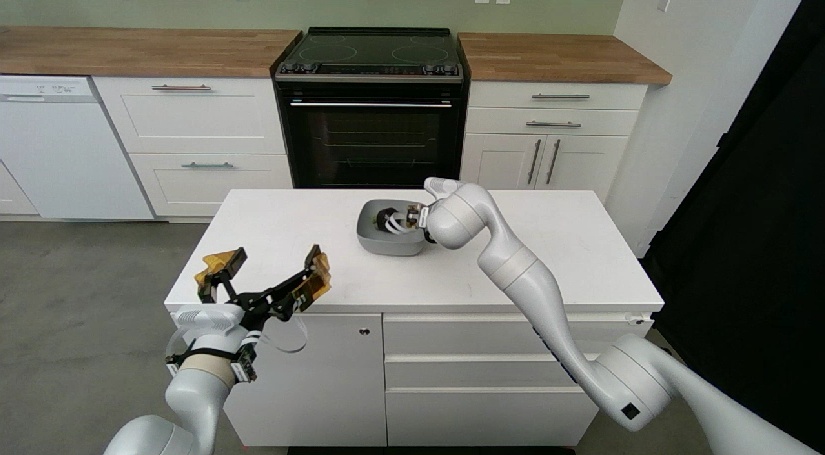} \\

    \multirow{1}{*}[3.3em]{\rotatebox[origin=c]{90}{\scriptsize Orange Bowl}} &
    \includegraphics[width=0.16\linewidth]{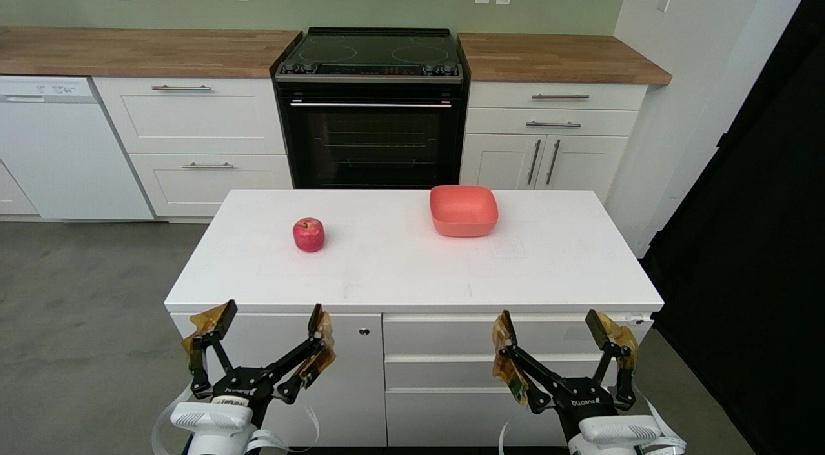} &
    \includegraphics[width=0.16\linewidth]{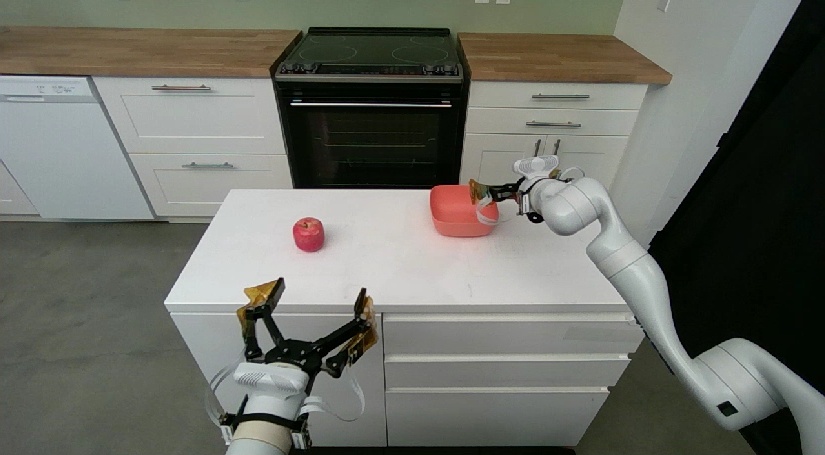} &
    \includegraphics[width=0.16\linewidth]{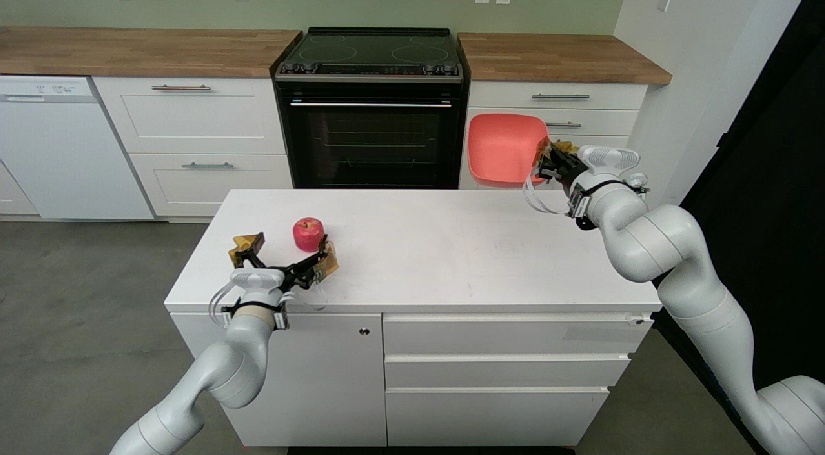} &
    \includegraphics[width=0.16\linewidth]{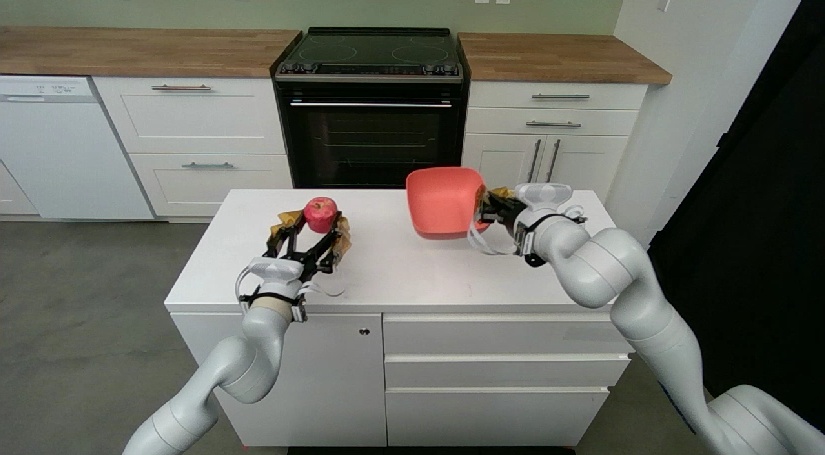} &
    \includegraphics[width=0.16\linewidth]{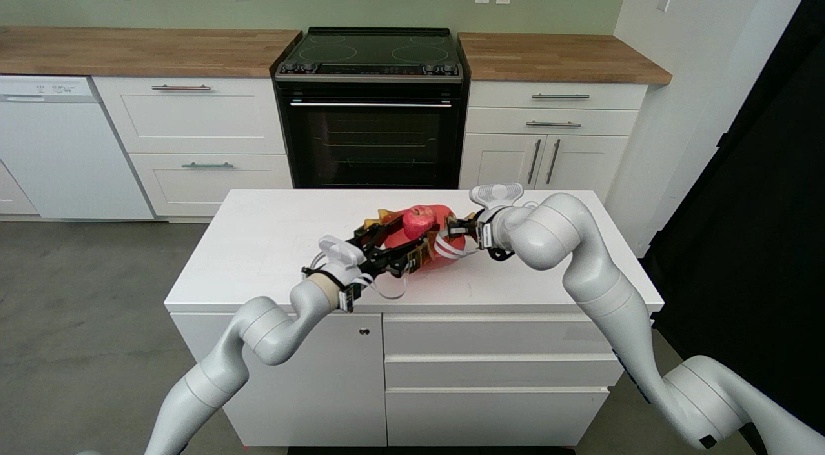} &
    \includegraphics[width=0.16\linewidth]{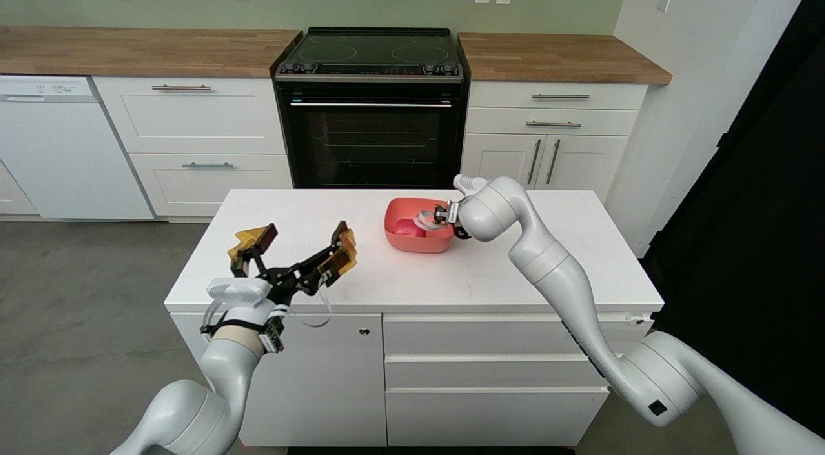} \\

    \multirow{1}{*}[3.2em]{\rotatebox[origin=c]{90}{\scriptsize Beige Table}} &
    \includegraphics[width=0.16\linewidth]{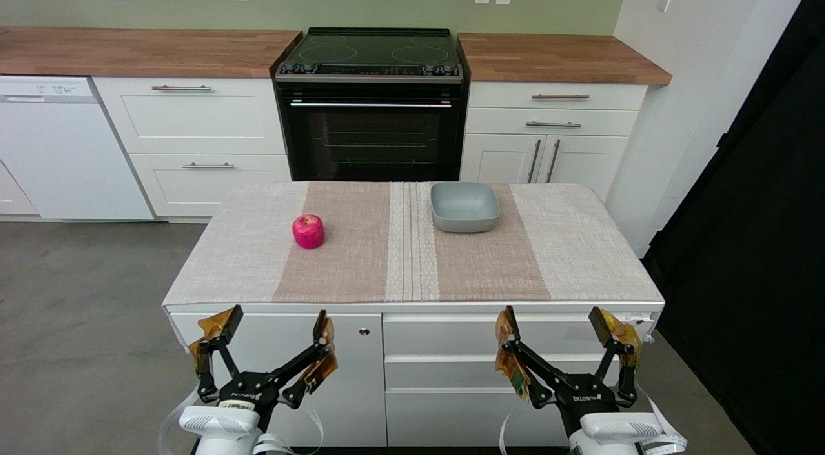} &
    \includegraphics[width=0.16\linewidth]{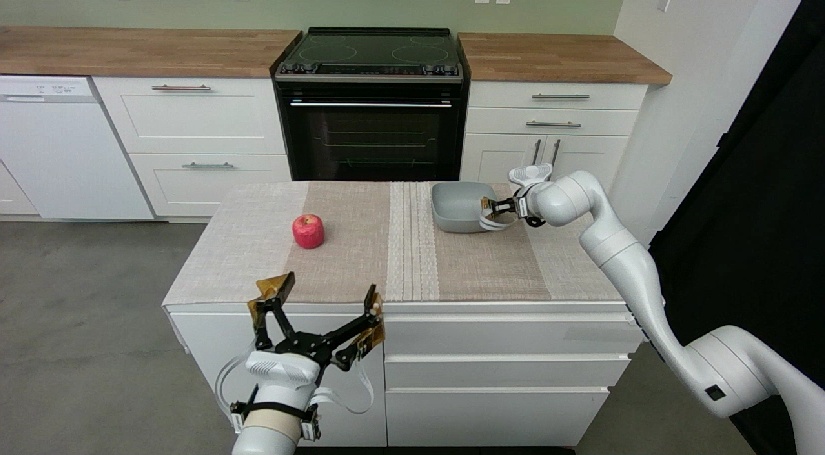} &
    \includegraphics[width=0.16\linewidth]{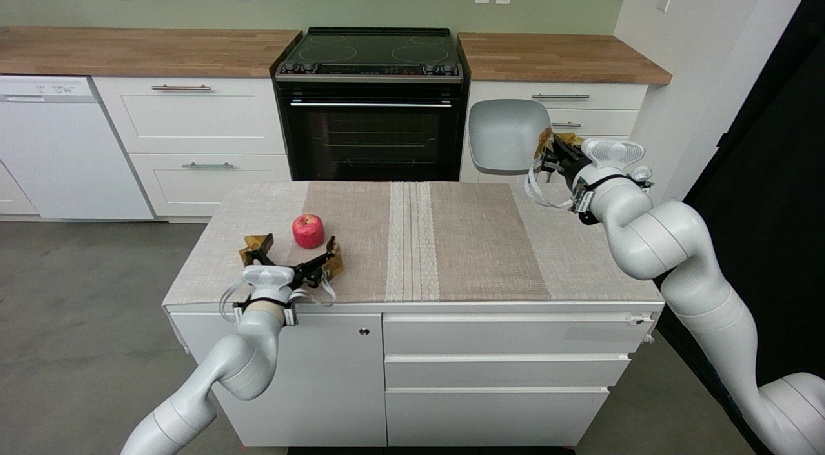} &
    \includegraphics[width=0.16\linewidth]{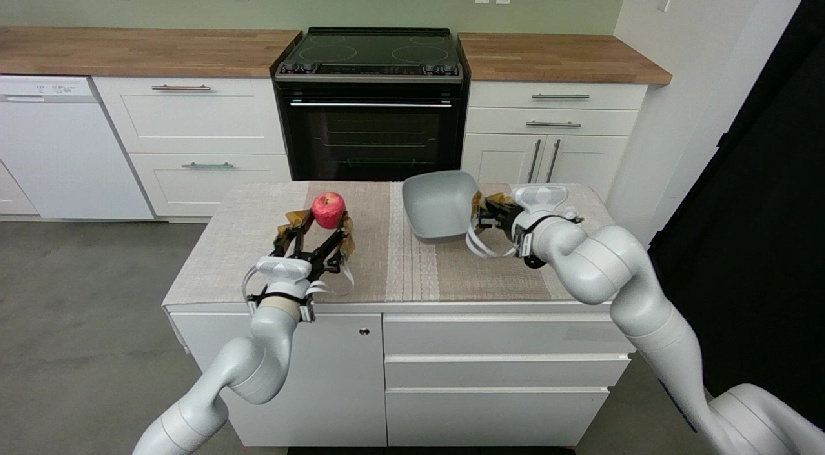} &
    \includegraphics[width=0.16\linewidth]{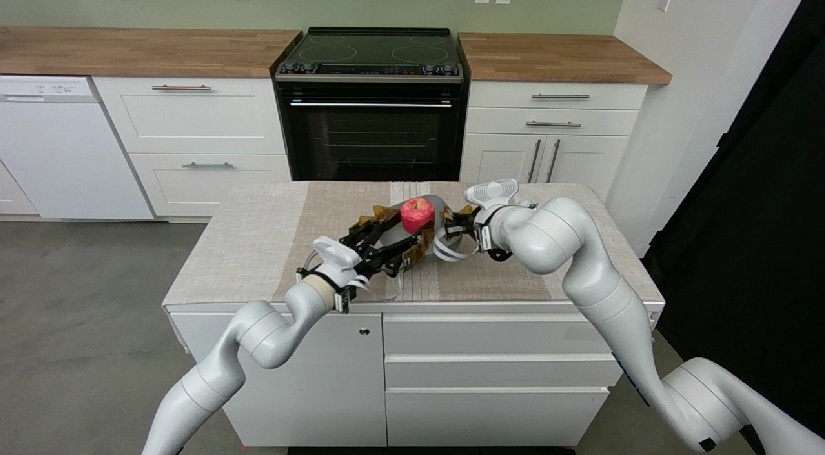} &
    \includegraphics[width=0.16\linewidth]{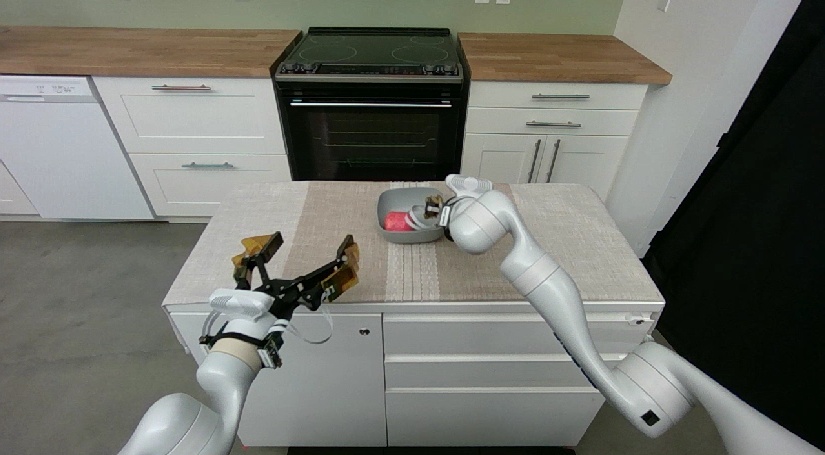} \\

    \multirow{1}{*}[3.2em]{\rotatebox[origin=c]{90}{\scriptsize Black Table}} &
    \includegraphics[width=0.16\linewidth]{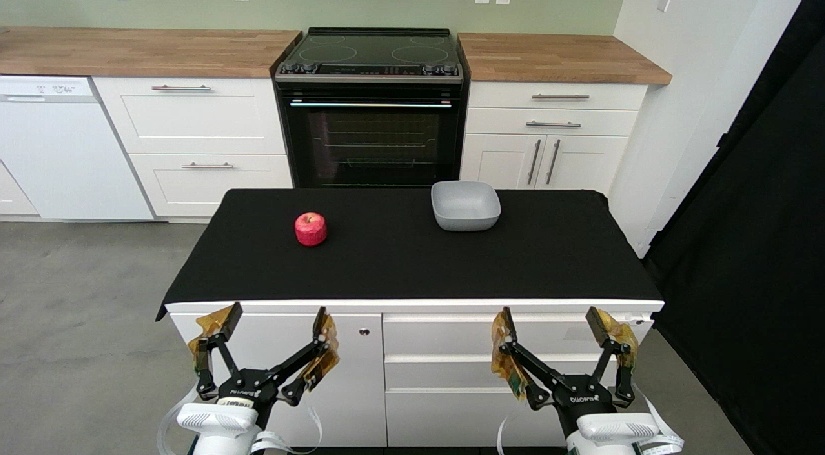} &
    \includegraphics[width=0.16\linewidth]{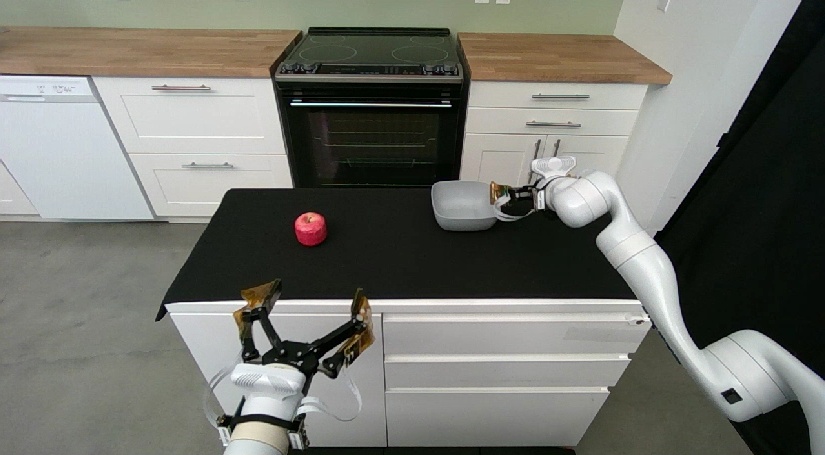} &
    \includegraphics[width=0.16\linewidth]{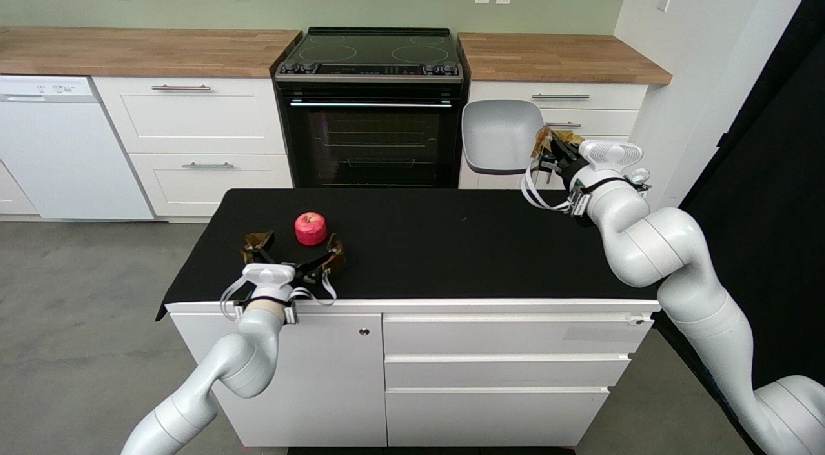} &
    \includegraphics[width=0.16\linewidth]{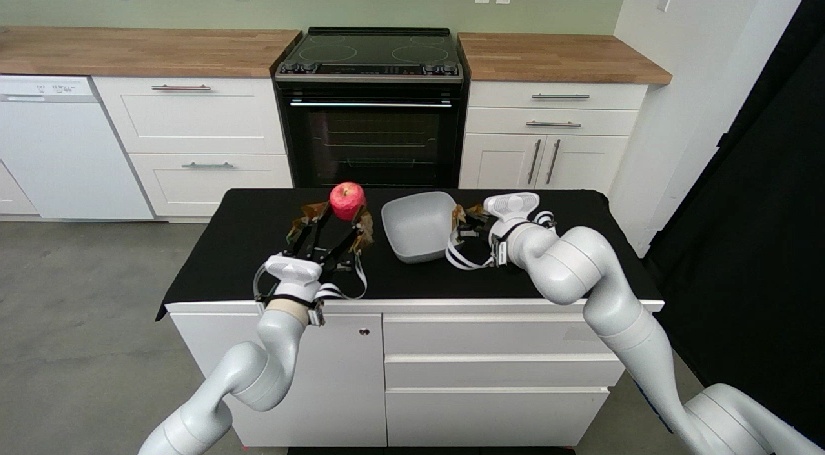} &
    \includegraphics[width=0.16\linewidth]{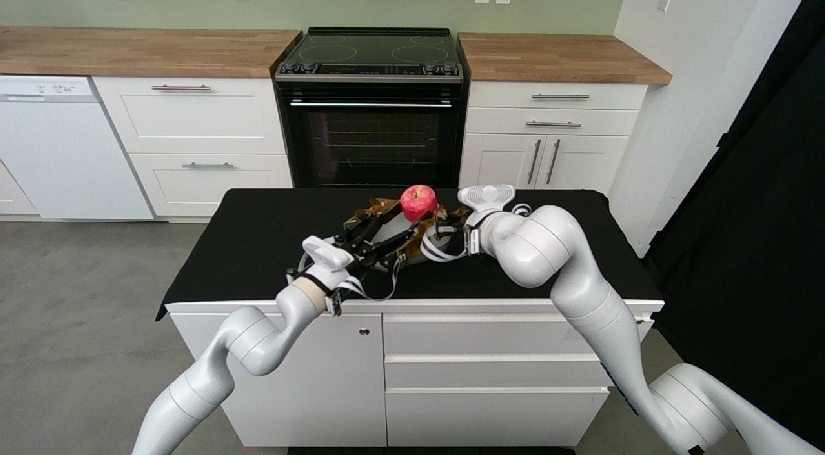} &
    \includegraphics[width=0.16\linewidth]{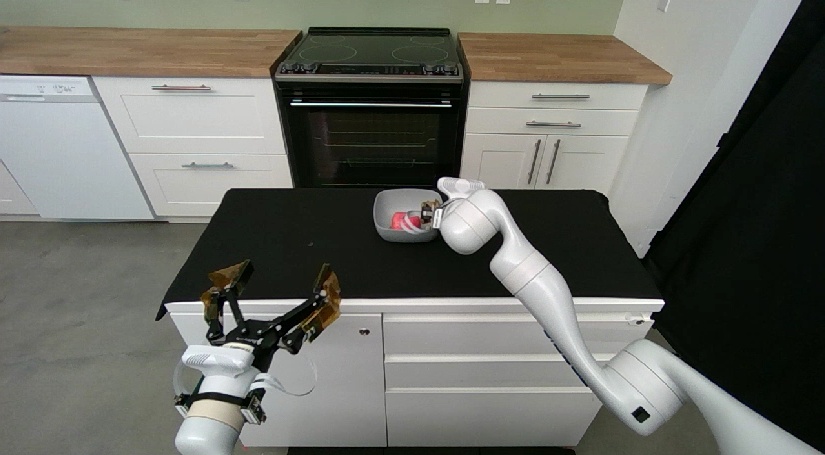} \\

    \multirow{1}{*}[2.8em]{\rotatebox[origin=c]{90}{\scriptsize Light On}} &
    \includegraphics[width=0.16\linewidth]{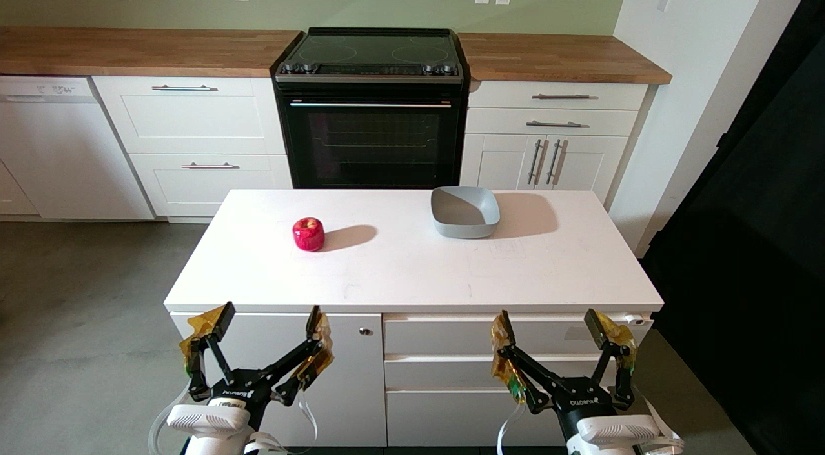} &
    \includegraphics[width=0.16\linewidth]{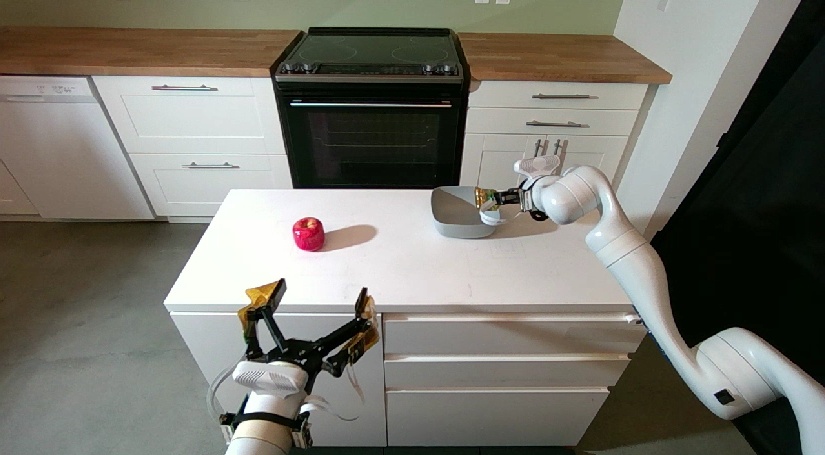} &
    \includegraphics[width=0.16\linewidth]{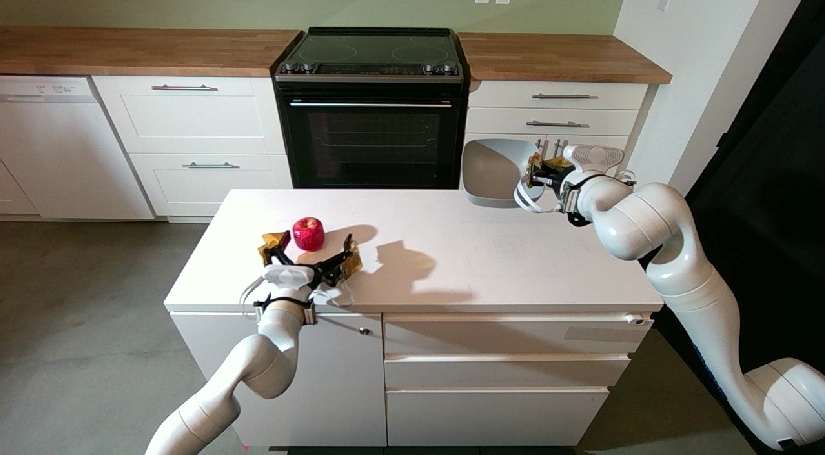} &
    \includegraphics[width=0.16\linewidth]{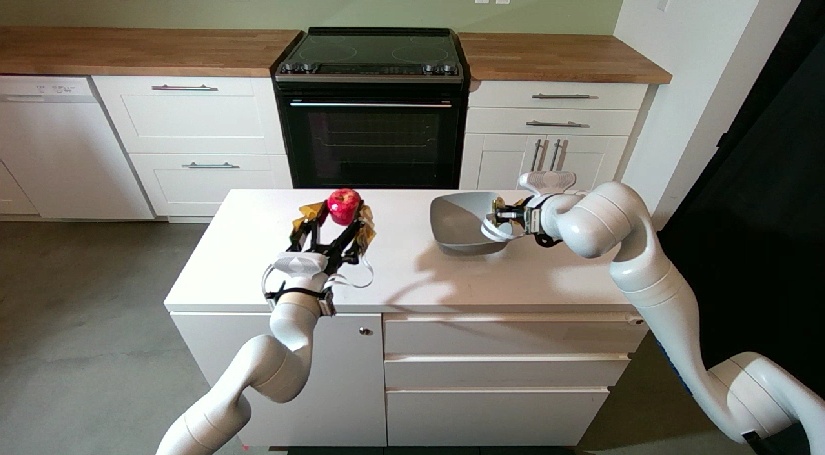} &
    \includegraphics[width=0.16\linewidth]{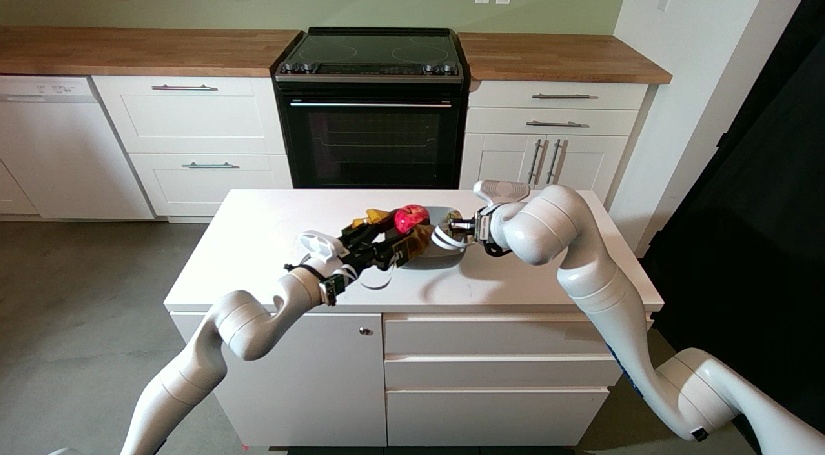} &
    \includegraphics[width=0.16\linewidth]{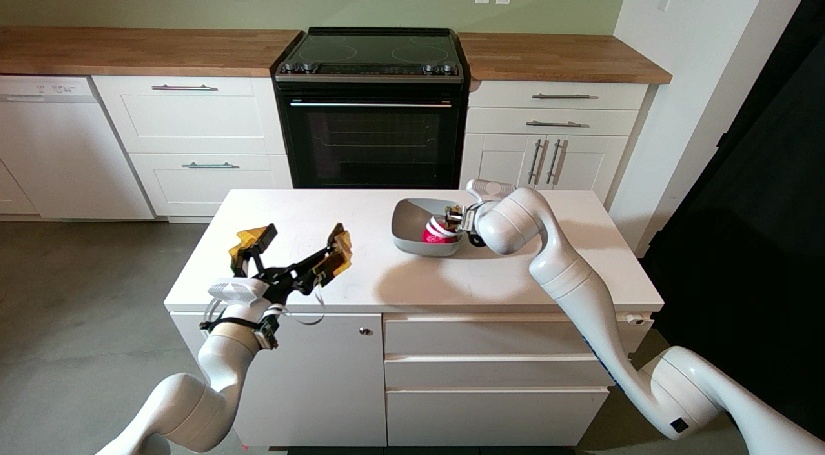} \\

    \multirow{1}{*}[3.2em]{\rotatebox[origin=c]{90}{\scriptsize Distractors}} &
    \includegraphics[width=0.16\linewidth]{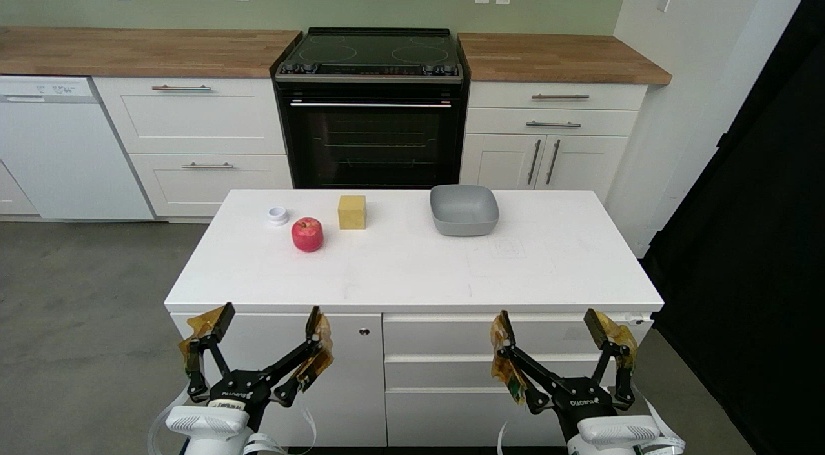} &
    \includegraphics[width=0.16\linewidth]{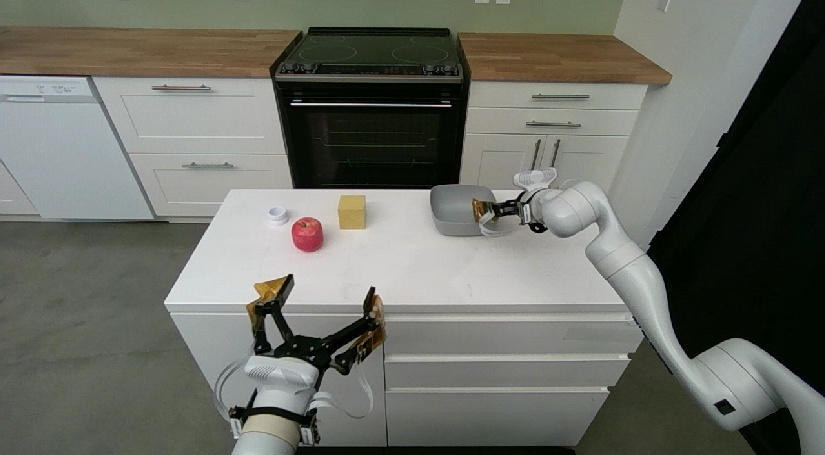} &
    \includegraphics[width=0.16\linewidth]{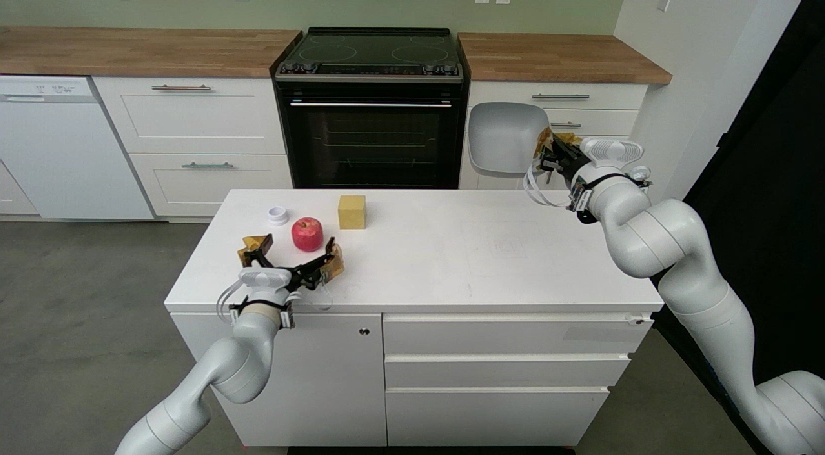} &
    \includegraphics[width=0.16\linewidth]{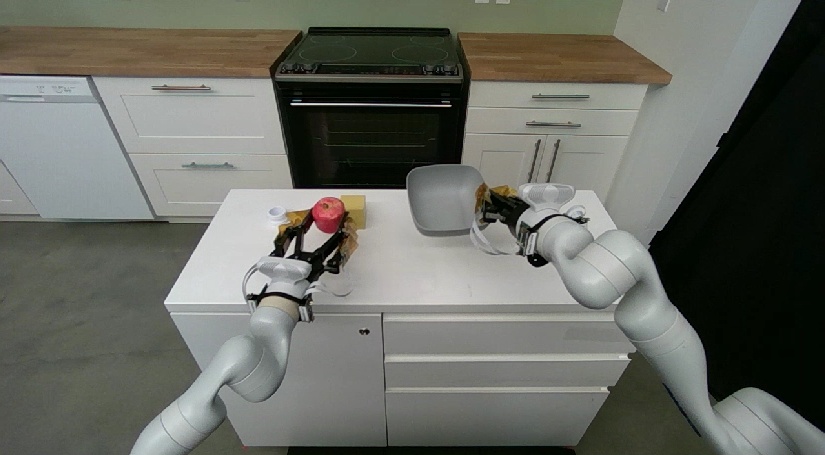} &
    \includegraphics[width=0.16\linewidth]{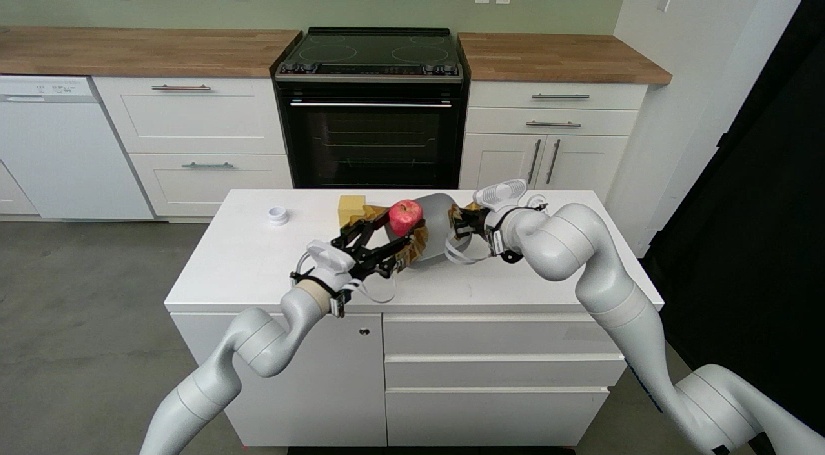} &
    \includegraphics[width=0.16\linewidth]{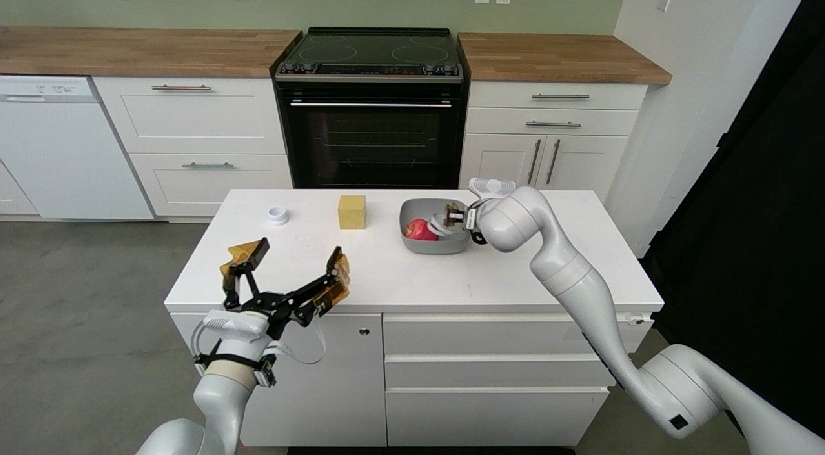} \\

    \multirow{1}{*}[3.6em]{\rotatebox[origin=c]{90}{\scriptsize Black Cabinet}} &
    \includegraphics[width=0.16\linewidth]{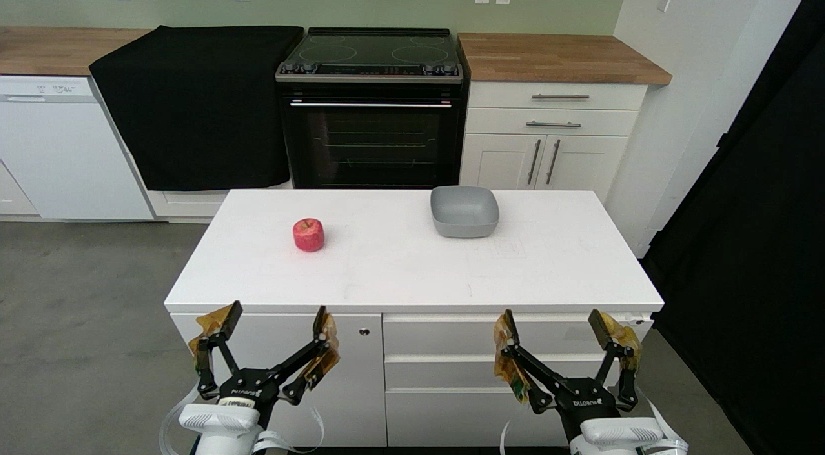} &
    \includegraphics[width=0.16\linewidth]{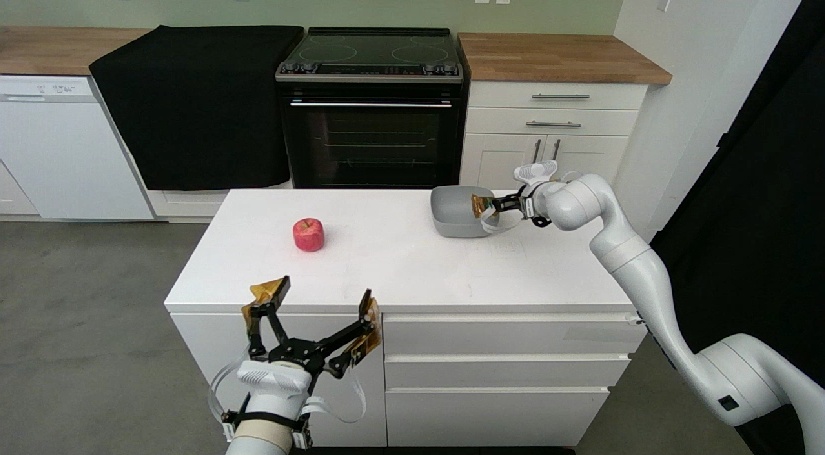} &
    \includegraphics[width=0.16\linewidth]{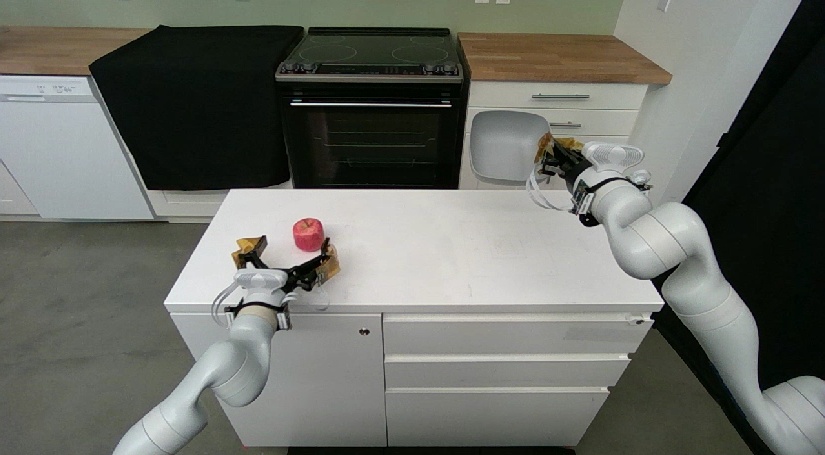} &
    \includegraphics[width=0.16\linewidth]{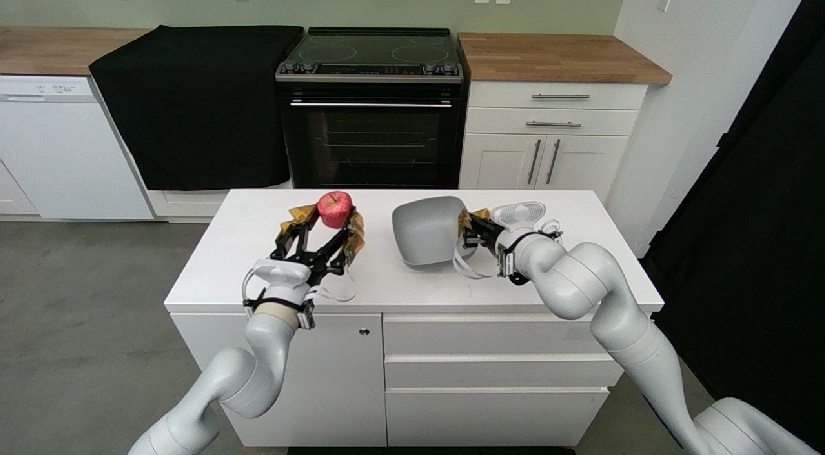} &
    \includegraphics[width=0.16\linewidth]{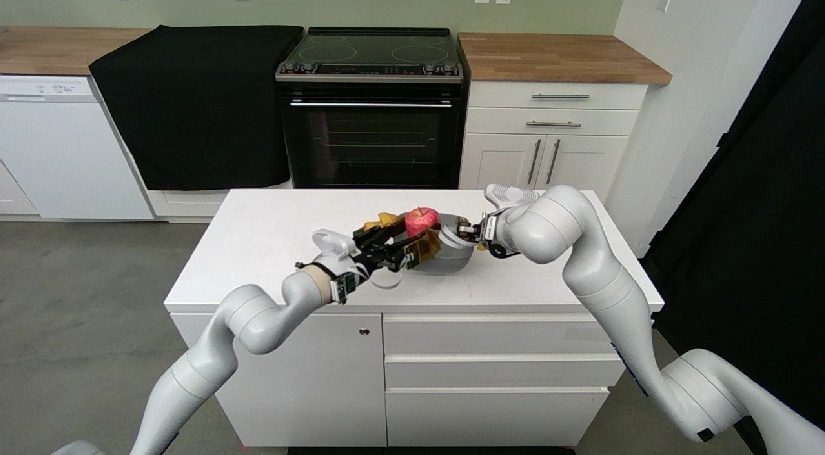} &
    \includegraphics[width=0.16\linewidth]{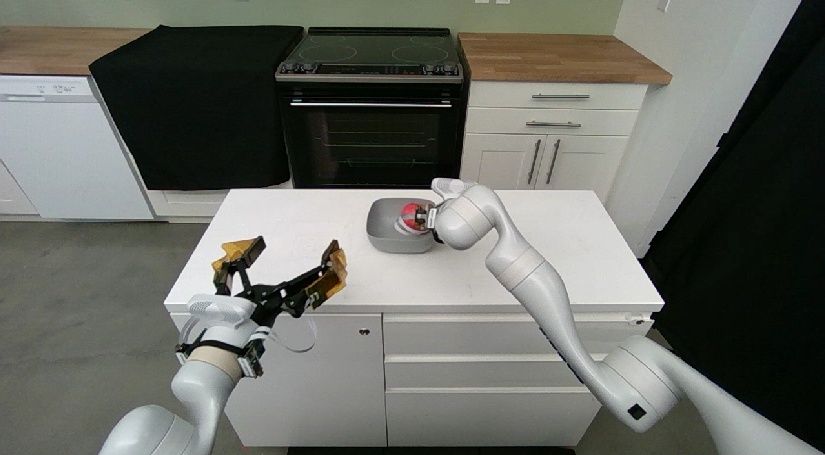} \\

    \multirow{1}{*}[3.6em]{\rotatebox[origin=c]{90}{\scriptsize Open Drawers}} &
    \includegraphics[width=0.16\linewidth]{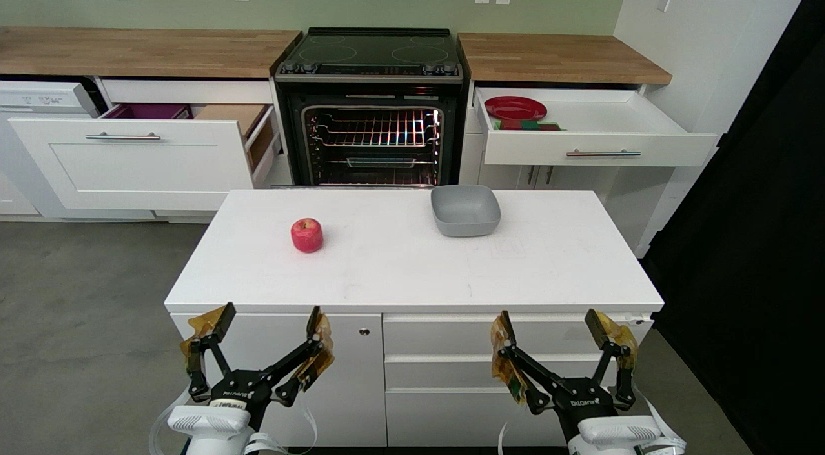} &
    \includegraphics[width=0.16\linewidth]{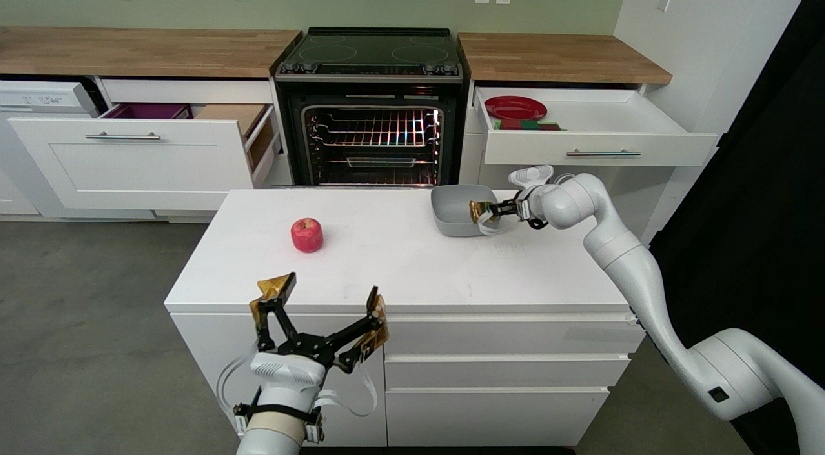} &
    \includegraphics[width=0.16\linewidth]{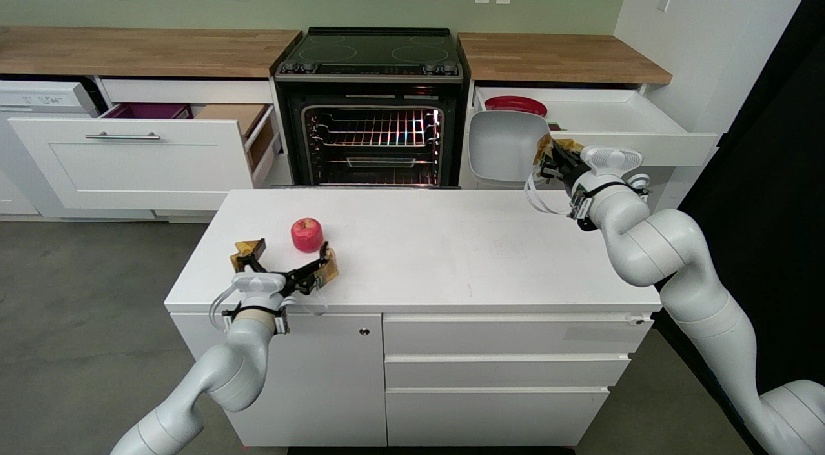} &
    \includegraphics[width=0.16\linewidth]{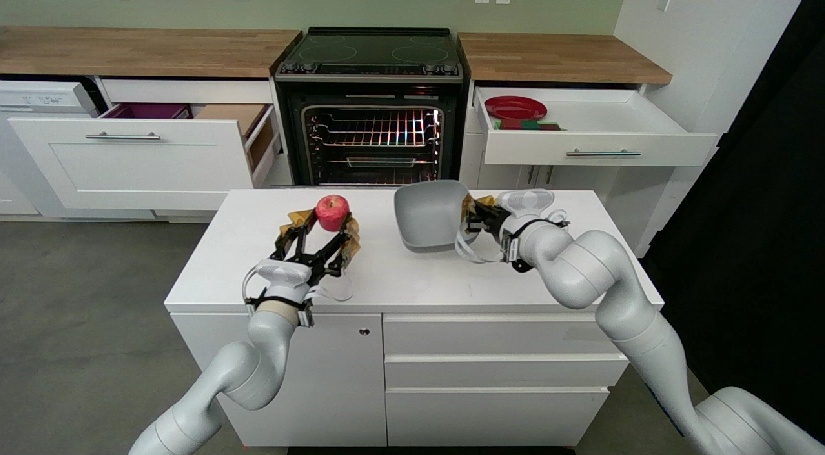} &
    \includegraphics[width=0.16\linewidth]{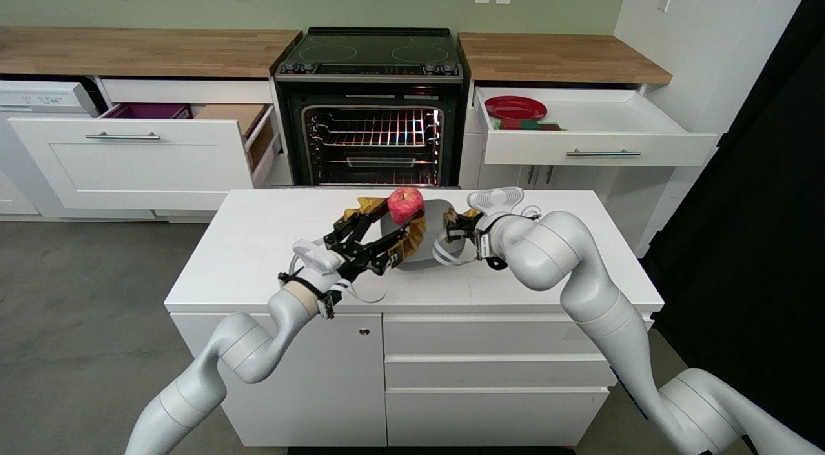} &
    \includegraphics[width=0.16\linewidth]{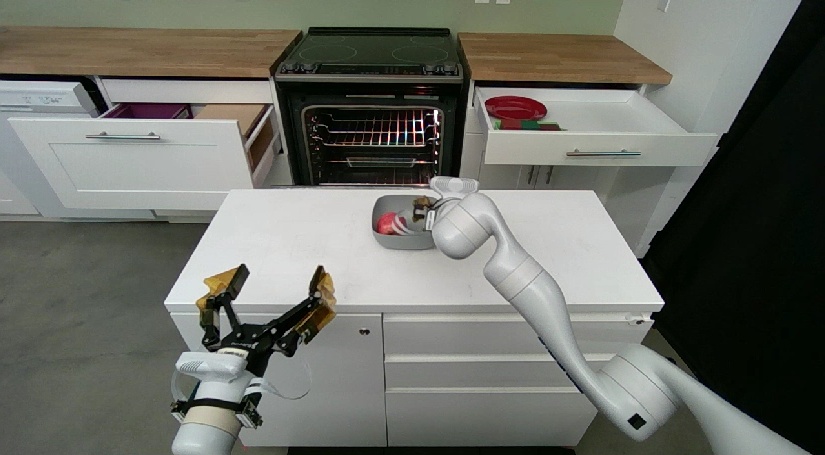} \\

    \multirow{1}{*}[2.6em]{\rotatebox[origin=c]{90}{\scriptsize Combo}} &
    \includegraphics[width=0.16\linewidth]{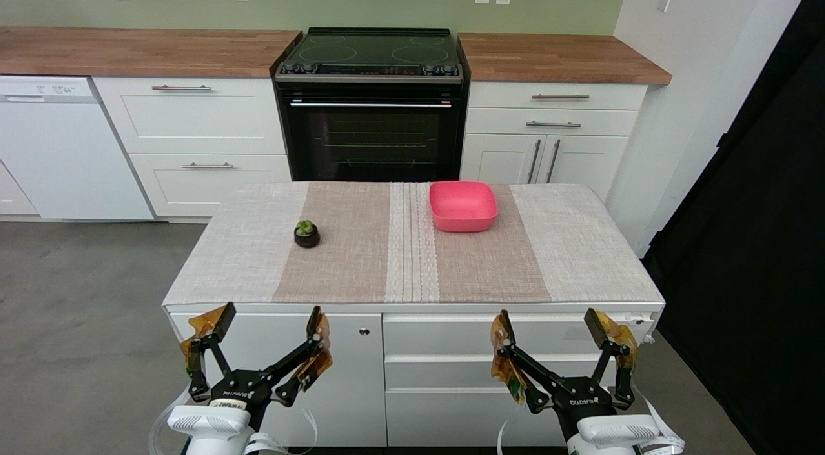} &
    \includegraphics[width=0.16\linewidth]{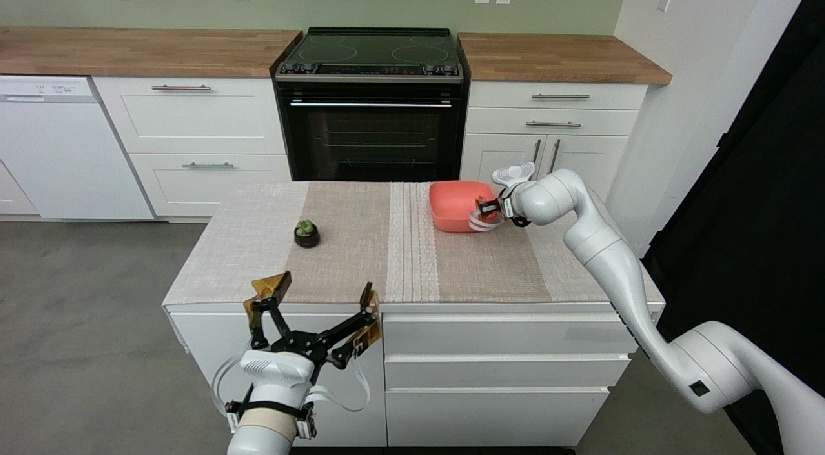} &
    \includegraphics[width=0.16\linewidth]{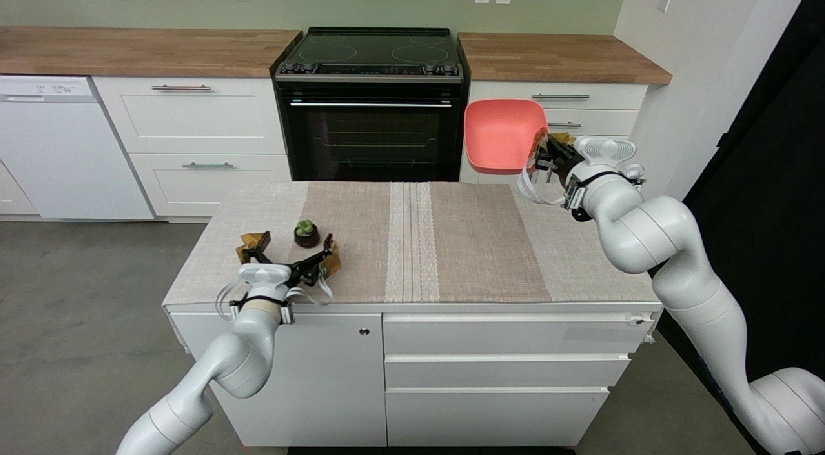} &
    \includegraphics[width=0.16\linewidth]{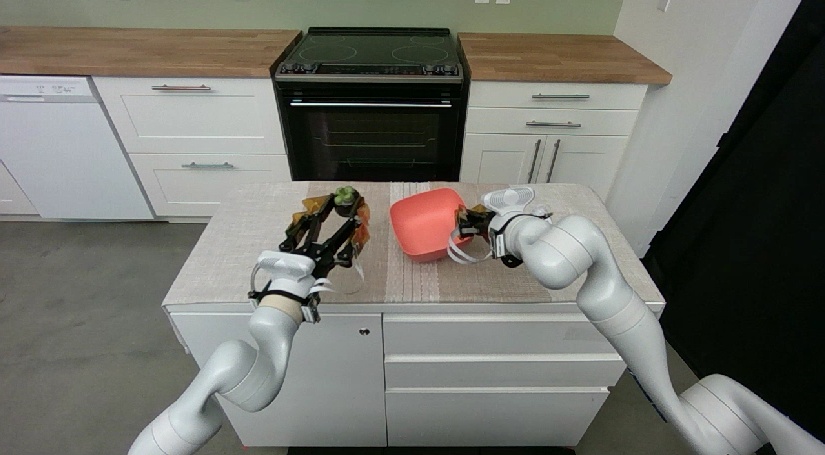} &
    \includegraphics[width=0.16\linewidth]{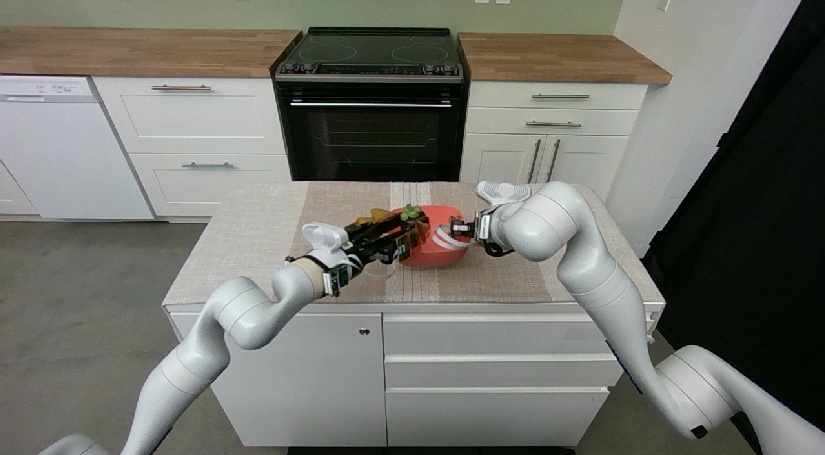} &
    \includegraphics[width=0.16\linewidth]{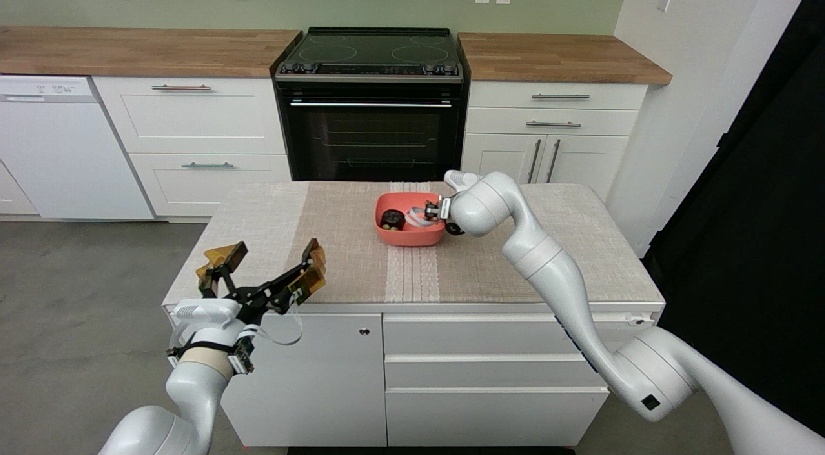} \\
  \end{tabular}

  \caption{\textbf{[Cosmos-Transfer2.5] Real-Robot Policy Rollouts.} We present sample [Cosmos-Transfer2.5-2B] policy rollouts under the base setting and nine unseen test-time scenarios.}
  \label{fig:real_robot_rollouts}
\end{figure*}

In \cref{fig:real_robot_gallery} (bottom two rows), we present a few examples of diverse and realistic synthetic videos used for visual augmentation. These examples illustrate variations in apple and bowl colors, table appearances with realistic textures, as well as diverse lighting conditions, object shadows, and background changes. For each of the 100 original demonstration videos, we randomly generate five synthetic variants for augmentation. The rest of the training data (i.e., actions and joint states) remain unchanged, while only the input images are augmented.

\subsubsection{Experiments and Results} 
We perform real-robot experiments under varied test-time object and environment conditions. Beyond the base setting, which mirrors the training configuration, we evaluate nine novel scenarios: (1) replacing the apple with a purple mangosteen, (2) replacing the gray bowl with an orange bowl, (3) placing a beige tablecloth, (4) placing a black tablecloth, (5) adding a spotlight to the robot’s left, (6) adding distractor objects on the table, (7) changing the left-side background cabinet to black, (8) opening the background drawers and oven door, and (9) a challenging combination of the first three modifications. Notably, while the first five variations may fall within the range of our diverse prompt augmentations, the subsequent three represent clear out-of-distribution shifts, and the final composite condition poses an especially challenging test. See \cref{fig:real_robot_rollouts} (leftmost column) for an overview of all ten test settings.

\begin{table}[htb!]
\caption{\textbf{Real-Robot Quantitative Evaluation.} We test the base, baseline, and proposed (a policy trained with [Cosmos-Transfer2.5-2B] augmented observations) on 10 test scenarios.}
\centering
\label{tab:transfer_real_robot}
\resizebox{\textwidth}{!}{%
\begin{tabular}{l|cccccccccc|c}
\toprule
 & Base & Mangosteen & Orange Bowl & Beige Table & Black Table & Light On & Distractors & Black Cabinet & Open Drawers & Combo & \textbf{Total} \\
\midrule
Base      & 1/3 & 0/3 & 0/3 & 0/3 & 0/3 & 0/3 & 0/3 & 0/3 & 0/3 & 0/3 & ~~1/30 \\
Baseline  & \textbf{3/3} & 0/3 & 2/3 & 0/3 & 0/3 & 0/3 & 0/3 & 0/3 & 0/3 & 0/3 & ~~5/30 \\
Proposed & \textbf{3/3} & \textbf{3/3} & \textbf{3/3} & \textbf{1/3} & \textbf{1/3} & \textbf{2/3} & \textbf{3/3} & \textbf{2/3} & \textbf{3/3} & \textbf{3/3} & \textbf{24/30} \\
\bottomrule
\end{tabular}}
\end{table}

We compare our trained [Cosmos-Transfer2.5-2B] diffusion policy against two policies:
\begin{enumerate}
\item a base policy trained solely on 100 teleoperation videos, and 
\item a baseline policy trained with standard image-based data augmentation techniques (\eg, random adjustments of brightness, contrast, saturation, and hue; gamma correction; salt-and-pepper noise; histogram equalization; random blurring or sharpening).
\end{enumerate}
For the baseline policy, augmentations are applied on-the-fly during training to maximize input diversity. Example augmented images are shown in \cref{fig:real_robot_gallery} (top row). While standard image-based augmentation is a commonly used technique to improve test-time visual robustness, it cannot perform semantic edits such as changing object colors, environment appearances, or lighting conditions, which [Cosmos-Transfer2.5-2B] can naturally address.

\Cref{tab:transfer_real_robot} summarizes our policy performance against the base and baseline policies. For each test scenario, we perform three trials and fix the object pose and environment configuration to ensure fair comparisons. The [Cosmos-Transfer2.5-2B]-augmented policy achieves 24 successes out of 30 trials, clearly outperforming both baselines. It demonstrates markedly higher robustness and generalization to novel test-time object and environment changes.

The base policy, trained only on the base setting, fails to generalize to novel settings and performs poorly even on the base setting due to subtle, human-imperceptible scene variations. The baseline policy, using standard image augmentations, succeeds in just one case, highlighting the limitations of basic transformations for these challenging scenarios.

\Cref{fig:real_robot_rollouts} visualizes successful rollouts across all ten test cases. Despite occasional failures (\eg, imprecise grasps), the results indicate that [Cosmos-Transfer2.5] provides a promising, lightweight, and effective pipeline for synthetic data generation in robotics.

\subsection{Cosmos-Transfer2.5 for Driving Simulation}
\label{subsec::cosmos-auto}

We extend [Cosmos-Predict2.5-2B] from single-view to multi-view world generation, resulting in [Cosmos-
Predict2.5-2B/auto/multiview]. In addition, just as we extended [Cosmos-Predict2.5-2B] into [Cosmos-Transfer2.5-2B] by adding a control branch, we similarly augment the multi-view version. This yields [Cosmos-Transfer2.5-2B/auto/multiview], a conditional, multi-view world generation model capable of generating consistent scenes across multiple viewpoints.

\subsubsection{Model Architecture}

To generate multiple 720p views, we re-purpose the latent temporal dimension by concatenating multiple views along it, effectively treating views as sequential frames. To remain within memory limits while still benefiting from FSDP and context parallelism, we reduce the latent temporal dimension to 8, which allows us to fit up to 7 views simultaneously.

\begin{figure}[!t]
    \centering
    \begin{minipage}[c]{0.38\textwidth}
        \centering
        {
          \setlength{\fboxsep}{0pt}
          \fbox{\includegraphics[width=\linewidth]{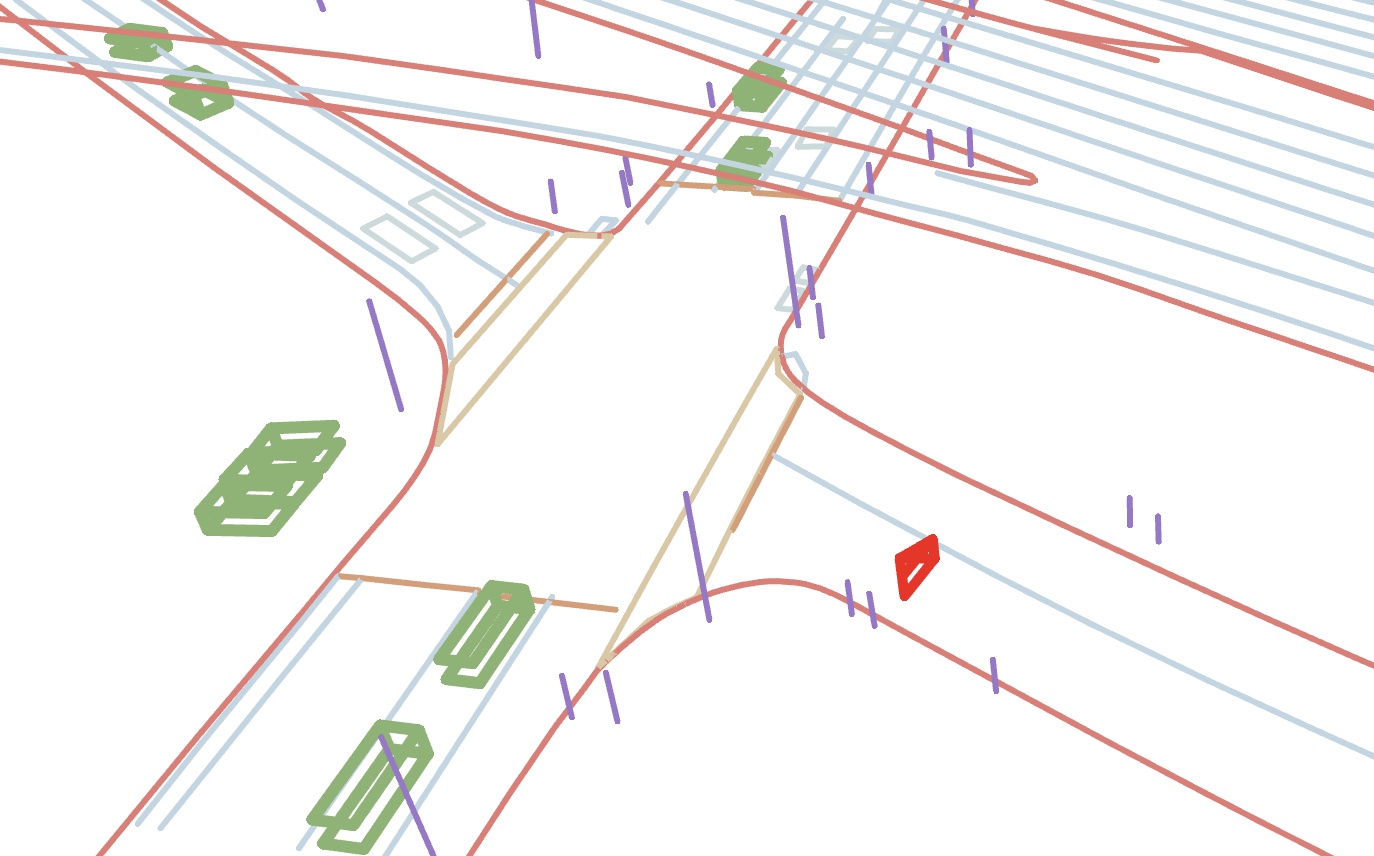}}
        }
        {\small 3D vector map and actors.
        
        The red frustum represents the front wide camera on the ego-vehicle, which has just exited the intersection.}
    \end{minipage}
    \hspace{0.0em}
    \begin{minipage}[c]{0.60\textwidth}
        \centering
        \includegraphics[width=\linewidth]{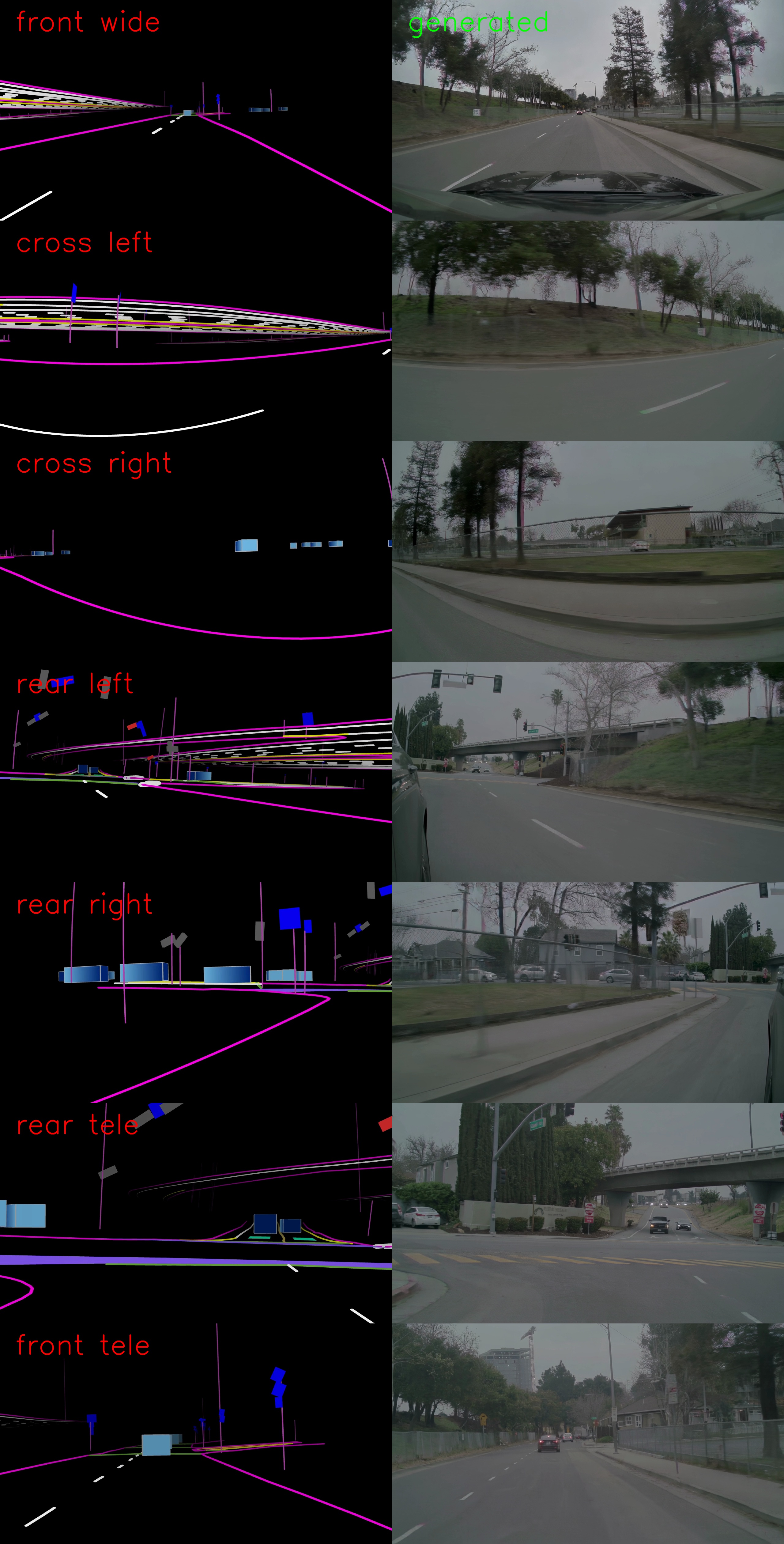}
    \end{minipage}
    \caption{Generated multi-view frames from [Cosmos-Transfer2.5-2B/auto/multiview]. The multi-view 720p control videos for driving simulation consist of HD map elements like lanes, road markings, poles, traffic signals, traffic lights (with state), all of which can represent complex road topologies (including overpasses) as well as actors represented as cuboids. Each cuboid is color-coded based on a coarse class ontology (e.g., truck, vehicle, pedestrian), and is also shaded to differentiate between the front and back.}
     \label{fig:transfer2mv_control_signals}
\end{figure}

Each view is encoded (and decoded) independently by the tokenizer. Once encoded, we concatenate in the latent channel dimension a compact per-view learnt embedding (of size 7) before passing it through the DiT network. We apply 3D-factorized Rotary Position Embeddings (RoPE) and cross-attention with text embeddings, following the same design as in [Cosmos-Predict2.5-2B]. Although we concatenate the views in the latent temporal dimension, we construct the RoPE embeddings separately per view. Each view can also be conditioned by one or more frames, and in the case of  [Cosmos-Transfer2.5-2B/auto/multiview], each view can additionally be controlled by a separate control signal, as shown in \cref{fig:transfer2mv_control_signals}.

\begin{table*}[htb!]
\centering
\small
\caption{Evaluation of visual metrics of on generated multi-view videos from RDS-HQ-HL dataset \citep{ren2025cosmos}. We use FVD StyleGAN, FVD I3D, and FID for visual quality \citep{stylegan_v} and TSE and CSE \citep{sampson1987} for multi-view consistency.}
\label{tab:transfer-multiview}
\begin{tabular}{ccccccc}
\toprule
\textbf{Model} &
\textbf{FVD StyleGAN \(\downarrow\)} &
\textbf{FVD I3D \(\downarrow\)} &
\textbf{FID \(\downarrow\)} &
\textbf{TSE \(\downarrow\)} &
\textbf{CSE \(\downarrow\)} \\
\midrule
Predict2.5-2B/auto/mv & \textbf{23.060} & \textbf{25.308} & \textbf{12.095} & 0.948 & \textbf{1.903} \\
Predict1-7B-Sample-AV & 63.685 & 69.613 & 25.341 & 0.930 & 2.631 \\
\midrule
Transfer2.5-2B/auto/multiview & \textbf{24.222} & \textbf{25.692} & \textbf{20.022} & 1.246 & 2.310 \\

Transfer1-7B-Sample-AV  & 56.606 & 60.660 & 22.633 & 1.017 & 1.835 \\
\midrule
Real Videos (Reference) & - & - & - & 1.193 & 1.832 \\
\bottomrule
\end{tabular}
\end{table*}

\begin{table*}[htb!]
\centering
\small
\caption{Evaluation of lane and bounding box detection on multi-view generated videos from RDS-HQ-HL dataset \citep{ren2025cosmos}. We use LET-AP/APL/APH for cuboid metrics \citep{hung2024let3daplongitudinalerrortolerant}, and F1, x-coordinate rMSE and accuracy for detection, regression and classification scores of lane detection.}
\label{tab:transfer-multiview-ground-truth}
\begin{tabular}{ccccccc}
\toprule
\multirow{2}{*}{Model} &
\multicolumn{3}{c}{Cuboids} &
\multicolumn{3}{c}{Lanes} \\
\cmidrule(lr){2-4} \cmidrule(lr){5-7}
 & LET-AP \(\uparrow\) &
   LET-APL \(\uparrow\) &
   LET-APH \(\uparrow\) &
   F1 \(\uparrow\) &
   x-error (far) \(\downarrow\) &
   Category Acc. \(\uparrow\)\\
\midrule
Transfer2.5-2B/auto/multiview & \textbf{0.394} & \textbf{0.254} & \textbf{0.383} & \textbf{0.637} & \textbf{0.487} & \textbf{0.904}  \\
Transfer1-7B-Sample-AV  & 0.243 & 0.154 & 0.236 & 0.604 & 0.524 & 0.899 \\
\midrule
Real Videos (Reference) & 0.476 & 0.319 & 0.462 & 0.637 & 0.480 & 0.905 \\
\bottomrule
\end{tabular}
\end{table*}

\begin{figure}[!tp]
\centering

\setlength{\extrarowheight}{1pt}
\renewcommand{\arraystretch}{1.0}

\begin{tabular}{@{}m{0.03\linewidth} m{0.48\linewidth} m{0.48\linewidth}@{}}
   & \centering (1) & \centering (2) \tabularnewline \hline

   \rotatebox{90}{Transfer1-7B-Sample-AV} &
   \includegraphics[width=.9\linewidth]{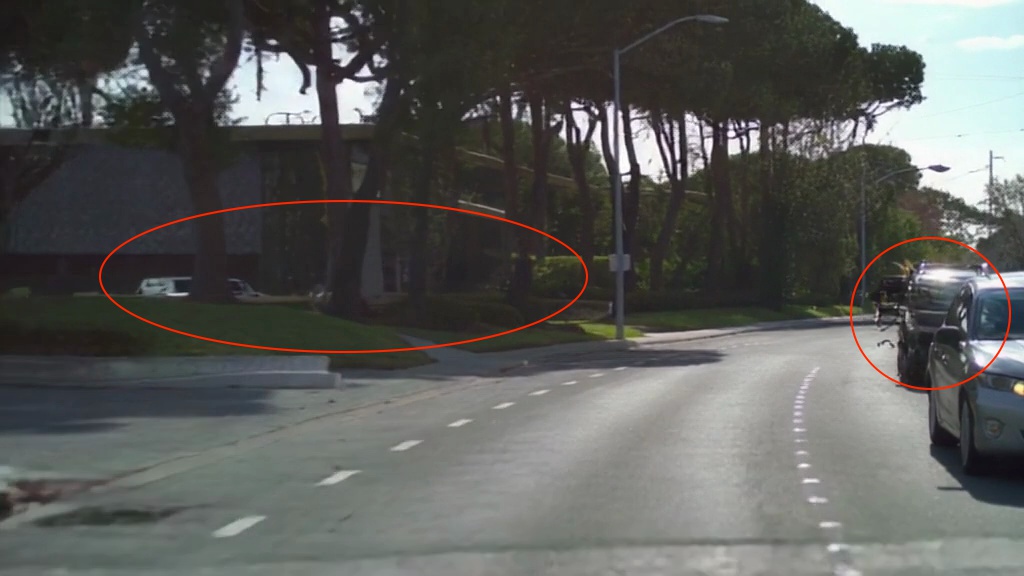} &
   \includegraphics[width=.9\linewidth]{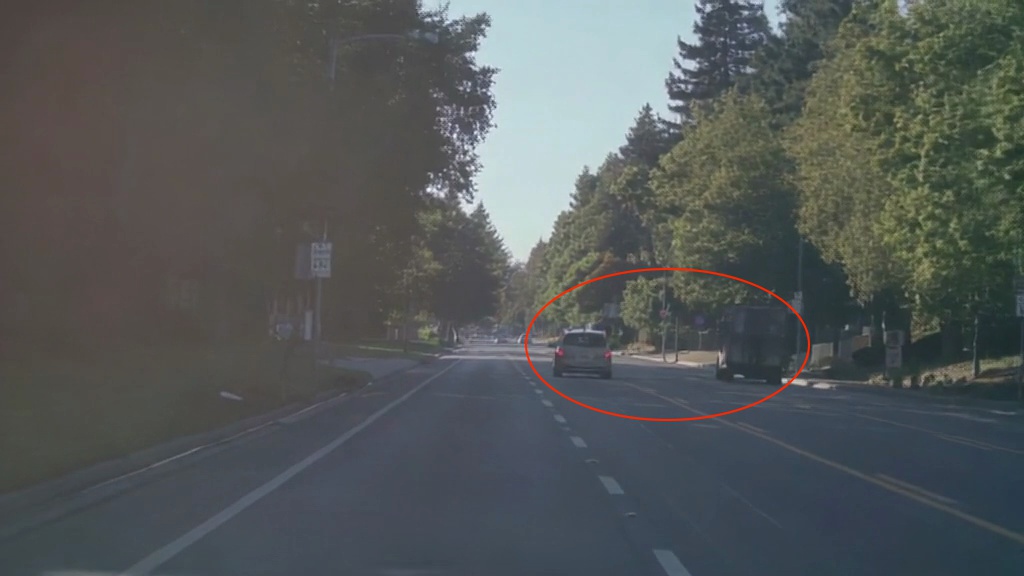} \\

   \rotatebox{90}{Transfer2.5-2B/drive/mv} &
   \includegraphics[width=.9\linewidth]{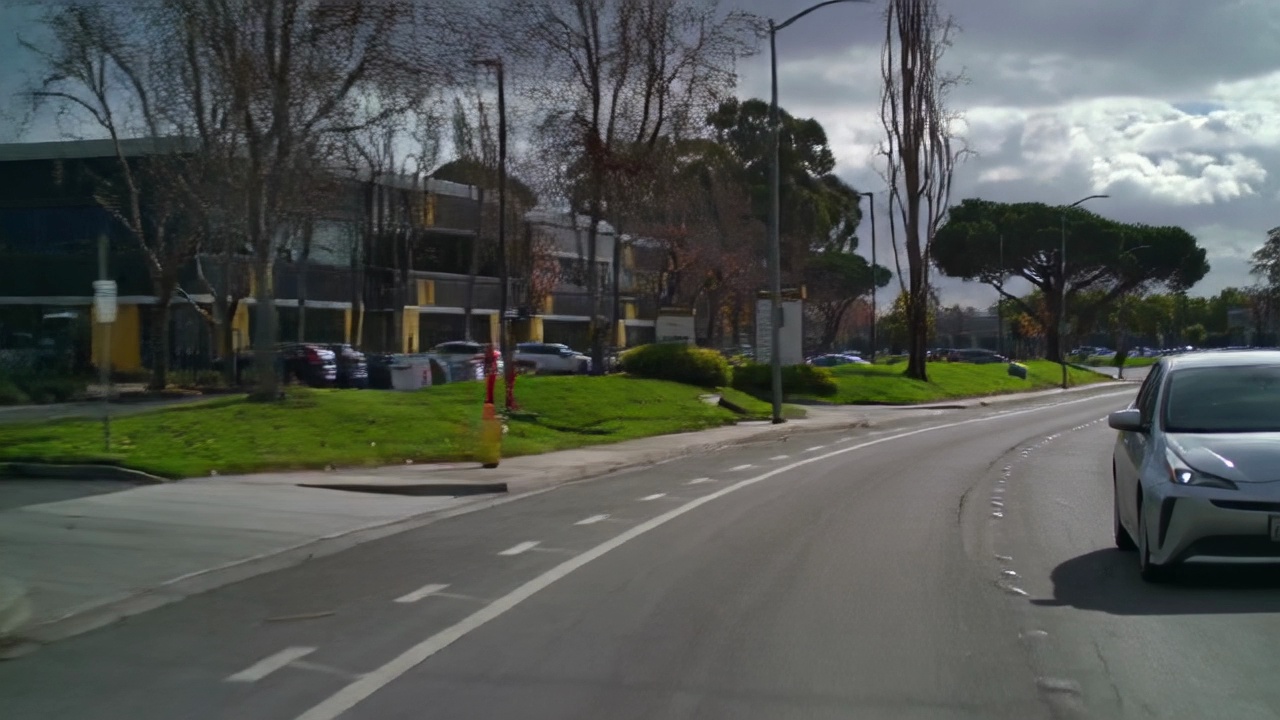} &
   \includegraphics[width=.9\linewidth]{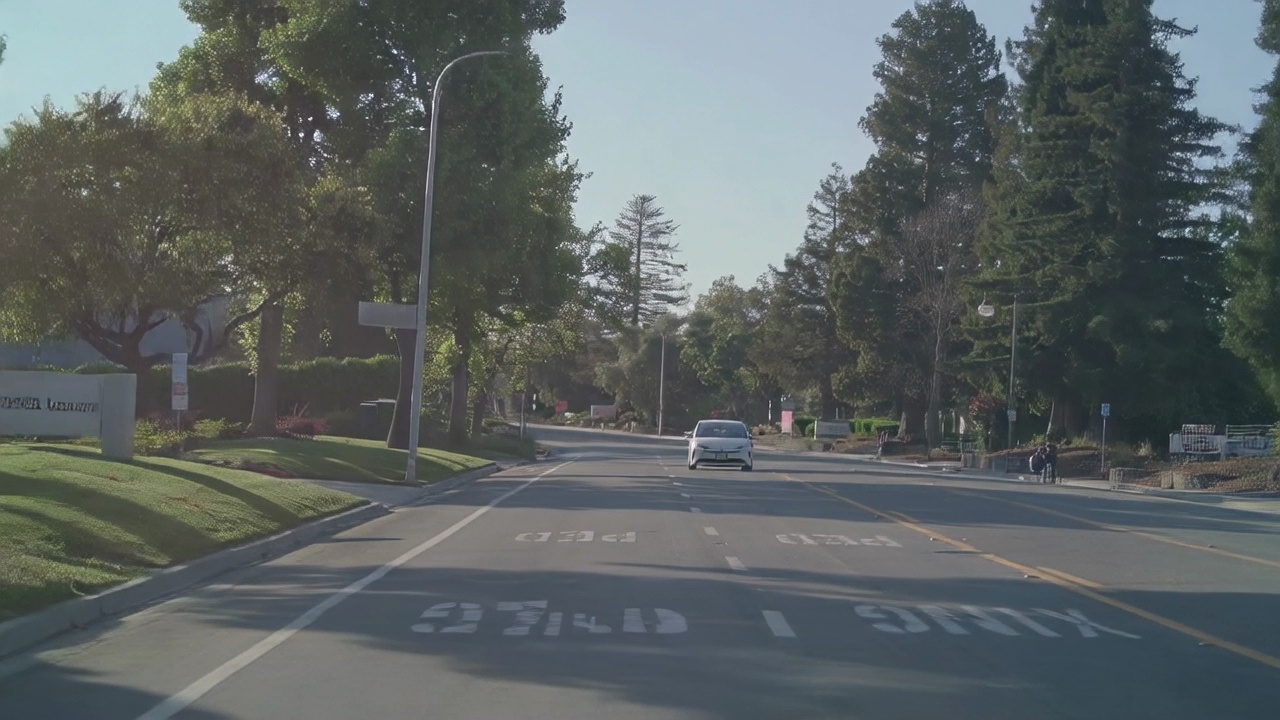} \\

   \rotatebox{90}{Control} &
   \includegraphics[width=.9\linewidth]{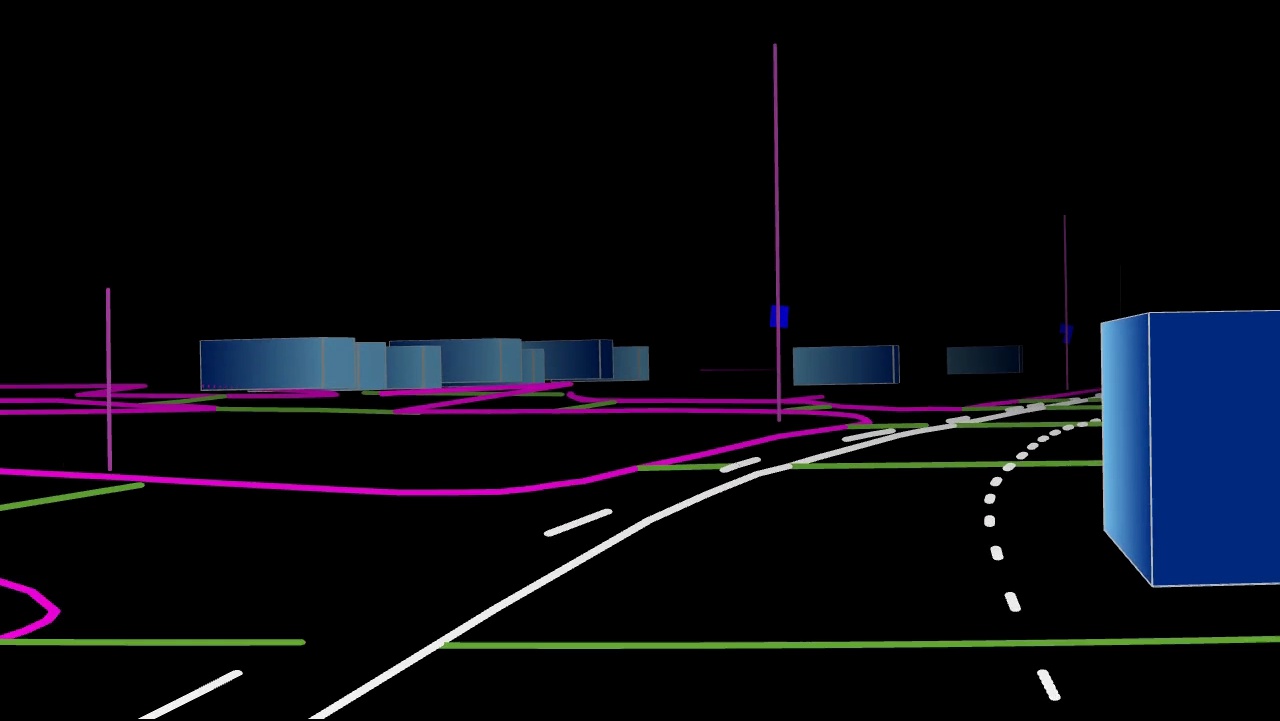} &
   \includegraphics[width=.9\linewidth]{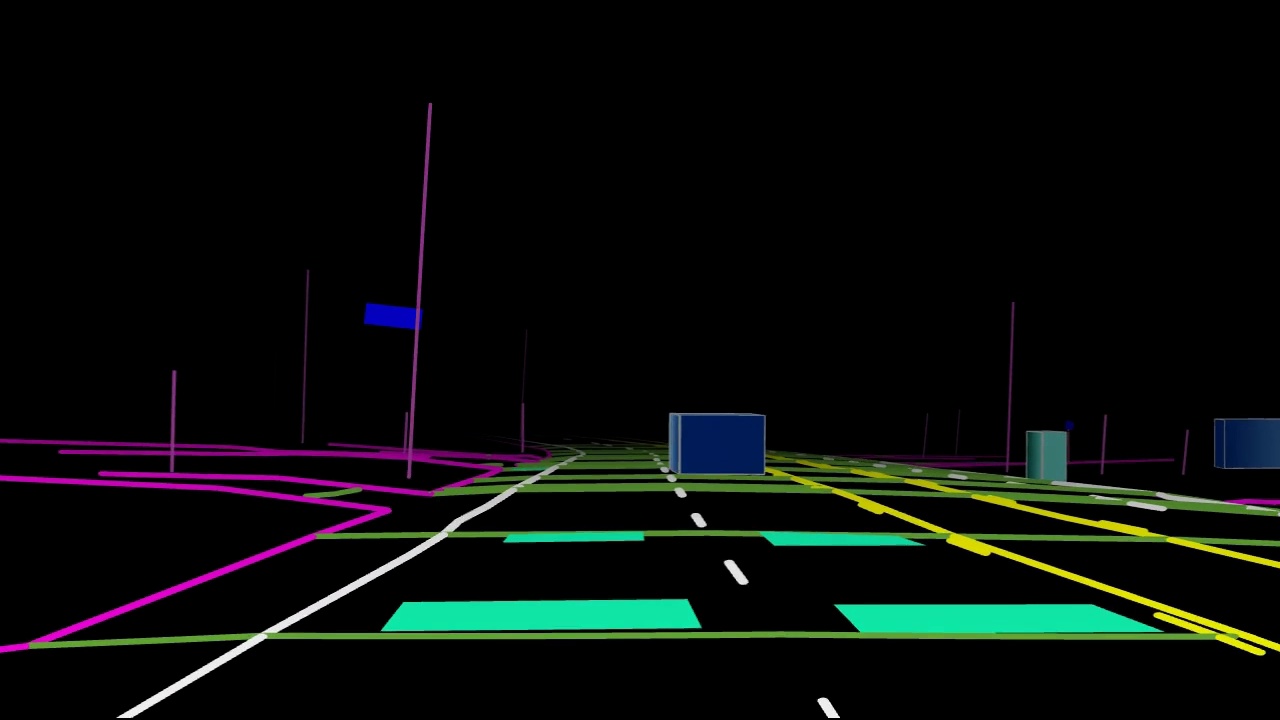} \\
\end{tabular}

\caption{Comparative controlled generations between [Cosmos-Transfer1-7B-Sample-AV] and [Cosmos-Transfer2.5-2B/auto/multiview]. In example (1), we can observe that [Cosmos-Transfer1-7B-Sample-AV] hallucinates a distorted black car behind the silver vehicle, which is described neither in the text prompt nor in the control video. We can also observe the lack of alignment to the control signal when generating the parked vehicles behind the grassy mounds. In example (2), we can observe that [Cosmos-Transfer1-7B-Sample-AV] renders the vehicle in the central lane driving on the wrong side of the street with an incorrect orientation, as well as a truck instead of a pedestrian close to the sidewalk. All these inconsistencies are resolved in [Cosmos-Transfer2.5-2B/auto/multiview].}
\label{fig:transfer_multiview}
\end{figure}

\subsubsection{Training Datasets}

For [Cosmos-Predict2.5-2B/auto/multiview], we curate a multi-view captioned dataset of 1.5 million clips, each containing 20-second-long scenarios with 7 synchronized cameras recording at 30FPS (front-wide, front-tele, front-left, front-right, rear-left, rear-right, rear-tele). To facilitate training with text conditioning, we generate captions at 150-frame intervals using Qwen2.5-7B-Instruct with three different lengths (short, medium, and long).

For [Cosmos-Transfer2.5-2B/auto/multiview], we project HD maps and dynamic objects in the scene onto the seven camera views as the control input, and we name it ``world scenario map'' \cref{fig:transfer2mv_control_signals}. The world scenario map includes map elements like lane lines, poles, road boundaries, traffic lights, etc., and is augmented with dynamic 3D bounding boxes that indicate the positions of vehicles and pedestrians. Each object type is color-coded, and the bounding boxes are shaded according to the direction of motion, providing both semantic and motion cues. 

To train the control net, we use the RDS-HQ dataset \citep{ren2025cosmos}, which consists of 140,000 20-second multi-view driving scenes and HD map metadata covering a diverse set of traffic scenarios. Compared to the original control videos used in this work, the new world scenario map improves the following aspects: firstly, it has fine-grained controls of lane line types (e.g., dashed line, dotted line, double yellow line), whose colors and geometry patterns are directly rendered into the control video. Secondly, the bounding boxes of dynamic objects are occlusion-aware and heading-aware, providing more accurate control signals for the model learning. 

\subsubsection{Experiments and Results}

We train [Cosmos-Predict2.5-2B/auto/multiview] for 2 epochs on the 1.5m clip dataset, using a global batch size of 64 and context parallelism of 8. We denoise 203 frames (29 per view) using 30 FPS video. For [Cosmos-Transfer2.5-2B/auto/multiview], we subsample the video and control inputs to 10FPS.

For evaluation, we use a 1000 multi-view clip dataset in RQS-HQ \citep{ren2025cosmos}, with HD map, as well as human-labeled lanes and cuboids. These clips are disjoint from the prior two datasets used in training.
As shown in \cref{tab:transfer-multiview}, we observe a significant boost (up to 2.3x) in FVD/FID scores while remaining competitive in temporal and cross-camera Sampson error.

To test adherence to the control signals, we measure the detection performance of 3D-cuboid and lane detection models on generated videos, and compare these with the ground truth labels. Following the protocol described in \citep{ren2025cosmos}, we use a monocular 3D lane detector, LATR \citep{luo2023latr3dlanedetection}, for evaluating 3D lane detection tasks, and a temporal 3D object detector, BEVFormer \citep{li2022bevformerlearningbirdseyeviewrepresentation}, for evaluating 3D cuboid detection tasks. 
As shown in \cref{tab:transfer-multiview-ground-truth}, we observe a substantial improvement (up to 60\%) in detection metrics compared to Transfer1-7B-Sample-AV \citep{cosmos_transfer1}. See \cref{fig:transfer_multiview} for visual comparisons of [Cosmos-Transfer-7B-AV-Sample] versus [Cosmos-Transfer2.5-2B/auto/multiview].

\subsection{Multi-view Generation with Camera Control}
\label{subsec::camera-control}

We develop [Cosmos-Transfer2.5-2B/robot/multiview], a camera-controllable multi-view world generation model built on top of [Cosmos-Predict2.5-2B]. Unlike standard single-view generation, this model takes a video from a reference view and synthesizes additional videos from multiple target viewpoints defined by camera trajectories. Such a setting is especially valuable in robotics, where it enables mapping a humanoid robot’s egocentric head-camera view to the gripper views on its two hands, useful for robotic manipulation simulation, where the robot must reason about objects beyond its direct line of sight. By generating consistent views that fill in occluded regions, the model provides a richer and more complete representation of the scene, enabling more reliable perception, planning, and control in real-world settings.

\begin{table}[htb!]

\caption{Camera Control Comparison between [Cosmos-Predict2.5] and [Cosmos-Predict1]. } 
\label{tab:multicam_comp_2.5v1}
\centering
\begin{threeparttable}
\resizebox{.85\textwidth}{!} 
{
\begin{tabular}{cccccc}
\toprule
 Model & Camera Views & Condition & Type & Resolution \\
 \midrule
 Cosmos-Predict1  & 1 & text + image condition & future prediction & 720p \\
 Cosmos-Predict2.5 & 3 & text + video condition & video re-rendering & 720p \\
\bottomrule
\end{tabular}
}
\end{threeparttable}
\end{table}

Given a source video and a set of $N$ target camera trajectories, each specified by extrinsic-intrinsic parameters, our objective is to synthesize $N$ target videos, each corresponding to a distinct virtual camera trajectory. We assume a standard pinhole camera model to project 3D scene points into 2D image coordinates. A comparison of camera control between [Cosmos-Predict1] and [Cosmos-Predict2.5] is provided in~\cref{tab:multicam_comp_2.5v1}.

\begin{figure}[!tp]
    \centering
\includegraphics[width=.95\textwidth]{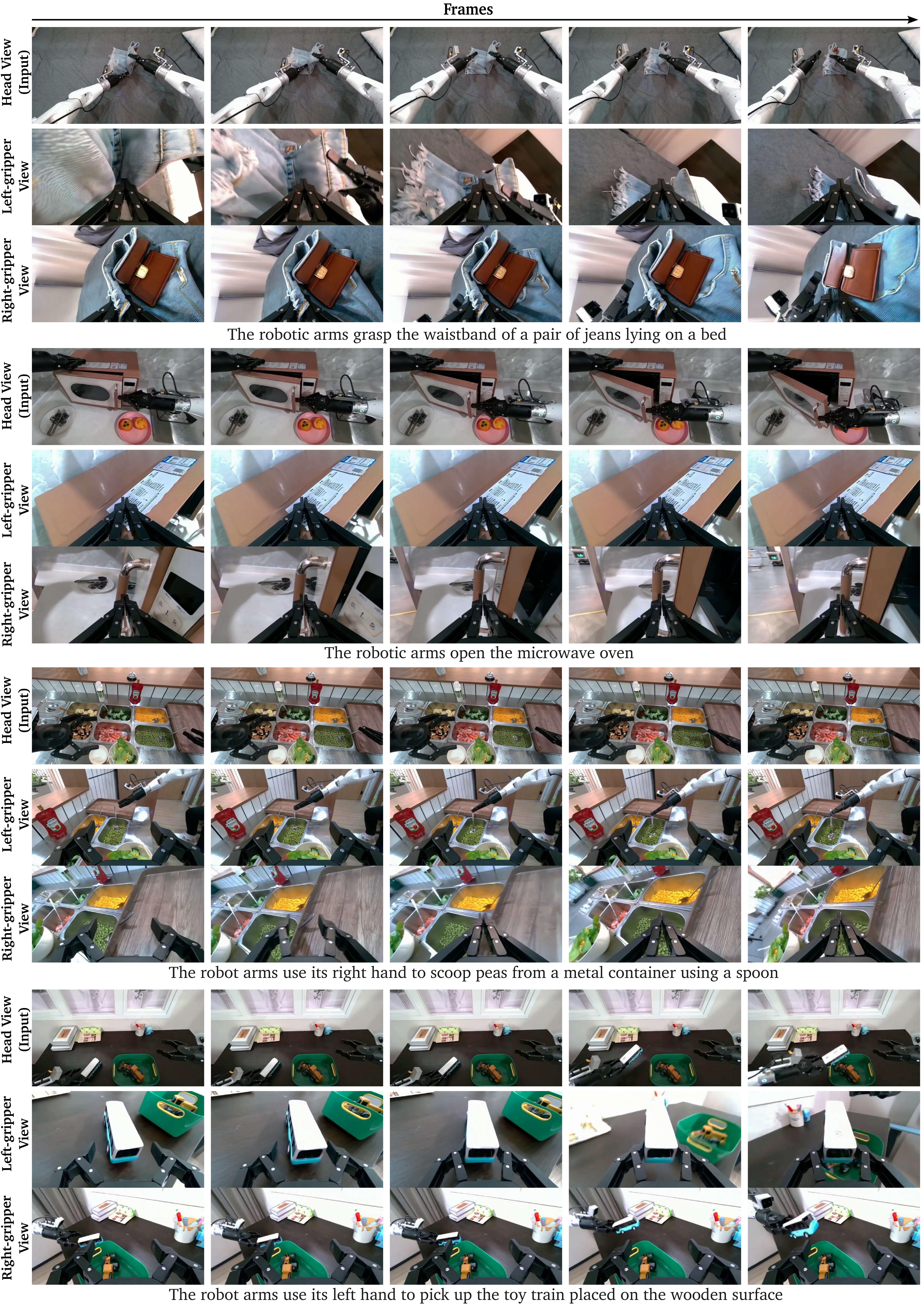}
    \caption{
    [Cosmos-Transfer2.5-2B/robot/multiview-agibot] generates temporally synchronized robotic manipulation videos from the left and right gripper viewpoints, conditioned on the head-view input.
    }
\label{fig:multicam_demo_agibot}
\end{figure}
\begin{figure}[!tp]
    \centering
    \includegraphics[width=.95\textwidth]{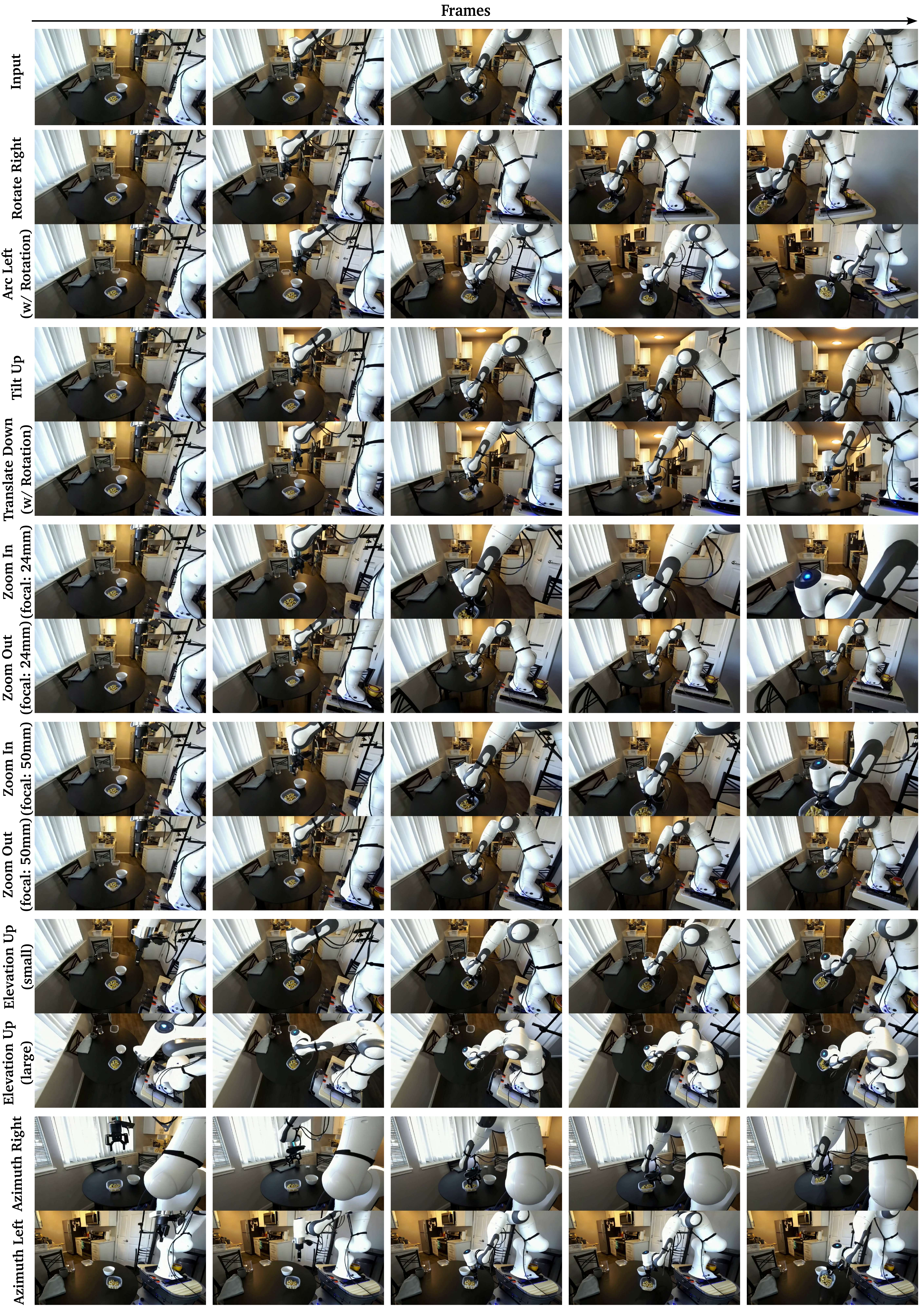}
    \caption{
    [Cosmos-Transfer2.5-2B/robot/multiview] synthesizes synchronized videos under basic dynamic and static camera transformations, conditioned on the third-view robotic manipulation input.
    }
\label{fig:multicam_demo_basic}
\end{figure}

\noindent \textbf{Architecture.}
We tokenize both source and target videos and concatenate their tokens along the temporal dimension. Since the encoder downsamples videos temporally by a factor of 4, we also sample camera parameters (intrinsic and extrinsic) every 4 frames to maintain temporal alignment with the latent features. Target cameras are represented as Plücker raymaps~\citep{sitzmann2021light}, where pixels are mapped to 6D ray representations and subsequently patchified. A camera projection layer is introduced to align the raymap representation with the dimensionality of the video latents. The resulting raymap tokens are then added to the video tokens prior to the self-attention operation, enabling the DiT to incorporate camera pose information. During training, we freeze all the layers except the self-attention layers and the camera projection layer.

\noindent \textbf{Training Datasets.} We train our model on the following datasets:
\begin{itemize}
    \item Agibot~\citep{bu2025agibot}: A robot dataset contain $\sim$1,000,000 episodes. We sample 145,820 episodes, each providing 3 video views with precise camera pose information.
    
    \item MultiCamVideo~\citep{bai2025recammaster}: A large-scale synthetic dataset comprising ~136,000 episodes of human motion captured with dynamic camera trajectories.

    \item SynCamVideo~\citep{bai2024syncammaster}: A complementary synthetic dataset containing ~34,000 episodes similar in content to MultiCamVideo but with fixed novel camera viewpoints, enabling evaluation under static multi-view settings.
    
\end{itemize}

\noindent \textbf{Experiments.}
We adopt [Cosmos-Predict2.5-2B] as the backbone model and further post-train two variants as follows. Both variants generate outputs at 720p resolution. To address out-of-memory issues during training and inference, we employ context parallelism across multiple GPUs.

\begin{itemize}
    \item \textbf{[Cosmos-Transfer2.5-2B/robot/multiview-agibot]}: Fine-tuned on the Agibot dataset. Given a head-view robotic manipulation video as input, it synthesizes synchronized videos from the left and right gripper perspectives, as illustrated in~\cref{fig:multicam_demo_agibot}.
    
    \item \textbf{[Cosmos-Transfer2.5-2B/robot/multiview]}: Fine-tuned on MultiCamVideo and SynCamVideo datasets. Conditioned on a third-view video, it generates two synchronized videos under basic camera transformations, such as left/right rotations, left/right arcs, zoom in/out, azimuth shifts, elevation changes, and distance variations, while allowing for dynamic focal length adjustments, as shown in~\cref{fig:multicam_demo_basic}.
\end{itemize}

\begin{figure}[!th]
    \centering
    \includegraphics[width=.96\textwidth]{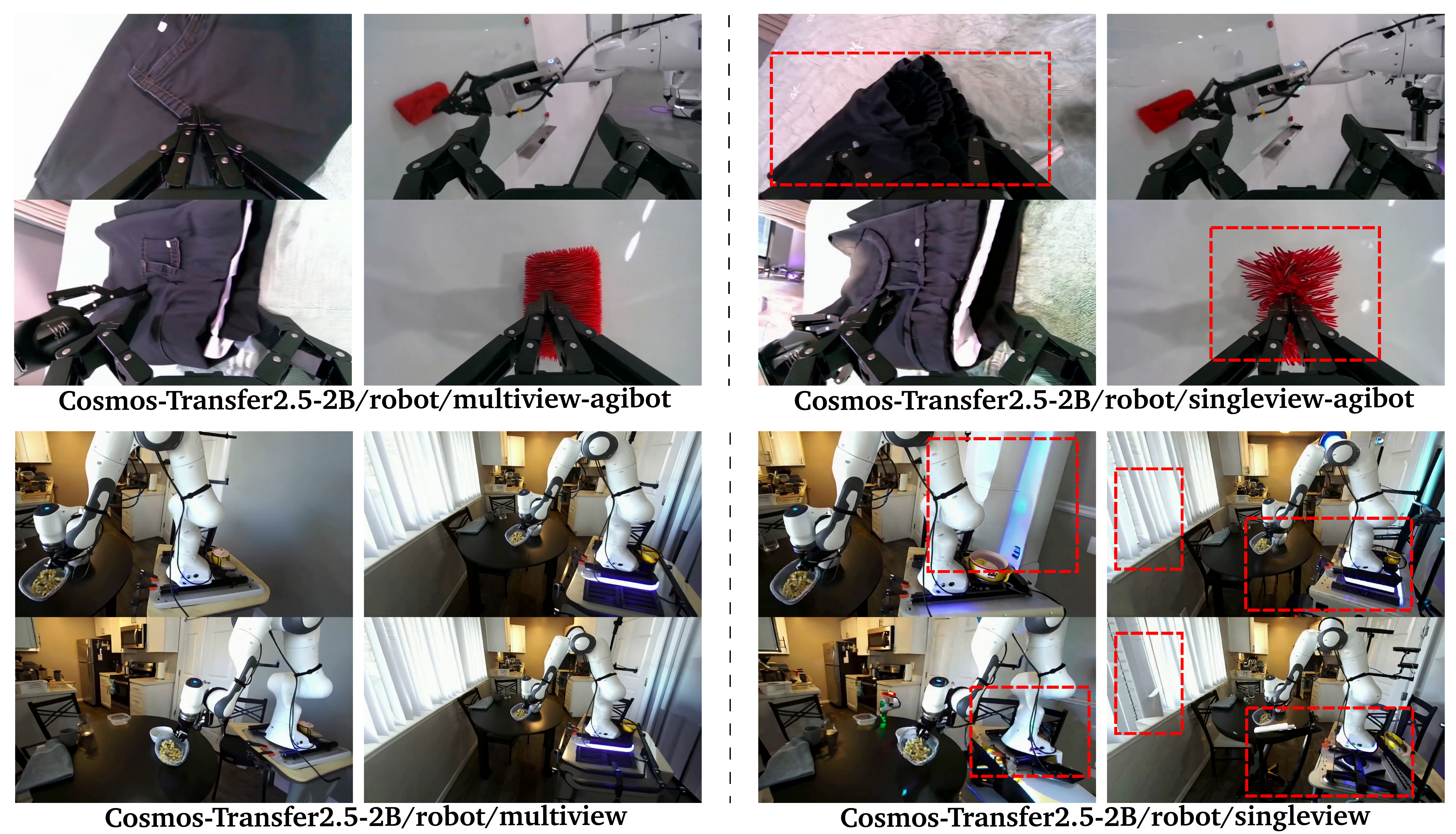}
    \caption{
    \textbf{View synchronization comparison.} [Cosmos-Transfer2.5-2B/robot/multiview] generates more coherent videos across multiple viewpoints, compared with the single-view targeted baseline (the red dotted box highlights the inconsistent parts).
    }
    \label{fig:multicam_comparison}
\end{figure}

We further evaluate the generated synchronized videos along two dimensions: (1) camera trajectory error. including rotation error and translation, which measures the error between predicted camera poses from ViPE~\citep{huang2025vipe} on the generated videos and the corresponding ground-truth poses, and (2) cross-view consistency, quantified by the Sampson error between pairs of generated views~\citep{cosmos_v1}.
Specifically, we conduct experiments on 80 validation videos with 16 camera trajectories using [Cosmos-Transfer2.5-2B/
robot/multiview]. For the baseline, we implement a single-view-to-single-view variant ([Cosmos-Transfer2.5-2B/robot/singleview]) by restricting the number of target views to a single view. As illustrated in~\cref{fig:multicam_comparison} and~\cref{tab:multicam_eval}, [Cosmos-Transfer2.5-2B/robot/multiview] achieves significantly better cross-view consistency than its single-view counterpart, while maintaining comparable camera trajectory accuracy.

\begin{table}[htb!]

\caption{\textbf{Multi-Camera Video Generation Evaluation}. We evaluate both our model and the baseline on 80 in-the-wild robotic manipulation videos across 16 diverse camera trajectories. Best is bolded.} 
\label{tab:multicam_eval}
\centering
\begin{threeparttable}
\resizebox{.86\textwidth}{!} 
{
\begin{tabular}{cccccc}
\toprule
&  \multicolumn{2}{c}{ Camera Accuracy } & \multicolumn{1}{c}{ View Synchronization} \\
\cmidrule(lr){2-3} 
\cmidrule(lr){4-4} 
Model & TransErr $\downarrow$ & RotErr (rad) $\downarrow$ &  Sampson Error (px) $\downarrow$ \\
\midrule
Cosmos-Transfer2.5-2B/robot/singleview & \textbf{0.08} & \textbf{0.19} & 26.61  \\
Cosmos-Transfer2.5-2B/robot/multiview &  \textbf{0.08} & 0.20  & \textbf{19.73}  \\
\bottomrule
\end{tabular}
}
\end{threeparttable}
\end{table}

\begin{figure}[!tp]
    \centering
\includegraphics[width=.92\textwidth]{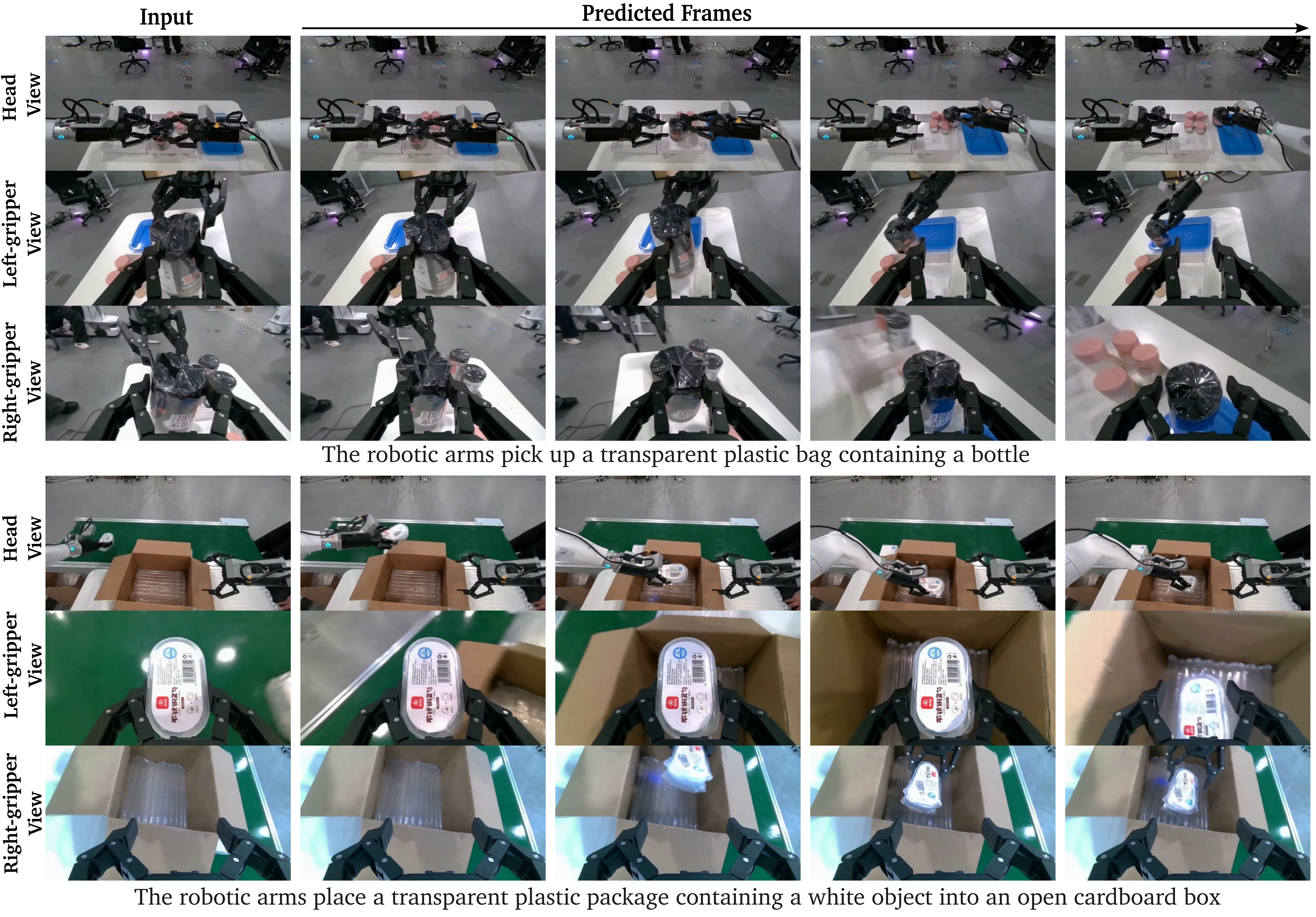}
    \caption{
    [Cosmos-Predict2.5-2B/robot/multiview-agibot] generates synchronized robotic manipulation videos conditioned on single-frame input of 3-camera views and their corresponding camera trajectories.
    }
\label{fig:multicam_predict_agibot}
\end{figure}

Beyond [Cosmos-Transfer2.5-2B/robot/multiview], we further develop \textbf{[Cosmos-Predict2.5-2B/robot/multiview-agibot]}, which takes three single images (two gripper views and one head view) along with their corresponding camera trajectories as input to generate three robotic manipulation videos, as shown in~\cref{fig:multicam_predict_agibot}. The model is trained on Agibot dataet and adopts the same architecture as [Cosmos-Transfer2.5-2B/robot/multiview]. This task facilitates the generation of diverse data for training robotic policies.

\subsection{Synthetic Data Generation for VLA training}
\label{subsec::gr00tdream}

World models show significant potential as planners and simulators for robotic manipulation. After post-training on a large video dataset of real demonstrations where robots perform actions from natural language instructions, [Cosmos-Predict2.5] can generate realistic videos of robots executing unseen commands. We can then extract pseudo-action sequences from these videos using either a latent action model or an inverse-dynamics model (IDM)~\citep{jang2025dreamgen}. This renders samples with vision (generated videos), language (instructions), and action (generated pseudo actions) annotations for VLA training. We can leverage this paradigm to generate diverse synthetic VLA training data that augments real demonstrations, thereby improving the generalization capabilities of a VLA model.

We post-train [Cosmos-Predict2.5-14B] and evaluate its performance on the synthetic VLA training data generation task using the DreamGen benchmark~\citep{jang2025dreamgen}. DreamGen examines three key dimensions of generalization---object, behavior, and environment---and employs automated evaluation with vision-language models such as Qwen-VL-2.5~\citep{qwen2p5vl} and GPT-4o~\citep{hurst2024gpt}. The benchmark specifically measures whether the generated videos accurately follow task instructions involving unseen objects, novel behaviors, or new environments.

From~\cref{tab:gr00tdream}, we found that the resulting post-trained model, [Cosmos-Predict2.5-14B/robot/gr00tdream-gr1], achieved the highest instruction-following scores on the GR1 humanoid robot dataset~\citep{jang2025dreamgen}. It outperformed competing models including Hunyuan~\citep{kong2024hunyuanvideo}, CogVideoX~\citep{yang2024cogvideox}, and WAN 2.1~\citep{wan2025}, particularly in object and environment generalization. These results highlight [Cosmos-Predict2.5-14B]’s adaptability through post-training and its ability to generate contextually accurate robot videos that faithfully realize natural language instructions.

\begin{table}[ht]
\centering
\caption{ \textbf{DreamGen Bench Statistics and Results}.
GPT represents the evaluation from GPT4o, and Qwen represents the evaluation from Qwen2.5VL. All the models are SFT models. The best is bold and the second best is underlined. [Cosmos-Predict2-14B/robot/gr00tdream-gr1] is an earlier version of [Cosmos-Predict], which did not use [Cosmos-Reason1] for text embedding.}
\label{tab:gr00tdream}
\small
\begin{tabular}{l|cc|cc|cc}
\toprule
\multicolumn{7}{c}{\textbf{DreamGen Bench GR1 Instruction Following}} \\
\midrule
& \multicolumn{2}{c|}{\textbf{Object}} & \multicolumn{2}{c|}{\textbf{Behavior}} & \multicolumn{2}{c}{\textbf{Env}} \\
& GPT & Qwen & GPT & Qwen & GPT & Qwen \\
\midrule
Hunyuan & 38.0 & 26.0 & 38.3 & 10.6 & 27.6 & 27.6 \\
CogVideoX & 72.0 & 38.0 & 44.0 & 28.0 & \underline{55.2} & 41.4 \\
WAN2.1 & 72.0 & 58.0 & \textbf{72.3} & 55.3 & 48.3 & \underline{65.5} \\
Cosmos-Predict2-14B/robot/gr00tdream-gr1 & \underline{90.0} & \underline{62.0} & 59.6 & \textbf{61.7} & \textbf{69.0} & \underline{65.5} \\
Cosmos-Predict2.5-14B/robot/gr00tdream-gr1 & \textbf{91.8} & \textbf{69.4} & \underline{70.2} & \underline{59.6} & \textbf{69.0} & \textbf{69.0} \\
\bottomrule
\end{tabular}
\end{table}

\subsection{Action-Conditioned World Generation}
\label{subsec::action-cond}

We extend [Cosmos-Predict2.5] from pure video generation to action-conditioned video generation, resulting in [Cosmos-Predict2.5-2B/robot/action-cond]. The model takes as input a single conditional image together with a sequence of robot actions, and generates a chunk of future frames that follow the provided action sequence. To produce full trajectories, generation is carried out in an autoregressive manner, where each chunk is predicted conditioned on the last generated frame.

Because actions represent a new modality not present during pre-training, we introduce additional modules for conditioning. Specifically, we add an action embedder MLP that maps each action into a tensor. Instead of injecting this tensor directly, we incorporate it by adding it to the timestamp embeddings of the DiT modules.

\begin{table}[ht]
    \setlength{\tabcolsep}{16.12pt}
    \small
    \captionsetup{justification=centering}
    \caption{Evaluation of action-conditioned video prediction on Bridge dataset.}
    \centering
    \begin{tabular}{rcccc}
        \toprule
        Method & {PSNR} $\uparrow$ & {SSIM} $\uparrow$ & {Latent L2} $\downarrow$ & {FVD} $\downarrow$ \\
        \midrule
        \makecell[r]{Cosmos-Predict1-7B-Video2World-\\Sample-ActionCond} & {21.14} & {0.82} & {0.32} & {190} \\
        \midrule
        \makecell[r]{Cosmos-Predict2.5-2B/robot/action-cond} & \textbf{24.95} & \textbf{0.85} & \textbf{0.28} & \textbf{146} \\
        \bottomrule
    \end{tabular}
    \label{tab:action_conditioned_quantative}
\end{table}
\begin{figure*}[th!]
    \centering
    \setlength{\tabcolsep}{2pt}
    \renewcommand{\arraystretch}{0.9} 
    \begin{tabular}{cccccc} 
         \footnotesize{Input frame} & & \multicolumn{4}{c}{\footnotesize Predicted frames} \\
         \includegraphics[width=0.19\textwidth]{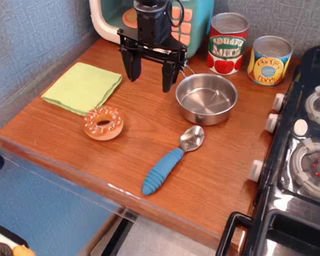} & &
         \includegraphics[width=0.19\textwidth]{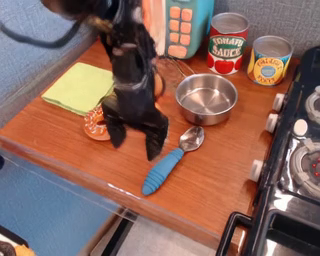} &
         \includegraphics[width=0.19\textwidth]{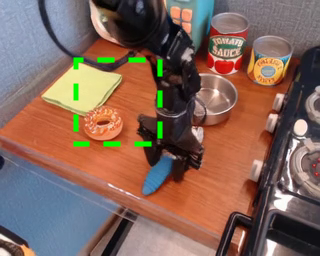} &
         \includegraphics[width=0.19\textwidth]{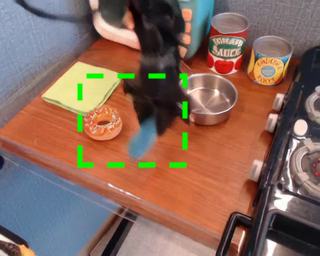} &
         \includegraphics[width=0.19\textwidth]{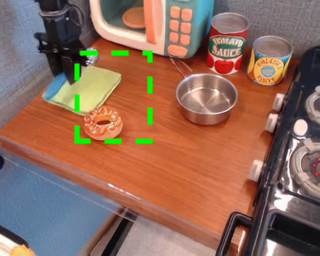} \\
         \multicolumn{6}{c}{\footnotesize Cosmos-Predict2.5-2B/robot/action-cond} \\
         \includegraphics[width=0.19\textwidth]{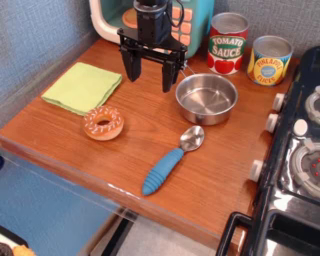} & &
         \includegraphics[width=0.19\textwidth]{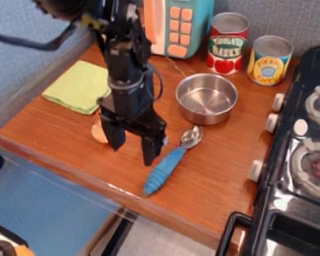} &
         \includegraphics[width=0.19\textwidth]{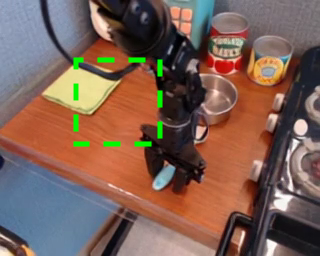} &
         \includegraphics[width=0.19\textwidth]{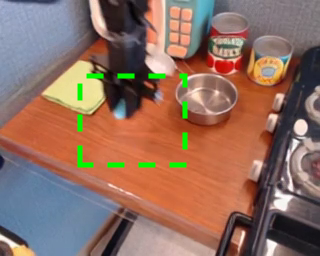} &
         \includegraphics[width=0.19\textwidth]{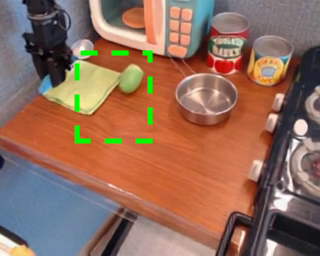} \\
         \multicolumn{6}{c}{\footnotesize Cosmos-Predict1-7B-Video2World-Sample-ActionCond} \\
         \includegraphics[width=0.19\textwidth]{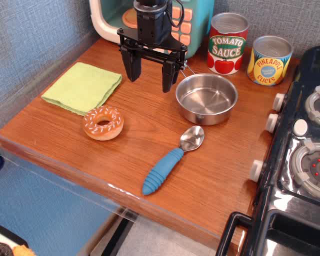} & &
         \includegraphics[width=0.19\textwidth]{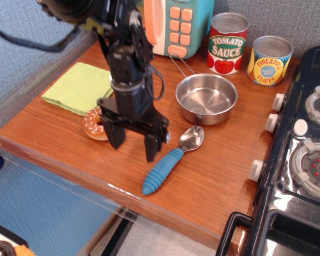} &
         \includegraphics[width=0.19\textwidth]{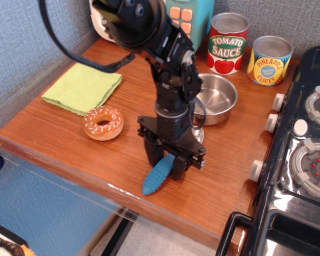} &
         \includegraphics[width=0.19\textwidth]{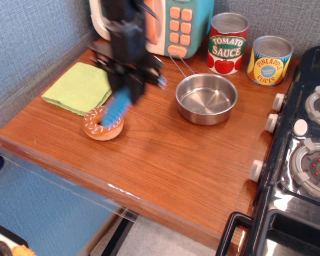} &
         \includegraphics[width=0.19\textwidth]{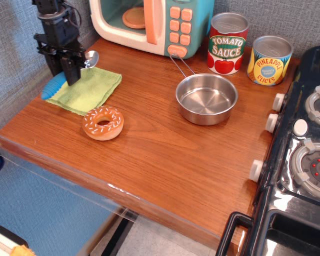} \\
         \multicolumn{6}{c}{\footnotesize Ground Truth (GT)} \\
    \end{tabular}
    \caption{\textbf{Action-conditioned video prediction samples on the Bridge dataset.} Comparison of predicted rollouts from [Cosmos-Predict2.5-2B/robot/action-cond] and [Cosmos-Predict1-7B-Video2World-Sample-ActionCond] against the ground-truth frames. [Cosmos-Predict2.5-2B/robot/action-cond] demonstrates better object permanence. The green dotted box highlights the parts with the object permanence issues.}
    \label{fig:action_conditioned_qualitative}
\end{figure*}

\noindent \textbf{Experiments.}
We conduct experiments using the public Bridge dataset~\citep{walke2023bridgedata} following prior work~\citep{zhu2024irasim}. The dataset contains approximately 20,000 episodes of third-person videos capturing a robot arm performing various tasks in a kitchen environment. Each video has a resolution of $320\times256$ and is recorded at 5 FPS. Corresponding to each frame, the robot action is represented as a 7-dimensional vector in the gripper coordinate space: $(\Delta x, \Delta y, \Delta z, \Delta \theta_r, \Delta \theta_p, \Delta \theta_y, \mbox{GripperWidth})$, which specifies the relative displacement, rotation, and width of the gripper.

To evaluate the quality of video generation, we randomly sample 100 episodes from the official Bridge test set and generate videos for them, comparing the results against the ground-truth videos. We use [Cosmos-Predict1-7B-Video2World-Sample-ActionCond] as a baseline for comparison.

The quantitative metrics, summarized in \cref{tab:action_conditioned_quantative}, include PSNR, SSIM, Latent L2~\citep{zhu2024irasim}, and FVD. As shown, the [Cosmos-Predict2.5-2B/robot/action-cond] models outperform the baseline across all metrics. Selected predicted video frames are presented in~\cref{fig:action_conditioned_qualitative}, highlighting the high quality of the predictions relative to the ground-truth frames. 

\begin{table}[ht]
    \setlength{\tabcolsep}{16.12pt} 
    \small
    \captionsetup{justification=centering}
    \caption{Ablation study on the Bridge dataset. The results show that incorporating action conditioning with time embeddings yields better action-conditioned video generation performance.}
    \centering
    \begin{tabular}{rcccc}
        \toprule
        Method & {PSNR} $\uparrow$ & {SSIM} $\uparrow$ & {Latent L2} $\downarrow$ & {FVD} $\downarrow$ \\
        \midrule
        \makecell[r]{Cosmos-Predict2.5-2B/robot/action-cond\\with TimeEmbedding (proposed)} & \textbf{24.95} & \textbf{0.85} & \textbf{0.28} & \textbf{146} \\
        \midrule
        \makecell[r]{Cosmos-Predict2.5-2B/robot/action-cond\\with CrossAtten} & {24.41} & {0.84} & \textbf{0.28} & {159} \\
        \midrule
        \makecell[r]{Cosmos-Predict2.5-2B/robot/action-cond\\with ChannelConcat} & 23.11 & 0.78 & 0.35 & 267 \\
        \bottomrule
    \end{tabular}
    \label{tab:action_conditioned_ablation}
\end{table}

We further investigate various methods for incorporating action conditioning. In addition to applying it through time embeddings, we also explore two alternatives: (1) cross-attention within the DiT blocks and (2) channel concatenation. The results are presented in \cref{tab:action_conditioned_ablation}.

\section{Related Work}
\label{sec::related_work}

\textbf{World Models.} Recent years have seen growing interest in world models that learn to predict future states from current observations and potential actions, enabling more efficient decision-making and planning \citep{ha2018world}. Research in this area has evolved into two primary paradigms for modeling world dynamics. The first focuses on learning predictive models in abstract, latent representation spaces \citep{ha2018world,hafner2019dream, assran2025v, chen2025planning}. These approaches aim to compress high-dimensional sensory inputs into compact, learned state representations that preserve the essential structure of the environment, thereby enabling efficient and tractable planning. In contrast, the second paradigm, the one we adopt, centers on modeling world dynamics directly in pixel space through high-fidelity video prediction as a video generative model~\citep{sora,cosmos_v1,genie3}. These models simulate future observations frame-by-frame and can be extended to incorporate various control signals, such as camera pose, action sequences, and spatially dense inputs like a world scenario map. This retains rich, high-fidelity information, making our models effective synthetic data generators for downstream policy learning, while also remaining flexible enough to be extended to support diverse control signals. In addition to these two dominant approaches, a third, emerging direction explores native 3D and 4D representations of world states, using either neural scene representations or physically grounded simulators \citep{singer2023text, watson2024controlling, zhao2024genxd, liu2025dynamicscaler, li2025wonderplay,nasiriany2024robocasa}. These models aim to provide a more structured and geometric-aware understanding of the environment.

\textbf{Video Generative Models.} Video generative models represent a rapidly advancing frontier in generative AI. In recent years, several powerful closed-source systems---such as Sora \citep{sora}, Kling \citep{Kling}, \citep{gen3}, Hailuo \citep{minimax}, MovieGen \citep{polyak2024movie}, Seedance \citep{gao2025seedance}, Veo \citep{veo3}, and Waver \citep{zhang2025waver}---have demonstrated remarkable progress in general-purpose video generation. Despite their impressive capabilities, the proprietary nature of these models poses a significant barrier to research and downstream applications. The lack of access to model weights and training code prevents the broader community from fine-tuning, extending, or adapting these models for specialized use cases such as autonomous driving and robotics. In contrast, the emergence of open-source video generation models, including Wan \citep{wan2025}, LTX \citep{hacohen2024ltx}, and Hunyuan \citep{kong2024hunyuanvideo}, has fostered greater transparency and accessibility. These models enable reproducible research and community-driven innovation. However, most remain optimized for general-purpose content creation and often fall short in domains requiring precise, fine-grained control over object dynamics, interactions, and physical consistency—capabilities essential for advancing physical AI. [Cosmos-Predict1]~\citep{cosmos_v1} represents the first open-source video generative model explicitly tailored for physical AI applications. In this work, we further enhance its capabilities by training it on a high-quality, domain-specific dataset curated for the complexities of physical reasoning. Additionally, we integrate a text encoder based on [Cosmos-Reason1]~\citep{azzolini2025cosmos}, our vision-language foundation model designed specifically for physical AI tasks. This integration significantly improves the model's ability to generate physically plausible and controllable video sequences conditioned on natural language descriptions.

\textbf{Foundation World Model for Physical AI.} Most existing world models, whether closed-source or open-source, regardless of technical approaches, focus on general content generation in the digital world, e.g., movies and computer games. The introduction of [Cosmos-Predict1]~\citep{cosmos_v1} and [Cosmos-Transfer1]~\citep{cosmos_transfer1} brings the first batch of open-source world models in Physical AI. It has facilitated the development of open evaluation benchmarks for both general-purpose world generation \citep{duan2025worldscore, zhou2025vlm4d, zhao2025synthetic} and specialized domains such as physics \citep{bansal2024videophy, motamed2025generative, guo2025t2vphysbench, bansal2025videophy, bordes2025intphys} and Embodied AI \citep{yang2025embodiedbench, liao2025genie, yue2025ewmbench}. As foundation world models, [Cosmos-Predict1] was post-trained to enable new capabilities, including camera control \citep{ren2025gen3c}, motion trajectory control \citep{wang2025frame}, and video relighting \citep{he2025unirelight}. It has also been used as a synthetic data generation engine for robot policy training \citep{jang2025dreamgen, bjorck2025gr00t} and development of autonomous driving systems \citep{ren2025cosmos, fu2025llm}. In this work, we demonstrated the enhanced capabilities of [Cosmos-Predict2.5] for VLA model training, robot policy training/validation, autonomous driving simulation, and robotic manipulation. [Cosmos-Transfer2.5] also improved upon its predecessors for long-horizon video translations and closed-loop simulation. We hope the open-source of [Cosmos-Predict2.5] and [Cosmos-Transfer2.5] can continue facilitating development and innovation within the Physical AI community. 

\section{Conclusion}
\label{sec::conclusion}

We presented [Cosmos-Predict2.5] and [Cosmos-Transfer2.5], the latest Cosmos video world foundation models for Physical AI. Leveraging large-scale curated video datasets, flow-matching training, improved text embedding, domain-specific post-training, and reinforcement learning, our models achieve leading results on Physical AI benchmarks. Beyond benchmarks, we demonstrated their effectiveness in robotics and autonomous driving, where high-fidelity synthetic video is essential. By releasing models and code, we aim to establish Cosmos as a world foundation model platform for a simulation-first ecosystem that advances Physical AI and bridges the gap between simulation and real-world deployment.

\clearpage
\appendix
\section{Contributors and Acknowledgments}
\label{sec::contributors}

\subsection{Contributors} 

\begin{multicols}{4}
\setlength{\parindent}{0pt}
\footnotesize
\raggedright
Arslan Ali \\
Junjie Bai \\
Maciej Bala \\
Yogesh Balaji \\
Aaron Blakeman \\
Tiffany Cai \\
Jiaxin Cao \\
Tianshi Cao \\
Elizabeth Cha \\
Yu-Wei Chao \\
Prithvijit Chattopadhyay \\
Mike Chen \\
Yongxin Chen \\
Yu Chen \\
Shuai Cheng \\
Yin Cui \\
Jenna Diamond \\
Yifan Ding \\
Jiaojiao Fan \\
Linxi Fan \\
Liang Feng \\
Francesco Ferroni \\
Sanja Fidler \\
Xiao Fu \\
Ruiyuan Gao \\
Yunhao Ge \\
Jinwei Gu \\
Aryaman Gupta \\
Siddharth Gururani \\
Imad El Hanafi \\
Ali Hassani \\
Zekun Hao \\
Jacob Huffman \\
Joel Jang \\
Pooya Jannaty \\
Jan Kautz \\
Grace Lam \\
Xuan Li \\
Zhaoshuo Li \\
Maosheng Liao \\
Chen-Hsuan Lin \\
Tsung-Yi Lin \\
Yen-Chen Lin \\
Huan Ling \\
Ming-Yu Liu \\
Xian Liu \\
Yifan Lu \\
Alice Luo \\
Qianli Ma \\
Hanzi Mao \\
Kaichun Mo \\
Seungjun Nah \\
Yashraj Narang \\
Abhijeet Panaskar \\
Lindsey Pavao \\
Trung Pham \\
Morteza Ramezanali \\
Fitsum Reda \\
Scott Reed \\
Xuanchi Ren \\
Haonan Shao \\
Yue Shen \\
Stella Shi \\
Shuran Song \\
Bartosz Stefaniak \\
Shangkun Sun \\
Shitao Tang \\
Sameena Tasmeen \\
Lyne Tchapmi \\
Wei-Cheng Tseng \\
Jibin Varghese \\
Andrew Z. Wang \\
Hao Wang \\
Haoxiang Wang \\
Heng Wang \\
Ting-Chun Wang \\
Fangyin Wei \\
Jiashu Xu \\
Dinghao Yang \\
Xiaodong Yang \\
Haotian Ye \\
Seonghyeon Ye \\
Xiaohui Zeng \\
Jing Zhang \\
Qinsheng Zhang \\
Kaiwen Zheng \\
Andrew Zhu \\
Yuke Zhu \\
\end{multicols}

\subsection{Acknowledgments}

\begin{multicols}{4}
\setlength{\parindent}{0pt}
\footnotesize
\raggedright
Chris Alexiuk \\
Jon Allen \\
Charles Anderson \\
Vince Auletta \\
Joshua Bapst \\
Sanchit Bhattacharjee \\
Alexis Bjorlin \\
Dan Blick \\
Aditi Bodhankar \\
Jeremy Bottleson \\
HJ Chen \\
Rui Chen \\
Tiffany Chen \\
Wenkai Chen \\
Nuttapong Chentanez \\
Himanshu Chodhary \\
Tae Eun Choe \\
Jeana Choi \\
Ashley Chow \\
Guillermo Garcia Cobo \\
Marek Dabek \\
John Dickinson \\
Naomi Eigbe \\
Damien Fagnou \\
Dominik Farhan \\
Amol Fasale \\
Mohamed Fawzy \\
Sergiy Fefilatyev \\
TJ Galda \\
Abhinav Garg \\
Vlad Getselevich \\
Pengfei Guo \\
Aryaman Gupta \\
Brett Hamilton \\
Mohammad Harrim \\
Milos Hasan \\
Nathan Hayes-Roth \\
Yufan He \\
Logan Herche \\
Sophia Huang \\
Spencer Huang \\
Ryan Ji \\
Yangqing Jia \\
Brendan Johnson \\
Jaehyun Jung \\
Sagar Karale \\
Artur Kasymov \\
Arnav Khanna \\
Will Kim \\
Carsten Kolve \\
Reka Kovacs \\
Christopher Labis \\
Christian Laforte \\
Ti Leggett \\
Alice Li \\
Xiang Li \\
Edy Lim \\
Daniel Lindsey \\
Aditya Mahajan \\
Nikolay Matveiev \\
Amanda Moran \\
Jashojit Mukherjee \\
Apurv Naman \\
Nigel Nelson \\
Merlin Nimier-David \\
Sangmin Oh \\
Ruben Ohana \\
Shubham Pachori \\
Sravan Patchala \\
Mahesh Patekar \\
Mitesh Patel \\
Joel Pennington \\
Sahil Ramani \\
Vaibhav Ranglani \\
Wojciech Rymer \\
Amirmojtaba Sabour \\
Jun Saito \\
Akul Santhosh \\
Chris Schultz \\
Alexander Schwarz \\
Gautham Sholingar \\
Mateusz Sieniawski \\
Rajat Vikram Singh \\
Hariharan Srinivasan \\
Javier Gamazo Tejero \\
Ruchik Thaker \\
Peter Udvardi \\
Gandhi Vaithilingam \\
George Vine \\
Thomas Volk \\
Raju Wagwani \\
Mengmeng Xiao \\
Ning Xu \\
Lixiao Yang \\
Wei Yang \\
Jianhe Yuan \\
Itai Zadok \\
Zheng Zeng \\
Suyue Zhang \\
Hongning Zhao \\
Fengzhe Zhou \\
Andrew Zhu \\
\end{multicols}

\clearpage
\setcitestyle{numbers}
\bibliographystyle{plainnat}
\bibliography{main}

@article{liu2025improving,
  title={Improving video generation with human feedback},
  author={Liu, Jie and Liu, Gongye and Liang, Jiajun and Yuan, Ziyang and Liu, Xiaokun and Zheng, Mingwu and Wu, Xiele and Wang, Qiulin and Qin, Wenyu and Xia, Menghan and others},
  journal={arXiv preprint arXiv:2501.13918},
  year={2025}
}

@article{guo2025deepseek,
  title={Deepseek-R1: Incentivizing reasoning capability in llms via reinforcement learning},
  author={Guo, Daya and Yang, Dejian and Zhang, Haowei and Song, Junxiao and Zhang, Ruoyu and Xu, Runxin and Zhu, Qihao and Ma, Shirong and Wang, Peiyi and Bi, Xiao and others},
  journal={arXiv preprint arXiv:2501.12948},
  year={2025}
}

@article{schulman2017proximal,
  title={Proximal policy optimization algorithms},
  author={Schulman, John and Wolski, Filip and Dhariwal, Prafulla and Radford, Alec and Klimov, Oleg},
  journal={arXiv preprint arXiv:1707.06347},
  year={2017}
}

@article{ouyang2022training,
  title={Training language models to follow instructions with human feedback},
  author={Ouyang, Long and Wu, Jeffrey and Jiang, Xu and Almeida, Diogo and Wainwright, Carroll and Mishkin, Pamela and Zhang, Chong and Agarwal, Sandhini and Slama, Katarina and Ray, Alex and others},
  journal={NeurIPS},
  year={2022}
}

@inproceedings{gao2025diffusion,
  title={Diffusion models and gaussian flow matching: Two sides of the same coin},
  author={Gao, Ruiqi and Hoogeboom, Emiel and Heek, Jonathan and De Bortoli, Valentin and Murphy, Kevin Patrick and Salimans, Tim},
  booktitle={The Fourth Blogpost Track at ICLR 2025},
  year={2025}
}

@inproceedings{wang2025comprehensivestudydecoderonlyllms,
  title={A Comprehensive Study of Decoder-Only LLMs for Text-to-Image Generation},
  author={Wang, Andrew Z and Ge, Songwei and Karras, Tero and Liu, Ming-Yu and Balaji, Yogesh},
  booktitle={CVPR},
  year={2025}
}

@inproceedings{dover,
  title={Exploring video quality assessment on user generated contents from aesthetic and technical perspectives},
  author={Wu, Haoning and Zhang, Erli and Liao, Liang and Chen, Chaofeng and Hou, Jingwen and Wang, Annan and Sun, Wenxiu and Yan, Qiong and Lin, Weisi},
  booktitle={CVPR},
  year={2023}
}

@inproceedings{koala,
  title={Koala-36m: A large-scale video dataset improving consistency between fine-grained conditions and video content},
  author={Wang, Qiuheng and Shi, Yukai and Ou, Jiarong and Chen, Rui and Lin, Ke and Wang, Jiahao and Jiang, Boyuan and Yang, Haotian and Zheng, Mingwu and Tao, Xin and others},
  booktitle={CVPR},
  year={2025}
}

@misc{delta_lake_databricks_2019,
  author       = {{Databricks}},
  title        = {Delta Lake: Open-source storage framework that enables building Lakehouses},
  year         = {2019},
  howpublished = {\url{https://delta.io/}},
  note         = {Open-source project, Delta Lake}
}

@misc{milvus_zilliz_2019,
  author       = {{Zilliz}},
  title        = {Milvus: An open-source vector database for scalable similarity search},
  year         = {2019},
  howpublished = {\url{https://milvus.io/}},
  note         = {Open-source project, Milvus}
}

@article{chen2023importance,
  title={On the importance of noise scheduling for diffusion models},
  author={Chen, Ting},
  journal={arXiv preprint arXiv:2301.10972},
  year={2023}
}

@misc{bloc97_ntkaware_scaled_rope_2023,
  author       = {{bloc97}},
  title        = {NTK-Aware Scaled RoPE allows LLaMA models to have extended (8k+) context size without any fine-tuning and minimal perplexity degradation},
  year         = {2023},
  howpublished = {\url{https://www.reddit.com/r/LocalLLaMA/comments/14lz7j5/ntkaware_scaled_rope_allows_llama_models_to_have/}},
  note         = {Reddit post, r/LocalLLaMA}
}

@article{peng2023yarn,
  title={Yarn: Efficient context window extension of large language models},
  author={Peng, Bowen and Quesnelle, Jeffrey and Fan, Honglu and Shippole, Enrico},
  journal={arXiv preprint arXiv:2309.00071},
  year={2023}
}

@article{atzmon2024edify,
  title={Edify image: High-quality image generation with pixel space laplacian diffusion models},
  author={Atzmon, Yuval and Bala, Maciej and Balaji, Yogesh and Cai, Tiffany and Cui, Yin and Fan, Jiaojiao and Ge, Yunhao and Gururani, Siddharth and Huffman, Jacob and Isaac, Ronald and others},
  journal={arXiv preprint arXiv:2411.07126},
  year={2024}
}

@inproceedings{hoogeboom2023simple,
  title={simple diffusion: End-to-end diffusion for high resolution images},
  author={Hoogeboom, Emiel and Heek, Jonathan and Salimans, Tim},
  booktitle={ICML},
  year={2023}
}

@article{lipman2022flow,
  title={Flow matching for generative modeling},
  author={Lipman, Yaron and Chen, Ricky TQ and Ben-Hamu, Heli and Nickel, Maximilian and Le, Matt},
  journal={arXiv preprint arXiv:2210.02747},
  year={2022}
}

@article{karras2022elucidating,
  title={Elucidating the design space of diffusion-based generative models},
  author={Karras, Tero and Aittala, Miika and Aila, Timo and Laine, Samuli},
  journal={NeurIPS},
  year={2022}
}

@article{wan2025,
  title={Wan: Open and advanced large-scale video generative models},
  author={Wan, Team and Wang, Ang and Ai, Baole and Wen, Bin and Mao, Chaojie and Xie, Chen-Wei and Chen, Di and Yu, Feiwu and Zhao, Haiming and Yang, Jianxiao and others},
  journal={arXiv preprint arXiv:2503.20314},
  year={2025}
}

@article{jang2025dreamgen,
  title={DreamGen: Unlocking Generalization in Robot Learning through Video World Models},
  author={Jang, Joel and Ye, Seonghyeon and Lin, Zongyu and Xiang, Jiannan and Bjorck, Johan and Fang, Yu and Hu, Fengyuan and Huang, Spencer and Kundalia, Kaushil and Lin, Yen-Chen and others},
  journal={arXiv preprint arXiv:2505.12705},
  year={2025}
}

@article{azzolini2025cosmos,
  title={Cosmos-reason1: From physical common sense to embodied reasoning},
  author={NVIDIA},
  journal={arXiv preprint arXiv:2503.15558},
  year={2025}
}

@inproceedings{chen2025videodepth,
  title={Video depth anything: Consistent depth estimation for super-long videos},
  author={Chen, Sili and Guo, Hengkai and Zhu, Shengnan and Zhang, Feihu and Huang, Zilong and Feng, Jiashi and Kang, Bingyi},
  booktitle={CVPR},
  year={2025}
}

@article{ravi2024sam,
  title={Sam 2: Segment anything in images and videos},
  author={Ravi, Nikhila and Gabeur, Valentin and Hu, Yuan-Ting and Hu, Ronghang and Ryali, Chaitanya and Ma, Tengyu and Khedr, Haitham and R{\"a}dle, Roman and Rolland, Chloe and Gustafson, Laura and others},
  journal={arXiv preprint arXiv:2408.00714},
  year={2024}
}

@article{liu2023grounding,
  title={Grounding dino: Marrying dino with grounded pre-training for open-set object detection},
  author={Liu, Shilong and Zeng, Zhaoyang and Ren, Tianhe and Li, Feng and Zhang, Hao and Yang, Jie and Li, Chunyuan and Yang, Jianwei and Su, Hang and Zhu, Jun and others},
  journal={arXiv preprint arXiv:2303.05499},
  year={2023}
}

@article{ren2024grounded,
  title={Grounded sam: Assembling open-world models for diverse visual tasks},
  author={Ren, Tianhe and Liu, Shilong and Zeng, Ailing and Lin, Jing and Li, Kunchang and Cao, He and Chen, Jiayu and Huang, Xinyu and Chen, Yukang and Yan, Feng and others},
  journal={arXiv preprint arXiv:2401.14159},
  year={2024}
}

@inproceedings{bu2025agibot,
  title={Agibot world colosseo: A large-scale manipulation platform for scalable and intelligent embodied systems},
  author={Bu, Qingwen and Cai, Jisong and Chen, Li and Cui, Xiuqi and Ding, Yan and Feng, Siyuan and Gao, Shenyuan and He, Xindong and Hu, Xuan and Huang, Xu and others},
  booktitle={IROS},
  year={2025}
}

@article{khazatsky2024droid,
  title={Droid: A large-scale in-the-wild robot manipulation dataset},
  author={Khazatsky, Alexander and Pertsch, Karl and Nair, Suraj and Balakrishna, Ashwin and Dasari, Sudeep and Karamcheti, Siddharth and Nasiriany, Soroush and Srirama, Mohan Kumar and Chen, Lawrence Yunliang and Ellis, Kirsty and others},
  journal={arXiv preprint arXiv:2403.12945},
  year={2024}
}

@inproceedings{bai2024syncammaster,
  title={Syncammaster: Synchronizing multi-camera video generation from diverse viewpoints},
  author={Bai, Jianhong and Xia, Menghan and Wang, Xintao and Yuan, Ziyang and Fu, Xiao and Liu, Zuozhu and Hu, Haoji and Wan, Pengfei and Zhang, Di},
  booktitle={ICLR},
  year={2025}
}

@inproceedings{bai2025recammaster,
  title={Recammaster: Camera-controlled generative rendering from a single video},
  author={Bai, Jianhong and Xia, Menghan and Fu, Xiao and Wang, Xintao and Mu, Lianrui and Cao, Jinwen and Liu, Zuozhu and Hu, Haoji and Bai, Xiang and Wan, Pengfei and others},
  booktitle={ICCV},
  year={2025}
}

@article{sitzmann2021light,
  title={Light field networks: Neural scene representations with single-evaluation rendering},
  author={Sitzmann, Vincent and Rezchikov, Semon and Freeman, Bill and Tenenbaum, Josh and Durand, Fredo},
  journal={NeurIPS},
  year={2021}
}

@inproceedings{rasley2020deepspeed,
  title={Deepspeed: System optimizations enable training deep learning models with over 100 billion parameters},
  author={Rasley, Jeff and Rajbhandari, Samyam and Ruwase, Olatunji and He, Yuxiong},
  booktitle={KDD},
  year={2020}
}

@inproceedings{
   liang2025torchtitan,
   title={TorchTitan: One-stop PyTorch native solution for production ready {LLM} pretraining},
   author={Wanchao Liang and Tianyu Liu and Less Wright and Will Constable and Andrew Gu and Chien-Chin Huang and Iris Zhang and Wei Feng and Howard Huang and Junjie Wang and Sanket Purandare and Gokul Nadathur and Stratos Idreos},
   booktitle={ICLR},
   year={2025}
}

@article{lu2024simplifying,
  title={Simplifying, stabilizing and scaling continuous-time consistency models},
  author={Lu, Cheng and Song, Yang},
  journal={arXiv preprint arXiv:2410.11081},
  year={2024}
}

@inproceedings{wang2024internvideo2,
  title={Internvideo2: Scaling foundation models for multimodal video understanding},
  author={Wang, Yi and Li, Kunchang and Li, Xinhao and Yu, Jiashuo and He, Yinan and Chen, Guo and Pei, Baoqi and Zheng, Rongkun and Wang, Zun and Shi, Yansong and others},
  booktitle={ECCV},
  year={2024}
}

@article{qwen2p5vl,
  title={Qwen2.5-VL Technical Report},
  author={Bai, Shuai and Chen, Keqin and Liu, Xuejing and Wang, Jialin and Ge, Wenbin and Song, Sibo and Dang, Kai and Wang, Peng and Wang, Shijie and Tang, Jun and others},
  journal={arXiv preprint arXiv:2502.13923},
  year={2025}
}

@inproceedings{walke2023bridgedata,
  title={Bridgedata v2: A dataset for robot learning at scale},
  author={Walke, Homer Rich and Black, Kevin and Zhao, Tony Z and Vuong, Quan and Zheng, Chongyi and Hansen-Estruch, Philippe and He, Andre Wang and Myers, Vivek and Kim, Moo Jin and Du, Max and others},
  booktitle={CoRL},
  year={2023},
}

@article{cosmos_v1,
  title={Cosmos world foundation model platform for physical ai},
  author={NVIDIA},
  journal={arXiv preprint arXiv:2501.03575},
  year={2025}
}

@article{cosmos_transfer1,
  title={Cosmos-Transfer1: Conditional World Generation with Adaptive Multimodal Control}, 
  author={NVIDIA},
  year={2025},
  journal={arXiv preprint arXiv:2503.14492},
}

@article{hassani2025generalized,
  title={Generalized Neighborhood Attention: Multi-dimensional Sparse Attention at the Speed of Light},
  author={Hassani, Ali and Zhou, Fengzhe and Kane, Aditya and Huang, Jiannan and Chen, Chieh-Yun and Shi, Min and Walton, Steven and Hoehnerbach, Markus and Thakkar, Vijay and Isaev, Michael and others},
  journal={arXiv preprint arXiv:2504.16922},
  year={2025}
}

@inproceedings{esser2024scaling,
  title={Scaling rectified flow transformers for high-resolution image synthesis},
  author={Esser, Patrick and Kulal, Sumith and Blattmann, Andreas and Entezari, Rahim and M{\"u}ller, Jonas and Saini, Harry and Levi, Yam and Lorenz, Dominik and Sauer, Axel and Boesel, Frederic and others},
  booktitle={ICML},
  year={2024}
}

@article{yang2024model,
  title={Model merging in llms, mllms, and beyond: Methods, theories, applications and opportunities},
  author={Yang, Enneng and Shen, Li and Guo, Guibing and Wang, Xingwei and Cao, Xiaochun and Zhang, Jie and Tao, Dacheng},
  journal={arXiv preprint arXiv:2408.07666},
  year={2024}
}

@inproceedings{wortsman2022model,
  title={Model soups: averaging weights of multiple fine-tuned models improves accuracy without increasing inference time},
  author={Wortsman, Mitchell and Ilharco, Gabriel and Gadre, Samir Ya and Roelofs, Rebecca and Gontijo-Lopes, Raphael and Morcos, Ari S and Namkoong, Hongseok and Farhadi, Ali and Carmon, Yair and Kornblith, Simon and others},
  booktitle={ICML},
  year={2022}
}

@article{yadav2023ties,
  title={Ties-merging: Resolving interference when merging models},
  author={Yadav, Prateek and Tam, Derek and Choshen, Leshem and Raffel, Colin A and Bansal, Mohit},
  journal={NeurIPS},
  year={2023}
}

@inproceedings{yu2024language,
  title={Language Models are Super Mario: Absorbing Abilities from Homologous Models as a Free Lunch},
  author={Yu, Le and Yu, Bowen and Yu, Haiyang and Huang, Fei and Li, Yongbin},
  booktitle={ICML},
  year={2024}
}

@inproceedings{vuong2023open,
  title={Open x-embodiment: Robotic learning datasets and rt-x models},
  author={Vuong, Quan and Levine, Sergey and Walke, Homer Rich and Pertsch, Karl and Singh, Anikait and Doshi, Ria and Xu, Charles and Luo, Jianlan and Tan, Liam and Shah, Dhruv and others},
  booktitle={Towards Generalist Robots: Learning Paradigms for Scalable Skill Acquisition@ CoRL2023},
  year={2023}
}

@article{wu2024robomind,
  title={Robomind: Benchmark on multi-embodiment intelligence normative data for robot manipulation},
  author={Wu, Kun and Hou, Chengkai and Liu, Jiaming and Che, Zhengping and Ju, Xiaozhu and Yang, Zhuqin and Li, Meng and Zhao, Yinuo and Xu, Zhiyuan and Yang, Guang and others},
  journal={arXiv preprint arXiv:2412.13877},
  year={2024}
}

@misc{1XTechno4,
author = {1X Technologies},
  title = {1X Technologies | Safe humanoids for the home},
  url = "https://www.1x.tech/",
  year = {2025}
}

@article{chen2025planning,
  title={Planning with Reasoning using Vision Language World Model},
  author={Chen, Delong and Moutakanni, Theo and Chung, Willy and Bang, Yejin and Ji, Ziwei and Bolourchi, Allen and Fung, Pascale},
  journal={arXiv preprint arXiv:2509.02722},
  year={2025}
}

@article{ha2018world,
  title={World models},
  author={Ha, David and Schmidhuber, J{\"u}rgen},
  journal={arXiv preprint arXiv:1803.10122},
  year={2018}
}

@article{assran2025v,
  title={V-jepa 2: Self-supervised video models enable understanding, prediction and planning},
  author={Assran, Mido and Bardes, Adrien and Fan, David and Garrido, Quentin and Howes, Russell and Muckley, Matthew and Rizvi, Ammar and Roberts, Claire and Sinha, Koustuv and Zholus, Artem and others},
  journal={arXiv preprint arXiv:2506.09985},
  year={2025}
}

@article{hafner2019dream,
  title={Dream to control: Learning behaviors by latent imagination},
  author={Hafner, Danijar and Lillicrap, Timothy and Ba, Jimmy and Norouzi, Mohammad},
  journal={arXiv preprint arXiv:1912.01603},
  year={2019}
}

@misc{sora,
  title = {SORA},
  author = {OpenAI},
  year = {2024},
  url = {https://openai.com/sora/}
}

@misc{genie3,
  title={Genie 3: A new frontier for world models},
  author={Ball, Philip J and Bauer, J and Belletti, F and others},
  year={2025}
}

@article{li2025wonderplay,
  title={WonderPlay: Dynamic 3D Scene Generation from a Single Image and Actions},
  author={Li, Zizhang and Yu, Hong-Xing and Liu, Wei and Yang, Yin and Herrmann, Charles and Wetzstein, Gordon and Wu, Jiajun},
  journal={arXiv preprint arXiv:2505.18151},
  year={2025}
}

@article{ren2025cosmos,
  title={Cosmos-Drive-Dreams: Scalable Synthetic Driving Data Generation with World Foundation Models},
  author={Ren, Xuanchi and Lu, Yifan and Cao, Tianshi and Gao, Ruiyuan and Huang, Shengyu and Sabour, Amirmojtaba and Shen, Tianchang and Pfaff, Tobias and Wu, Jay Zhangjie and Chen, Runjian and others},
  journal={arXiv preprint arXiv:2506.09042},
  year={2025}
}

@article{yolox2021,
  title={YOLOX: Exceeding YOLO Series in 2021},
  author={Ge, Zheng and Liu, Songtao and Wang, Feng and Li, Zeming and Sun, Jian},
  journal={arXiv preprint arXiv:2107.08430},
  year={2021}
}

@article{jiang2023rtmpose,
  title={Rtmpose: Real-time multi-person pose estimation based on mmpose},
  author={Jiang, Tao and Lu, Peng and Zhang, Li and Ma, Ningsheng and Han, Rui and Lyu, Chengqi and Li, Yining and Chen, Kai},
  journal={arXiv preprint arXiv:2303.07399},
  year={2023}
}

@inproceedings{ren:iclr2025,
 title     = {Diffusion Policy Policy Optimization},
 author    = {Allen Z. Ren and Justin Lidard and Lars Lien Ankile and Anthony Simeonov and Pulkit Agrawal and Anirudha Majumdar and Benjamin Burchfiel and Hongkai Dai and Max Simchowitz},
 booktitle = {ICLR},
 year      = {2025},
}

@article{sundaralingam:arxiv2023,
 author    = {Balakumar Sundaralingam and Siva Kumar Sastry Hari and Adam Fishman and Caelan Garrett and Karl Van Wyk and Valts Blukis and Alexander Millane and Helen Oleynikova and Ankur Handa and Fabio Ramos and Nathan Ratliff and Dieter Fox},
 title     = {{cuRobo}: Parallelized Collision-Free Minimum-Jerk Robot Motion Generation},
 journal   = {arXiv preprint arXiv:2310.17274},
 year      = {2023},
}

@INPROCEEDINGS{chi:rss2023,
 AUTHOR    = {Cheng Chi AND Siyuan Feng AND Yilun Du AND Zhenjia Xu AND Eric Cousineau AND Benjamin CM Burchfiel AND Shuran Song},
 TITLE     = {Diffusion Policy: Visuomotor Policy Learning via Action Diffusion},
 BOOKTITLE = {RSS},
 YEAR      = {2023},
}

@article{zhu2024irasim,
  title={IRASim: Learning Interactive Real-Robot Action Simulators},
  author={Zhu, Fangqi and Wu, Hongtao and Guo, Song and Liu, Yuxiao and Cheang, Chilam and Kong, Tao},
  journal={arXiv preprint arXiv:2406.14540},
  year={2024}
}

@misc{kling,
  title = {Kling},
  author = {KuaiShou},
  year = {2024},
  url = {https://klingai.com/}
}

@misc{gen3,
  title = {Gen 3},
  author = {Runway},
  year = {2024},
  url = {https://runwayml.com/research/introducing-gen-3-alpha}
}

@misc{minimax,
  title = {Hailuo},
  author = {MiniMax},
  year = {2024},
  url = {https://hailuoai.com/video}
}

@misc{veo3,
  title = {Veo 3},
  author = {Google DeepMind},
  year = {2025},
  month = {5},
  url = {https://deepmind.google/technologies/veo/veo-3/}
}

@article{gao2025seedance,
  title={Seedance 1.0: Exploring the Boundaries of Video Generation Models},
  author={Gao, Yu and Guo, Haoyuan and Hoang, Tuyen and Huang, Weilin and Jiang, Lu and Kong, Fangyuan and Li, Huixia and Li, Jiashi and Li, Liang and Li, Xiaojie and others},
  journal={arXiv preprint arXiv:2506.09113},
  year={2025}
}

@article{polyak2024movie,
  title={Movie gen: A cast of media foundation models},
  author={Polyak, Adam and Zohar, Amit and Brown, Andrew and Tjandra, Andros and Sinha, Animesh and Lee, Ann and Vyas, Apoorv and Shi, Bowen and Ma, Chih-Yao and Chuang, Ching-Yao and others},
  journal={arXiv preprint arXiv:2410.13720},
  year={2024}
}

@article{zhang2025waver,
  title={Waver: Wave Your Way to Lifelike Video Generation},
  author={Zhang, Yifu and Yang, Hao and Zhang, Yuqi and Hu, Yifei and Zhu, Fengda and Lin, Chuang and Mei, Xiaofeng and Jiang, Yi and Yuan, Zehuan and Peng, Bingyue},
  journal={arXiv preprint arXiv:2508.15761},
  year={2025}
}

@article{hacohen2024ltx,
  title={Ltx-video: Realtime video latent diffusion},
  author={HaCohen, Yoav and Chiprut, Nisan and Brazowski, Benny and Shalem, Daniel and Moshe, Dudu and Richardson, Eitan and Levin, Eran and Shiran, Guy and Zabari, Nir and Gordon, Ori and others},
  journal={arXiv preprint arXiv:2501.00103},
  year={2024}
}

@article{huang2025vipe,
  title={Vipe: Video pose engine for 3d geometric perception},
  author={Huang, Jiahui and Zhou, Qunjie and Rabeti, Hesam and Korovko, Aleksandr and Ling, Huan and Ren, Xuanchi and Shen, Tianchang and Gao, Jun and Slepichev, Dmitry and Lin, Chen-Hsuan and others},
  journal={arXiv preprint arXiv:2508.10934},
  year={2025}
}

@article{kong2024hunyuanvideo,
  title={Hunyuanvideo: A systematic framework for large video generative models},
  author={Kong, Weijie and Tian, Qi and Zhang, Zijian and Min, Rox and Dai, Zuozhuo and Zhou, Jin and Xiong, Jiangfeng and Li, Xin and Wu, Bo and Zhang, Jianwei and others},
  journal={arXiv preprint arXiv:2412.03603},
  year={2024}
}

@article{he2025unirelight,
  title={UniRelight: Learning Joint Decomposition and Synthesis for Video Relighting},
  author={He, Kai and Liang, Ruofan and Munkberg, Jacob and Hasselgren, Jon and Vijaykumar, Nandita and Keller, Alexander and Fidler, Sanja and Gilitschenski, Igor and Gojcic, Zan and Wang, Zian},
  journal={arXiv preprint arXiv:2506.15673},
  year={2025}
}

@article{wang2025frame,
  title={Frame In-N-Out: Unbounded Controllable Image-to-Video Generation},
  author={Wang, Boyang and Chen, Xuweiyi and Gadelha, Matheus and Cheng, Zezhou},
  journal={arXiv preprint arXiv:2505.21491},
  year={2025}
}

@inproceedings{ren2025gen3c,
  title={Gen3c: 3d-informed world-consistent video generation with precise camera control},
  author={Ren, Xuanchi and Shen, Tianchang and Huang, Jiahui and Ling, Huan and Lu, Yifan and Nimier-David, Merlin and M{\"u}ller, Thomas and Keller, Alexander and Fidler, Sanja and Gao, Jun},
  booktitle={CVPR},
  year={2025}
}

@article{bjorck2025gr00t,
  title={Gr00t n1: An open foundation model for generalist humanoid robots},
  author={Bjorck, Johan and Casta{\~n}eda, Fernando and Cherniadev, Nikita and Da, Xingye and Ding, Runyu and Fan, Linxi and Fang, Yu and Fox, Dieter and Hu, Fengyuan and Huang, Spencer and others},
  journal={arXiv preprint arXiv:2503.14734},
  year={2025}
}

@article{bordes2025intphys,
  title={IntPhys 2: Benchmarking Intuitive Physics Understanding In Complex Synthetic Environments},
  author={Bordes, Florian and Garrido, Quentin and Kao, Justine T and Williams, Adina and Rabbat, Michael and Dupoux, Emmanuel},
  journal={arXiv preprint arXiv:2506.09849},
  year={2025}
}

@article{zhao2025synthetic,
  title={Are Synthetic Videos Useful? A Benchmark for Retrieval-Centric Evaluation of Synthetic Videos},
  author={Zhao, Zecheng and Song, Selena and Chen, Tong and Chen, Zhi and Sadiq, Shazia and Luo, Yadan},
  journal={arXiv preprint arXiv:2507.02316},
  year={2025}
}

@article{yue2025ewmbench,
  title={Ewmbench: Evaluating scene, motion, and semantic quality in embodied world models},
  author={Yue, Hu and Huang, Siyuan and Liao, Yue and Chen, Shengcong and Zhou, Pengfei and Chen, Liliang and Yao, Maoqing and Ren, Guanghui},
  journal={arXiv preprint arXiv:2505.09694},
  year={2025}
}

@article{liao2025genie,
  title={Genie envisioner: A unified world foundation platform for robotic manipulation},
  author={Liao, Yue and Zhou, Pengfei and Huang, Siyuan and Yang, Donglin and Chen, Shengcong and Jiang, Yuxin and Hu, Yue and Cai, Jingbin and Liu, Si and Luo, Jianlan and others},
  journal={arXiv preprint arXiv:2508.05635},
  year={2025}
}

@article{zhou2025vlm4d,
  title={Vlm4d: Towards spatiotemporal awareness in vision language models},
  author={Zhou, Shijie and Vilesov, Alexander and He, Xuehai and Wan, Ziyu and Zhang, Shuwang and Nagachandra, Aditya and Chang, Di and Chen, Dongdong and Wang, Xin Eric and Kadambi, Achuta},
  journal={arXiv preprint arXiv:2508.02095},
  year={2025}
}

@article{bansal2025videophy,
  title={Videophy-2: A challenging action-centric physical commonsense evaluation in video generation},
  author={Bansal, Hritik and Peng, Clark and Bitton, Yonatan and Goldenberg, Roman and Grover, Aditya and Chang, Kai-Wei},
  journal={arXiv preprint arXiv:2503.06800},
  year={2025}
}

@article{guo2025t2vphysbench,
  title={T2vphysbench: A first-principles benchmark for physical consistency in text-to-video generation},
  author={Guo, Xuyang and Huo, Jiayan and Shi, Zhenmei and Song, Zhao and Zhang, Jiahao and Zhao, Jiale},
  journal={arXiv preprint arXiv:2505.00337},
  year={2025}
}

@article{motamed2025generative,
  title={Do generative video models understand physical principles?},
  author={Motamed, Saman and Culp, Laura and Swersky, Kevin and Jaini, Priyank and Geirhos, Robert},
  journal={arXiv preprint arXiv:2501.09038},
  year={2025}
}

@article{duan2025worldscore,
  title={Worldscore: A unified evaluation benchmark for world generation},
  author={Duan, Haoyi and Yu, Hong-Xing and Chen, Sirui and Fei-Fei, Li and Wu, Jiajun},
  journal={arXiv preprint arXiv:2504.00983},
  year={2025}
}

@article{yang2025embodiedbench,
  title={Embodiedbench: Comprehensive benchmarking multi-modal large language models for vision-driven embodied agents},
  author={Yang, Rui and Chen, Hanyang and Zhang, Junyu and Zhao, Mark and Qian, Cheng and Wang, Kangrui and Wang, Qineng and Koripella, Teja Venkat and Movahedi, Marziyeh and Li, Manling and others},
  journal={arXiv preprint arXiv:2502.09560},
  year={2025}
}

@article{bansal2024videophy,
  title={Videophy: Evaluating physical commonsense for video generation},
  author={Bansal, Hritik and Lin, Zongyu and Xie, Tianyi and Zong, Zeshun and Yarom, Michal and Bitton, Yonatan and Jiang, Chenfanfu and Sun, Yizhou and Chang, Kai-Wei and Grover, Aditya},
  journal={arXiv preprint arXiv:2406.03520},
  year={2024}
}

@misc{li2022bevformerlearningbirdseyeviewrepresentation,
      title={BEVFormer: Learning Bird's-Eye-View Representation from Multi-Camera Images via Spatiotemporal Transformers}, 
      author={Zhiqi Li and Wenhai Wang and Hongyang Li and Enze Xie and Chonghao Sima and Tong Lu and Qiao Yu and Jifeng Dai},
      year={2022},
      eprint={2203.17270},
      archivePrefix={arXiv},
      primaryClass={cs.CV},
      url={https://arxiv.org/abs/2203.17270}, 
}

@misc{luo2023latr3dlanedetection,
      title={LATR: 3D Lane Detection from Monocular Images with Transformer}, 
      author={Yueru Luo and Chaoda Zheng and Xu Yan and Tang Kun and Chao Zheng and Shuguang Cui and Zhen Li},
      year={2023},
      eprint={2308.04583},
      archivePrefix={arXiv},
      primaryClass={cs.CV},
      url={https://arxiv.org/abs/2308.04583}, 
}

@article{fu2025llm,
  title={LLM-based Realistic Safety-Critical Driving Video Generation},
  author={Fu, Yongjie and Zha, Ruijian and Tian, Pei and Di, Xuan},
  journal={arXiv preprint arXiv:2507.01264},
  year={2025}
}

@article{singer2023text,
  title={Text-to-4d dynamic scene generation},
  author={Singer, Uriel and Sheynin, Shelly and Polyak, Adam and Ashual, Oron and Makarov, Iurii and Kokkinos, Filippos and Goyal, Naman and Vedaldi, Andrea and Parikh, Devi and Johnson, Justin and others},
  journal={arXiv preprint arXiv:2301.11280},
  year={2023}
}

@inproceedings{liu2025dynamicscaler,
  title={Dynamicscaler: Seamless and scalable video generation for panoramic scenes},
  author={Liu, Jinxiu and Lin, Shaoheng and Li, Yinxiao and Yang, Ming-Hsuan},
  booktitle={CVPR},
  year={2025}
}

@article{watson2024controlling,
  title={Controlling space and time with diffusion models},
  author={Watson, Daniel and Saxena, Saurabh and Li, Lala and Tagliasacchi, Andrea and Fleet, David J},
  journal={arXiv preprint arXiv:2407.07860},
  year={2024}
}

@article{zhao2024genxd,
  title={Genxd: Generating any 3d and 4d scenes},
  author={Zhao, Yuyang and Lin, Chung-Ching and Lin, Kevin and Yan, Zhiwen and Li, Linjie and Yang, Zhengyuan and Wang, Jianfeng and Lee, Gim Hee and Wang, Lijuan},
  journal={arXiv preprint arXiv:2411.02319},
  year={2024}
}

@misc{hung2024let3daplongitudinalerrortolerant,
      title={LET-3D-AP: Longitudinal Error Tolerant 3D Average Precision for Camera-Only 3D Detection}, 
      author={Wei-Chih Hung and Vincent Casser and Henrik Kretzschmar and Jyh-Jing Hwang and Dragomir Anguelov},
      year={2024},
      eprint={2206.07705},
      archivePrefix={arXiv},
      primaryClass={cs.CV},
      url={https://arxiv.org/abs/2206.07705}, 
}

@misc{stylegan_v,
    title={StyleGAN-V: A Continuous Video Generator with the Price, Image Quality and Perks of StyleGAN2},
    author={Ivan Skorokhodov and Sergey Tulyakov and Mohamed Elhoseiny},
    journal={arXiv preprint arXiv:2112.14683},
    year={2021}
}

@article{sampson1987,
title = {Fitting conic sections to “very scattered” data: An iterative refinement of the bookstein algorithm},
journal = {Computer Graphics and Image Processing},
year = {1982},
issn = {0146-664X},
doi = {https://doi.org/10.1016/0146-664X(82)90101-0},
url = {https://www.sciencedirect.com/science/article/pii/0146664X82901010},
author = {Paul D Sampson},
}

@article{yang2024cogvideox,
  title={Cogvideox: Text-to-video diffusion models with an expert transformer},
  author={Yang, Zhuoyi and Teng, Jiayan and Zheng, Wendi and Ding, Ming and Huang, Shiyu and Xu, Jiazheng and Yang, Yuanming and Hong, Wenyi and Zhang, Xiaohan and Feng, Guanyu and others},
  journal={arXiv preprint arXiv:2408.06072},
  year={2024}
}

@article{hurst2024gpt,
  title={Gpt-4o system card},
  author={Hurst, Aaron and Lerer, Adam and Goucher, Adam P and Perelman, Adam and Ramesh, Aditya and Clark, Aidan and Ostrow, AJ and Welihinda, Akila and Hayes, Alan and Radford, Alec and others},
  journal={arXiv preprint arXiv:2410.21276},
  year={2024}
}

@article{PAI-Bench,
  title={PAI-Bench: A Comprehensive Benchmark For Physical AI},
  author={Zhou, Fengzhe and Huang, Jiannan and Li, Jialuo and Ramanan, Deva and Shi, Humphrey},
  journal={arXiv preprint arXiv:2512.01989},
  year={2025}
}

@article{nasiriany2024robocasa,
  title={Robocasa: Large-scale simulation of everyday tasks for generalist robots},
  author={Nasiriany, Soroush and Maddukuri, Abhiram and Zhang, Lance and Parikh, Adeet and Lo, Aaron and Joshi, Abhishek and Mandlekar, Ajay and Zhu, Yuke},
  journal={arXiv preprint arXiv:2406.02523},
  year={2024}
}

@article{zheng2025large,
  title={Large Scale Diffusion Distillation via Score-Regularized Continuous-Time Consistency},
  author={Zheng, Kaiwen and Wang, Yuji and Ma, Qianli and Chen, Huayu and Zhang, Jintao and Balaji, Yogesh and Chen, Jianfei and Liu, Ming-Yu and Zhu, Jun and Zhang, Qinsheng},
  journal={arXiv preprint arXiv:2510.08431},
  year={2025}
}

@article{ye2025data,
  title={Data-regularized Reinforcement Learning for Diffusion Models at Scale},
  author={Ye, Haotian and Zheng, Kaiwen and Xu, Jiashu and Li, Puheng and Chen, Huayu and Han, Jiaqi and Liu, Sheng and Zhang, Qinsheng and Mao, Hanzi and Hao, Zekun and others},
  journal={arXiv preprint arXiv:2512.04332},
  year={2025}
}

\end{document}